\documentclass[10pt,twocolumn]{article}

\usepackage[pagenumbers]{cvpr} 

\usepackage{graphicx}
\usepackage{amsmath}
\usepackage{amssymb}
\usepackage{booktabs}

\usepackage[pagebackref,breaklinks,colorlinks]{hyperref}

\usepackage[capitalize]{cleveref}
\crefname{section}{Sec.}{Secs.}
\Crefname{section}{Section}{Sections}
\Crefname{table}{Table}{Tables}
\crefname{table}{Tab.}{Tabs.}

\def\cvprPaperID{8849} 
\def\confName{CVPR}
\def\confYear{2023}

\usepackage[utf8]{inputenc} 
\usepackage[T1]{fontenc}    
\usepackage{hyperref}       
\usepackage{url}            
\usepackage{amsfonts}       
\usepackage{nicefrac}       
\usepackage{microtype}      
\usepackage{xcolor}         

\usepackage{microtype}

\usepackage[toc,page,header]{appendix}
\usepackage{minitoc} 
\usepackage{pifont} 

\usepackage{xcolor}  
\definecolor{ForestGreen}{HTML}{228B22}
\definecolor{NavyBlue}{HTML}{00008B}

\DeclareGraphicsExtensions{.pdf,.jpeg,.jpeg}
\graphicspath{{Images/}} 
\usepackage{rotating} 
\usepackage{dblfloatfix} 

\usepackage{multirow}
\usepackage{afterpage} 
\usepackage[normalem]{ulem} 

\usepackage[noadjust]{cite} 

\usepackage{algorithmic}
\usepackage[ruled, vlined]{algorithm2e}
\DontPrintSemicolon

\usepackage{diagbox}
\usepackage{amsthm}
\usepackage{mathtools}
\usepackage{extarrows} 
\usepackage{amsfonts}
\usepackage{mathrsfs} 
\usepackage{physics} 
\usepackage{bm} 
\usepackage{array} 
\usepackage{relsize} 

\usepackage{url}

\usepackage{caption}
\usepackage{placeins} 

\newcommand{\pd}{p_d}

\newcommand{\pg}{p_g}
\newcommand{\pz}{p_z}

\newcommand{\mbbI}{\mathbb{I}}

\newcommand{\bmc}{\bm{c}}

\newcommand{\x}{\bm{x}}

\newcommand{\y}{\bm{y}}
\newcommand{\z}{\bm{z}}

\newcommand{\mcalC}{\mathcal{C}}

\newcommand{\mcalN}{\mathcal{N}}

\newcommand{\mfrakD}{\mathfrak{D}}
\newcommand{\rmd}{\mathrm{d}}

\makeatletter
\newcommand{\subalign}[1]{%
  \vcenter{%
    \Let@ \restore@math@cr \default@tag
    \baselineskip\fontdimen10 \scriptfont\tw@
    \advance\baselineskip\fontdimen12 \scriptfont\tw@
    \lineskip\thr@@\fontdimen8 \scriptfont\thr@@
    \lineskiplimit\lineskip
    \ialign{\hfil$\m@th\scriptstyle##$&$\m@th\scriptstyle{}##$\hfil\crcr
      #1\crcr
    }%
  }%
}
\makeatother

\usepackage{transparent} 

\usepackage{array}
\newcolumntype{P}[1]{>{\centering\arraybackslash}p{#1}}

\title{Spider GAN: Leveraging Friendly Neighbors to Accelerate GAN Training}

\author{%
  Siddarth Asokan\thanks{Siddarth Asokan is funded by the Qualcomm Innovation Fellowship, and the Robert Bosch Center for Cyber-Physical Systems  Ph.D. Fellowship.} \\
  Robert Bosch Center for Cyber-Physical Systems\\
  Indian Institute of Science\\
  Bengaluru - 50012, India\\
  \texttt{siddartha@iisc.ac.in} \\
  \and
    Chandra Sekhar Seelamantula\\
  Department of Electrical Engineering\\
  Indian Institute of Science\\
  Bengaluru - 50012, India\\
  \texttt{css@iisc.ac.in} \\
}

\begin{document}
\doparttoc 
\faketableofcontents

\maketitle

\begin{abstract}
Training Generative adversarial networks (GANs) stably is a challenging task. The generator in GANs transform noise vectors, typically Gaussian distributed, into realistic data such as images. In this paper, we propose a novel approach for training GANs with images as inputs, but without enforcing any pairwise constraints. The intuition is that images are more structured than noise, which the generator can leverage to learn a more robust transformation. The process can be made efficient by identifying closely related datasets, or a ``friendly neighborhood'' of the target distribution, inspiring the moniker, {\it Spider GAN}. To define friendly neighborhoods leveraging proximity between datasets, we propose a new measure called the {\it signed inception distance} (SID), inspired by the polyharmonic kernel. We show that the Spider GAN formulation results in faster convergence, as the generator can {\it discover} correspondence even between seemingly unrelated datasets, for instance, between Tiny-ImageNet and CelebA faces. Further, we demonstrate {\it cascading} Spider GAN, where the output distribution from a pre-trained GAN generator is used as the input to the subsequent network. Effectively, transporting one distribution to another in a cascaded fashion until the target is learnt --  a new flavor of transfer learning.  We demonstrate the efficacy of the {\it Spider} approach on DCGAN, conditional GAN, PGGAN, StyleGAN2 and StyleGAN3. The proposed approach achieves state-of-the-art Fr\'echet inception distance (FID) values, with one-fifth of the training iterations, in comparison to their baseline counterparts  on high-resolution small datasets such as MetFaces, Ukiyo-E Faces and AFHQ-Cats.

\end{abstract}

\section{Introduction}
Generative adversarial networks (GANs)~\cite{SGAN14} are designed to model the underlying distribution of a target dataset (with underlying distribution \(\pd\)) through a {\it min-max} optimization between the generator \(G\) and the discriminator \(D\) networks. The generator transforms an input \(\z\sim\pz\), typically Gaussian or uniform distributed, into a generated sample \(G(\z) \sim \pg\).  The discriminator is trained to classify samples drawn from \(\pg\) or \(\pd\) as real or fake. The optimal generator is the one that outputs images that confuse the discriminator.  \par

\begin{figure*}[!th]
\begin{center}
  \begin{tabular}[b]{P{.4\linewidth}|P{.4\linewidth}}
        \includegraphics[width=1\linewidth]{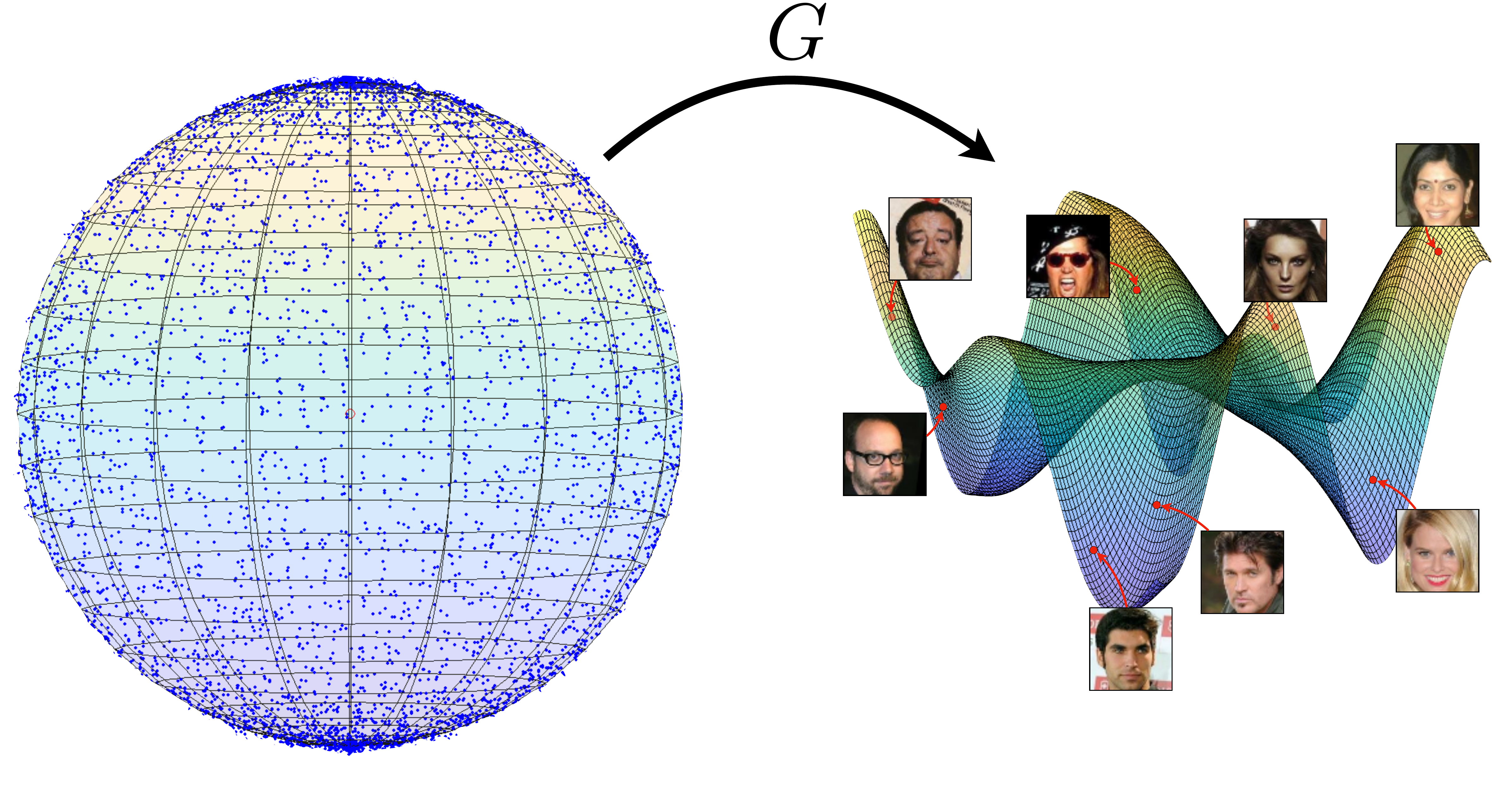} & 
          \includegraphics[width=1\linewidth]{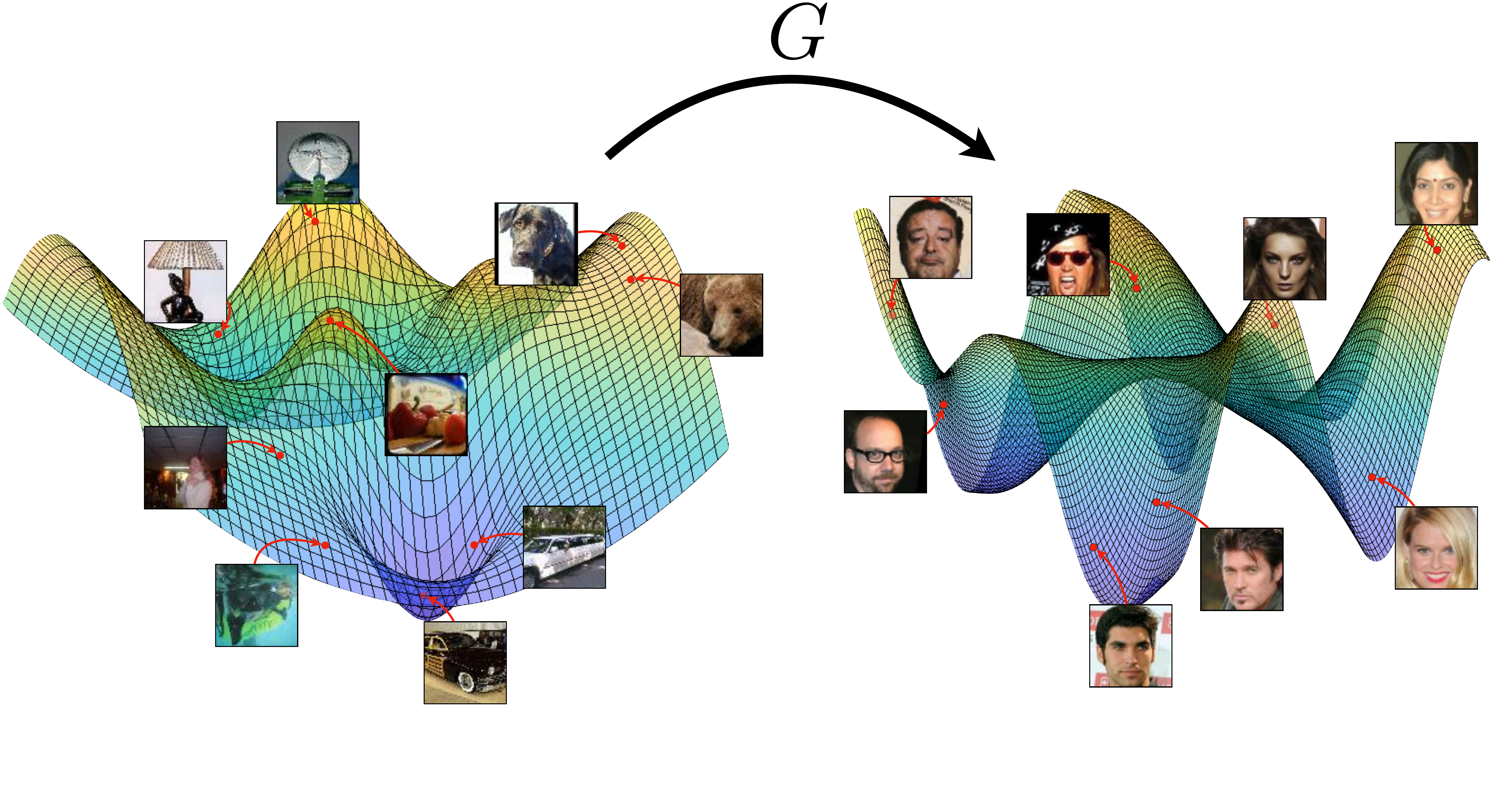}  \\[-5pt]
          (a) Classical GANs & (b) Spider GAN
  \end{tabular} 
  \caption[]{(\includegraphics[height=0.009\textheight]{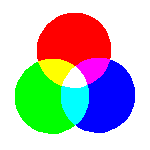} Color online) A comparison of design philosophies of the standard GANs and Spider GAN. (a) A prototypical GAN transforms high-dimensional Gaussian data, which is concentrated at the surface of hyperspheres in \(n\)-D, into an image distribution comprising a union of low-dimensional manifolds embedded in a higher-dimensional space. (b) The Spider GAN generator aims to learn a simpler transformation between two closely related data manifolds in an unconstrained manner, thereby accelerating convergence.} 
   \vspace{-2.em}
  \label{Fig_SpiderGAN}  
  \end{center}
\end{figure*}



\noindent {\it {\bfseries Inputs to the GAN generator:}} The input distribution plays a definitive role in the quality of GAN output. Low-dimensional latent vectors have been shown to help disentangle the representations and control features of the target being learnt~\cite{DisentangleGAN17,OTInterpol19}. Prior work on optimizing the latent distribution in GANs has been motivated by the need to improve the quality of interpolated images. Several works have considered replacing the Gaussian prior with Gaussian mixtures,  Gamma, non-parametric distributions, etc~\cite{DeliGAN17,SamplingPrior16,Gamma18,Cauchy19,Googol19,NonPara19}. Alternatively, the GAN generator can be trained with the latent-space distribution of the target dataset, as learnt by variational autoencoders~\cite{VAEGAN16,DCVAE21}. However, such approaches are not in conformity with the low-dimensional manifold structure of real data. Khayatkhoei {\it et al.}~\cite{DMGAN18} attributed the poor quality of the interpolates to the disjoint structure of data distribution in high-dimensions, which motivates the need for an  informed choice of the input distribution.  \par

\noindent{\it {\bfseries GANs and image-to-image translation:}} GANs that accept images as input fall under the umbrella of {\it image translation}. Here, the task is to modify particular features of an image, either within domain (style transfer) or across domains (domain adaptation). Examples for in-domain translation include changing aspects of face images, such as the expression, gender, accessories, etc.~\cite{InfoGAN16,StyleGAN19,FaceEdit20}, or modifying the illumination or seasonal characteristics of natural scenes~\cite{Seasons17}.  On the other hand, {\it domain adaptation tasks} aim at transforming the image from one style to another. Common applications include simulation to real-world translation~\cite{PixelDA17,ADDA17,DomainAda18,ImgExtra21}, or translating images across styles of artwork~\cite{GANilla20,Pix2Pix18,CycleGAN17}. While the supervised Pix2Pix framework~\cite{Pix2Pix18}
originally proposed training GANs with pairs of images drawn from the source and target domains, semi-supervised and unsupervised extensions~\cite{CycleGAN17,UNIT17,DualingGAN17,CyCADA18,DivPix2Pix18,MUNIT18} tackle the problem in an unpaired setting, and introduce modifications such as cycle-consistency or the addition of regularization functionals to the GAN loss to maintain a measure of consistency between images. Existing domain-adaptation GANs~\cite{FewShotDA21,StarGANv220} enforce cross-domain consistency to retain visual similarity. Ultimately, these approaches rely on enforcing some form of coupling between the source and the target via feature-space mapping. \par

\section{The Proposed Approach: Spider GAN} \label{Sec_Approach}
We propose the Spider GAN formulation motivated by the low-dimensional disconnected manifold structure of data~\cite{ManifoldHyp16,DMGAN18,SubGAN18,noGANsLand20}. Spider GANs lie at the cross-roads between classical GANs and image-translation GANs. As opposed to optimizing the latent parametric prior, we hypothesize that providing the generator with closely related image source datasets, (dubbed the {\it friendly neighborhood}, leading to the moniker {\it Spider GAN}) will result in superior convergence of the GAN. Unlike image translation tasks, the Spider GAN generator is agnostic to individual input-image features, and is allowed to {\it discover} implicit structure in the mapping from the source distribution to the target. Figure~\ref{Fig_SpiderGAN} depicts the design philosophy of Spider GAN juxtaposed with the classical GAN training approach. \par
The choice of the {\it input dataset} affects the generator's ability to learn a stable and accurate mapping. Intuitively, if the GAN has to be trained to learn the distribution of {\it street view house numbers }(SVHN)~\cite{SVHN}, the MNIST~\cite{MNIST} dataset proves to be a better initialization of the input space than standard densities such as the uniform or Gaussian. It is a well known result that, for a given mean and variance, the Gaussian has maximum entropy, while for a given support (say, \([-1,1]\) when training with re-normalized images), the uniform distribution has maximum entropy~\cite{InfoTheory05}. However, image datasets are highly structured, and possess lower entropy~\cite{GenTop17}. Therefore, one could interpret the generative modeling of images using GANs as effectively one of entropy minimization~\cite{InfoGAN16}. We argue that choosing a low entropy input distribution that is structurally closer to the target would lead to a more efficient generator transformation, thereby accelerating the training process. Existing image-translation approaches aim to maintain semantic information, for example, translating a specific instance of the digit `2' in the MNIST dataset to the SVHN style. However, the Spider GAN formulation neither enforces nor requires such constraints. Rather, it allows for an implicit structure in the source dataset to be used to learn the target efficiently. It is entirely possible for the {\it Trouser} class in Fashion-MNIST~\cite{FMNIST} to map to the digit `1' in MNIST due to structural similarity. Thus, the scope of Spider GAN is much wider than image translation. 
\subsection{Our Contributions} \label{Sec_Contributions} 
In Section~\ref{Sec_FriendlyNb}, we discuss the central focus in Spider GANs: defining what constitutes a {\it friendly neighborhood}. Preliminary experiments suggest that, while the well known Fr{\'e}chet  inception distance (FID)~\cite{TTGAN18} and kernel inception distance (KID)~\cite{DemistifyMMD18} are able to capture visual similarity, they are unable to quantify the diversity of samples in the underlying manifold. We therefore propose a novel distance measure to evaluate the input to GANs, one that is motivated by electrostatic potential fields and charge neutralization between the (positively charged) target data samples and (negatively charged) generator samples~\cite{CoulombGAN18,PolyLSGAN22}, named {\it signed inception distance} (SID) (Section~\ref{Sec_SID}). An implementation of SID atop the Clean-FID~\cite{CleanFID21} backbone is available at \url{https://github.com/DarthSid95/clean-sid}. We identify {\it friendly neighborhoods} for multiple classes of standard image datasets such as MNIST, Fashion MNIST, SVHN, CIFAR-10~\cite{CIFAR10}, Tiny-ImageNet~\cite{TIN}, LSUN-Churches~\cite{LSUN}, CelebA~\cite{CelebA}, and Ukiyo-E Faces~\cite{UkiyoE}. We present experimental validation on training the {\it Spider} variant of DCGAN~\cite{DCGAN} (Section~\ref{Sec_Exp}) and show that it results in up to 30\% improvement in terms of FID, KID and cumulative SID of the converged models. The {\it Spider} framework is lightweight and can be extended to any GAN architecture, which we demonstrate via class-conditional learning with the {\it Spider} variant of auxiliary classifier GANs (ACGANs)~\cite{ACGAN17} (Section~\ref{Sec_Exp}). The source code for Spider GANs built atop the DCGAN architecture are available at \url{https://github.com/DarthSid95/SpiderDCGAN}.  We also present a novel approach to transfer learning using Spider GANs by feeding the output distribution of a pre-trained generator to the input of the subsequent stage (Section~\ref{Sec_Cascade}). Considering progressively growing GAN (PGGAN)~\cite{PGGAN18} and StyleGAN~\cite{StyleGAN220,ADAStyleGAN20,StyleGAN321} architectures, we show that the corresponding {\it Spider} variants achieve competitive FID scores in one-fifth of the training iterations on FFHQ~\cite{StyleGAN19} and AFHQ-Cats~\cite{StarGANv220}, while achieving state-of-the-art FID on high-resolution small-sized datasets such as Ukiyo-E Faces and MetFaces~\cite{ADAStyleGAN20} (Section~\ref{Sec_SpiderPGStyleGAN}). The source code for implementing Spider StyleGANs is available at \url{https://github.com/DarthSid95/SpiderStyleGAN}.

\subsection{Related Works}\label{Sec_HighDim}
The choice of the input distribution in GANs determines the quality of images generated by feeding the generator interpolated points, which in turn is determined by the probability of the interpolated points lying on the manifold. High-dimensional Gaussian random vectors are concentrated on the surface of a hypersphere ({\it Gaussian annulus theorem}~\cite{RaviKannan213}), akin to a  {\it soap bubble}, resulting in interpolated points that are less likely to lie on the manifold. Alternatives such as the Gamma~\cite{Gamma18} or Cauchy~\cite{Cauchy19} prior result in superior performance over interpolated points, while Singh {\it et al.}~\cite{NonPara19}  derive a non-parametric prior that minimized the divergence between the input and the midpoint distributions. \par
A well known result in high-dimensional data analysis is that structured datasets are embedded in a low-dimensional manifold with an {\it intrinsic dimensionality}  (\(n_{\mfrakD}\)) significantly lower than the ambient dimensionality \(n\)~\cite{GenTop17}. For instance, in MNIST, \(n=784\), while \( n_{\mfrakD}\approx12\)~\cite{MNIST11D05}. Feng {\it et al.}\cite{HomeoGAN21} showed that the mismatch between \(n_{\mfrakD}\) of the generator input and its output adversely affects performance. Although in practice, estimating \(n_{\mfrakD}\) may not always be possible~\cite{MNIST11D05,DMGAN18,EstimateID17}, these results justify picking input distributions that are structurally similar to the target. In instance-conditioned GANs~\cite{ICGAN21}, the target data is modeled as clusters on the data manifold to improve learning.  \par
The philosophy of cascading Spider GAN generators runs in parallel to input optimization in transfer learning with GANs, such as Mine GAN~\cite{MineGAN20} where {\it mining} networks are implemented that transform the input distribution of the GAN nonlinearly to learn the target samples better. Kerras {\it et al.}~\cite{ADAStyleGAN20} showed that transfer learning improves the performance of GANs on small datasets, and observed empirically that transferring weights from models trained on visually diverse data lead to better performance of the target model.

\section{Where is the Friendly Neighborhood?}  \label{Sec_FriendlyNb}

We now consider various distance measures between datasets that can be used to identify the friendly neighborhood/source dataset in Spider GANs. While the most direct approach is to compare the intrinsic dimensions of the manifolds, such approaches are either computationally intensive~\cite{IntrinsicDim16}, or do not scale with sample size~\cite{MNIST11D05,EstimateID17}. We observed that the friendly neighbors detected by such approach did not correlate well experimentally, and therefore, defer discussions on such methods to Appendix~\ref{App_FriendNbd}. \par

Based on the approach advocated by Wang {\it et al.}~\cite{TransferringGANs18} to identify pre-trained GAN networks for transfer learning, we initially considered FID and KID to identify {\it friendly neighbors}. We use the FID to measure the distance between the source (generator input) and the target data distributions. A source that has a lower FID is closer to the target and will serve as a better input to the generator. The first four columns of Table~\ref{Table_FID_SID} present FID scores between the standard datasets we consider in this paper. The \textcolor{ForestGreen}{\bf first}, \textcolor{blue}{\bf second} and \textcolor{red}{\bf third} {\it friendly neighbors} (color coded) of a target dataset are the source datasets with the lowest three FIDs.  As observed from~Table~\ref{Table_FID_SID}, a limitation is that the FID of a dataset with itself is not always zero, which is counterintuitive for a distance measure. In cases such as CIFAR-10 or TinyImageNet, this is indicative of the variability in the dataset, and in Ukiyo-E Faces, this is due to limited availability of data samples, which has been shown to negatively affect FID estimation~\cite{PR, DemistifyMMD18}. 
FID satisfies {\it reciprocity}, {\it i.e.,} it identifies datasets as being mutually close to each other, such as CIFAR-10 and Tiny-ImageNet. However, preliminary experiments on training Spider GAN using FID to identify friendly neighbors showed that the relative diversity between datasets is not captured. Given a source, learning a less diverse target distribution is easier (cf. Section~\ref{Sec_Exp} and Appendix~\ref{App_SpiderDCGAN}). These issues are similar to the observations made by Kerras {\it et al.}~\cite{ADAStyleGAN20} in the context of weight transfer. This can be understood via an example --- fitting a multimodal target Gaussian having 10 modes would be easier with a 20-component source distribution than a 5-component one.

\begin{table*}[t!]
\fontsize{7}{12}\selectfont
\begin{center}
\caption{A comparison of FID and CSID\(_{m}\) between popular training datasets for \(m = \lfloor\frac{n}{2}\rfloor\). The rows represent the source and the columns correspond to the target. The \textcolor{ForestGreen}{\bf first}, \textcolor{blue}{\bf second} and \textcolor{red}{\bf third}  {\it friendly neighbors} of the target are the sources with the three lowest FID, or lowest positive CSID values, respectively. CSID is superior to FID, as it assigns negative values to sources that are less diverse than the target. MNIST and Fashion-MNIST are \textcolor{gray}{shown in gray} to denote scenarios where grayscale images are not valid sources for the color-image targets.} \label{Table_FID_SID}  \vskip-0.1in
\begin{tabular}{P{2.22cm}||P{1.1cm}|P{1.32cm}|P{1.52cm}|P{1.1cm}||P{1.1cm}|P{1.32cm}|P{1.52cm}|P{1.1cm}}
\toprule
\multirow{2}{*}{\backslashbox{Source}{Target}}& \multicolumn{4}{c||}{\(\mathrm{FID}\,(\mathrm{Source}\,,\,\mathrm{Target})\)}&\multicolumn{4}{c}{\(\mathrm{CSID}_{m}(\mathrm{Source}\,\|\,\mathrm{Target})\)} \\
\cline{2-9}
& MNIST & CIFAR-10 & TinyImageNet &Ukiyo-E & MNIST & CIFAR-10 & TinyImageNet &Ukiyo-E  \\[3pt]
\hline\hline
MNIST 			& 1.2491 & \textcolor{black}{{\transparent{0.3}258.246}} & \textcolor{black}{{\transparent{0.3}264.250}} & \textcolor{black}{{\transparent{0.3}398.280}} & 0.1863 & \textcolor{black}{{\transparent{0.3}29.298}} & \textcolor{black}{{\transparent{0.3}9.436}} & \textcolor{black}{{\transparent{0.3}201.550}} \\[1pt]
F-MNIST   		& \textcolor{ForestGreen}{\bf176.813} & \textcolor{black}{{\transparent{0.4}188.367}} & \textcolor{black}{{\transparent{0.4}197.057}} & \textcolor{black}{{\transparent{0.4}387.049}} & \textcolor{ForestGreen}{\bf162.962} & \textcolor{black}{{\transparent{0.4}19.051}} & \textcolor{black}{{\transparent{0.4}-2.5571}} & \textcolor{black}{{\transparent{0.4}191.010}} \\[1pt] \hline
SVHN  			& \textcolor{blue}{\bf236.707} & \textcolor{blue}{\bf168.615} & \textcolor{blue}{\bf189.133} & 372.444 & 212.473 & \textcolor{red}{\bf34.534} & \textcolor{red}{\bf21.668} & 214.507 \\[1pt]
CIFAR-10 		& \textcolor{red}{\bf259.045} & 5.0724 & \textcolor{ForestGreen}{\bf64.3941} & 303.694 & 221.337 & -0.1487 & -7.109 & 198.991 \\[1pt]
TinyImageNet 		& 264.309 & \textcolor{ForestGreen}{\bf64.0312} & 6.4854 & \textcolor{ForestGreen}{\bf257.078} & 230.916 & \textcolor{ForestGreen}{\bf12.892} & 0.6743 & \textcolor{blue}{\bf197.447} \\[1pt]
CelebA 			& 360.773 & 303.490 & \textcolor{red}{\bf250.735} & \textcolor{red}{\bf301.108} & \textcolor{blue}{\bf204.794} & \textcolor{blue}{\bf23.685} & \textcolor{ForestGreen}{\bf8.829} & \textcolor{ForestGreen}{\bf184.170} \\[1pt]
Ukiyo-E 		& 396.791 & 300.511 & 254.102 & 5.9137 & 250.226 & 39.793 & \textcolor{blue}{\bf18.727} & 0.5494\\[1pt]
Church 			& 350.708 &  \textcolor{red}{\bf294.982} & 254.991 & \textcolor{blue}{\bf267.638} & \textcolor{red}{\bf212.452} & -4.655 & -23.115 & \textcolor{red}{\bf198.750} \\
\bottomrule
\end{tabular}
\end{center}
\vskip-2em
\end{table*}

\subsection{The Signed Inception Distance (SID)} \label{Sec_SID}
Given the limitations of FID discussed above, we propose a novel signed distance for measuring the proximity between two distributions. The distance is ``signed'' in the sense that it can also take negative values. Further, it is not symmetric. The distance is also practical to compute because it is expressed in terms of the samples drawn from the distributions. The proposed distance draws inspiration from the improved precision-recall scores of GANs~\cite{ImprovedPR19} and the potential-field interpretation in Coulomb GANs~\cite{CoulombGAN18} and Poly-LSGAN~\cite{PolyLSGAN22}. Consider batches of samples drawn from distributions \(\mu_p\) and \(\mu_q\), given by \(\mfrakD_p = \{ \tilde{\bmc}_i\}_{i=1}^{N_p}\) and  \(\mfrakD_q = \{ \bmc_j\}_{j=1}^{N_q}\), respectively. Given a test vector \(\x \in \mathbb{R}^n\), consider the Coulomb GAN discriminator~\cite{CoulombGAN18}:
\begin{align}
f(\x) = \frac{1}{N_p} \sum_{\substack{i=1\\\tilde{\bmc}_i\sim \mu_p}}^{N_p}\!\!\Phi(\x,\tilde{\bmc}_i) -  \frac{1}{N_q} \sum_{\substack{j=1\\\bmc_j\sim \mu_q}}^{N_q}\!\! \Phi(\x,\bmc_j),
\label{Eqn_CoulombPot}
\end{align}
where \(\Phi\) is the polyharmonic kernel~\cite{PolyLSGAN22,PolyFunctions}:
\begin{align*}
 \Phi(\x,\y)\!=\!\kappa_{m,n}\begin{cases}
\|\x\!-\!\y\|^{2m-n}, & \subalign{&\text{if}~2m-n < 0 \\& \text{or}~n~\text{is odd,}}\\
\|\x\!-\!\y\|^{2m-n}\ln(\|\x\!-\!\y\|), & \subalign{&\text{if}~2m-n \geq 0~ \\& \text{and}~n~\text{is even,}} 
\end{cases}\!\!,
\end{align*}
and \(\kappa_{m,n}\) is a positive constant, given the order \(m\) and dimensionality \(n\).  The higher-order generalization gives us more flexibility and numerical stability in computation. We use \(m \approx \lfloor \frac{n}{2} \rfloor \) as a stable choice, while ablation studies on choosing \(m\) are given in Appendix~\ref{App_GANsSID} \par

From the perspective of electrostatics, for \(\mu_p = \pg\) and \(\mu_q = \pd\), \(f(\x)\) in Equation~\eqref{Eqn_CoulombPot} treats the target data as negative charges, and generator samples as positive charges.  The quality of \(\mu_p\) in approximating/matching \(\mu_q\) is measurable by computing the effect of the net charge present in any chosen volume around the target \(\mu_q\) on a test charge \(\x\). Consider a hypercube \(\mcalC_{q,r}\) of side length \(r\), centered around \(\mu_q\) with test charges \(\{\x_{\ell}\}_{\ell = 1}^{M_{\x}}\), \(\x_{\ell}\in\mcalC_{q,r}\). To analyze the average behavior of target and generated samples in \(\mcalC_{q,r}\), we draw \(\x_{\ell}\) uniformly within \(\mcalC_{q,r}\). We consider \(N_p = N_q = N\) for simplicity. We now define the {\it signed distance} of \(\mu_p\) from \(\mu_q\) as the negative of \(f(\x)\), summed over a uniform sampling of points over \(\mcalC_{q,r}\), {\it i.e.} \(\mathrm{SD}_{m,r}(\mu_p\|\mu_q) \) is given by:
\begin{align}
\frac{1}{NM_{\x}}\sum_{\substack{\ell=1\\\tilde{\x}_{\ell}\in\mcalC_{q,r}}}^{M_{\x}} \!\!\!\Bigg(   \sum_{\substack{j=1\\\bmc_j\sim\mu_q}}^{N} \Phi(\x_{\ell},\bmc_j) - \sum_{\substack{i=1\\\tilde{\bmc}_i\sim\mu_p}}^{N} \Phi(\x_{\ell},\tilde{\bmc}_i)\Bigg).
\label{Eqn_SD}
\end{align}
 Similar to the improved precision and recall (IPR) metrics, \(\mathrm{SD}_{m,r}(\mu_p\|\mu_q)\) is asymmetrical, {\it i.e.,}  \(\mathrm{SD}_{m,r}(\mu_p\|\mu_q) \neq \mathrm{SD}_{m,r}(\mu_q\|\mu_p)\). When \(\mathrm{SD}_{m,r}(\mu_p\|\mu_q)<0\), on the average, samples from \(\mu_q\) are relative more spread out than those drawn from \(\mu_p\) with respect to \(\mcalC_{q,r}\), and vice versa. When \(\mu_p = \mu_q\), we have \(\mathrm{SD}_{m,r}(\mu_p\|\mu_q)\approx0\). Illustrations of these three scenarios are provided in  Appendix~\ref{App_SIDGaussians}. \par

\begin{figure*}[!bth]
\begin{center}
  \begin{tabular}[b]{P{.22\linewidth}P{.22\linewidth}P{.22\linewidth}P{.13\linewidth}}
       (a) MNIST  & (b) CIFAR-10   & (c) Tiny-ImageNet \\[-1pt]
    \includegraphics[width=0.99\linewidth]{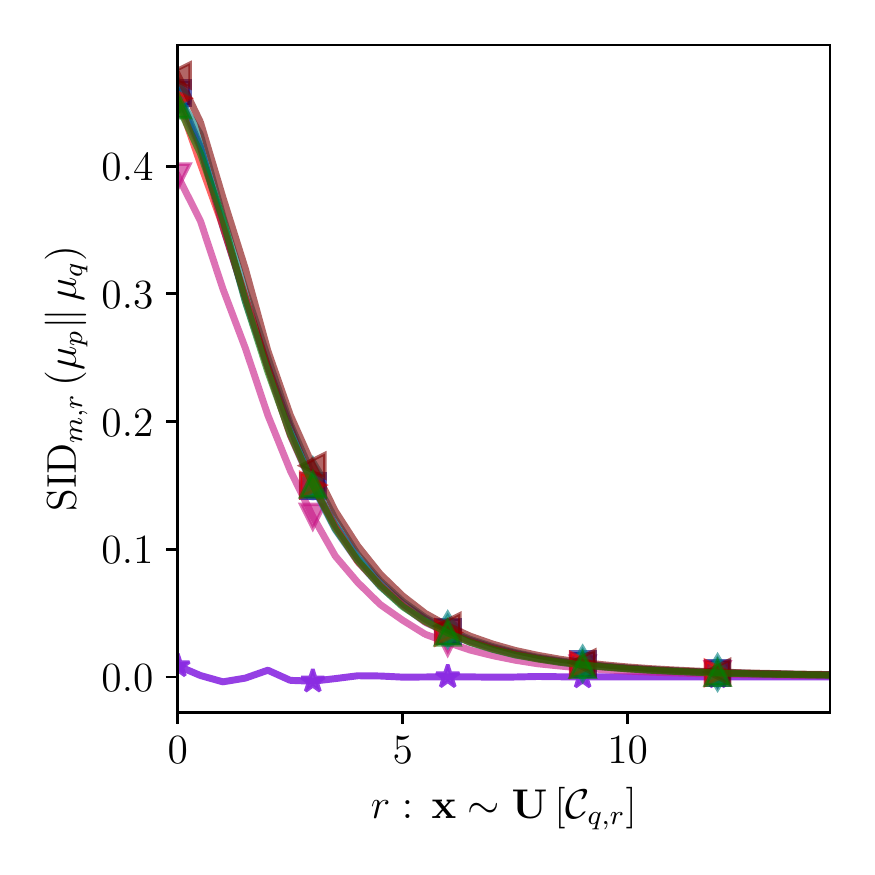} & 
    \includegraphics[width=0.99\linewidth]{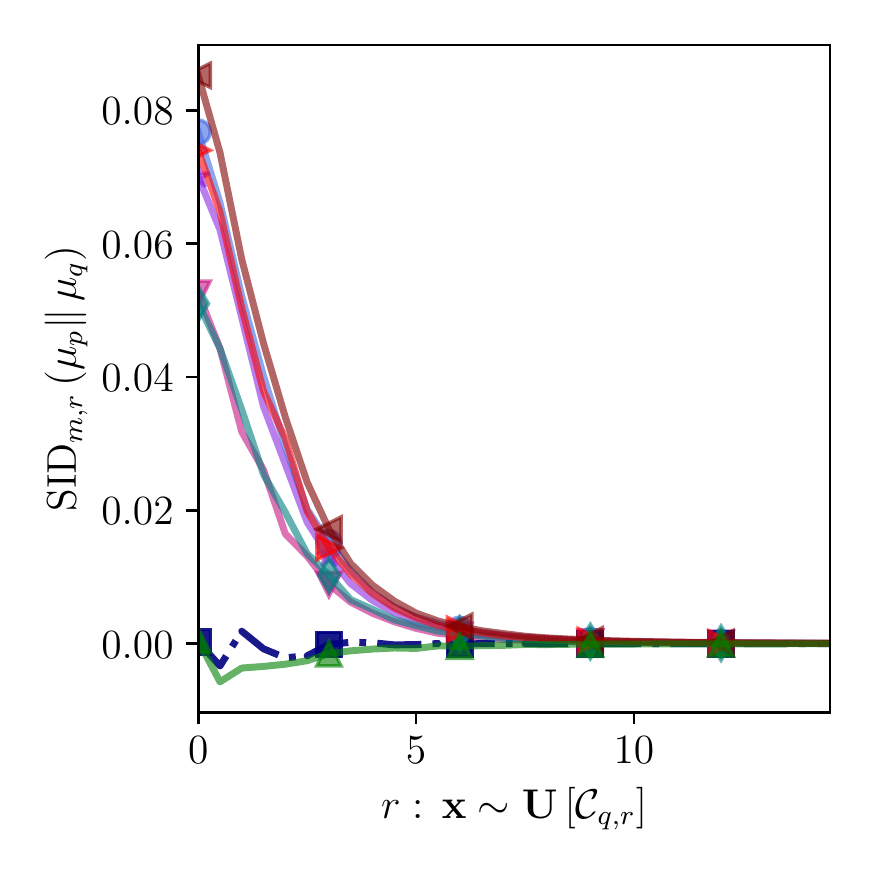} & 
    \includegraphics[width=0.99\linewidth]{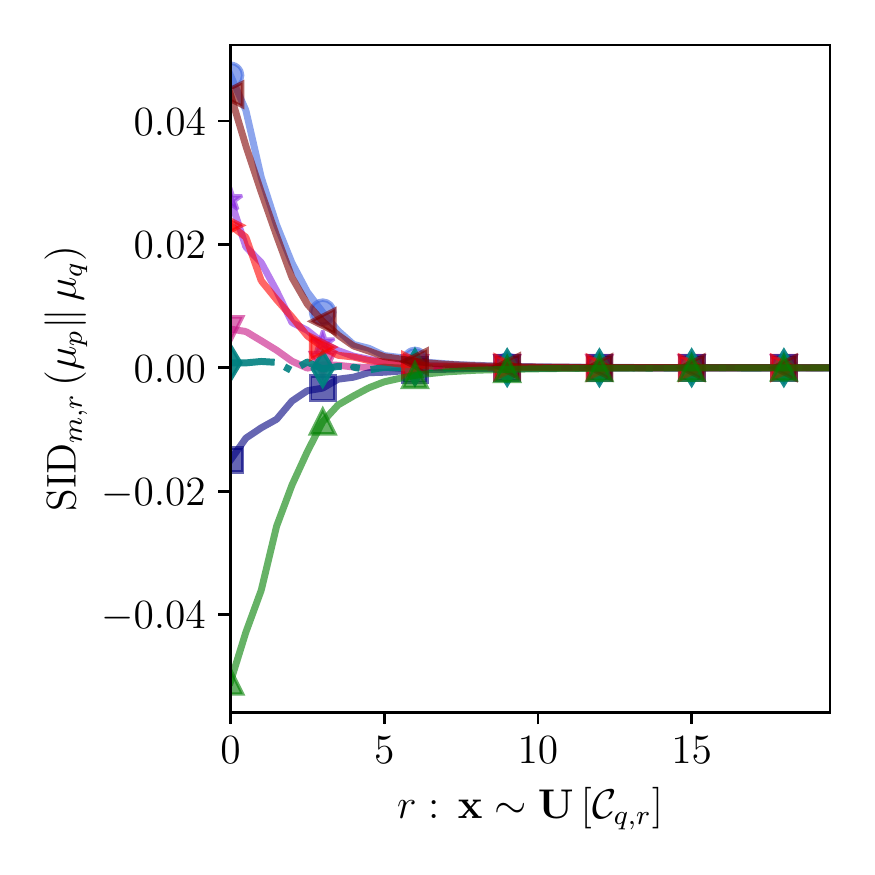}  &
    \includegraphics[width=0.99\linewidth]{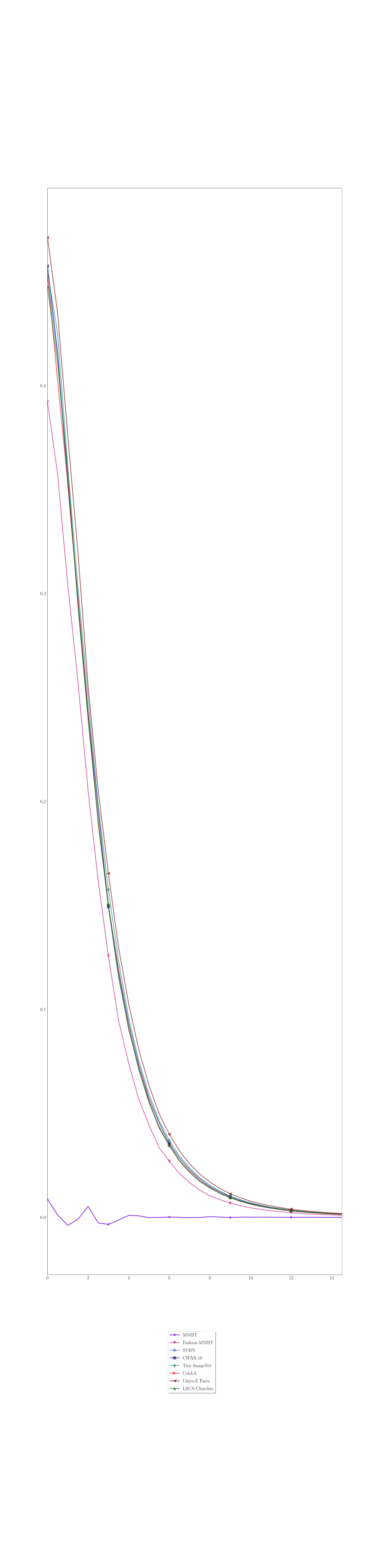}  \\[-11pt]
  \end{tabular} 
\caption[]{(\includegraphics[height=0.009\textheight]{Rgb.png} Color online) \(\mathrm{SID}_{m,r}\) as a function of the hyper-cube length \(r\). We observe that Fashion-MNIST is the closest to MNIST, while Tiny-ImageNet and SVHN are closest to CIFAR-10. Fashion-MNIST and CelebA are friendly neighbors of Tiny-ImageNet.}
\label{Fig_SIDImages}  
\end{center}
\vskip-1.9em
\end{figure*}

In practice, similar to the standard GAN metrics, the computation of \(\mathrm{SD}\) can be made practical and efficient on higher-resolution images by evaluating the measure on the feature-space of the images learnt by the pre-trained InceptionV3~\cite{InceptionV3} network mapping \(\psi(\bmc)\). This results in the {\it signed inception distance} \(\mathrm{SID}_{m,r}(\mu_p\|\mu_q) \) given by:
\begin{align}
 \frac{1}{NM_{\x}}\!\!\!\!\sum_{\substack{\ell=1\\\x_{\ell}\in\mcalC^{\prime}_{q,r}}}^{M_{\x}}\!\!\!\!\!\Bigg(\!\!\sum_{\substack{j=1\\\bmc_j\sim\mu_q}}^{N}\!\Phi\left(\x_{\ell},\psi\!\left(\bmc_j\right)\right)\!-\!\!\! \sum_{\substack{i=1\\\tilde{\bmc}_i\sim\mu_p}}^{N} \!\!\Phi\!\left(\x_{\ell},\psi\!\left(\tilde{\bmc}_i\right)\right)\!\!\!\Bigg),
\label{Eqn_SID}
\end{align}
where \(\mcalC^{\prime}_{q,r}\) denotes the hypercube of side \(r\) centered on the transformed distribution \(\psi(\mu_q)\). To begin with, we find \(\sigma_q = \max\{\mathrm{diag}(\Sigma_q)\}\), where in turn, \(\Sigma_q\) is the covariance matrix of the samples in \(\mfrakD_q\). We define the hypercube \(\mcalC^{\prime}_{q,r}\) as having side \(r = \sigma_q\) along each dimension and centered around the mean of \(\mu_q\). To compare two datasets, we plot \(\mathrm{SID}_{m,r}(\mu_p\|\mu_q)\) as a function of \(r \in [\sigma_q, 100\,\sigma_q]\) varying \(r\) in steps of 0.5. SID comparison figures for a few representative target datasets are given in Figure~\ref{Fig_SIDImages}. We observe that, when two datasets are closely related, SID is close to zero even for small $r$. Datasets with lower diversity than the target have a negative SID, and vice versa. In order to quantify SID as a single number (akin to FID and KID) we consider SID, accumulated over all radii \(r\) (the cumulative SID or CSID, for short) given by: \(\mathrm{CSID}_{m} = \sum_r \mathrm{SID}_{m,r}\). The last four columns of Table~\ref{Table_FID_SID} presents CSID for \(m=\lfloor\frac{n}{2}\rfloor\) for the various datasets considered. We observe that CSID is highly correlated with FID when the source is more diverse than the target, while it is able to single out sources that lack diversity, which FID cannot. These results quantitatively verify the empirical {\it closeness} observed when transfer-learning across datasets~\cite{ADAStyleGAN20}. Additional experiments and ablation studies on SID are given in Appendices~\ref{App_FriendNbd} and~\ref{App_SID}. \par

\noindent {\it {\bfseries Picking the Friendliest Neighbor:}} While the various approaches to compare datasets generally suggest different {\it friendly neighbors}, we observe that the overall trend is consistent across the measures. For example, Tiny-ImageNet and CelebA are consistently friendly neighbors to multiple datasets. We show in Sections~\ref{Sec_Exp} and~\ref{Sec_Cascade} that choosing these datasets as the input indeed improves the GAN training algorithm. Both the proposed SID, and baseline FID/KID measures are relative in that they can only measure closeness between provided candidate datasets. Incorporating domain-awareness aids in the selection of appropriate input datasets between which SID can be compared. For example, all metrics identify Fashion-MNIST as a friendly neighbor when compared against color-image targets, although, as expected, the performance is sub-par in practice (cf. Section~\ref{Sec_Exp}). One would therefore discard MNIST and Fashion-MNIST when identifying friendly neighbors of color-image datasets. Although SID is superior to FID and KID in identifying less diverse source datasets, no single approach can always find the best dataset yet in all real-world scenarios. A pragmatic strategy is to compute various similarity measures between the target and visually/structurally similar datasets, and identify the closest one by voting.

\section{Experimental Validation}\label{Sec_Exp}
To demonstrate the Spider GAN philosophy, we train {\it Spider} DCGAN on MNIST, CIFAR-10, and \(256\times256\) Ukiyo-E Faces datasets using the input datasets mentioned in Section~\ref{Sec_FriendlyNb}. While encoder-decoder architectures akin to image-to-image translation GANs could also be employed, their performance does not scale with image dimensionality. Detailed ablation experiments are provided in Appendix~\ref{App_CAE}. The second aspect is the limited stochasticity of the input dataset, when its cardinality is lower than that of the target. In these scenarios, the generator would attempt to learn one-to-many mappings between images, thereby not modeling the target entirely.  For Spider DCGAN variants, the source data is resized to \(16\times16\), vectorized, and provided as input. Based on preliminary experimentation (cf. Appendix~\ref{App_NoisePerturb}), to improve the input dataset diversity, we consider a Gaussian mixture centered around the samples of the source dataset formed by adding zero-mean Gaussian noise with variance \(\sigma \approx 0.25\) to each source image. An alternative solution, based on pre-trained generators is presented in Section~\ref{Sec_Cascade}. We consider the Wasserstein GAN~\cite{WGAN17} loss with a one-sided gradient penalty~\cite{R1R218}. The training parameters are described in Appendix~\ref{App_ImpDetails}. In addition to FID and KID, we compare the GAN variants in terms of the cumulative SID (CSID\(_m\)) for \(m = \lfloor\frac{n}{2}\rfloor\) to demonstrate the viability of evaluating GANs with the proposed SID metric. 

\begin{figure*}[!t]
\begin{center}
  \begin{tabular}[b]{P{.005\linewidth}|P{.42\linewidth}|P{.42\linewidth}}
  \rotatebox{90}{~{\footnotesize Source } } & 
   \includegraphics[width=1\linewidth]{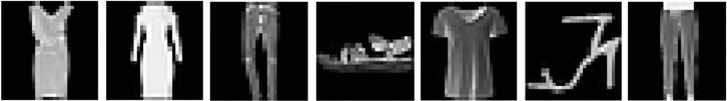} &
    \includegraphics[width=1\linewidth]{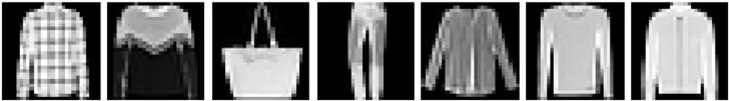}\\[-3pt]
    \rotatebox{90}{~{\footnotesize Target } } & 
   \includegraphics[width=1\linewidth]{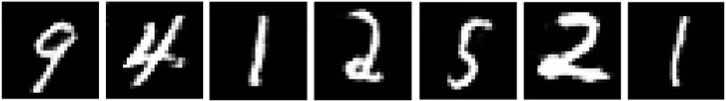} &
    \includegraphics[width=1\linewidth]{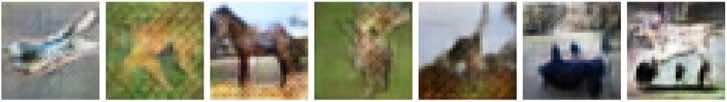}\\[-4pt]
    & (a) Fashion-MNIST to MNIST & (b) Fashion-MNIST to CIFAR-10 \\[-1pt]
     \hline \\[-9pt]
  \rotatebox{90}{~{\footnotesize Source } } & 
   \includegraphics[width=1\linewidth]{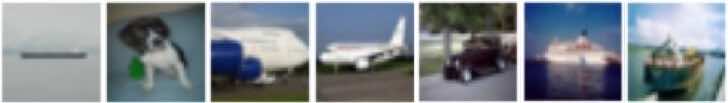} &
    \includegraphics[width=1\linewidth]{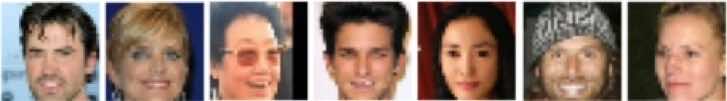}\\[-3pt]
    \rotatebox{90}{~{\footnotesize Target } } & 
   \includegraphics[width=1\linewidth]{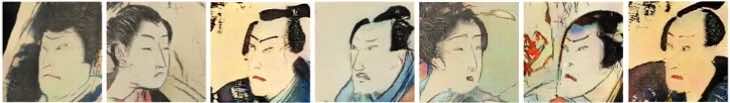} &
    \includegraphics[width=1\linewidth]{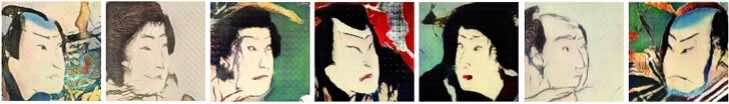} \\[-4pt]
    & (c) CIFAR-10 to Ukiyo-E Faces & (d) CelebA to Ukiyo-E Faces \\[-7pt]
  \end{tabular} 
  \caption[]{(\includegraphics[height=0.009\textheight]{Rgb.png} Color online) Figures depicting the implicit structure learnt by Spider GAN when transforming the source to the target. The network learns both visual, and implicit correspondences across datasets. For example, the {\it Trouser} class in Fashion-MNIST maps to the digit {\it 1} in MNIST, while the implicit structure is leveraged by the generator in transforming either CIFAR-10 or CelebA to Ukiyo-E Faces. A poor choice of the input distribution, for instance selecting Fashion-MNIST as the friendly neighbor of CIFAR-10, results in suboptimal learning. }
   \vspace{-1.2em}
  \label{Fig_IOPairs}  
  \end{center}
  \vskip-1pt
\end{figure*}

\begin{figure*}[t!]
\begin{center}
  \begin{tabular}[b]{P{.27\linewidth}P{.27\linewidth}P{.2558\linewidth}}
       (a) MNIST  & (b) CIFAR-10   & (c) Ukiyo-E\\[1pt]
    \includegraphics[width=1.05\linewidth]{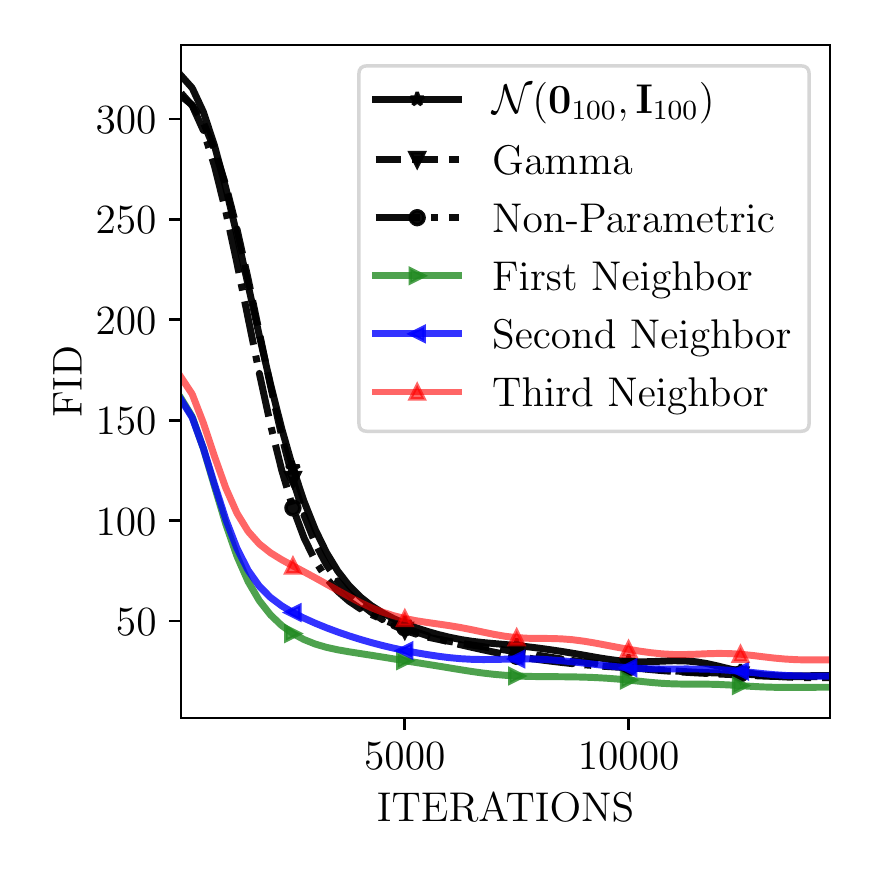} & 
    \includegraphics[width=1.04\linewidth]{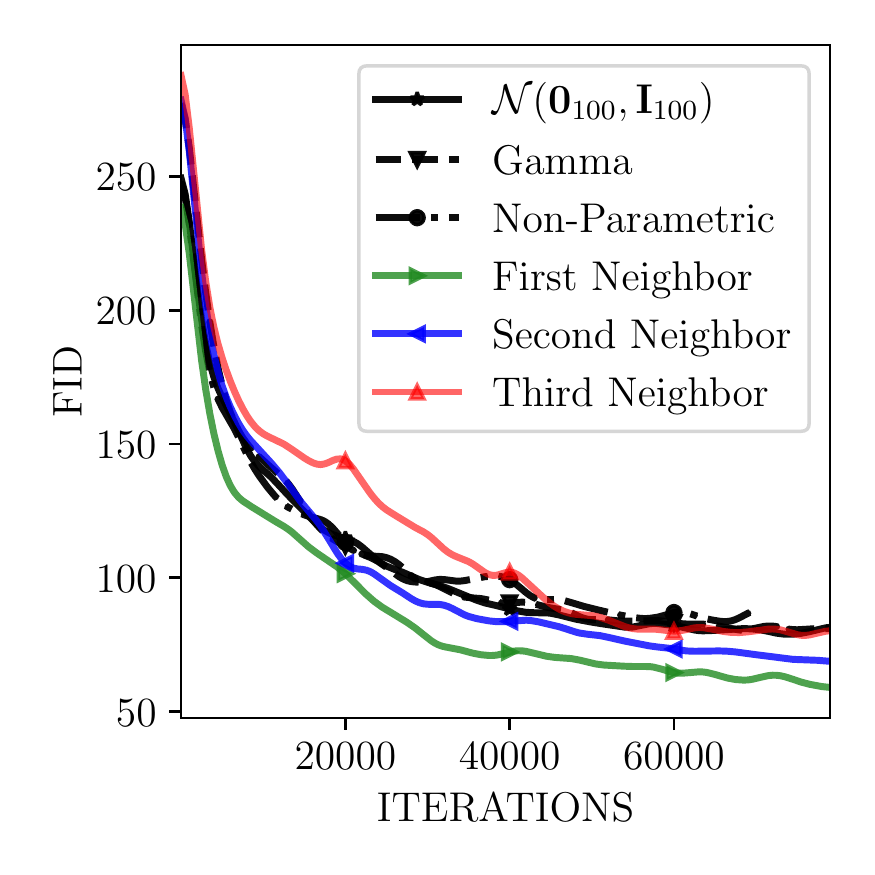} & 
    \includegraphics[width=1.09\linewidth]{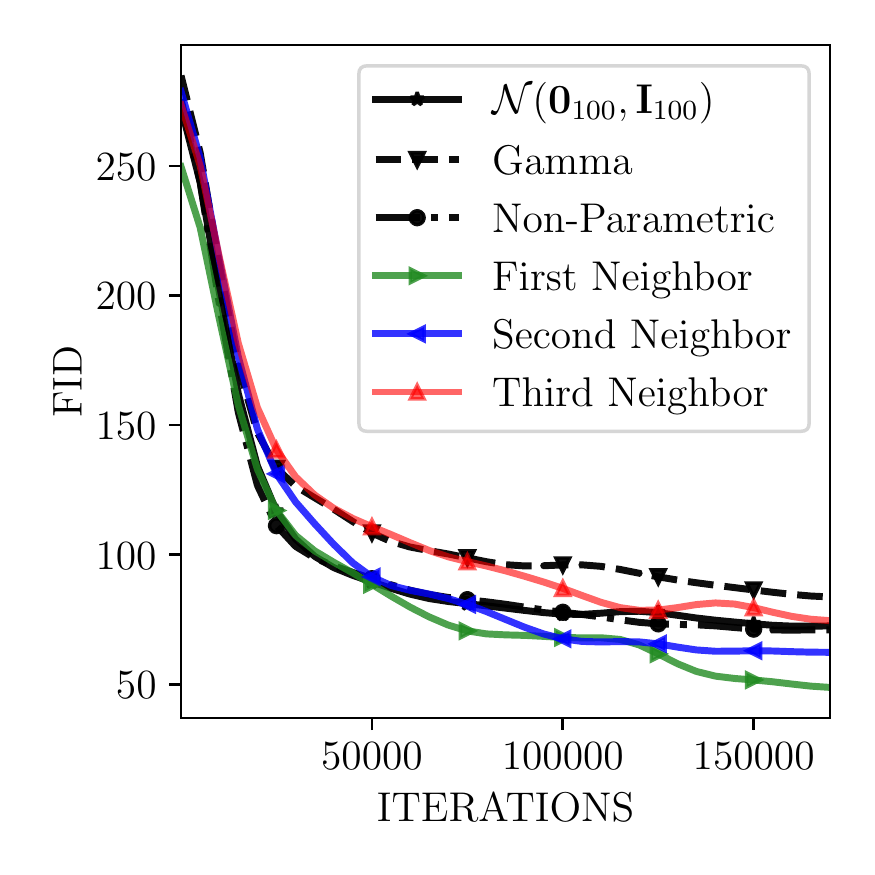}  \\[-4pt]
  \end{tabular} 
\caption[]{(\includegraphics[height=0.009\textheight]{Rgb.png} Color online) FID versus iterations for training baseline and Spider GAN with the \textcolor{ForestGreen}{\bf first}, \textcolor{blue}{\bf second} and \textcolor{red}{\bf third} friendly neighbors (color coded) identified by CSID (cf. Table~\ref{Table_FID_SID}). Using the friendliest neighbor results in the best (lowest) FID scores.On MNIST, Spider GAN variants saturate to a lower FID in an order of iterations faster than the baselines.}
\label{Fig_FID_vs_Iters}  
\end{center}
\vskip-1.5em
\end{figure*}

\begin{table*}[t!]
\fontsize{8.5}{12}\selectfont
\begin{center}
\caption{Comparison of FID, KID and the proposed CSID\(_m\) (with \(m =\lfloor\frac{n}{2}\rfloor\)) for the Spider DCGAN and baseline variants on MNIST, CIFAR-10, and Ukiyo-E Faces datasets. The first (\textcolor{ForestGreen}{\(\bm{\dagger}\)}), second (\textcolor{blue}{\(\bm{\ddagger}\)}) and third (\textcolor{red}{\(\star\)})  {\it friendly neighbors} (cf. CSID; Table~\ref{Table_FID_SID}) of the target are marked for cross-referencing against the {\bfseries first}, \uline{second} and {\it third} best FID/KID/CSID\(_m\) scores. Spider DCGAN, with {\it friendly neighborhood} input datasets outperform the baseline parametric and non-parametric priors, while a bad choice for the input results in a poorer performance.} \label{Table_FID}  \vskip-0.1in
\begin{tabular}{P{0.05cm}P{3.3cm}||P{0.95cm}|P{1.11cm}|P{0.95cm}||P{0.95cm}|P{1.11cm}|P{0.95cm}||P{0.95cm}|P{1.11cm}|P{0.95cm}}
\toprule 
\multicolumn{2}{c||}{\multirow{2}{*}{Input Distribution}}&\multicolumn{3}{c||}{MNIST} & \multicolumn{3}{c||}{CIFAR10}& \multicolumn{3}{c}{Ukiyo-E Faces} \\\cline{3-11}
&& FID & KID & CSID\(_m\) & FID & KID & CSID\(_m\) & FID & KID & CSID\(_m\)   \\
\hline\hline
\multirow{4}{*}{\rotatebox{90}{{\footnotesize\uline{ \enskip Baselines \enskip }} }}
&Gaussian~\cite{DCGAN} \((\mathbb{R}^{100})\)			& 21.49 & 0.0139 & 21.31 & 71.84 & 0.0619 & 19.90  & {\it 62.26} & 0.0535 & {\it 23.10}  \\
&Gamma~\cite{Gamma18} 	 \((\mathbb{R}^{100})\)		 & 21.15 & 0.0133 & 19.44 & 72.66 & 0.0483 & 19.87  & 70.02 & 0.0495 & 30.59  \\
&Non-Parametric~\cite{NonPara19}  \((\mathbb{R}^{100})\)	& 20.94 & 0.0137 & 20.78 & 74.90 & 0.0530 & 19.45  & 65.36 & \uline{0.0421} & 25.40  \\
&Gaussian  \((\mathbb{R}^{H\times W\times C})\)	& 42.44 & 0.0354& 32.20  & 73.00 & 0.0504 & 21.99  & 70.96 & 0.0501 & 35.30  \\[1pt] \hline
\multirow{8}{*}{\rotatebox{90}{{\footnotesize\uline{\quad\quad Spider DCGAN \quad\quad\enskip} } }}
&MNIST 			   		 & -- & -- & -- & 71.70 & 0.0535 & 21.83  & 68.87 & 0.0438 & 33.13  \\
&Fashion MNIST  			& \textcolor{ForestGreen}{\(\bm{\dagger}\)} {\bfseries 16.80}  & \textcolor{ForestGreen}{\(\bm{\dagger}\)} {\bfseries 0.0103} & \textcolor{ForestGreen}{\(\bm{\dagger}\)} {\bfseries 12.44} &  77.86 & 0.0550 & 28.85  & 72.431 &0.0455 & 36.21  \\
&SVHN  			    		& 27.17 & 0.0205 & {\it 17.23} & \textcolor{red}{\(\bm{\star}\)} 64.30 & \textcolor{red}{\(\bm{\star}\)} 0.0451 & \textcolor{red}{\(\bm{\star}\)}{\it 18.44} & 70.13 & 0.0482 & 25.06  \\
&CIFAR-10 				& 29.22 & 0.0220 & 24.96 &  -- & -- & --  & 70.55 & 0.0530 & 24.12  \\
&TinyImageNet 			& 32.66 & 0.0244 & 36.90 & \textcolor{ForestGreen}{\(\bm{\dagger}\)} {\bfseries 58.82} & \textcolor{ForestGreen}{\(\bm{\dagger}\)} {\bfseries 0.0305} &  \textcolor{ForestGreen}{\(\bm{\dagger}\)} {\bfseries 14.02}  &  \textcolor{blue}{\(\bm{\ddagger}\)} \uline{61.91} &  \textcolor{blue}{\(\bm{\ddagger}\)} {\it 0.0463} & \textcolor{blue}{\(\bm{\ddagger}\)} \uline{21.07}  \\
&CelebA 			    		& \textcolor{blue}{\(\bm{\ddagger}\)} {\it 20.55} & \textcolor{blue}{\(\bm{\ddagger}\)} {\it 0.0144} & \textcolor{blue}{\(\bm{\ddagger}\)} \uline{15.74} & \textcolor{blue}{\(\bm{\ddagger}\)}  \uline{60.09} &  \textcolor{blue}{\(\bm{\ddagger}\)} {\it 0.0434} & \textcolor{blue}{\(\bm{\ddagger}\)} \uline{17.68}  & \textcolor{ForestGreen}{\(\bm{\dagger}\)} {\bfseries 54.09} & \textcolor{ForestGreen}{\(\bm{\dagger}\)} {\bfseries 0.0408} &  \textcolor{ForestGreen}{\(\bm{\dagger}\)} {\bf 20.12}   \\
&Ukiyo-E 			    		& \uline{18.72} & \uline{0.0122} & 19.35 & 67.80 & 0.0463 & 19.90  & -- & -- & --   \\
&LSUN-Churches 			& \textcolor{red}{\(\bm{\star}\)} 30.67 & \textcolor{red}{\(\bm{\star}\)} 0.0228 &  \textcolor{red}{\(\bm{\star}\)} 30.61 & {\it 61.46} & \uline{0.0365} & 19.82  &  \textcolor{red}{\(\bm{\star}\)} 66.26 &  \textcolor{red}{\(\bm{\star}\)} 0.0496 & \textcolor{red}{\(\bm{\star}\)} 25.21   \\
\bottomrule
\end{tabular}
\end{center}
\vskip-2.5em
\end{table*}

\noindent{\it {\bfseries Results:}} We demonstrate the ability of Spider GAN to leverage the structure present in the source dataset. From the input-output pairs given in Figure~\ref{Fig_IOPairs}, we observe that, although trained in an unconstrained manner, the generator learns structurally motivated mappings. In the case when learning MNIST images with Fashion-MNIST as input, the generator has learnt to cluster similar classes, such as {\it Trousers} and the {\it 1} class, or the {\it Shoes} class and digit {\it 2}, which serendipitously are also visually similar. Even in scenarios where such pairwise similarity is not present, as in the case of generating Ukiyo-E Faces from CelebA or CIFAR-10, Spider GAN leverages implicit/latent structure to accelerate the generator convergence. Figure~\ref{Fig_FID_vs_Iters} presents FID as a function of iterations for each learning task for a few select target datasets. Spider GAN variants with {\it friendly neighborhood} inputs outperform the baseline models with parametric noise inputs, while also converging faster (up to an order in the case of MNIST). Table~\ref{Table_FID} presents the FID of the best-case models. In choosing a {\it friendly neighbor}, a poorly related dataset results in worse performance than the baselines, while a closely related input results in FID improvements of about 30\%. The poor performance of Fashion MNIST as a friendly neighbor to CIFAR-10 and Ukiyo-E faces datasets corroborate the observations made in Section~\ref{Sec_FriendlyNb}. We observe that CSID\(_m\) is generally in agreement with the performance indicated by FID/KID, making it a viable alternative in evaluating GANs.  Experiments on remaining source-target combinations are provided in Appendix~\ref{App_SpiderDCGAN}. 

\noindent  {\it {\bfseries  Extension to Class-conditional Learning}}: As a proof of concept, we developed the {\it Spider} counterpart to the auxiliary classifier GAN (ACGAN)~\cite{ACGAN17}, entitled Spider ACGAN. Here, the discriminator predicts the class label of the input in addition to the {\it real} versus {\it fake} classification. We consider two variants of the generator, one without class information, and the other with the class label provided as a fully-connected embedding to the input layer. While Spider ACGAN without generator embeddings is superior to the baseline Spider GAN in learning class-level consistency, mixing between the classes is not eliminated entirely. However, with the inclusion of class embeddings in the generator, the disentanglement of classes can be achieved in Spider ACGAN. Additional details are provided in Appendix~\ref{App_CCSpiderGAN}. Extensions of Spider GAN to larger class-conditional GAN models such as BigGAN~\cite{BIGGAN18}, and scenarios involving mismatch between the number of classes in the input and output datasets, are promising directions for future research.

\section{Cascading Spider GANs} \label{Sec_Cascade}
The DCGAN architecture employed in Section~\ref{Sec_Exp} does not scale well for generating high-resolution images. While training with image datasets has proven to improve the generated image quality, the improvement is accompanied by an additional memory requirement. While inference with standard GANs requires inputs drawn purely from random number generators, Spider DCGAN would require storing an additional dataset as input. To overcome this limitation, we propose a novel cascading approach, where the output distribution of a publicly available pre-trained generator is used as the input distribution to subsequent Spider GAN stages. The benefits are four-fold: First, the memory requirement is significantly lower (by an order or two), as only the weights of an input-stage generator network are required to be stored. Second, the issue of limited stochasticity in the input distribution is overcome, as infinitely many unique input samples can be drawn. Third, the network can be cascaded across architectures and styles, {\it i.e.,} one could employ a BigGAN input stage (trained on CIFAR-10, for example) to train a Spider StyleGAN network on ImageNet, or vice versa. {\it Essentially, no pre-trained GAN gets left behind}. Lastly, the cascaded Spider GANs can be coupled with existing transfer learning approaches to further improve the generator performance on small datasets~\cite{ADAStyleGAN20}.

\begin{table*}[!bht]
\begin{minipage}{.55\linewidth}
\fontsize{8}{12}\selectfont
\begin{center}
\caption{A comparison of the FID and KID values achieved by the PGGAN and StyleGAN2 baselines and their {\it Spider} variants, when trained on small datasets. A \(\star\) indicates scores computed on publicly available pre-trained models using the Clean-FID library~\cite{CleanFID21}. Spider StyleGAN2 achieves state-of-the-art FID and KID scores, while Spider PGGAN achieves performance comparable with the baseline StyleGAN methods. } \label{Table_SpiderPGGAN}  \vskip-0.05in
\begin{tabular}{P{2.95cm}|P{1.55cm}||P{0.5cm}|P{0.6cm}||P{0.5cm}|P{0.6cm}}
\toprule 
 \multirow{2}{*}{Architecture} &  \multirow{2}{*}{Input}   & \multicolumn{2}{c||}{Ukiyo-E Faces} & \multicolumn{2}{c}{MetFaces}  \\\cline{3-6} 
&& FID & KID & FID & KID \\
\hline\hline && \\[-12pt]
PGGAN~\cite{PGGAN18}					&  Gaussian & 69.03 & 0.0762 & 85.74& 0.0123  \\
Spider PGGAN {\bf (Ours)}		&  TinyImageNet  & 57.63 & 0.0161  & 45.32 & 0.0063 \\ \midrule
StyleGAN2\(^{\star}\) \cite{StyleGAN220} 	&  Gaussian & 56.74 & 0.0159 & 65.74& 0.0350 \\
StyleGAN2-ADA\(^{\star}\) \cite{ADAStyleGAN20} 	&  Gaussian& \uline{26.74} & \uline{0.0109} & \uline{18.75} & {\bf 0.0023} \\
Spider StyleGAN2 {\bf (Ours)}	&  TinyImageNet  & {\bf 20.44} & {\bf 0.0059}  & {\bf 15.60} & \uline{0.0026}  \\
Spider StyleGAN2 {\bf (Ours)}		&  AFHQ-Dogs  & 32.59 & 0.0269 & 29.82 & 0.0019 \\
\bottomrule
\end{tabular}
\end{center}
\end{minipage}
\hspace{0.15cm}
\begin{minipage}{.42\linewidth}
\fontsize{8}{12}\selectfont
\begin{center}
\caption{A comparison of StyleGAN2-ADA and StyleGAN3 variants in terms of FID, on learning FFHQ. A \({\dagger}\) indicates a reported score. Spider StyleGAN2-ADA performs on par with the state-of-the-art StyleGAN-XL (three-fold higher network complexity)~\cite{StyleGANXL22}, and outperforms variants with customized sampling techniques~\cite{PolStyleGAN22,MaGStyleGAN22}. } \label{Table_SpiderStyleGAN_FFHQ}  \vskip-0.05in
\begin{tabular}{P{3.65cm}|P{1.55cm}||P{0.6cm}}
\toprule \\[-15pt]
Architecture &  Input  & FID \\
\hline\hline && \\[-12pt]
StyleGAN-XL~\cite{StyleGANXL22} & Gaussian & {\bf 2.02}\(^{\dagger}\)  \\[-2pt]
Polarity-StyleGAN2~\cite{PolStyleGAN22} & Gaussian & 2.57\(^{\dagger}\)  \\[-2pt]
MaGNET-StyleGAN2~\cite{MaGStyleGAN22} & Gaussian & 2.66\(^{\dagger}\)  \\\midrule \\[-13pt]
StyleGAN2-ADA \cite{ADAStyleGAN20} 	&  Gaussian & 2.70\(^\dagger\)     \\[-2pt]
Spider StyleGAN2-ADA {\bf (Ours)}		&  TinyImageNet  & \uline{2.45}   \\[-2pt]
 Spider StyleGAN2-ADA {\bf (Ours)} 		&  AFHQ-Dogs  & 3.07   \\\midrule  \\[-13pt]
StyleGAN3-T \cite{StyleGAN321} 	&  Gaussian & 2.79\(^{\dagger}\)  \\[-2pt]
Spider StyleGAN3-T {\bf (Ours)} 	&  TinyImageNet & 2.86 \\[-2pt]
\bottomrule
\end{tabular}
\end{center}
\end{minipage}
\vskip-0.5em
\end{table*}
\begin{table*}[!bht]
\fontsize{9}{12}\selectfont
\begin{center}
\caption{A comparison of the FID and KID values achieved by the StyleGAN baselines and their {\it Spider} variants, when trained on the the AFHQ-Cats dataset, considering various training configurations. A \(\star\) indicates a score reported in the Clean-FID library~\cite{CleanFID21}. \(\dagger\) Karras {\it et al.} only report FID on the combined AFHQv2 dataset consisting of images from the {\it Dogs, Cats}, and {\it Wild-Animals} classes.  Spider StyleGAN2-ADA and Spider StyleGAN3 achieve FID and KID scores competitive with the baselines in a mere one-fifth of the training iterations, while Spider StyleGAN3 with weight transfer achieves state-of-the-art FID on AFHQ in one-fourth of the training iterations. } \label{Table_SpiderStyleGAN}  \vskip-0.05in
\begin{tabular}{P{4.6cm}|P{2.3cm}||P{2.8cm}|P{2cm}||P{0.6cm}|P{2cm}}
\toprule 
Architecture & Weight Transfer &  Input Distribution  & Training steps & FID & KID \((\times 10^{-3})\) \\[-2pt]
\hline\hline &&&&& \\[-12pt]
StyleGAN2-ADA~\cite{ADAStyleGAN20} 	& --  & Gaussian & 25000 & 5.13\(^\star\) & 1.54\(^\star\)  \\[-2pt]
StyleGAN3-T~\cite{StyleGAN321}		& --  & Gaussian & 25000 & 4.04\(^\dagger\) & --  \\[-2pt]
Spider StyleGAN3-T  {\bf (Ours)} 				& -- & AFHQ-Dogs & 5000 & 6.29 & 1.64  \\ \midrule\\[-13pt]
StyleGAN2-ADA~\cite{ADAStyleGAN20}	& FFHQ  & Gaussian & 5000 & 3.55 & \uline{0.35}  \\[-2pt]
Spider StyleGAN2-ADA {\bf (Ours)}	& FFHQ 	& Tiny-ImageNet & 1000 & 3.91 & 1.23  \\ \midrule\\[-13pt]
StyleGAN2-ADA~\cite{ADAStyleGAN20}	& AFHQ-Dogs  & Gaussian & 5000 & \uline{3.47}\(^\star\) & 0.37\(^\star\)  \\[-2pt]
Spider StyleGAN2-ADA  {\bf (Ours)}			& AFHQ-Dogs  & Tiny-ImageNet & 1500 & {\bf 3.07} & {\bf 0.29}    \\[-2pt]
Spider StyleGAN3-T {\bf (Ours)} 		&  AFHQ-Dogs  & Tiny-ImageNet  & 1000 & 3.86 & 1.01\\[-2pt]
\bottomrule 
\end{tabular}
\end{center}
\vskip-14pt
\end{table*}

\subsection{Spider Variants of PGGAN and StyleGAN} \label{Sec_SpiderPGStyleGAN}

We consider training the {\it Spider} variants of StyleGAN2~\cite{StyleGAN220} and progressively growing GAN (PGGAN)~\cite{PGGAN18} on small datasets, specifically the 1024-MetFaces and 1024-Ukiyo-E Faces datasets, and high-resolution FFHQ. We consider input from pre-trained GAN generators trained on the following two distributions (a) Tiny-ImageNet, based on CSID\(_m\), that suggest that it is a {\it friendly neighbor} to the targets; and (b) AFHQ-Dogs, which possesses structural similarity to the face datasets. The experimental setup is provided in Appendix~\ref{App_ExpPGGAN}, while evaluation metrics are described in Appendix~\ref{App_Metrics}. To maintain consistency with the reported scores for state-of-the-art baselines models, we report only FID/KID here, and defer comparisons on CSID\(_m\) to Appendix~\ref{App_ExpStyleGAN}. To isolate and assess the performance improvements introduced by the Spider GAN framework, we do not incorporate any augmentation or weight transfer~\cite{ADAStyleGAN20}.  Table~\ref{Table_SpiderPGGAN} shows the FID values obtained by the baselines and their {\it Spider} variants. Spider PGGAN performs on par with the baseline StyleGAN2 in terms of FID. Spider StyleGAN2 achieves state-of-the-art FID on both Ukiyo-E and MetFaces. \par
To incorporate transfer learning techniques, we consider (a) learning FFHQ considering StyleGAN with adaptive discriminator augmentation (ADA)~\cite{ADAStyleGAN20}; and (b) learning AFHQ-Cats considering both ADA and weight transfer~\cite{ADAStyleGAN20}. Spider StyleGAN2-ADA achieves FID scores on par with the state of the art, outperforming improved sampling techniques such as Polarity-StyleGAN2~\cite{PolStyleGAN22} and MaGNET-StyleGAN2~\cite{MaGStyleGAN22}. While StyleGAN-XL achieves marginally superior FID, it does so at the cost of a three-fold increase in network complexity~\cite{StyleGANXL22}. The FID and KID scores, and training configurations are described in Tables~\ref{Table_SpiderStyleGAN_FFHQ}-\ref{Table_SpiderStyleGAN}. Spider StyleGAN2-ADA and Spider StyleGAN3 achieve competitive FID scores with a mere one-fifth of the training iterations. The Spider StyleGAN3 model with weight transfer achieves a state-of-the-art FID of 3.07 on AFHQ-Cats, in a fourth of the training iterations as StyleGAN3 with weight transfer. Additional results are provided in Appendix~\ref{App_ExpStyleGAN}. 

\subsection{Understanding the Spider GAN Generator} \label{Sec_UnderstandSpiderGAN}

The idea of learning an optimal transformation between a pair of distributions has been explored in the context of optimal transport in {\it Schr{\"o}dinger bridge} diffusion models~\cite{DDPM20,ShroBridgeML21,DiffShroBridge21,FBShroBridge22}. The {\it closer} the two distributions are, the easier it is to learn a transport map between them. Spider GANs leverage underlying similarity, not necessarily visual, between datasets to improve generator learning. Similar discrepancies between visual features and those learnt by networks have been observed in ImageNet~\cite{ImageNet09} object classification~\cite{Harmonize22}. To shed more light on this intuition, consider a scenario where both the input and target datasets in Spider DCGAN are the same, with or without random noise perturbation. As expected, the generator learns an identity mapping, reproducing the input image at the output (cf. Appendix~\ref{App_Identity}). \par
\noindent   {\it {\bfseries Input Dataset Bias}}: Owing to the unpaired nature of training, Spider GANs do not enforce image-level structure to learn pairwise transformations. Therefore, the diversity of the source dataset (such as racial or gender diversity) does not affect the diversity in the learnt distribution.  Experiments on Spider DCGAN with varying levels of class-imbalance in the input dataset validate this claim (cf. Appendix~\ref{App_Bias}). \par
\noindent  {\it {\bfseries  Input-space Interpolation}}: Lastly, to understand the representations learnt by Spider GANs, we consider input-space interpolation. Unlike classical GANs, where the input noise vectors are the only source of control, in cascaded Spider GANs, interpolation can be carried out at two levels. Interpolating linearly between the noise inputs to the pre-trained GAN result in a set of interpolations of the intermediate image. Transforming these images through the Spider StyleGAN generator results in greater diversity in the output images, with sharper transitions between images. This is expected as interpolating on the Gaussian manifold is known to result in discontinuities in the generated images~\cite{Gamma18,Cauchy19}. Alternatively, for fine-grained tuning, linear interpolations of the intermediate input images can be carried out, resulting in smoother transitions in the output images. Images demonstrating this behavior are provided in Appendix~\ref{App_InterpolStyleGAN}. Qualitative experiments on input-space interpolation in Spider DCGAN and additional images are provided in Appendix~\ref{App_Interpol}. These results indicate that stacking Spider GAN stages yields varying levels of fineness in controlling features.

\section{Conclusions} \label{Sec_Conclusion}
We introduced the Spider GAN formulation, where we provide the GAN generator with an input dataset of samples from a closely related neighborhood of the target. Unlike image-translation GANs, there are no pairwise or cycle-consistency requirements in Spider GAN, and the trained generator learns a transformation from the underlying latent data distribution to the target data. While the {\it best} input dataset is a problem-specific design choice, we proposed approaches to identify promising friendly neighbors. We proposed a novel signed inception distance, which measures the relative diversity between two datasets. Experimental validation showed that Spider GANs, trained with closely related datasets, outperform baseline GANs with parametric input distributions, achieving state-of-the-art FID on Ukiyo-E Faces, MetFaces, FFHQ and AFHQ-Cats.\par 
While we focused on adaptive augmentation and weight transfer, incorporating other transfer learning approaches~\cite{MineGAN20,FewShotDA21,ElasticFewShot20} is a promising direction for future research. One could also explore extensions to vector quantized GANs~\cite{VQGAN21,ImpVQGAN22} or high-resolution class-conditional GANs~\cite{BIGGAN18,RebootACGAN21}. 

\newpage

{\small
\bibliographystyle{ieeetr}
\bibliography{CVPR2023_SpiderGAN_arXiv}
}

\onecolumn

\addcontentsline{toc}{section}{Appendix} 
\appendix
\part{Appendix} 
\vskip-10pt
\parttoc 
\section*{Overview of the Supplementary Material}\label{App_Intro}

The Supplementary Material comprises these appendices, the source codes of this project, consisting of the implementations of various Spider GAN variants and SID metric, and animations corresponding to (a) Evaluating the signed distance on Gaussian data; and (ii) Interpolation in the input, and intermediate stages of Spider StyleGAN.  The appendices contain additional discussions on identifying the friendly neighborhood in Spider GANs, ablation studies on SID, implementation details, and additional experiments on the Spider GAN variants considered in the {\it Main Manuscript}. 

\newpage 

\section{Baselines for Identifying the Friendly Neighborhood} \label{App_FriendNbd}
 
Approaches that compute the intrinsic dimensionality \(n_{\mfrakD}\) of a dataset are either computationally intensive~\cite{IntrinsicDim16} or do not scale with sample size~\cite{MNIST11D05,EstimateID17}. Campadelli {\it et al.}~\cite{IntrinsicDimEst15} presented a survey of various nearest-neighbor and maximum-likelihood estimators of \(n_{\mfrakD}\) for low-dimensional datasets. A well-known approach for computing \(n_{\mfrakD}\) is provided in the Davis-Kahan \(\sin\Theta\) theorem~\cite{SinTheta70}, which provides an upper bound on the distance between two subspaces in terms of the eigen-gap between them. A practically implementable version~\cite{DavisKahanPrac15} is based on the sample covariance matrix of the two datasets and their eigenvalues.  Along a parallel vertical, multiple works have derived convergence guarantees on the GAN training algorithms, given \(n_{\mfrakD}\) \cite{HowWellGANs21,DimEffectGANs21,IntDimGAN21}. We now discuss the Davis-Kahan \(\sin\Theta\) theorem, and compare its performance against the FID, KID and CSID\(_m\) approaches in terms of the friendliest neighbors picked by them.
 
 \subsection{The Davis-Kahan Theorem}
 
 
The Davis-Kahan \(\sin\Theta\) Theorem~\cite{SinTheta70} upper-bounds the distance between subspaces in terms of the eigen-gap between them.  Let \(\Sigma_p, \Sigma_q \in \mathbb{R}^{n\times n}\) denote the sample covariance matrices of two datasets \(\mfrakD_p\) and \(\mfrakD_q\), respectively, with \(\lambda_{1_p}\geq\lambda_{2_p}\geq\cdots\geq\lambda_{n_p}\) and \(\lambda_{1_q}\geq\lambda_{2_q}\geq\cdots\geq\lambda_{n_q}\) denoting their respective eigenvalues in order.  
Consider \(1\leq r \leq s \leq n\), and define \(d:=s-r+1\), \(\lambda_0 := \infty\) and \(\lambda_{n+1} := -\infty\). Consider the subspaces \(\mathcal{V}_p = \mathrm{span}\left\{v_r^p, v_{r+1}^p,\ldots,v_s^p\right\}\) and \(\mathcal{V}_q = \mathrm{span} \left\{ v_r^q, v_{r+1}^q,\ldots,v_s^q\right\}\) that are spanned by the eigenvectors of \(\Sigma_p\) and \(\Sigma_q\), respectively. The Davis-Kahan \(\sin\Theta\) theorem bounds the distance between the two subspaces \(\mathcal{V}_p\) and \(\mathcal{V}_q\) as follows:
\begin{align}
\| \sin\Theta(\mathcal{V}_p,\mathcal{V}_q)\|_{\textsc{F}} &\leq \frac{\|\Sigma_p - \Sigma_q\|_{\textsc{F}}}{\delta}, \label{Eqn_sin_org} \\
\text{where}~~\delta &= \inf\left\{ | \hat{\lambda} - \lambda|: \lambda \in \left[\lambda^q_s, \lambda^q_r \right],  \hat{\lambda} \in \left(-\infty, \hat{\lambda}^p_{s-1}\right]\,\bigcup\,\left[\hat{\lambda}^p_{r+1},\infty\right)\right\}. \nonumber
\end{align}
 As noted by Yu {\it et al.}~\cite{DavisKahanPrac15}, evaluating the infimum among all pairs of eigenvalues requires a huge computational overhead, particularly on high-dimensional data. They derived a loose, but computationally efficient upper bound:
\begin{align}
\left\| \sin\Theta(\mathcal{V}_p,\mathcal{V}_q) \right)\|_{\textsc{F}} \leq \frac{2 \min\left\{d^{\frac{1}{2}} \| \Sigma_p - \Sigma_q\|_{op}, \| \Sigma_p - \Sigma_q\|_{\textsc{F}} \right\}}{\min\left\{ \lambda^q_{r-1} - \lambda^q_r, \lambda^q_s - \lambda^q_{s+1}\right\}},
\label{Eqn_sin_new}
\end{align}
where \(\|\cdot\|_{op}\) and \(\|\cdot\|_{\textsc{F}}\) denote the operator and Frobenius norms, respectively. For large \(n\), the operator norm can be approximated by the \(\ell_{\infty}\) norm of the difference between the eigenvalues of \(\Sigma_p\) and \(\Sigma_q\)~\cite{DavisKahanPrac15}.
The form of the \(\sin\Theta\) distance in Equation~\eqref{Eqn_sin_new} replaces the infimum amongst all pairs with the minimum between only two pairs of eigenvalues, which requires less computation. \par

We now discuss a variant of the \(\sin\Theta\) distance between the subspaces spanned by two datasets. Since the intrinsic dimensionality of the data is not known priori, we compute the \(\sin\Theta\) distance for various choices of \(r\) and \(s\), and pick the {\it best} amongst them, which we call the \(\min\sin\Theta\) distance.

\noindent {\it {\bfseries The \(\min\sin\Theta\) Distance:}} Consider the space spanned by the (vectorized) images in the datasets. Since the pixel resolution of the images across datasets is not the same, it is appropriate to first rescale them to the same dimension, for instance, using bilinear interpolation. Depending on whether the rescaled image dimension is greater or smaller than the image dimension, there is a trade-off between the image quality (superior at higher resolution) and computational efficiency (superior at lower resolution). We found out experimentally that resizing all images to \(32\times 32\times 3\) is a viable compromise. We consider \(r = 1\) and compute the \(\sin\Theta(\mathcal{V}_p,\mathcal{V}_q;s)\), for \(s = 3,4,\ldots\lceil n/10\rceil\), where \( n = 3072 = 32\times32\times3\). The friendly neighborhood as indicated by the \(\min\sin\Theta\) distance is  \(\min_{s}\{\sin\Theta(\mathcal{V}_p,\mathcal{V}_q;s)\}\). In other words, the closest source dataset given all \(s\) is deemed the friendliest neighbor of the target.  \par
 
 \subsection{Comparison of Approaches for Identifying the Friendly Neighborhood}
 
 We compare the \(\min\sin\Theta\), FID, KID and CSID\(_m\) distances in terms of the friendliest neighbor predicted by these methods. FID, KID and CSID\(_m\) distances have been defined in Section~\ref{Sec_FriendlyNb}. Table~\ref{DatasetSinID} shows the \(\min\sin\Theta\) distance for the various datasets considered in Section~\ref{Sec_FriendlyNb}. We also present KID between the various datasets in Table~\ref{DatasetKID} of this document. Tables~\ref{DatasetFID} and~\ref{DatasetSSID} present the remaining combinations between datasets left out from the {\it Main Manuscript}. The \textcolor{ForestGreen}{\bf first}, \textcolor{blue}{\bf second} and \textcolor{red}{\bf third} {\it friendly neighbors} are color-coded for quick and easy identification. We observe across all datasets that, FID and KID are highly correlated in terms of the friendly neighbors they identify for a given target. CSID\(_m\) is also in agreement with the observations when the target is more diverse, but in scenarios such as TinyImageNet or CIFAR-10, it is able to indicate the less diverse sources as a poor input choice. The experiments on learning Tiny-ImageNet within the Spider GAN framework in Appendix~\ref{App_SpiderDCGAN} are more in agreement with the friendly neighbors identified by CSID\(_m\). \par
 
Across all distances, we observe that the results obtained on MNIST or Fashion-MNIST as the source do not correlate well with the experimental results (cf. Appendix~\ref{App_SpiderDCGAN}). This is attributed to the limitation of the Inception-Net embedding in handling grayscale images. Inception-Net operates on color images and offers limited performance on grayscale images. 
 
Table~\ref{DatasetSinID} shows that the \(\min\sin\Theta\) distance is unable to identify the friendliest neighbor accurately and consistently. For instance, the ordering of the top three neighbors on MNIST, CelebA or LSUN-Churches identified by using the \(\min\sin\Theta\) distance is not consistent with the ordering suggested by CSID\(_m\) and that verified experimentally. However, on the other datasets, \(\min\sin\Theta\) is worse than the InceptionNet approaches for identifying the friendliest neighborhood.

 \begin{table*}[!b]
 \fontsize{7.5}{12}\selectfont
 \begin{center}
 \caption{The best-case \(\min\sin\Theta(\cdot)\) distance between the spaces spanned by the eigenvectors of the source and target datasets. The rows represent the sources and the columns correspond to the target datasets. The \textcolor{ForestGreen}{\bf first}, \textcolor{blue}{\bf second} and \textcolor{red}{\bf third} {\it friendly neighbors} (color coded) of the target is the source with the three lowest \(\min\sin\Theta(\cdot)\) values is that column.  We observe that the {\it friendliest neighbor} identified by the \(\min\sin\Theta\) distance are generally not in agreement with those identified by FID, KID or CSID\(_m\).} \label{DatasetSinID}  \vskip0.1in
 \begin{tabular}{P{1.15cm}||P{1.cm}|P{1.2cm}|P{1.cm}|P{1.2cm}|P{1.2cm}|P{1.cm}|P{1.cm}|P{1.cm}}
 \toprule 
  \backslashbox{ Src}{ Tar}& MNIST & F-MNIST & SVHN &CIFAR-10 &T-ImgNet & CelebA & Ukiyo-E &  Church \\[3pt]
 \hline\hline
 MNIST 			& 0 & 60.74   &  \textcolor{gray}{63.25} &  \textcolor{gray}{85.73}  & \textcolor{gray}{43.19} & \textcolor{gray}{27.43} & \textcolor{gray}{23.79} & \textcolor{gray}{35.35} \\
 F-MNIST   		& 96.68  & 0 & \textcolor{gray}{69.01} & \textcolor{gray}{110.7} & \textcolor{gray}{53.77} & \textcolor{gray}{36.69} & \textcolor{gray}{45.02}  & \textcolor{gray}{48.29}  \\[2pt] \cline{4-9} \\[-10pt]
 SVHN  			& 79.91 & \textcolor{ForestGreen}{\bf 54.77} & 0 & 57.99 & 23.62 & 19.86 & 25.95 & 29.55 \\
 CIFAR-10 		& 72.16 & 58.56 & \textcolor{blue}{ \bf 35.97} & 0 & \textcolor{ForestGreen}{\bf 7.521} & \textcolor{blue}{ \bf 14.63} & 21.16 & \textcolor{blue}{ \bf 15.89} \\
 T-ImgNet 			& \textcolor{red}{ \bfseries 70.86} & \textcolor{blue}{ \bf 55.43} & \textcolor{ForestGreen}{\bf 30.67} & \textcolor{ForestGreen}{\bf 14.67} & 0 &\textcolor{ForestGreen}{\bf 13.97} & \textcolor{blue}{ \bf 20.05} & \textcolor{ForestGreen}{\bf 15.52}  \\
 CelebA 			& 72.13  & 60.62 & \textcolor{red}{ \bfseries 41.35} & \textcolor{red}{ \bfseries 45.74} & \textcolor{red}{ \bfseries 22.39} & 0 & \textcolor{ForestGreen}{\bf 19.16} & 23.48 \\
 Ukiyo-E 			& \textcolor{ForestGreen}{\bf54.09} & 59.30 & 43.08 & 52.75 & 25.65 & \textcolor{red}{ \bfseries 15.29}  & 0 & \textcolor{red}{ \bfseries 22.50} \\
 Church 			& \textcolor{blue}{ \bf 66.54}  & \textcolor{red}{ \bfseries 57.11} & 44.02 & \textcolor{blue}{ \bf 35.55} & \textcolor{blue}{ \bf 17.81} & 16.80 & \textcolor{red}{ \bfseries 20.19} & 0 \\
 \bottomrule
 \end{tabular}
 \end{center}
 \vskip50pt
 \end{table*}

 \begin{table*}[!b]
 \fontsize{8}{12}\selectfont
 \begin{center}
 \caption{A comparison of FID between popular training datasets. The rows correspond to the source (Src) and the columns correspond to the target (Tar). The \textcolor{ForestGreen}{\bf first}, \textcolor{blue}{\bf second} and \textcolor{red}{\bf third} {\it friendly neighbors} (color coded)  of the target are the sources with the three lowest FID values. FID fails to detect scenarios where the source possesses lower sample diverse that the target, as in the case of CIFAR-10 and LSUN-Church sources in comparison to the Tiny-ImageNet target. } 
 \label{DatasetFID}  
 \vskip0.1in
 \begin{tabular}{P{1.2cm}||P{1.cm}|P{1.2cm}|P{1.cm}|P{1.2cm}|P{1.2cm}|P{0.9cm}|P{1.cm}|P{0.9cm}}
 \toprule 
  \backslashbox{Src}{Tar}& MNIST & F-MNIST & SVHN & CIFAR-10 & T-ImgNet & CelebA & Ukiyo-E & Church \\[3pt]
 \hline\hline
 MNIST 			& 1.2491 & \textcolor{ForestGreen}{\bf 175.739} & \textcolor{gray}{234.850} & \textcolor{gray}{258.246} & \textcolor{gray}{264.250} & \textcolor{gray}{360.622} & \textcolor{gray}{398.280} & \textcolor{gray}{357.428} \\[3pt]
 F-MNIST  	& \textcolor{ForestGreen}{\bf 176.813} & 2.4936 &  \textcolor{gray}{212.619} & \textcolor{gray}{188.367} & \textcolor{gray}{197.057} & \textcolor{gray}{365.222} & \textcolor{gray}{387.049} & \textcolor{gray}{345.011} \\[2pt] \cline{4-9} \\[-10pt]
 SVHN  			& \textcolor{blue}{ \bf 236.707} & 214.262 & 3.4766 & \textcolor{blue}{ \bf 168.615} & \textcolor{blue}{ \bf 189.133} & 357.193 & 372.444 & 356.148 \\[3pt]
 CIFAR-10 			& \textcolor{red}{ \bfseries 259.045} & \textcolor{blue}{ \bf 188.710} & \textcolor{ForestGreen}{\bf 168.113} & 5.0724 & \textcolor{ForestGreen}{\bf 64.3941} & 305.528 & 303.694 & \textcolor{blue}{ \bf 256.207} \\[3pt]
 T-ImgNet 		& 264.309 & \textcolor{red}{ \bfseries 197.918} & \textcolor{blue}{ \bf 188.823} & \textcolor{ForestGreen}{\bf64.0312} & 6.4845 & \textcolor{ForestGreen}{\bf 251.198} & \textcolor{ForestGreen}{\bf 257.078} & \textcolor{ForestGreen}{\bf203.899} \\[3pt]
 CelebA 			& 360.773 & 364.586 & 357.383 & 303.490 & 250.735 & 2.5846 & \textcolor{red}{ \bfseries 301.108} & \textcolor{red}{ \bfseries 265.954} \\[3pt]
 Ukiyo-E 			& 396.791 & 387.088 & 372.557 & 300.511 & 254.102 & \textcolor{red}{ \bfseries 300.259} & 5.9137 & 267.624 \\[3pt]
 Church 			& 350.708 & 343.781 & \textcolor{red}{ \bfseries354.885} & \textcolor{red}{ \bfseries 254.991} & \textcolor{red}{ \bfseries 204.162} & \textcolor{blue}{ \bf 266.508} & \textcolor{blue}{ \bf 267.638} & 2.5085 \\[3pt]
 \bottomrule
 \end{tabular}
 \end{center}
 \vskip-5pt
 \end{table*}

 \begin{table*}[t!]
 \fontsize{7.5}{11}\selectfont
 \begin{center}
 \caption{KID between popular training datasets. The \textcolor{ForestGreen}{\bf first}, \textcolor{blue}{\bf second} and \textcolor{red}{\bf third} {\it friendly neighbors} (color coded) of the target (column) are the sources (rows) with the three lowest KID values. We observe that, akin to FID, the KID measure is also unable to compare the leave diversity between the source and target datasets, as is the case between Tiny-ImageNet and CIFAR-10. } \label{DatasetKID}  \vskip-0.05in
 \begin{tabular}{P{1.2cm}||P{1.cm}|P{1.2cm}|P{1.cm}|P{1.2cm}|P{1.2cm}|P{1.cm}|P{1.cm}|P{1.cm}}
 \toprule 
  \backslashbox{ Src}{ Tar}& MNIST & F-MNIST & SVHN &CIFAR-10 &T-ImgNet & CelebA & Ukiyo-E &  Church \\[3pt]
 \hline\hline
 MNIST 			& \(2\times10^{-6}\) & \textcolor{blue}{ \bf 0.1587} & \textcolor{gray}{0.2428} & \textcolor{gray}{0.2380} & \textcolor{gray}{0.2393} & \textcolor{gray}{0.4284} & \textcolor{gray}{0.5082} & \textcolor{gray}{0.4376} \\[3pt]
 F-MNIST  	& \textcolor{ForestGreen}{\bf0.1606} & \(1\times10^{-6}\) & \textcolor{gray}{0.1922} & \textcolor{gray}{0.1353} &  \textcolor{gray}{0.1578} & \textcolor{gray}{0.4291} & \textcolor{gray}{0.4751} & \textcolor{gray}{0.3963} \\[2pt] \cline{4-9} \\[-10pt]
 SVHN  			& 0.2458 & 0.1943 & \(2\times10^{-7}\) & \textcolor{blue}{ \bf 0.1377} & \textcolor{blue}{ \bf  \bfseries 0.1674} & 0.4059 & 0.4393 & 0.3962\\[3pt]
 CIFAR-10 			& \textcolor{red}{ \bfseries 0.2404} & \textcolor{ForestGreen}{\bf 0.1357} & \textcolor{ForestGreen}{\bf 0.1377} & \(6\times10^{-6}\) & \textcolor{ForestGreen}{\bf 0.0334} & \textcolor{red}{ \bfseries 0.3205} & \textcolor{red}{\bf 0.3229} & \textcolor{blue}{ \bf 0.2453} \\[3pt]
 T-ImgNet 		& \textcolor{blue}{ \bf 0.2397} & \textcolor{red}{ \bfseries 0.1579} & \textcolor{blue}{ \bf 0.1667} & \textcolor{ForestGreen}{\bf 0.0321} & \(8\times10^{-6}\) & \textcolor{ForestGreen}{\bf 0.2403} & \textcolor{ForestGreen}{\bf 0.2595} & \textcolor{ForestGreen}{\bf 0.1692} \\[3pt]
 CelebA 			& 0.4388 & 0.4265 & 0.4054 & 0.3165 & 0.2406 & \(7\times10^{-6}\) & 0.3620 & \textcolor{red}{ \bfseries 0.2856} \\[3pt]
 Ukiyo-E 			& 0.5064 & 0.4746 & 0.4408 & 0.3183 & 0.2568 & 0.3610 & \(2\times10^{-5}\) & 0.3022 \\[3pt]
 Church 			& 0.4379 & 0.3916 &  \textcolor{red}{\bfseries 0.3932} & \textcolor{red}{ \bfseries0.2408} & \textcolor{red}{ \bfseries0.1695} & \textcolor{blue}{ \bf 0.2857} & \textcolor{blue}{ \bf 0.3019} & \(3\times10^{-5}\) \\[3pt]
 \bottomrule
 \end{tabular}
 \end{center}
 \vskip-5pt
 \end{table*}

 \begin{table*}[t!]
 \fontsize{8}{12}\selectfont
 \begin{center}
 \caption{A comparison of CSID\(_{m}\) between popular training datasets for \(m = \lfloor\frac{n}{2}\rfloor\). The rows represent the source (Src) and the columns represent to the target (Tar). The \textcolor{ForestGreen}{\bf first}, \textcolor{blue}{\bf second} and \textcolor{red}{\bf third} {\it friendly neighbors} (color coded) of the target are the sources with the three lowest positive CSID\(_m\) values, respectively. CSID\(_m\) is superior to FID or KID, as it assigns negative values to source datasets that are less diverse than the target.} \label{DatasetSSID}  \vskip-0.05in
 \begin{tabular}{P{1.2cm}||P{1.cm}|P{1.2cm}|P{1.cm}|P{1.2cm}|P{1.2cm}|P{0.9cm}|P{1.cm}|P{0.9cm}}
 \toprule 
  \backslashbox{ Src}{ Tar}& MNIST & F-MNIST & SVHN &CIFAR-10 &T-ImgNet & CelebA & Ukiyo-E &  Church \\[3pt]
 \hline\hline
 MNIST 			& 0.1865 & \textcolor{ForestGreen}{\bf  21.886} &  \textcolor{gray}{37.227} & \textcolor{gray}{29.298} &  \textcolor{gray}{9.436} & \textcolor{gray}{198.714} & \textcolor{gray}{201.550} & \textcolor{gray}{205.322}  \\[3pt]
 F-MNIST   		& \textcolor{ForestGreen}{\bf 162.962} & 0.1097 & \textcolor{gray}{46.938} &\textcolor{gray}{19.051} & \textcolor{gray}{-0.5571} & \textcolor{gray}{167.840} & \textcolor{gray}{191.010} & \textcolor{gray}{181.458} \\[2pt] \cline{4-9} \\[-10pt]
 SVHN  			& \textcolor{red}{ \bfseries 212.473} & 77.357 & -0.0566 &  \textcolor{red}{\bf 34.534} & \textcolor{red}{\bf21.668} & 195.631 & 214.507 & 219.790 \\[3pt]
 CIFAR-10 			& 221.337 & \textcolor{red}{ \bfseries 65.426} & \textcolor{ForestGreen}{\bf 52.051} & -0.1478 & -7.109 & \textcolor{blue}{ \bfseries 180.491} & 198.991 & \textcolor{ForestGreen}{\bf 173.655} \\[3pt]
 T-ImgNet 			& 230.916 & 75.737 & \textcolor{red}{\bf67.902}&\textcolor{ForestGreen}{\bf 12.892} & 0.6743 & \textcolor{ForestGreen}{\bf 157.520}  & \textcolor{blue}{ \bfseries 197.447} & \textcolor{red}{ \bfseries184.977} \\[3pt]
 CelebA 			& \textcolor{blue}{\bf 204.794} & 68.828 & 65.299 & \textcolor{blue}{ \bfseries 23.685} & \textcolor{ForestGreen}{\bf 8.829} & 0.6241 & \textcolor{ForestGreen}{\bf 184.170} & 191.927 \\[3pt]
 Ukiyo-E 			& 250.226 & 92.741 & 82.157 & 39.792 & \textcolor{blue}{ \bfseries 18.727} & 191.930 & 0.5494 & \textcolor{blue}{\bf 180.697} \\[3pt]
 Church 			& 212.452 & \textcolor{blue}{\bf 48.676} & \textcolor{blue}{\bf56.136} & -4.655 & -23.115 &  \textcolor{red}{ \bfseries185.740} &  \textcolor{red}{ \bfseries198.750} & -0.5258 \\[3pt]
 \bottomrule
 \end{tabular}
 \end{center}
 \vskip-7pt
 \end{table*}
 
\begin{figure*}[!t]
\begin{center}
  \begin{tabular}[b]{P{.47\linewidth}P{.47\linewidth}}
   \multicolumn{ 2}{c}{\includegraphics[width=0.95\linewidth]{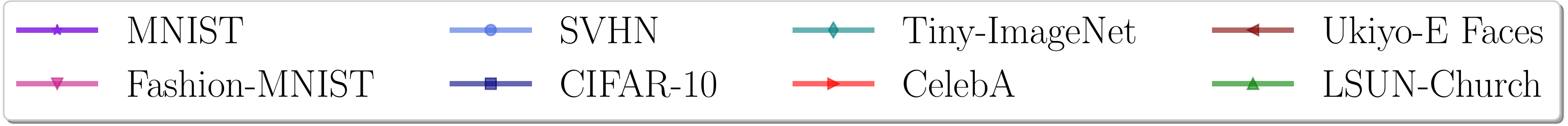} } \\[10pt]
       (a) Fashion-MNIST  & (b) SVHN \\[2pt]
    \includegraphics[width=0.99\linewidth]{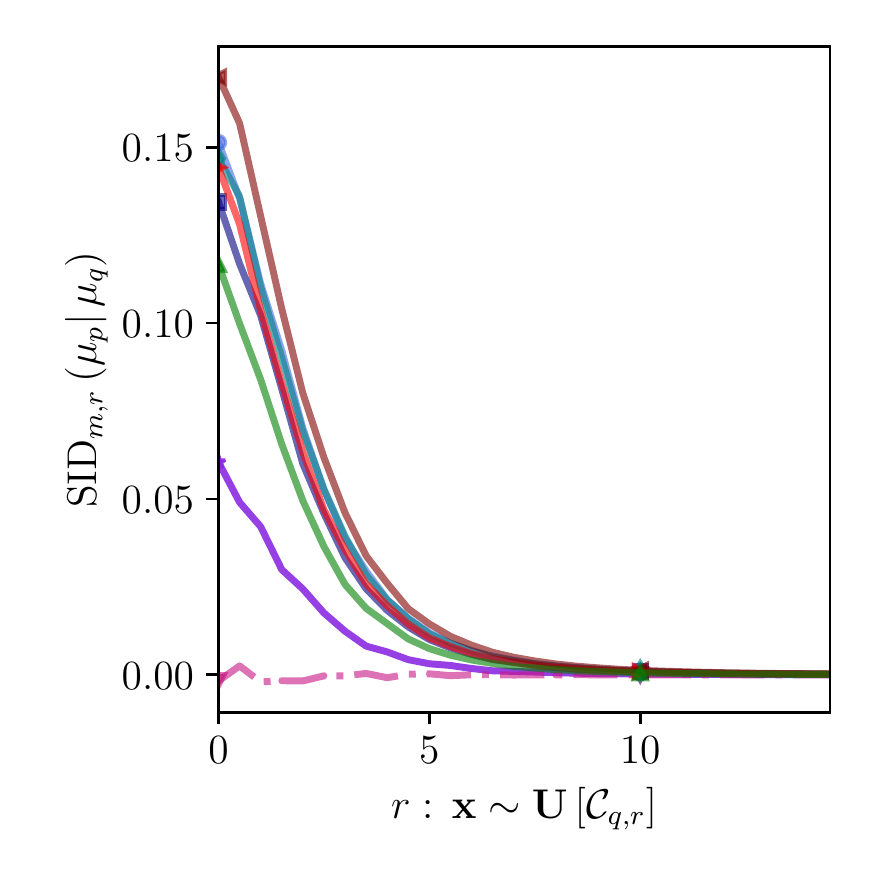} & 
    \includegraphics[width=0.99\linewidth]{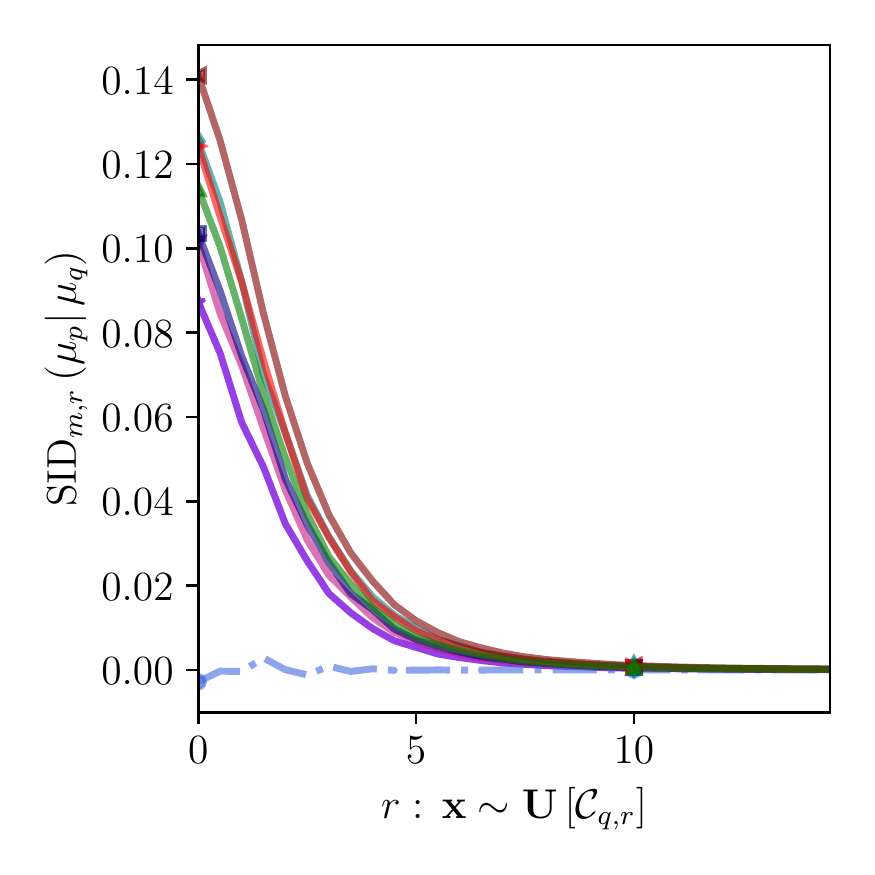}  \\[8pt]
        (c) CelebA  & (c) Ukiyo-E Faces \\[2pt]
    \includegraphics[width=0.99\linewidth]{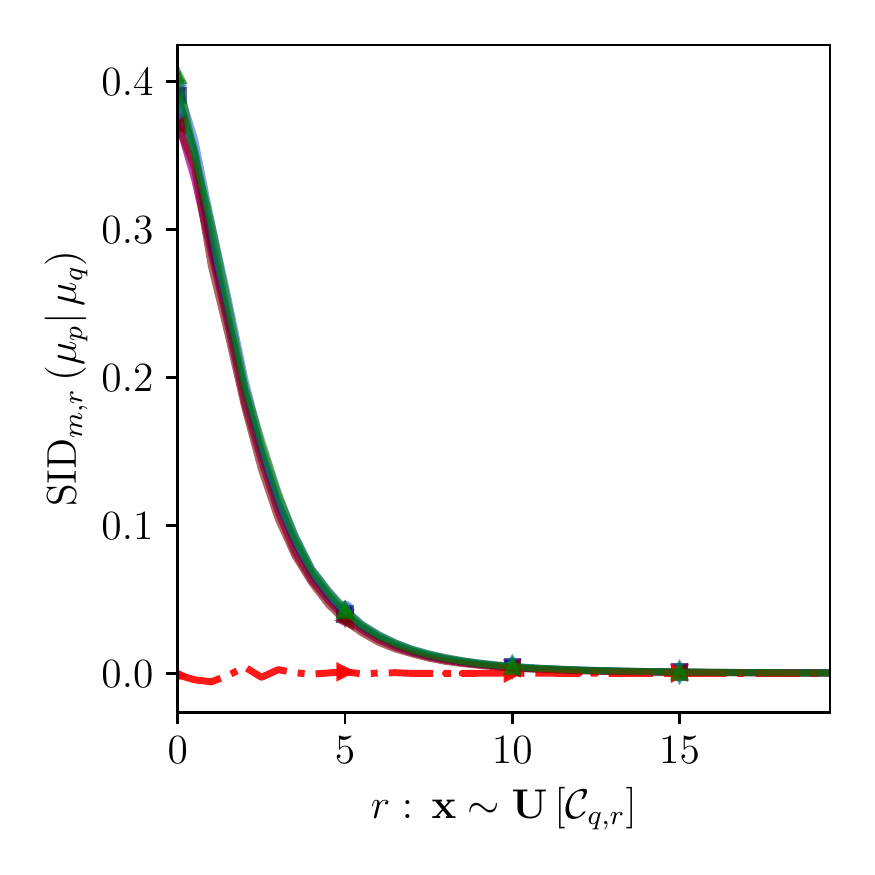} & 
    \includegraphics[width=0.99\linewidth]{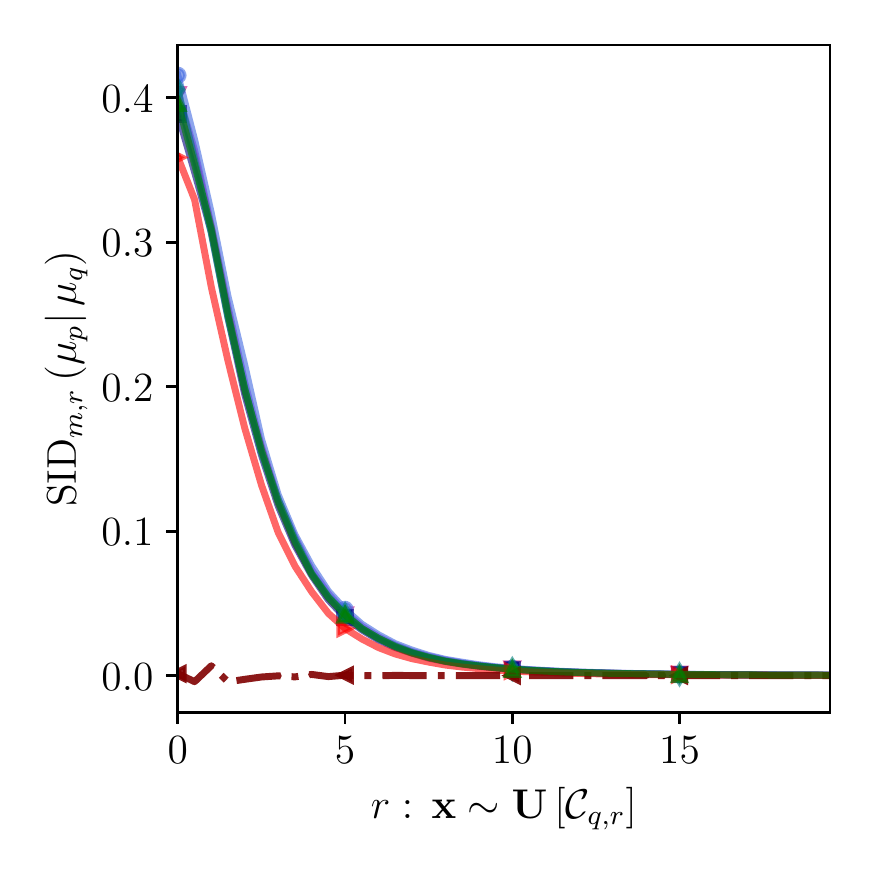}  \\[-2pt]
  \end{tabular} 
\caption[]{\(\mathrm{SID}_{m,r}\) as a function of the hyper-cube length \(r\). We observe that MNIST is the closest neighbor to both Fashion-MNIST and SVHN, while CelebA is marginally closer to Ukiyo-E than the other baselines considered. In scenarios such as case when the target is CelebA or Ukiyo-E Faces, the SID curve alone cannot be used to conclude the {\it friendliest neighbor} of a target dataset, and the area under the curve, CSID\(_{m}\) is more informative (cf. Table~\ref{DatasetSSID}) .}
\label{Fig_SID_Rest}  
\end{center}
\vskip-1em
\end{figure*}

\FloatBarrier
\newpage
 \section{The Signed Inception Distance (SID) } \label{App_SID}
 In this appendix, we derive a favorable theoretical guarantee of the SID metric, discuss the algorithm for the computation of SID with relevant ablation experiments on synthetic Gaussian and image datasets. 
  \subsection{Asymptotic Behavior of the Signed Distance}
Without loss of generality, consider the {\it signed distance} presented in Equation~\eqref{Eqn_SD}:
 \begin{align*}
\mathrm{SD}_{m,r}(\mu_p\|\mu_q) =  \frac{1}{M_{\x}}\sum_{\substack{\ell=1\\\tilde{\x}_{\ell}\in\mcalC_{q,r}}}^{M_{\x}} \left\{  \frac{1}{N_q} \sum_{\substack{j=1\\\bmc_j\sim\mu_q}}^{N_q} \Phi(\x_{\ell},\bmc_j) - \frac{1}{N_p} \sum_{\substack{i=1\\\tilde{\bmc}_i\sim\mu_p}}^{N_p} \Phi(\x_{\ell},\tilde{\bmc}_i)\right\}.
\end{align*}
Asymptotically, when infinite samples are drawn from the test space, \(\mcalC_{q,r}\), we get
 \begin{align*}
\mathrm{SD}_{m,r}(\mu_p\|\mu_q) &=  \lim_{M_{\x}\rightarrow \infty} \left\{\frac{1}{M_{\x}}\sum_{\substack{\ell=1\\\tilde{\x}_{\ell}\in\mcalC_{q,r}}}^{M_{\x}} \left\{  \frac{1}{N_q} \sum_{\substack{j=1\\\bmc_j\sim\mu_q}}^{N_q} \Phi(\x_{\ell},\bmc_j) - \frac{1}{N_p} \sum_{\substack{i=1\\\tilde{\bmc}_i\sim\mu_p}}^{N_p} \Phi(\x_{\ell},\tilde{\bmc}_i)\right\} \right\}. \\
&= \kappa \int_{\x\in\mcalC_{q,r}}  \Big\{\frac{1}{N_q} \sum_{\substack{j=1\\\bmc_j\sim\mu_q}}^{N_q} \Phi(\x,\bmc_j) - \frac{1}{N_p} \sum_{\substack{i=1\\\tilde{\bmc}_i\sim\mu_p}}^{N_p} \Phi(\x,\tilde{\bmc}_i)\Big\}~\rmd\x,
\end{align*}
for some positive constant \(\kappa\). Similarly, when the number of centers drawn from \(\mu_p\) and \(\mu_q\) tends to infinity, the inner summations can be replaced with their corresponding expectations, resulting in
 \begin{align*}
\mathrm{SD}_{m,r}(\mu_p\|\mu_q) &= \kappa \int_{\x\in\mcalC_{q,r}}  \lim_{N_{q}\rightarrow \infty}\Big\{\frac{1}{N_q} \sum_{\substack{j=1\\\bmc_j\sim\mu_q}}^{N_q} \Phi(\x,\bmc_j) \Big\} -\lim_{N_{p}\rightarrow \infty}\Big\{ \frac{1}{N_p} \sum_{\substack{i=1\\\tilde{\bmc}_i\sim\mu_p}}^{N_p} \Phi(\x,\tilde{\bmc}_i)\Big\}~\rmd\x. \\
& = \kappa^{\prime}\int_{\x\in\mcalC_{q,r}} \left( \int_{\y}  \Phi(\x,\y)\mu_q(\y)~\rmd\y  - \int_{\y} \Phi(\x,\y) \mu_p(\y)~\rmd\y \right)~\rmd\x. \\
& = \kappa^{\prime} \int_{\x\in\mcalC_{q,r}} \left( \mathbb{E}_{\y\sim\mu_q} \left[ \Phi(\x,\y)\right]  - \mathbb{E}_{\y\sim\mu_p} \left[\Phi(\x,\y)\right]  \right)~\rmd\x.
\end{align*}
Recall that the samples \(\x_{\ell}\) are drawn uniformly at random from \(\mcalC_{q,r}\)  (cf. Section~\ref{Sec_SID}). This allows us to replace the outer integral with another expectation, resulting in
 \begin{align*}
\mathrm{SD}_{m,r}(\mu_p\|\mu_q) &=  \mathbb{E}_{\x\sim\mcalC{q,r},\y\sim\mu_q} \left[ \Phi(\x,\y)\right]  - \mathbb{E}_{\x\sim\mcalC{q,r},\y\sim\mu_p} \left[\Phi(\x,\y)\right].
\end{align*}
The above result links the SD to kernel statistics and provides the asymptotic guarantee that when the two distributions \(\mu_p\) and \(\mu_q\) coincide, {\it i.e.,} \(\mu_p=\mu_q\), and therefore, \(\mathrm{SD}_{m,r}(\mu_p\|\mu_p) := 0\).
 \subsection{SID Computation}
 The procedure to compute the signed distance between the samples drawn from two distributions is given in Algorithm~\ref{Algo_SD}. While the algorithm is easily implementable for low-dimensional data, an extension to practical settings with images necessitates computing Inception embeddings over batches of samples. The signed distance (SD) computed over Inception embeddings is called SID. To extend the SID computation algorithm for evaluating GANs, we consider \(\mfrakD_q\),  the target dataset, and \(\mfrakD_p\), samples drawn from the generator. We set \(|\mfrakD_q| = |\mfrakD_p| = 5000\). For each \(r\), we average \(\mathrm{SID}_{m,r}\) over batches of size \(N_B = 100\). This allows for efficient computation of the Inception features for high-resolution images. Algorithm~\ref{Algo_SID} presents this modified approach for evaluating GANs with SID. We implement the SID computation atop the publicly available Clean-FID~\cite{CleanFID21} library. Similar to the Clean-FID framework, SID can be computed between any two image folders using the Clean-FID backend. As a result, the InceptionV3 mapping and resizing functions are consistent with the existing Clean-FID approach. Details regarding the public release of the Python + TensorFlow/PyTorch library for SID computation are discussed in Appendix~\ref{App_GitHub} of this document.

  \begin{algorithm}[!t]
    \caption{Signed distance (SD) between two distributions} \label{Algo_SD}
    \KwIn{Source data \(\mfrakD_p = \{ \tilde{\bmc}_i~|~i=1,2,3,\ldots,N_p;~\tilde{\bmc}_i\sim \mu_p\}\), kernel order \(m\), dimensionality \(n\), \\ \quad \quad \quad Target data \(\mfrakD_q = \{ \bmc_j~|~j=1,2,3,\ldots,N_q;~\bmc_j\sim \mu_q\}\), max radius \(r_{\max}\), step size \(\eta\), \\ \quad \quad \quad \,batch size \(M_{\x}\)}
    {\bf Compute:} $ \mu_q = \mathrm{mean}\left(\bmc_j \sim \mfrakD_q\right); \Sigma_q = \mathrm{covariance}\left(\bmc_j \sim \mfrakD_q\right) $ \;
    \For{\(k = 0, 1, 2, ... r_{\max}\)}{
    	{\bf Compute:} Hypercube length \(r = \eta\times k\times\max(\mathrm{diag}(\Sigma_q))  \) \\ 
    	{\bf Define:}  Hypercube \( \mcalC_{q,r} \): Center = \(\mu_q\)\\
	{\bf Sample:}  \(\x_{\ell} \sim \mathrm{Uniform}\left[\mcalC_{q,r} \right];~\ell = 1,2,3,\ldots,M_{\x}\) \\
	{\bf Compute:} \(\mathrm{SD}_{m,r}(\mu_p\|\mu_q)\) based on Equation~\eqref{Eqn_SD}
    }
\KwOut{Plot of \(\mathrm{SD}_{m,r}\) versus \(r\)}
\end{algorithm}

 \begin{algorithm}[!t]
    \caption{Signed inception distance (SID) between the generator output and target data} \label{Algo_SID}
    \KwIn{Target data \(\mfrakD_q = \{ \bmc_j~|~j=1,2,\ldots,N_q;~\bmc_j\sim \mu_q\}\), kernel order \(m\), dimensionality \(n\), \\ \quad \quad \quad max radius \(r_{\max}\), step size \(\eta\), hypercube sample batch size \(M_{\x}\), Generator G,  \\ \quad \quad \quad  Generator batch size \(N_B\), Inception model \(\psi\).} 

    {\bf Compute:} $ \mu_q = \mathrm{mean}\left(\bmc_j \sim \mfrakD_q\right); \Sigma_q = \mathrm{covariance}\left(\bmc_j \sim \mfrakD_q\right) $ \;
    \For{\(k = 0, 1, 2, ... r_{\max}\)}{
    	\For{ Batches \(i = 1, 2, \ldots \frac{N_q}{N_B}\)}{
    	{\bf Sample:}  \(\z_{\ell} \sim p_Z(\z);~\ell = 1,2,3,\ldots,N_{B}\) --  Generator inputs\\
	{\bf Sample:}  \(\tilde{\bmc}_{\ell} \sim G(\z);~\ell = 1,2,3,\ldots,N_{B}\) -- Generator outputs \\
	{\bf Sample:}  \(\bmc_{j} \sim \mfrakD_q;~\ell = 1,2,3,\ldots,N_{B}\) -- Target data samples \\
	{\bf Compute:} \(\psi(\tilde{\bmc}_{\ell})\) and \(\psi(\bmc_{j})\)  -- Inception embeddings of generator output and target data. \\
    	{\bf Compute:} Hypercube length \(r = \eta\times k\times\max(\mathrm{diag}(\Sigma_q))  \) \\ 
    	{\bf Define:}  Hypercube \( \mcalC_{q,r} \): Center = \(\mu_q\)\\
	{\bf Sample:}  \(\x_{\ell} \sim \mathrm{Uniform}\left[\mcalC_{q,r} \right];~\ell = 1,2,3,\ldots,M_{\x}\) \\
	{\bf Compute:} \(\mathrm{SID}_{m,r}\) between \(\psi(\tilde{\bmc}_{\ell})\) and \(\psi(\bmc_{j})\) based on Equation~\eqref{Eqn_SID}
	}
    }
    {\bf Compute:} \(\text{CSID}_{m} = \sum_r \text{SID}_{m,r}(\mu_p\|\mu_q) \)    \\
\KwOut{Plot of \(\mathrm{SID}_{m,r}\) versus \(r\); Measure \(\text{CSID}_{m}\)}
\end{algorithm}

 \subsection{Experiments on Gaussian Data} \label{App_SIDGaussians}
To begin with, we present results on computing the signed distance (SD) for various representative Gaussian and Gaussian mixture source and target distributions. \par
\begin{figure*}[!b]
\begin{center}
  \begin{tabular}[b]{P{.75\linewidth}}
   \includegraphics[width=0.5\linewidth]{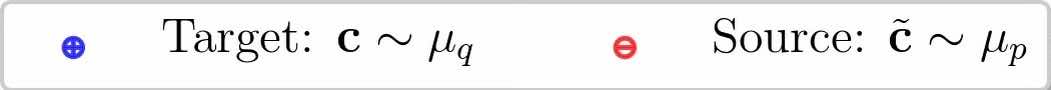}  \\
    \includegraphics[width=0.99\linewidth]{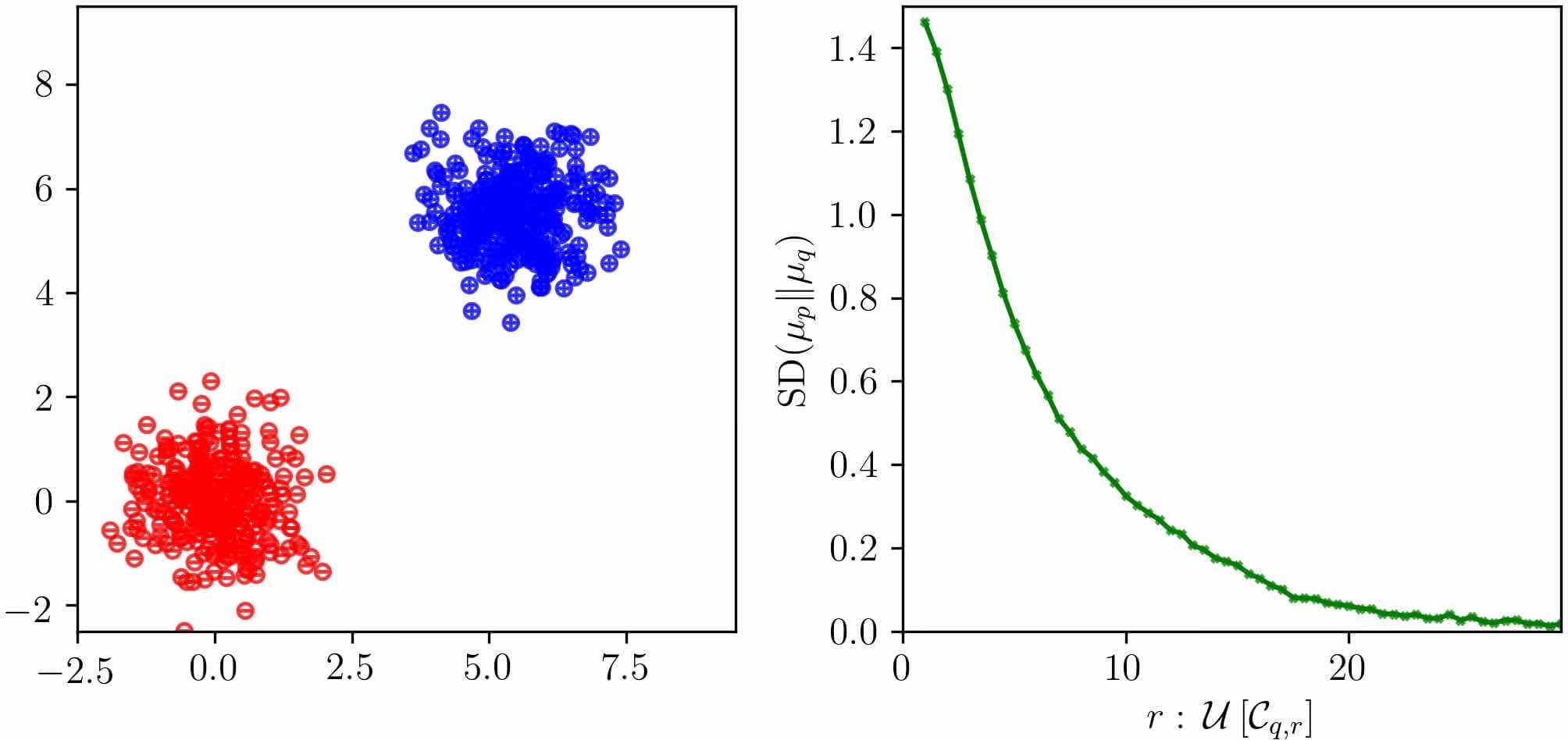}  \\[-3pt]
    (a)   \\[3pt]
    \includegraphics[width=0.99\linewidth]{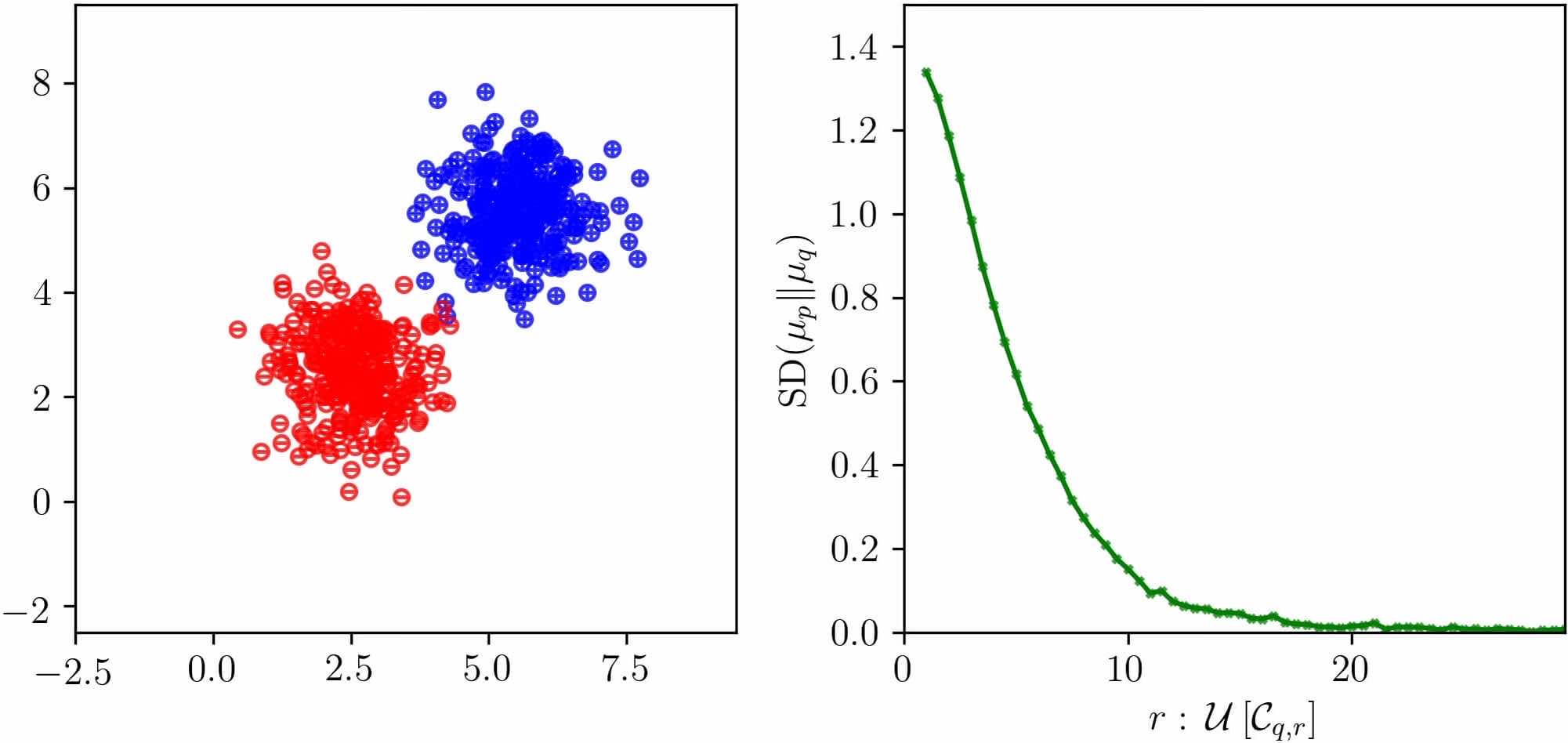}  \\[-3pt]
    (b)   \\[3pt]
    \includegraphics[width=0.99\linewidth]{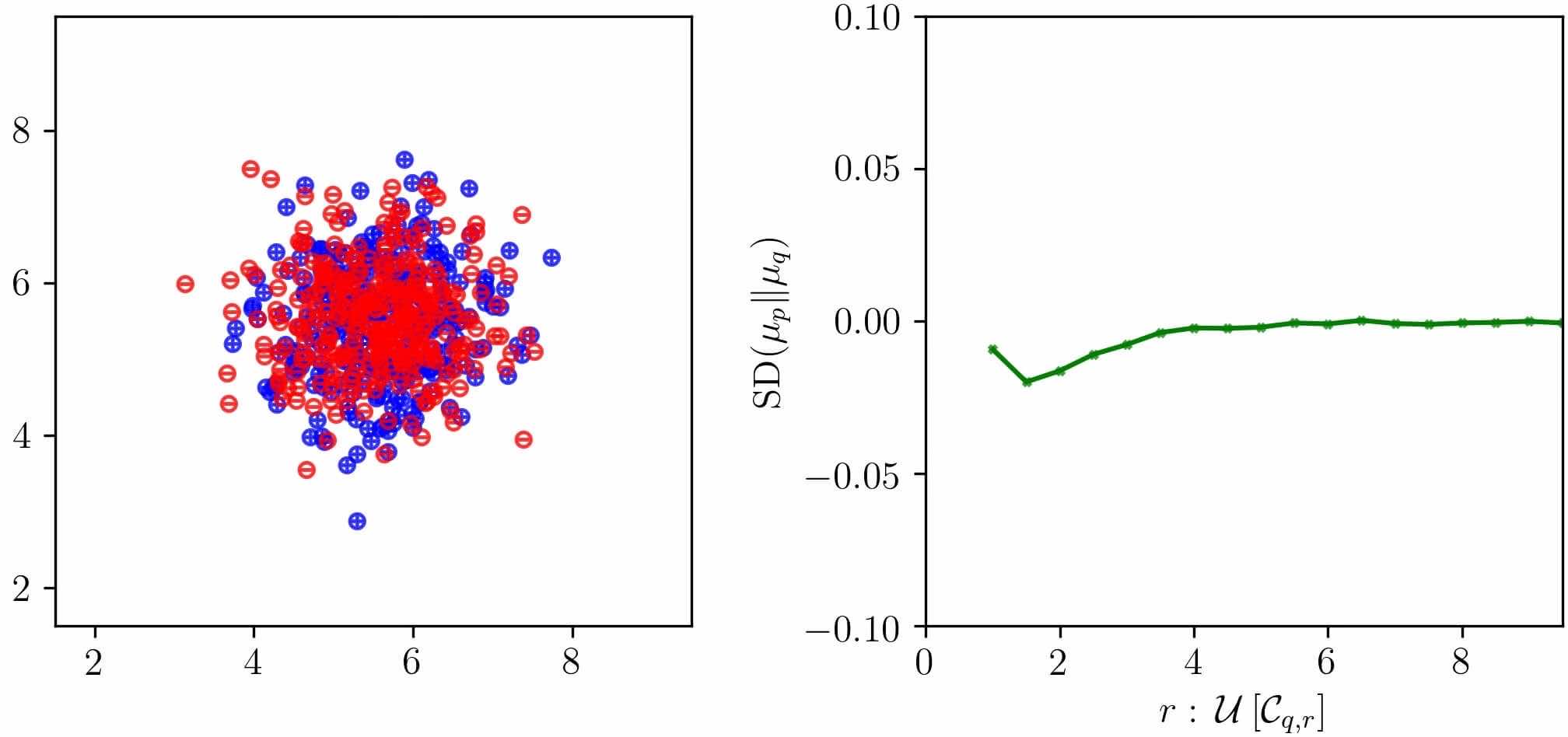}  \\[-3pt]
    (c)   \\[3pt]
  \end{tabular} 
\caption[]{Plots of the signed distance \(\mathrm{SD}_{m,r}\) between a source Gaussian \(\mu_p = \mathcal{N}(\bm{m}, 0.75\mathbb{I}_2)\) from a target Gaussian \(\mu_q = \mathcal{N}(5.5 \bm{1}_2, 0.75\mathbb{I}_2)\) for (a) \(\bm{m}_p =  \bm{0}_2\); (b) \(\bm{m}_p =  2.5\bm{1}_2\) ; and (c) \(\bm{m}_p =  5.5\bm{1}_2\). The closer the source Gaussian is to the target, the faster \(\mathrm{SD}_{m,r}(\mu_p\|\mu_q)\) approaches zero. When the two distributions coincide,  \(\mathrm{SD}_{m,r}(\mu_p\|\mu_q)\) is zero for all \(r\).  }
\label{Fig_SDGaussians_VarSame}  
\end{center}
\vskip-1em
\end{figure*}

 Figures~\ref{Fig_SDGaussians_VarSame}(a)-(c) present the visualization of SD versus \(r\) for a Gaussian target distribution with \(\mu_q = \mathcal{N}(5.5 \bm{1}_2, 0.75\mathbb{I}_2)\), where \( \bm{1}_2\) denotes a 2-D vector with all entries equal to 1. Consider the scenario where the source and target Gaussians possess the same variance, but different means. Consider three different sources \(\mu_p = \mathcal{N}(\bm{m}_p, \Sigma_p)\),  given by: (a) \(\bm{m}_p =  \bm{0}_2\) and \( \Sigma_p = 0.75\mathbb{I}_2\); (b) \(\bm{m}_p =  2.5\bm{1}_2\) and \( \Sigma_p = 0.75\mathbb{I}_2\); and (c) \(\bm{m}_p=  5.5\bm{1}_2\) and \( \Sigma_p = 0.75\mathbb{I}_2\). We observe that, when the source is far away from the target, SD is positive-valued and gradually approaches zero. When the two distributions are identical, SD is zero for all \(r\). In the context of identifying a friendly neighbor, a closer source dataset is expected to converge faster to zero than one that is far away. \par
\begin{figure*}[!b]
\begin{center}
  \begin{tabular}[b]{P{.75\linewidth}}
   \includegraphics[width=0.5\linewidth]{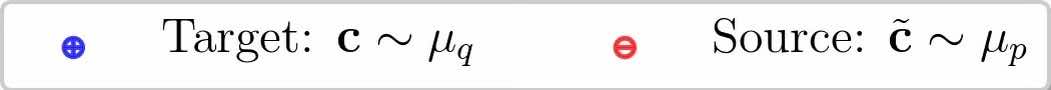}  \\
    \includegraphics[width=0.99\linewidth]{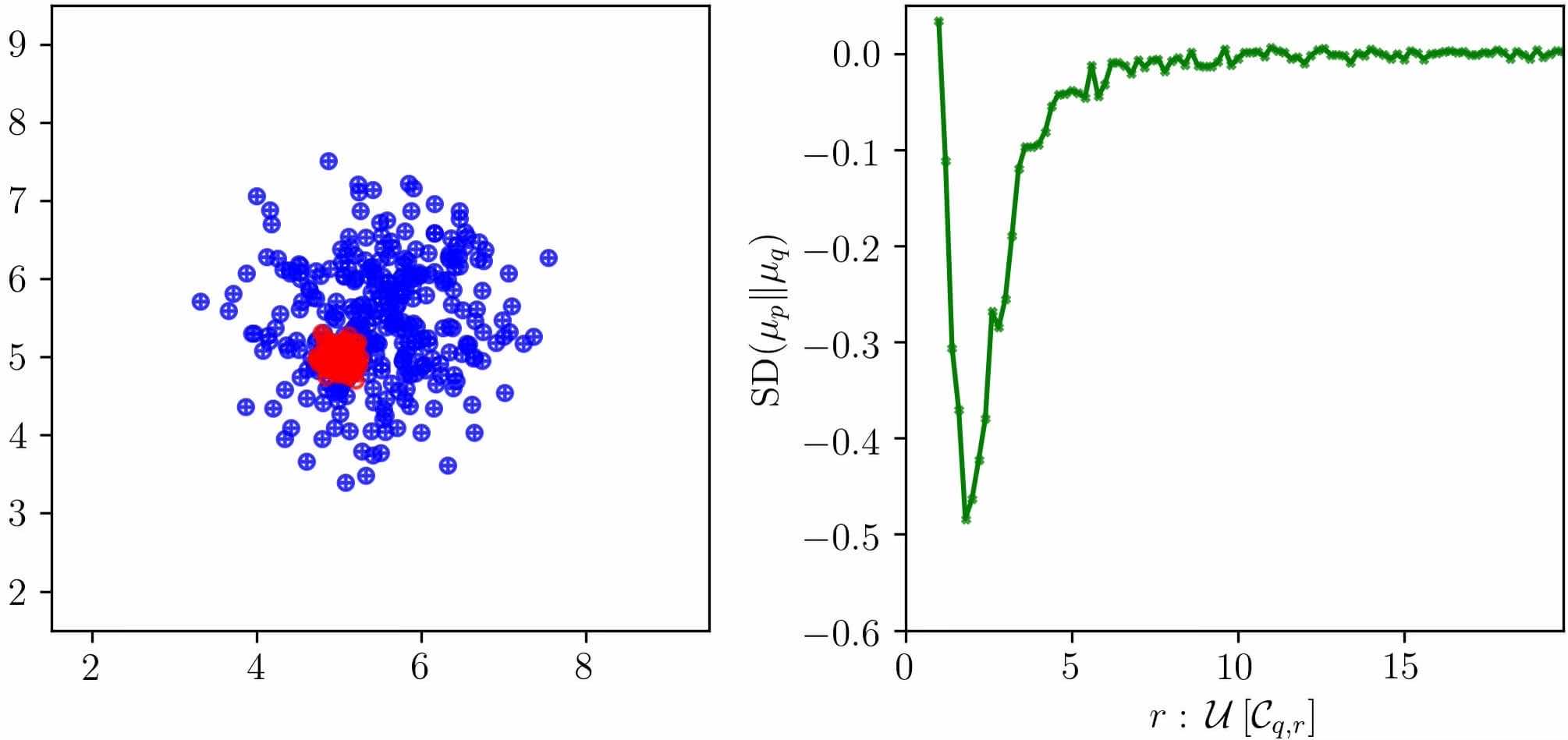}  \\[-3pt]
    (a)   \\[3pt]
    \includegraphics[width=0.99\linewidth]{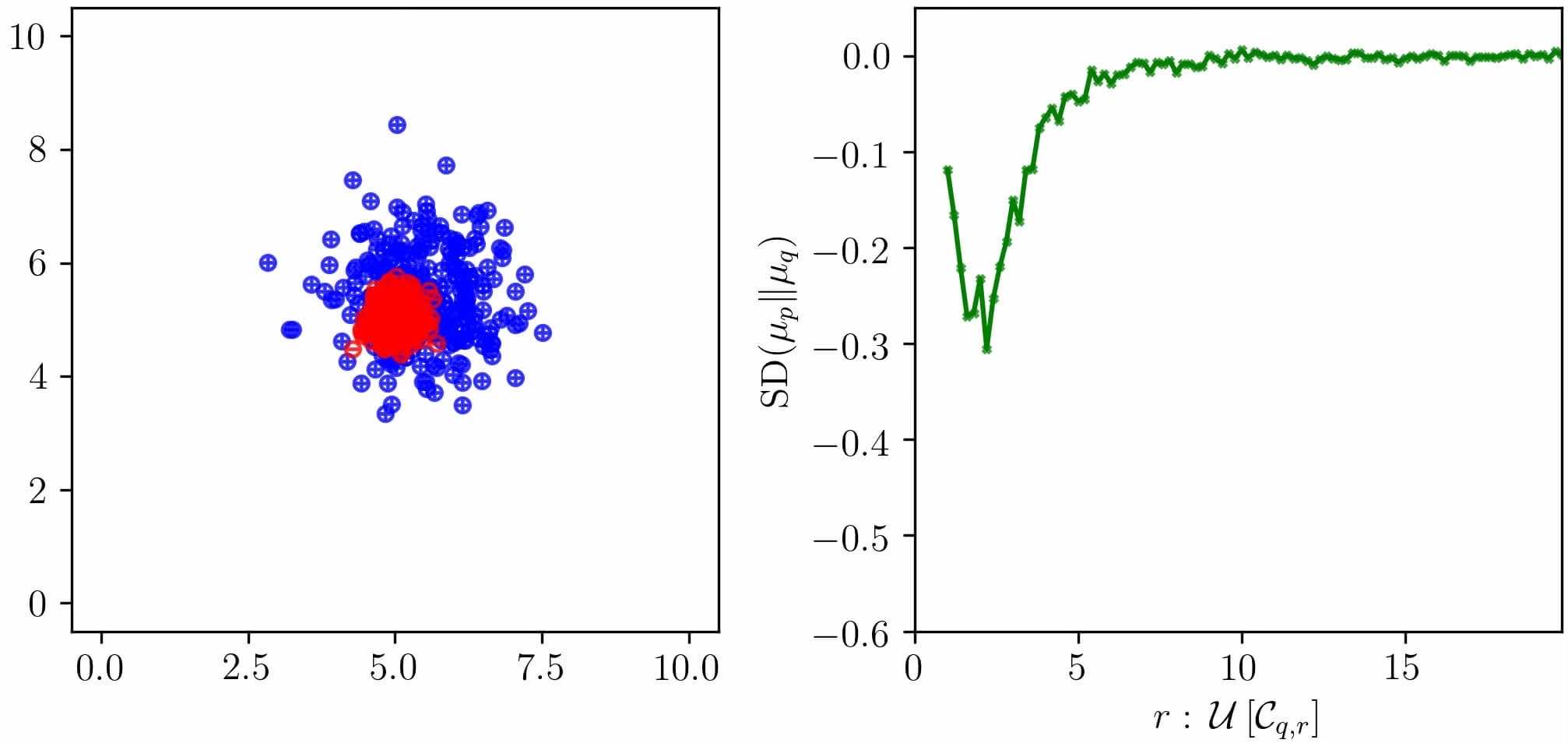}  \\[-3pt]
    (b)   \\[3pt]
    \includegraphics[width=0.99\linewidth]{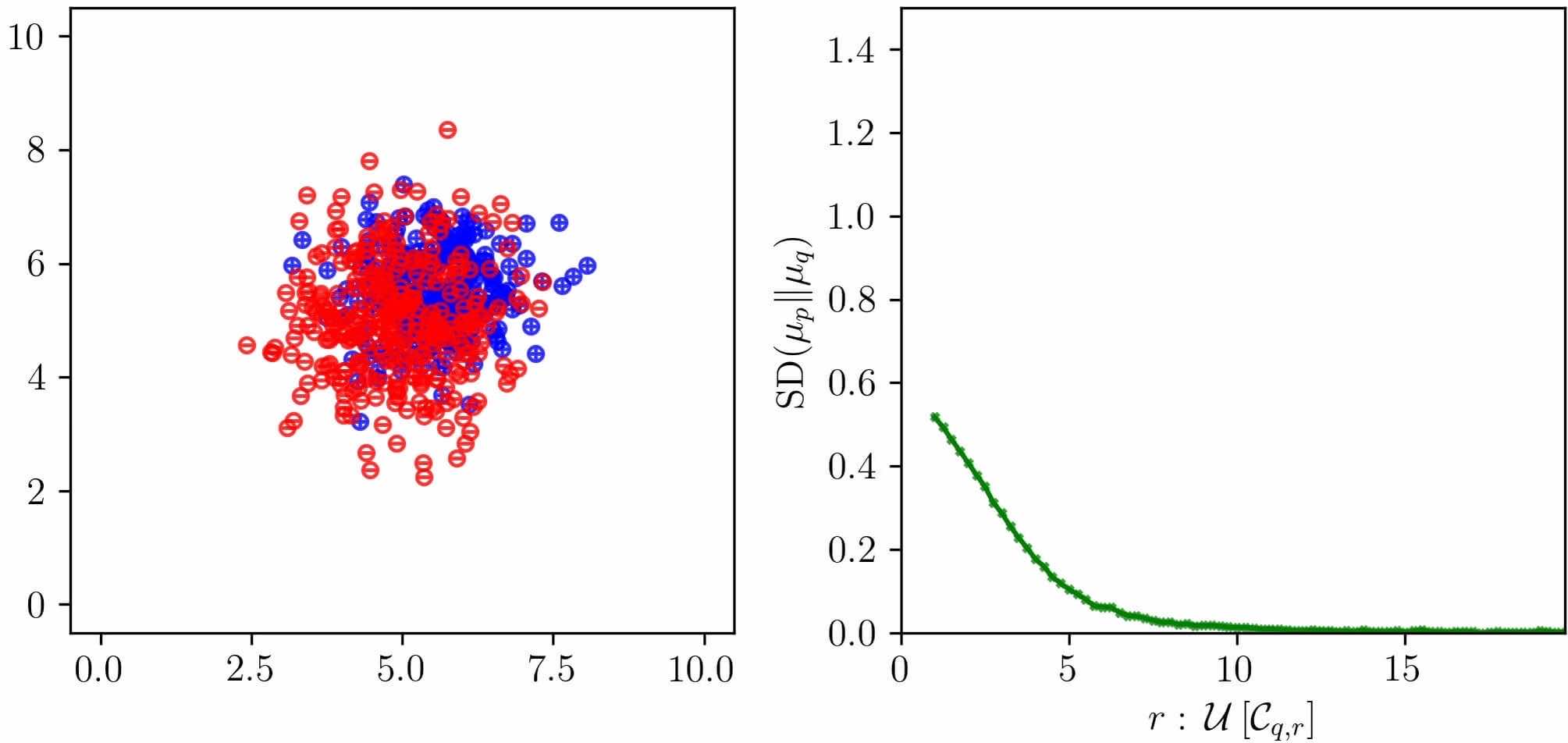}  \\[-3pt]
    (c)   \\[3pt]
  \end{tabular} 
\caption[]{Plots comparing the signed distance \(\mathrm{SD}_{m,r}\) between a source Gaussian \(\mu_p = \mathcal{N}(5.5 \bm{1}_2, 0.75\mathbb{I}_2)\) and a target Gaussian \(\mu_q = \mathcal{N}(5.5 \bm{1}_2, \Sigma_p)\) for (a) \( \Sigma_p = 0.1\mathbb{I}_2\); (b) \( \Sigma_p = 0.25\mathbb{I}_2\); and (c) \( \Sigma_p = \mathbb{I}_2\). When the source Gaussian overlaps with the target but with a smaller variance, \(\mathrm{SD}_{m,r}\) is negative. However, if the source has a larger variance than the target, \(\mathrm{SD}_{m,r}\) is positive. }
\label{Fig_SDGaussians_MeanSame}  
\end{center}
\vskip-1em
\end{figure*}

Figures~\ref{Fig_SDGaussians_MeanSame}(a)-(c) present the results for the other scenario where the mean is fixed, but the variances are different. Consider the same target as before, but with the following source distributions: (a) \(\bm{m}_p =  5\bm{1}_2\) and \( \Sigma_p = 0.1\mathbb{I}_2\); (b) \(\bm{m}_p =  5\bm{1}_2\) and \( \Sigma_p = 0.25\mathbb{I}_2\); and (c) \(\bm{m}_p =  \bm{1}_2\) and \( \Sigma_p = \mathbb{I}_2\). We observe that when the spread of the source is smaller than the target, SD initially goes negative, and subsequently converges to zero once the hypercube \(\mathcal{C}_{q,r}\) encompasses the source. On the other hand, when the spread of the source is greater than that of the target (as desired for identifying friendly neighbors), SD is always positive, and converges to zero faster if the relative spread between the source and target is smaller. \par

\begin{figure*}[!b]
\begin{center}
  \begin{tabular}[b]{P{.75\linewidth}}
   \includegraphics[width=0.5\linewidth]{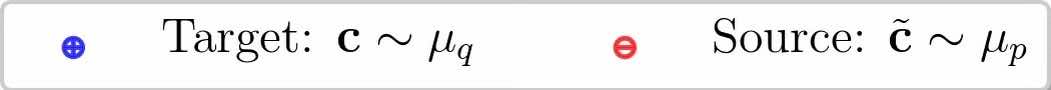}  \\
    \includegraphics[width=0.99\linewidth]{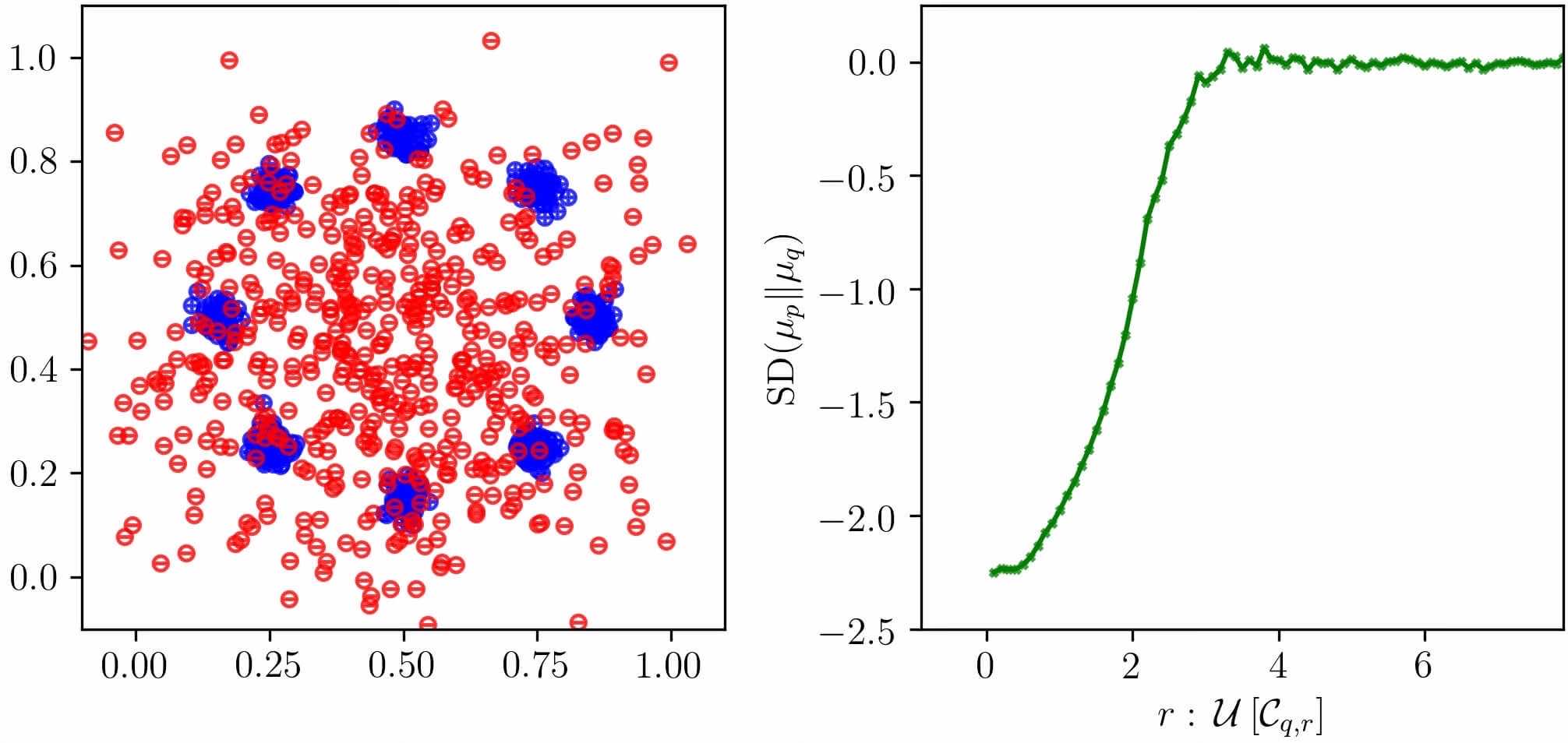}  \\[-3pt]
    (a)   \\[1pt]
    \includegraphics[width=0.99\linewidth]{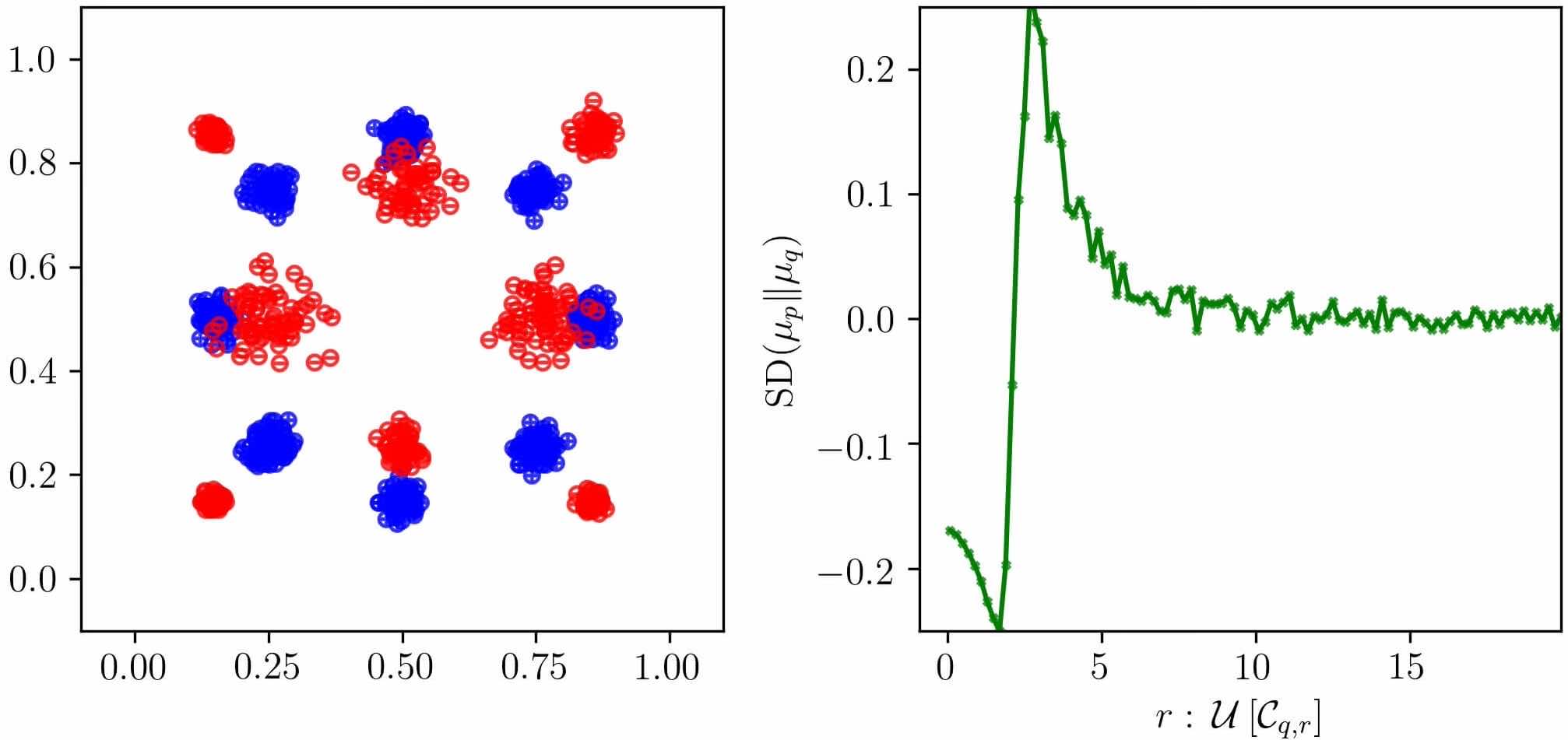}  \\[-3pt]
    (b)   \\[1pt]
    \includegraphics[width=0.99\linewidth]{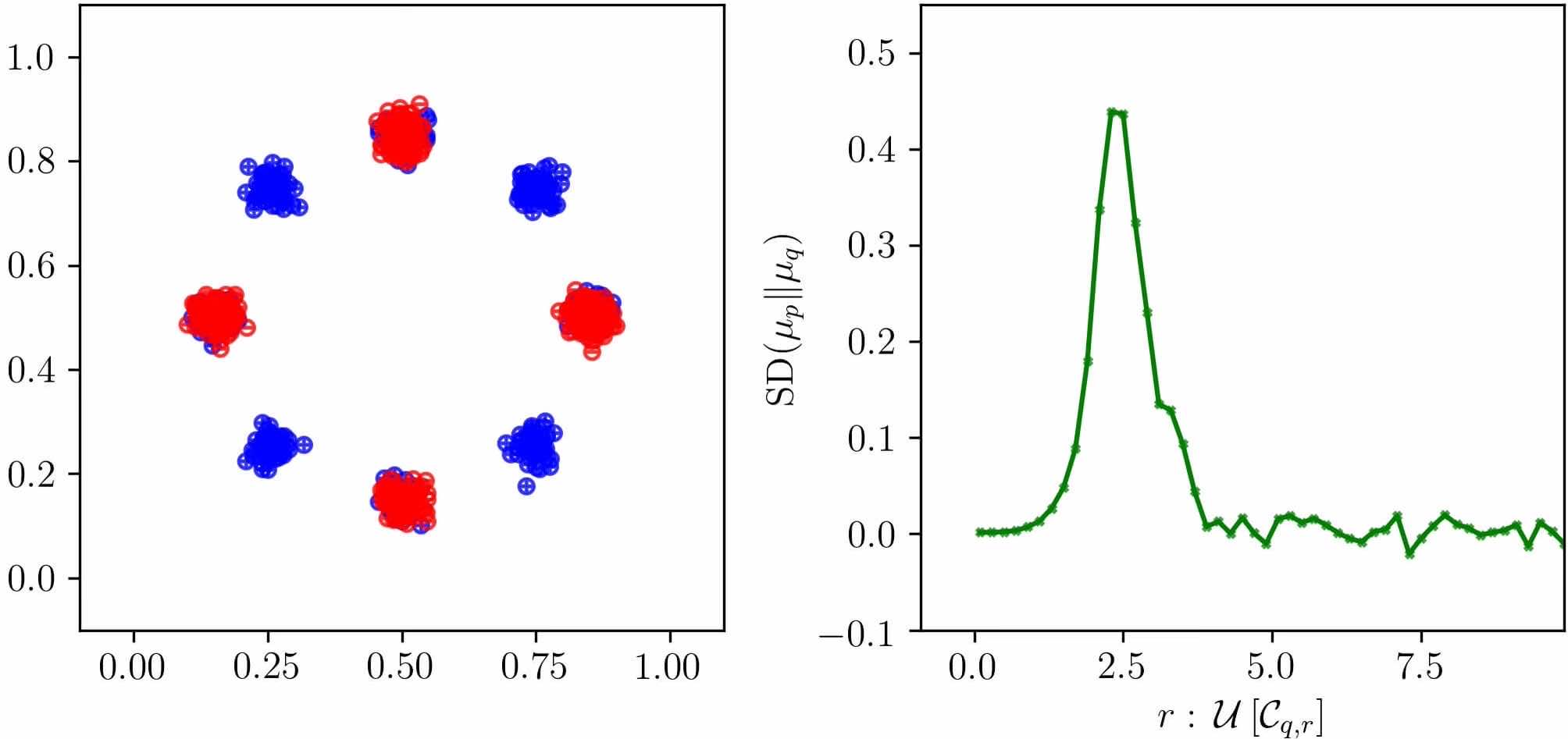}  \\[-3pt]
    (c)   \\[-5pt]
  \end{tabular} 
\caption[]{Plots comparing the signed distance \(\mathrm{SD}_{m,r}\) when the target is a Gaussian mixture density. (a) Unimodal Gaussian source with identical first and second moments as the target; \(\mathrm{SD}_{m,r}\) is negative as the source has lower diversity than the target. (b) A Gaussian mixture distinct from the target; \(\mathrm{SD}_{m,r}\) flips sign based on the relative concentrations of the source and target samples. (c) A mode-collapsed source results in a non-zero \(\mathrm{SD}_{m,r}\) although FID and KID between these distributions would be zero. }
\label{Fig_SDGMMs}  
\end{center}
\vskip-1em
\end{figure*}

To evaluate SD on a Gaussian mixture target, consider an 8-component Gaussian mixture model (GMM) with means drawn from \([0,1]\times[0,1]\) and identical covariance matrices \( 0.02\mathbb{I}_2\). Consider three source distributions: (a) A Gaussian with first and second moments matching that of the target; (b) An 8-component GMM distinct from the target; and (c) A 4-component GMM that has mode-collapsed on to some of the modes of the target. Figures~\ref{Fig_SDGMMs}(a)-(c) present these three scenarios and the associated SD versus \(r\). For Scenario (a), although the mean and covariance of both the source and target are identical, we observe that SD is negative, as the two distributions do not have a large overlap, preventing the positive and negative charges from cancelling each other. In Scenario (b), SD is able to capture the change in concentration between \(\mu_p\) and \(\mu_q\), indicated by the sudden sign change in SD. When \(\mu_p\) converges to a few modes of the target \(\mu_q\), SD is not zero for all \(r\), which indicates that the two distributions are not identical. In this scenario, however, FID between the two distribution would be close to zero as they have approximately the same first and second moments. 
  
Animations pertaining to these experiments are available at \url{https://github.com/DarthSid95/clean-sid}. \par

 \subsection{Evaluating GANs with SID} \label{App_GANsSID}
 
\begin{figure*}[!b]
\begin{center}
  \begin{tabular}[b]{P{.45\linewidth}P{.45\linewidth}}
   \multicolumn{ 2}{c}{\includegraphics[width=0.75\linewidth]{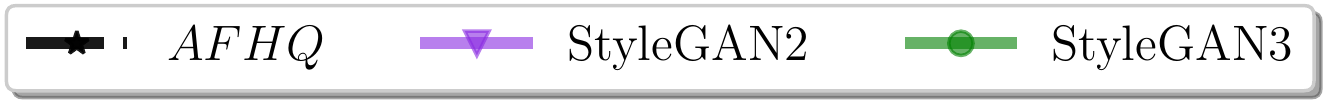} } \\
    \includegraphics[width=0.99\linewidth]{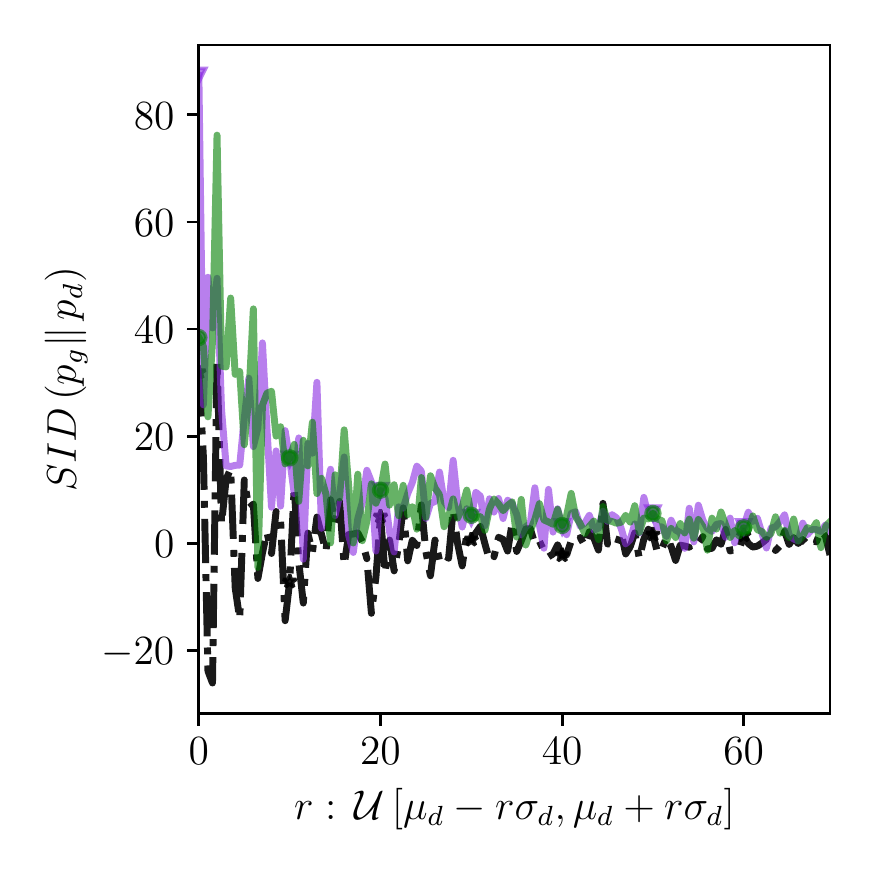} & 
    \includegraphics[width=0.99\linewidth]{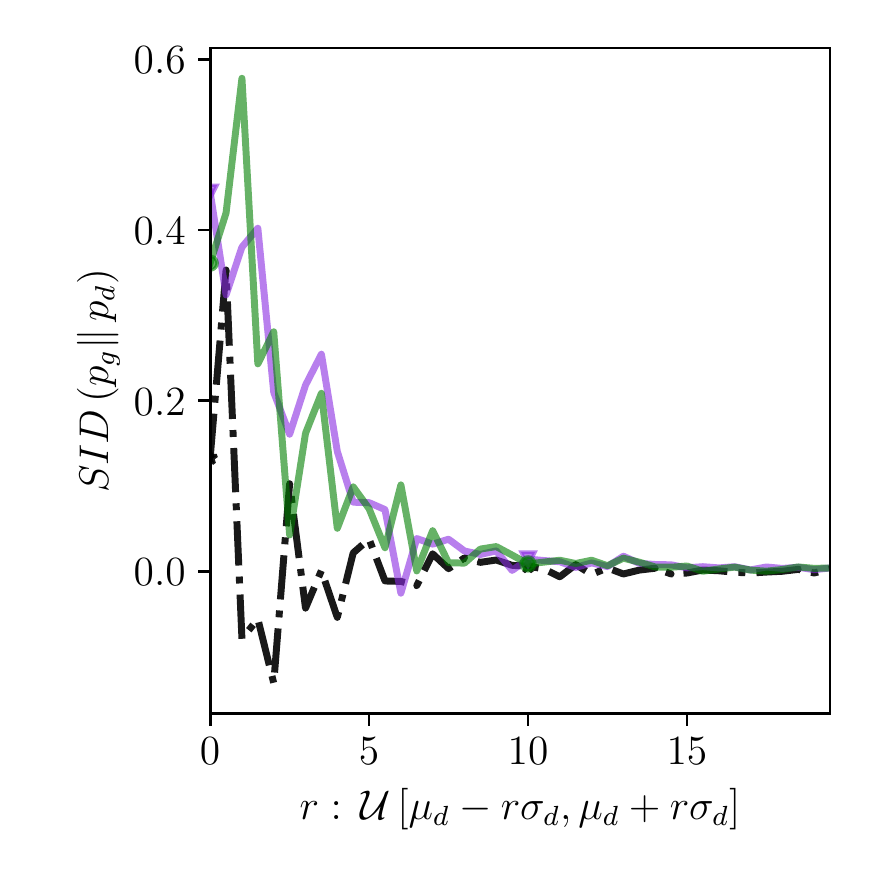}  \\[-1pt]
    (a) \(2m-n =1\)  & (b) \(2m-n =-1\) \\[1pt]
    \includegraphics[width=0.99\linewidth]{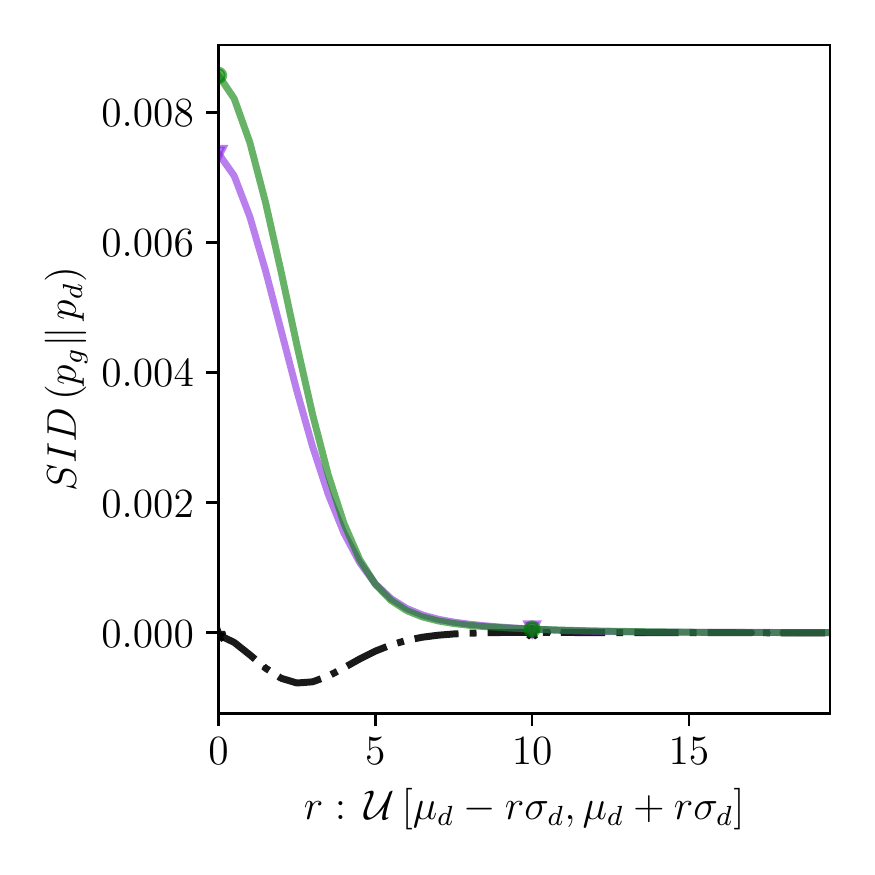} & 
    \includegraphics[width=0.99\linewidth]{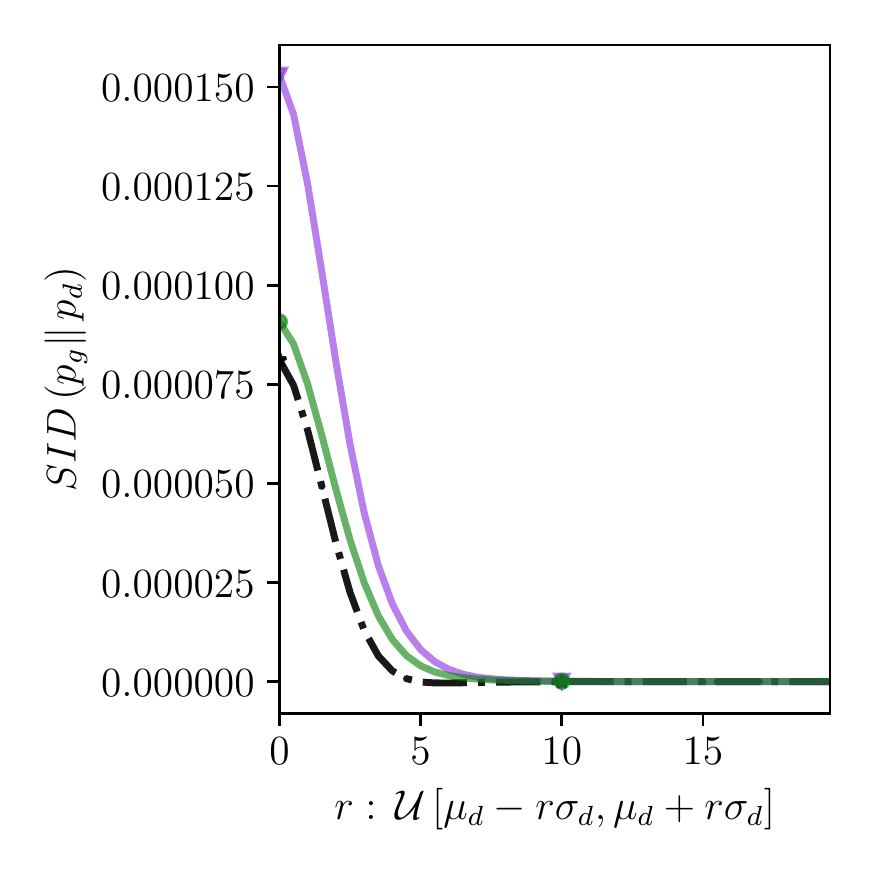}  \\[-1pt]
    (c) \(2m-n =-3\)  & (c) \(2m-n =-5\) \\[1pt]
  \end{tabular} 
\caption[]{SID versus \(r\) for multiple choices of \(2m-n\) for the case when the target dataset is \(512\times512\) Animal Faces HQ. Images generated by StyleGAN2 and StyleGAN3 are compared with the AFHQ dataset as the target. We observe that StyleGAN3 has a performance comparable to StyleGAN2 for higher orders \(m\). Convergence for lower orders is indicative of superior performance, as the penalty for mismatch between the source and target distributions increases with decrease in the order. The SID for StyleGAN3 closely matches the SID of the target data with itself for \(2m-n = -5\), indicating superior performance to StyleGAN2. This finding is in agreement with the comparison between StyleGAN2 and StyleGAN3 in terms of FID/KID reported in \cite{CleanFID21}. }
\label{Fig_AFHQSID}  
\end{center}
\vskip-1em
\end{figure*}

We consider evaluating pre-trained models with the SID measure to compare the performance with FID and KID. As a demonstration, we consider StyleGAN2~\cite{ADAStyleGAN20} and StyleGAN3~\cite{StyleGAN321} models with weights trained on \(512\times512\) high-quality Animal Faces (AFHQ) dataset~\cite{ADAStyleGAN20}. As a reference/benchmark, we also consider SID of the AFHQ dataset with itself. We consider orders in the range \(m = \lfloor \frac{n}{2}\rfloor-3, \lfloor \frac{n}{2}\rfloor-2, \ldots \lfloor \frac{n}{2}\rfloor+2\). Figure~\ref{Fig_AFHQSID} shows SID for select orders, comparing StyleGAN2 and StyleGAN3. For positive orders, we flip the sign of SID to maintain consistency with the interpretations developed for the negative order. Across all test cases, we observe that StyleGAN3 outperforms StyleGAN2, as suggested by the FID and KID values~\cite{CleanFID21}. As the order \(m\) reduces, GAN models with lower FID/KID/CSID\(_m\) approach zero more rapidly, which can be used to quantity the relative performance of converged GAN models. For \(m <  \lfloor \frac{n}{2}\rfloor-3\) numerical instability causes SID to approach zero and for \( m> \lfloor \frac{n}{2}\rfloor+2\) numerical instability blows up SID computation. While these experiments serve to demonstrate the feasibility in evaluating pre-trained GAN models with CSID\(_m\), comparisons between Spider DCGAN and the corresponding baselines are provided in Section~\ref{Sec_Exp} and Appendix~\ref{App_SpiderDCGAN} of this {\it Supporting Document}, while comparisons of Spider StyleGANs and baseline StyleGANs on FFHQ and MetFaces is provided in Appendix~\ref{App_ExpStyleGAN}. \par
\begin{figure*}[!b]
\begin{center}
  \begin{tabular}[b]{P{.3\linewidth}P{.3\linewidth}P{.3\linewidth}}
 	Source generator output & Target generator output & SID \\[5pt]
	\includegraphics[width=0.99\linewidth]{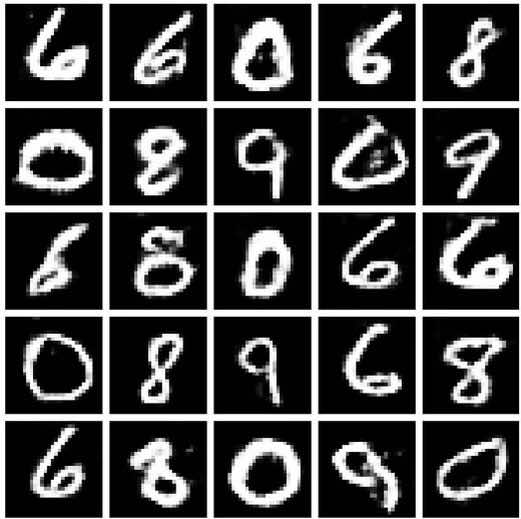} & 
    \includegraphics[width=0.99\linewidth]{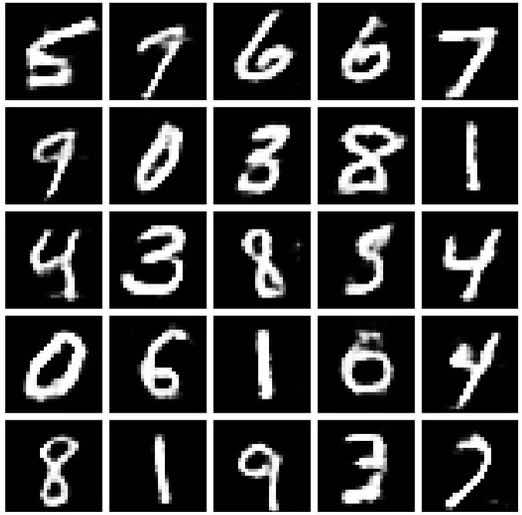} & 
   \includegraphics[width=0.99\linewidth]{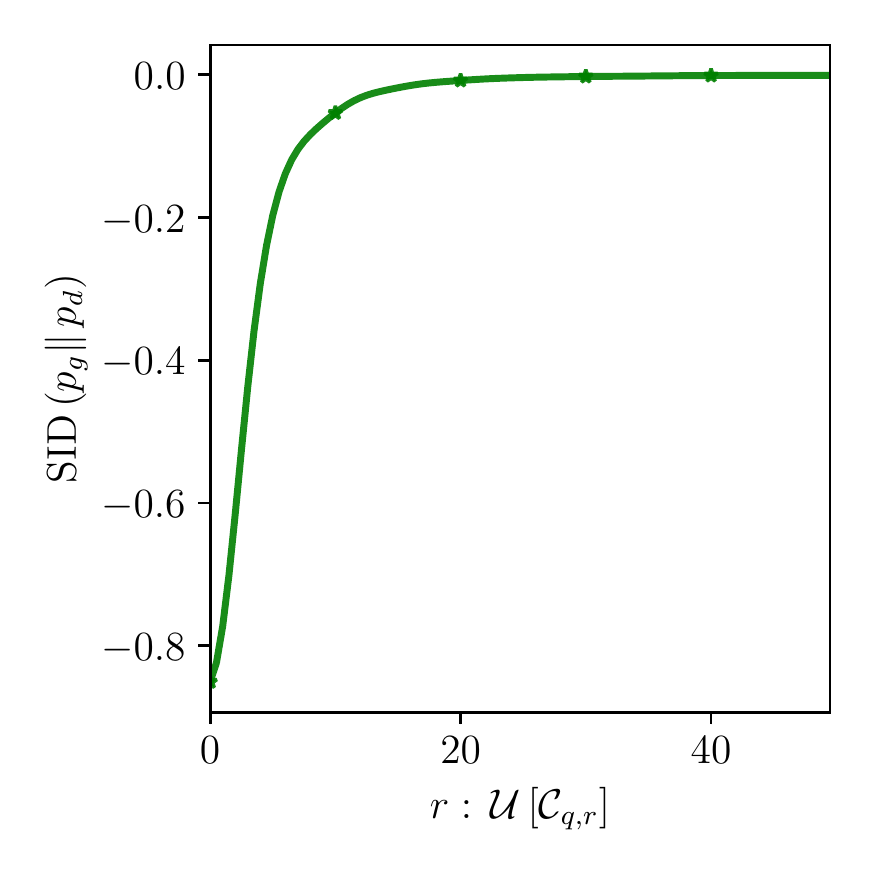}    \\[5pt]\hline&&\\
    \includegraphics[width=0.99\linewidth]{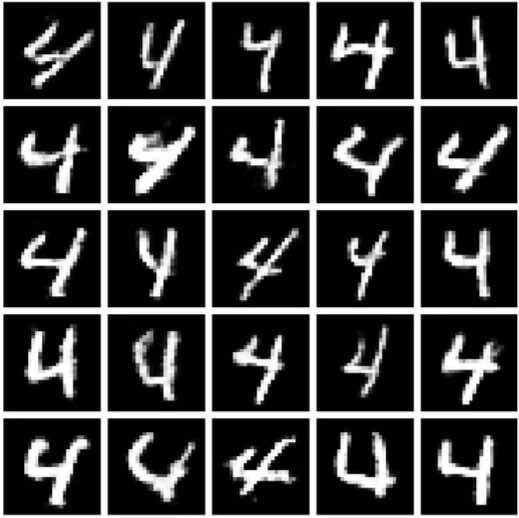} & 
    \includegraphics[width=0.99\linewidth]{Figure10_MNIST_Full.jpeg} &
   \includegraphics[width=0.99\linewidth]{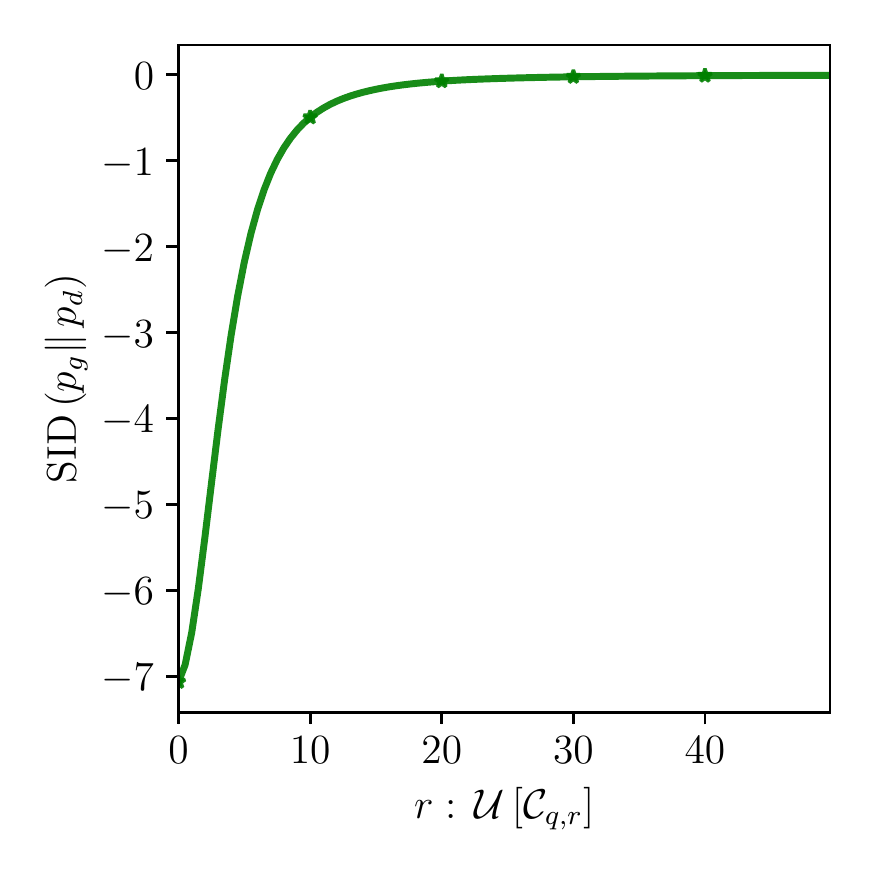}  \\[5pt]\hline&&\\
    \includegraphics[width=0.99\linewidth]{Figure10_MNIST_Four.jpeg} & 
   \includegraphics[width=0.99\linewidth]{Figure10_MNIST_Round.jpeg} & 
   \includegraphics[width=0.99\linewidth]{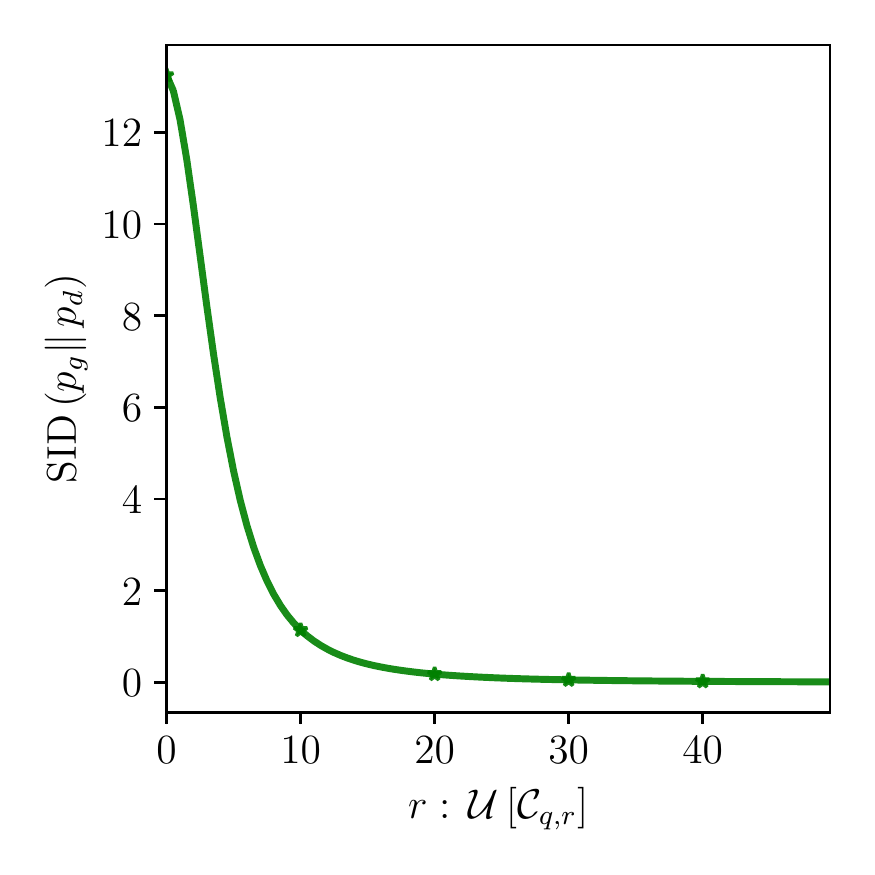}    \\[5pt]\hline&&\\
  \end{tabular} 
\caption[]{SID versus \(r\) when the source and target samples are drawn from GAN generators trained on various subsets of MNIST. When the source generator has mode-collapsed, either to a single digit or a subset of digits, the corresponding SID is negative. When comparing two mode-collapsed generators, the SID will be positive as the distributions of the Inception embeddings are less likely to overlap.}
\label{Fig_MNISTSID}  
\end{center}
\vskip-1em
\end{figure*}

SID can also be used to compare the relative performance of GAN generators. Consider three GANs trained on the MNIST dataset where one generator has learnt the distribution accurately, while the other two have {\it mode-collapsed} on to a subset of the classes (specifically, digits 0,8,6 and 9) or a single class (digit 4) of the target dataset. Figure~\ref{Fig_MNISTSID} presents samples output by these generators and the SID versus \(r\) plot for the corresponding pair of generators. We observe that, when the reference generator has learnt the target accurately, the SID of a test generator's output with respect to the reference will always be negative, as the test generator has less diversity. However, the SID between the output of two  generators that have mode-collapsed would be positive if there is no overlap between the classes they have collapsed to. This could be used to evaluate GANs with ensemble-generators~\cite{DMGAN18}, where each network is trained to learn a different mode/class.

\FloatBarrier
\FloatBarrier

\section{Implementation Details} \label{App_ImpDetails}
We provide details on the experimental setup, evaluation metrics and computational resources employed in the various experiments reported in the {\it Main Manuscript} and this {\it Supporting Document}. 
 \subsection{Experimental Setup} \label{App_ExpSetup}
\noindent {\it {\bfseries Spider DCGAN:}} The experiments presented in Section~\ref{Sec_Exp} consider the DCGAN~\cite{DCGAN} architecture for the generator and discriminator. For the baseline GANs, the parametric input is drawn from \(\mathbb{R}^{100}\). We consider the Gaussian, Gamma~\cite{Gamma18} and non-parametric~\cite{NonPara19} input distributions drawn from \(\mathbb{R}^{100}\) as baselines. In the case of Spider GAN, we conducted experiments by resizing the input data to \(16\times 16\times 3\). To bridge the gap between the two noise variants, we also consider Gaussian noise drawn from \(\mathbb{R}^{16\times16\times3}\) provided as input in a similar fashion to the datasets. We did not observe improvement in performance with higher-resolution images for the input dataset. The images are vectorized and provided as input to the generator. Both Spider DCGAN and the baselines are trained on the Wasserstein GAN~\cite{WGAN17} loss with a stable version of the gradient penalty~\cite{R1R218} enforced only on samples drawn from \(\pd\). The choice was motivated by its successful usage in baseline StyleGAN2 and StyleGAN3 variants.  \par
The networks are trained on batches of 100 samples. The Adam~\cite{Adam} optimizer is used with a learning rate \( \eta = 2\times10^{-4} \), and the exponential decay parameters for the first and second moments are \( \beta_1 = 0.5\) and \(\beta_2 = 0.999 \), respectively. The implementation was carried out using TensorFlow 2.0 \cite{TF}. The networks are trained for \(15\times10^3\) iterations on MNIST and Fashion-MNIST, \(10^4\) iterations on SVHN and CIFAR-10, and \(3\times10^4\) iterations on Celeb-A, Ukiyo-E and Tiny-ImageNet learning tasks. \par

\noindent {\it {\bfseries Spider PGGAN:}} The publicly available PGGAN GitHub repository (URL:~\url{https://github.com/tkarras/progressive_growing_of_gans}) was extended to incorporate the {\it Spider} framework. The implementation was carried out using TensorFlow 2.0 \cite{TF}. The input distributions are drawn from PGGAN models, trained on Tiny-ImageNet  images of resolution \(16\times16\times3\). The input PGGAN was trained for \(12\times10^3\) iterations. Samples drawn from the input PGGAN are resized to  \(14\times14\times3\), vectorized, and provided as input to the cascaded Spider PGGAN layer. \par

\noindent {\it {\bfseries Spider StyleGAN:}} The publicly available, PyTorch 1.10~\cite{PyTorch} based StyleGAN3 GitHub repository (URL:~\url{https://github.com/NVlabs/stylegan3}) was extended to incorporate the {\it Spider} framework, allowing for the implementation of both StyleGAN2, StyleGAN2-ADA and StyleGAN3 variants. The input distributions are drawn from StyleGAN2-ADA models, trained on (a) Tiny-ImageNet images of resolution \(16\times16\times3\); and (b) Images from the AFHQ-Dogs dataset, resized to \(16\times16\times3\).   The input StyleGAN was trained for \(25\times10^3\) iterations in both cases. We considered the following two input transformations to obtained 512-dimensional input vectors: (i)  Samples drawn from the input StyleGAN are averaged across the color channels, resized to  \(16\times32\times1\), vectorized, and provided as input to the cascaded layer; and (ii) Samples drawn from the input StyleGAN are averaged across the color channels, resized to  \(23\times23\times1\) and vectorized. The vectors are truncated to 512 entries, and provided as input to the cascaded stage. We did not observe a significant difference in performance when considering either of the two configurations. As in classical StyleGANs, the cascaded StyleGAN network transforms the input dataset to the latent \(\mathcal{W}\)-space, and subsequently learn the target. Spider StyleGANs are trained with transformation-(i) on FFHQ and AFHQ-Cats data, while transformation-(ii) is used to train the Spider StyleGAN variants on Ukiyo-E faces and MetFaces. 

  \subsection{Evaluation Metrics} \label{App_Metrics}
  
To draw a fair comparison with the baseline approaches, we evaluate various Spider GAN and baseline models in terms of their FID, KID and CSID\(_m\). We also compare the interpolation quality of the networks based on the sharpness of the interpolated images.  \par
\noindent {\it {\bfseries Fr\'echet Inception Distance (FID)}}: Proposed by Heusel {\it et al.}~\cite{TTGAN18}, FID can be used to quantify how {\it real} samples generated by GANs are. FID is computed as the Wasserstein distance between Gaussian distributed embeddings of the generated and target images. To compute the image embedding, we consider the InceptionV3~\cite{InceptionV3} model without the topmost layer, loaded with weights for the ImageNet~\cite{ImageNet09} classification task. Images are resized to \(299\times299\times3\) and given as input to these networks. Grayscale images are replicated across the color channels. FID is computed by assuming a Gaussian prior on the embeddings of real and fake images. The means and covariances are estimated using \(10,000\) samples. The publicly available TensorFlow based {\it Clean-FID} library~\cite{CleanFID21} is used to compute FID. As noted by Parmar {\it et al.}~\cite{CleanFID21}, the Clean-FID is generally found to be a few points higher than those computed through base PyTorch and TensorFlow implementations. Our implementation of the DCGAN baselines~\cite{Gamma18, NonPara19} also exhibit similar offsets between the reported FID and those computed by Clean-FID. However, in our experiments, we were able to reproduce the scores reported in~\cite{CleanFID21} for PGGAN and StyleGAN architectures fairly accurately.  \par

\noindent {\it {\bfseries  Kernel Inception Distance (KID)}}: The kernel inception distance ~\cite{DemistifyMMD18} is an unbiased alternative to FID. The KID computes the squared maximum-mean discrepancy (MMD) between the InceptionV3 embeddings of data in \(\mathbb{R}^n\). The embeddings are computed as in the FID case. The third-order polynomial kernel \( \mathcal{K}(\x,\y) = \left(\frac{1}{n}\x^{\mathrm{T}}\y + 1 \right)^3\) is used to compute the MMD over a batch of 5000 samples. As in the case of FID, to maintain consistency, we use the {\it Clean-FID}~\cite{CleanFID21} library implementation of KID. \par

\noindent {\it {\bfseries  Image Interpolation and Sharpness}}: In order to compare the performance of GAN for generating unseen images, we evaluate the output of the generator when the interpolated points between two input distribution samples are provided to the generator. We use the sharpness metric introduces by Tolstikhin {\it et al.}~\cite{WAE18} in the context of Wasserstein autoencoders. The edge-map of an image is obtained using the Laplacian operator. The average sharpness of the images is then defined as the variance in pixel intensities on the edge-map, averaged over batches of \(50,000\) images. In the case of of baseline GAN, the inputs are interpolated points between random samples drawn from the parametric noise distribution, while in the case of Spider GAN, the interpolation between two images from the input dataset are fed to the generator.  

\subsection{Computational Resources} 
All experiments on low-resolution \((\leq 32\times32\times3)\) images with the DCGAN architecture were conducted on workstations with one of two configuration: (a) \(4\times\) NVIDIA 2080Ti GPUs with 11 GB visual RAM (VRAM) each, and 256 GB system RAM; and (b) \(2\times\) NVIDIA 3090 GPUs with 24 GB VRAM and 256 GB system RAM.  The high-resolution experimentation involving PGGAN or StyleGAN was carried out on workstations with one of the two configurations: (i) NVIDIA DGX with \(8\times\) Tesla V100 GPUs with 32 GB VRAM each, and 512 GB system RAM; and (ii)  \(8\times\) NVIDIA A6000 GPUs with 48 GB VRAM each, and 512 GB system RAM. The memory requirements and training times for StyleGAN and PGGAN variants are on par with training times reported for the baselines~\cite{PGGAN18,StyleGAN321}. 


\begin{figure*}[!thb]
\begin{center}
  \begin{tabular}[b]{P{.47\linewidth}|P{.47\linewidth}}
  DCGAN  & CAE \\[1pt]
      \includegraphics[width=0.99\linewidth]{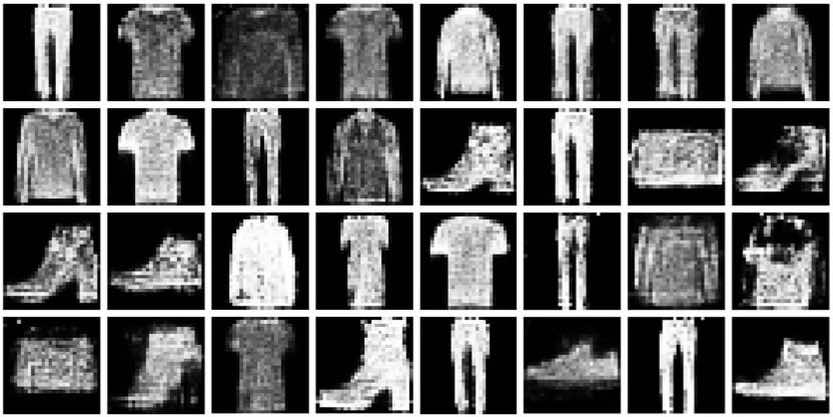} & 
    \includegraphics[width=0.99\linewidth]{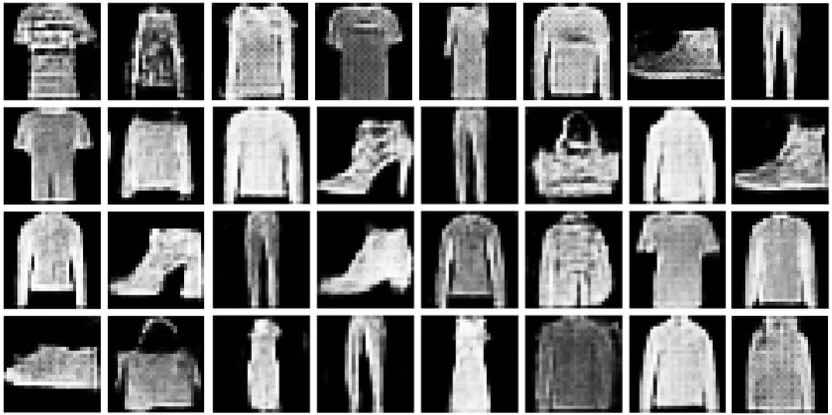}  \\[5pt]
    \includegraphics[width=0.99\linewidth]{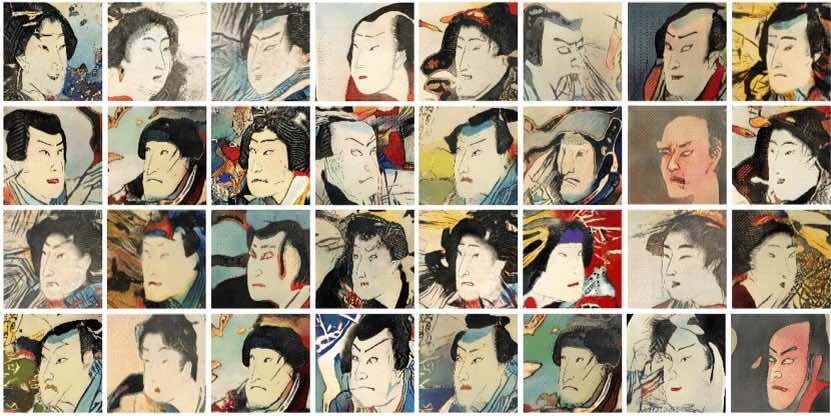} & 
    \includegraphics[width=0.99\linewidth]{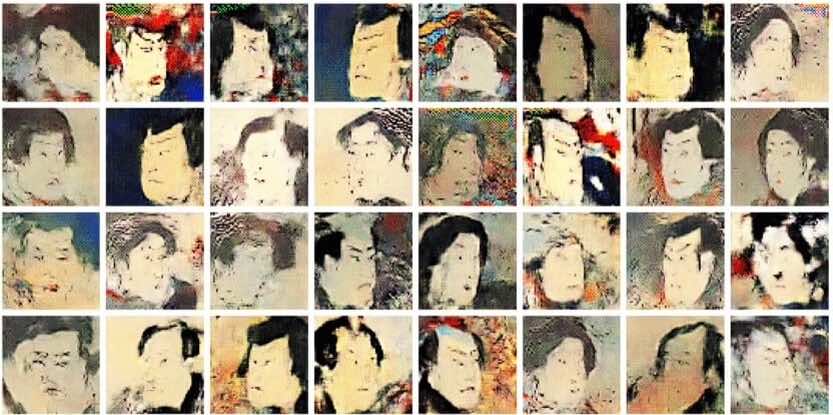}  \\[5pt]
  \end{tabular} 
\caption[]{Images generated by Spider GAN on Fashion-MNIST and Ukiyo-E Faces, given the friendliest neighbor input as identified by SID. Both CAE and DCGAN result in images of comparable visual quality on Fashion-MNIST. However, for high-resolution image generation on 256-dimensional Ukiyo-E Faces, the fully convolutional structure of the CAE generator result in images of poorer visual quality than those generated by DCGAN.}
\label{Fig_ArchCompare}  
\end{center}
\vskip-1em
\end{figure*}

\newpage

\begin{figure*}[!t]
\begin{center}
  \begin{tabular}[b]{P{.3\linewidth}P{.3\linewidth}P{.29\linewidth}}
   \multicolumn{ 3}{c}{\includegraphics[width=0.97\linewidth]{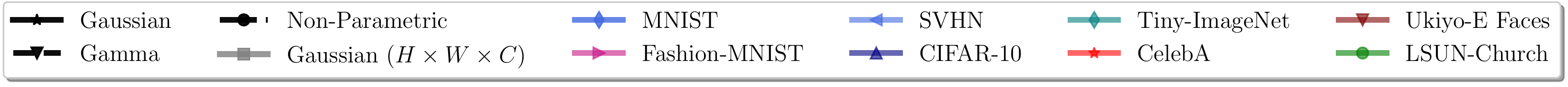} } \\
       (a) MNIST  & (b) CIFAR-10   & (c) Ukiyo-E\\[1pt]
    \includegraphics[width=1.0\linewidth]{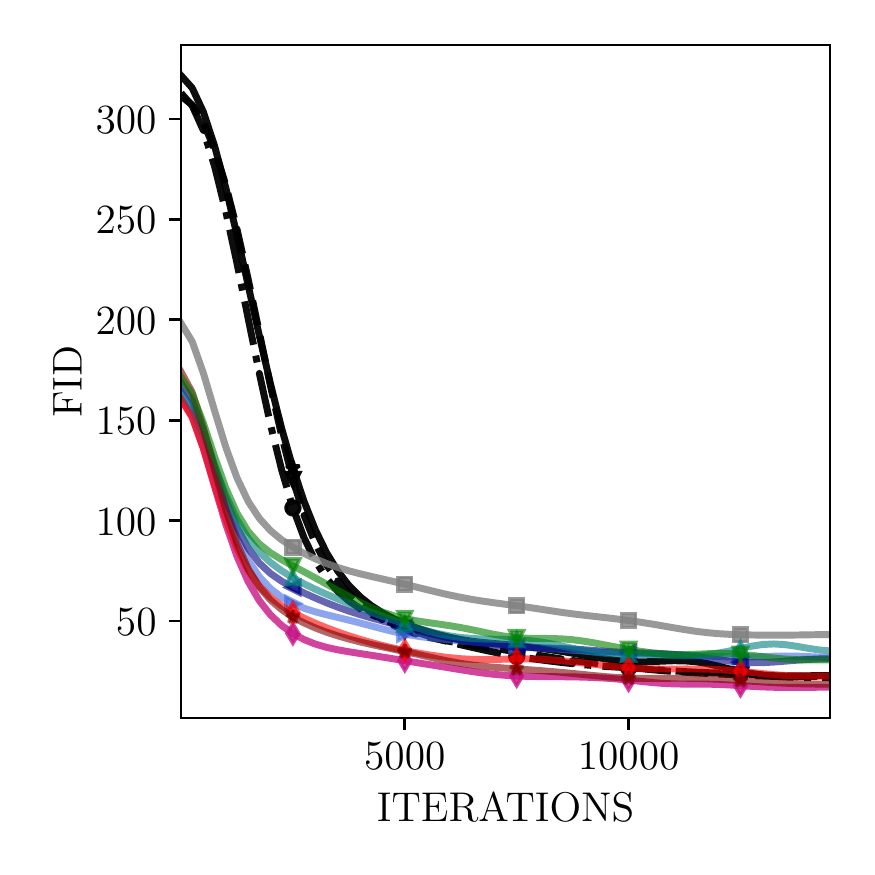} & 
    \includegraphics[width=1.0\linewidth]{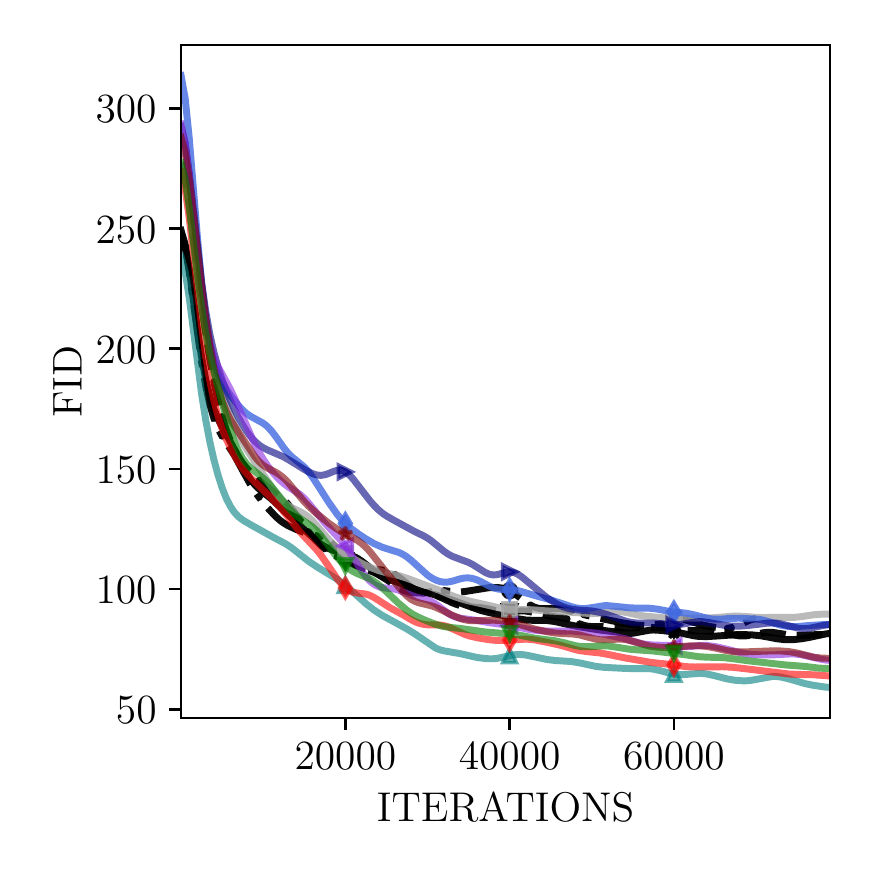} & 
    \includegraphics[width=1.0\linewidth]{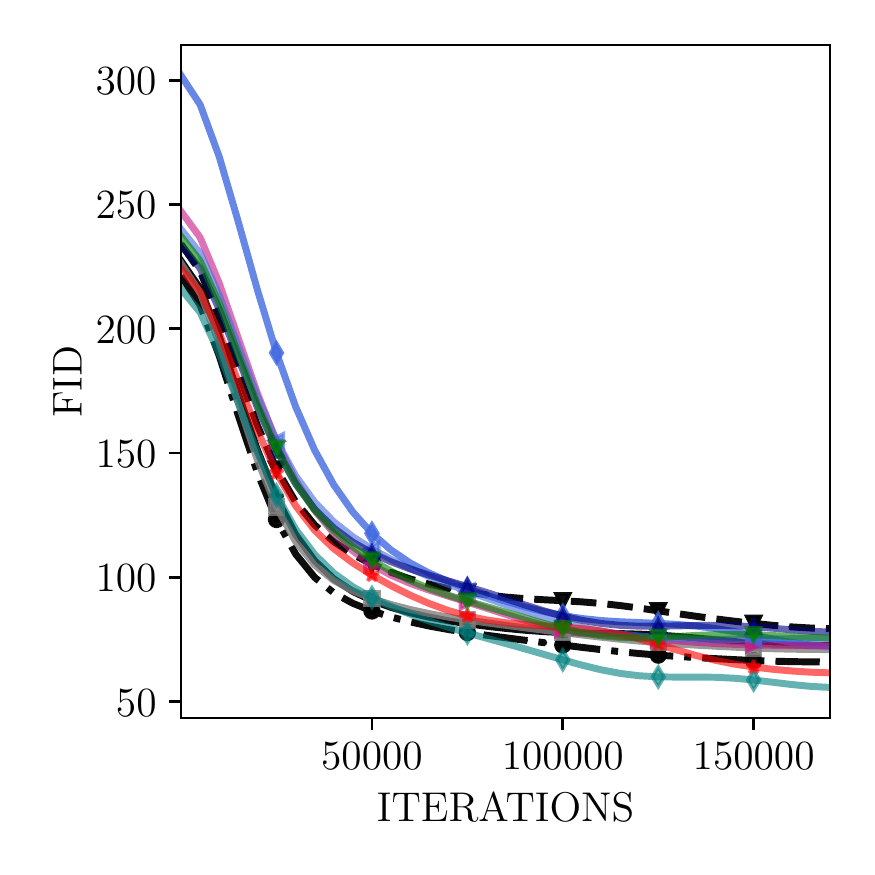}  \\[-6pt]
  \end{tabular} 
\caption[]{FID versus iterations for training baseline and Spider GAN variants. Spider GAN trained with the friendliest neighbor identified in Section~\ref{Sec_FriendlyNb} (of the {\it Main Manuscript}) result in the best (lowest) FID scores. On MNIST, Spider DCGAN approaches converge an order faster than the baseline counterparts.}
\label{Fig_FID_vs_Iters_All}  
\end{center}
\vskip-1.1em
\end{figure*}

\begin{table*}[t!]
\fontsize{7.5}{12}\selectfont
\begin{center}
\caption{Comparison of FID, KID and CSID\(_m\) for the Spider DCGAN and baseline variants on Fashion-MNIST, SVHN, Tiny-ImageNet, and CelebA datasets. Spider DCGANs with {\it friendly neighborhood} inputs outperform the baselines with parametric and non-parametric priors. The performance of Spider DCGAN with MNIST or Fashion-MNIST as the input is sub par when the target is a color-image dataset. } \label{Table_FIDKID_Rest}  \vskip-0.05in
\begin{tabular}{P{0.05cm}P{2.7cm}||P{0.65cm}|P{0.7cm}|P{0.65cm}||P{0.65cm}|P{0.7cm}|P{0.65cm}||P{0.65cm}|P{0.7cm}|P{0.65cm}||P{0.65cm}|P{0.7cm}|P{0.65cm}}
\toprule 
\multicolumn{2}{c||}{\multirow{2}{*}{Input Distribution}}&\multicolumn{3}{c||}{Fashion-MNIST} &\multicolumn{3}{c||}{SVHN} & \multicolumn{3}{c||}{Tiny-ImageNet}& \multicolumn{3}{c}{CelebA} \\\cline{3-14} 
\multirow{4}{*}{\rotatebox{90}{{\footnotesize\uline{Baselines}\quad \enskip \quad } }}
&& FID & KID  & CSID\(_m\) & FID & KID   & CSID\(_m\) & FID & KID   & CSID\(_m\) & FID & KID  & CSID\(_m\)   \\
\hline\hline
&Gaussian \((\mathbb{R}^{100})\)				& 76.60 & 0.0557 & 22.24 & 135.4 & 0.1245  & 30.02 & 89.94 & 0.0657 & 18.06 & 50.32 & 0.0554  & 24.31\\
&Gamma 	 \((\mathbb{R}^{100})\)		    		& 65.36 & 0.0513 & \uline{19.72} & 130.8 & 0.1181 & 27.13 &83.33 & 0.0536 & 14.63 & 40.69 & 0.0544 & 20.98 \\
&Non-Parametric  \((\mathbb{R}^{100})\)			& \uline{62.42} & \uline{0.0426} & 21.96 & 107.2 & 0.1053 & 33.52 &\uline{82.37} & \uline{0.0579} & 13.25 & 40.41 & 0.0543 & 72.18 \\
&Gaussian  \((\mathbb{R}^{H\times W\times C})\)	& 119.2 & 0.0905 & 28.96 & 113.7 & 0.1121 & 31.45 & 103.0 & 0.0844 & 15.62 & 83.61 & 0.0912 & 113.4 \\[1pt] \hline
\multirow{8}{*}{\rotatebox{90}{{\footnotesize\uline{\quad\quad\quad Spider GAN \quad\quad\quad} } }}
&MNIST 			   		 & {\bf 56.59} & {\bf 0.0387} & {\bf 18.50} & {\bf 95.71} & {\bf 0.0817} & {\bf 20.62} & 96.91 & 0.0669 & 14.95 & 40.78 & 0.0595 & 32.70 \\
&Fashion MNIST  			& -- & -- &-- & 115.0 & 0.1096 & 32.57 &108.8 & 0.0667 & \uline{13.06} & \uline{35.18} & 0.0574 & 23.98 \\
&SVHN  			    		& 79.14 & 0.0526 & 24.67 & -- & -- & -- & 98.11 & 0.0655 & 15.62 & 40.27 & 0.0575 & \uline{20.64} \\
&CIFAR-10 				& 92.60 & 0.0658 & 30.21 & \uline{101.8} & \uline{0.0998} & 32.40 &98.22 & 0.0642 & 17.90 & 36.16 & \uline{0.0508} & 22.16 \\
&TinyImageNet 			& 130.5 & 0.0883 & 22.26 & 111.7 & 0.1082 & 31.77 &-- & -- & -- & {\bfseries 29.47} & {\bfseries 0.0468} & {\bf 18.16} \\
&CelebA 			    		& 81.38 & 0.0604 & 24.73 & 108.9 & 0.1029 & \uline{22.77}  &{\bfseries 75.68} & {\bfseries 0.0511} & {\bf 12.42} & -- & -- & -- \\
&Ukiyo-E 			    		& 66.90 & 0.0475 & 23.29 & 114.8 & 0.1145 & 38.28 & 88.51 & 0.0612 & 16.01 & 39.41 & 0.0630 & 28.23 \\
&LSUN-Churches 			& 102.9 & 0.0774 & 33.87 & 106.8 & 0.1020 & 26.52 & 92.86 & 0.0697 & 15.98 & 53.01 & 0.0636 & 25.72 \\
\bottomrule
\end{tabular}
\end{center}
\vskip-2.05em
\end{table*}

 \section{Additional Experimentation on Spider GAN} 
 

We now discuss additional experimental results and ablation studies on various {\it Spider GAN} flavors presented in the {\it Main Manuscript}. One could also extend the {\it Spider} philosophy to VQGAN~\cite{VQGAN21,ImpVQGAN22} or diverse class-conditional models~\cite{BIGGAN18,RebootACGAN21}
 
 \subsection{Exploring Generator Architectures} \label{App_CAE}
  We now discuss the choice of the generator architecture in Spider GAN (cf. Section~\ref{Sec_Exp}). We consider two network architectures: 
\begin{itemize}
\itemsep-0.25em
\item {\bf DCGAN}: We consider standard DCGAN where the images from the friendly neighborhood are resized, vectorized, and provided as input to generator as described in Appendix~\ref{App_ExpSetup}.
\item {\bf  Convolutional autoencoder (CAE)}: In this setup, the images are resized to \(16\times 16\times 3\) and provided as input to convolutional layers to learn a low-dimensional latent representation.  The output image is generated by deconvolution layers.
\end{itemize}

The number of trainable parameters are fewer for the CAE architecture than the DCGAN approach in both cases. Figure~\ref{Fig_ArchCompare} shows the output images generated by these approaches considering the friendliest neighbor (as suggested by Tables~\ref{DatasetSinID}-\ref{DatasetSSID}) provided as input when learning the Fashion-MNIST and Ukiyo-E Faces datasets. We observe that the CAE based Spider GAN outperforms the DCGAN approach on Fashion-MNIST. However, on higher resolution images, multiple visual artifacts were found as a consequence of the fully convolutional architecture. We observed similar degradation in image quality when training Spider GAN with CAE on other high-resolution datasets such as CelebA. We therefore consider the DCGAN approach in the experiments presented in Section~\ref{Sec_Exp} and Appendix~\ref{App_SpiderDCGAN}.

\subsection{Additional Experiments on Spider DCGAN} \label{App_SpiderDCGAN}
\begin{figure*}[!thb]
\begin{center}
  \begin{tabular}[b]{P{.38\linewidth}|P{.38\linewidth}}
      \includegraphics[width=0.99\linewidth]{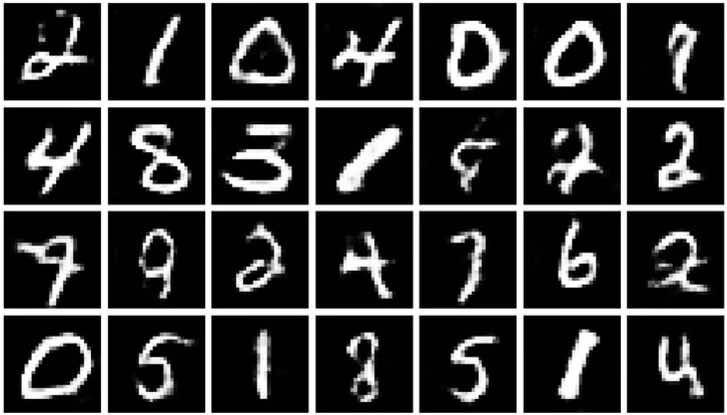} & 
    \includegraphics[width=0.99\linewidth]{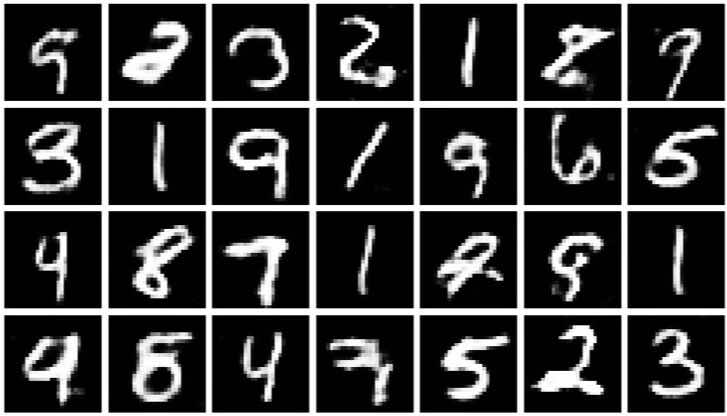}  \\[-1pt]
    (a) Gaussian input & (b) Gamma input \\[1pt] \hline
     \includegraphics[width=0.99\linewidth]{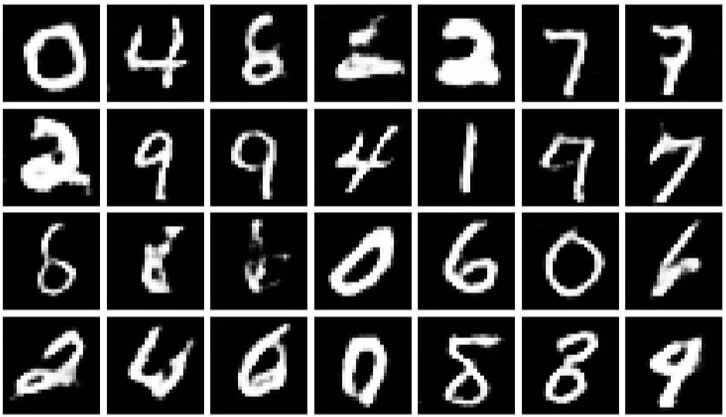} & 
    \includegraphics[width=0.99\linewidth]{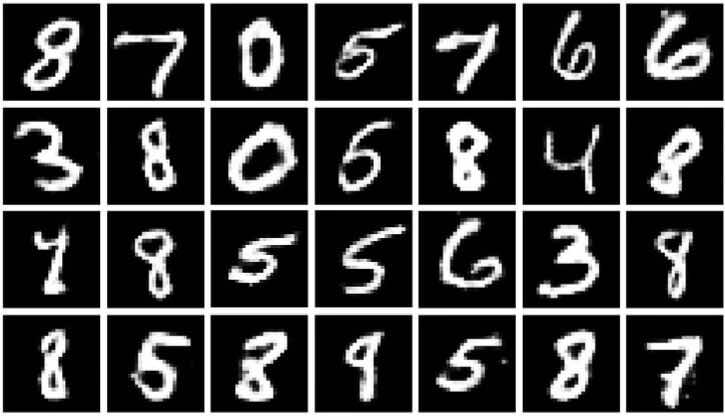}  \\[-1pt]
    (c) Non-Parametric input & (d) Fashion-MNIST input \\[1pt] \hline
     \includegraphics[width=0.99\linewidth]{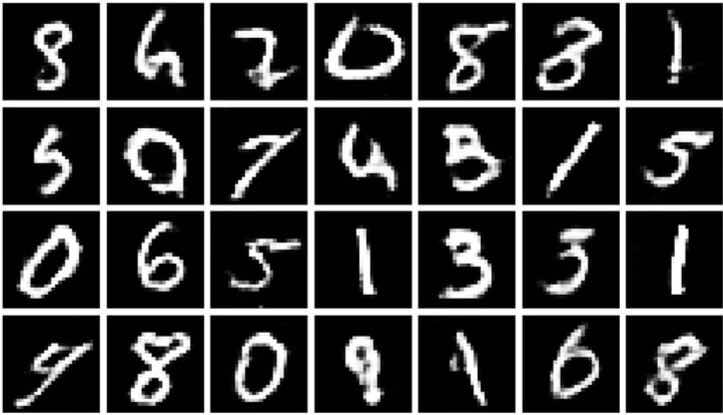} & 
    \includegraphics[width=0.99\linewidth]{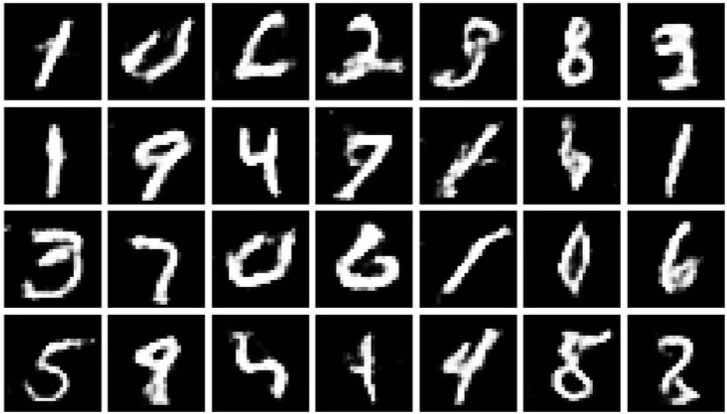}  \\[-1pt]
    (e) SVHN input & (f) CIFAR-10 input \\[1pt] \hline
     \includegraphics[width=0.99\linewidth]{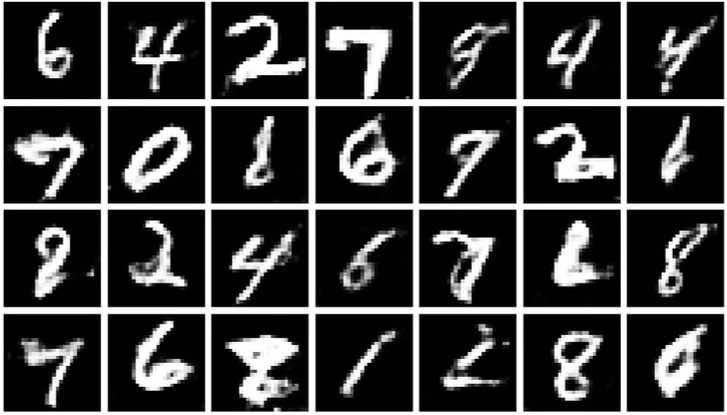} & 
    \includegraphics[width=0.99\linewidth]{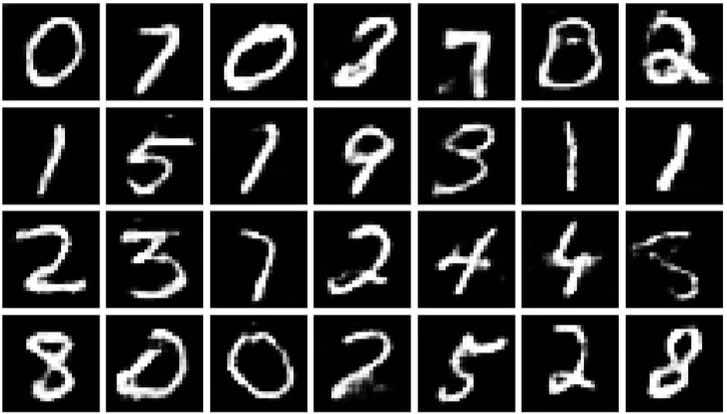}  \\[-1pt]
    (g) Tiny-ImageNet input & (h) CelebA input \\[1pt] \hline
     \includegraphics[width=0.99\linewidth]{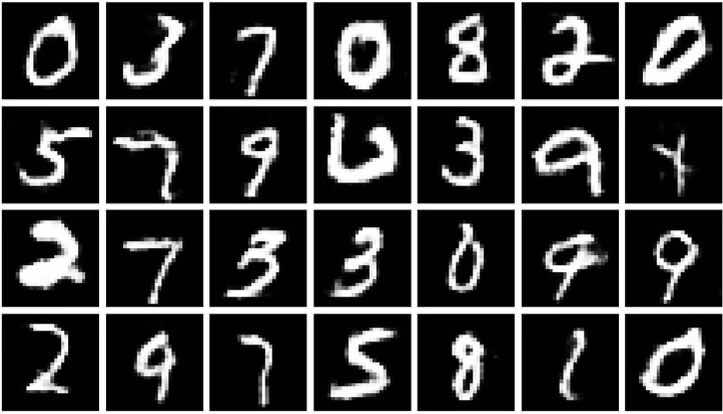} & 
    \includegraphics[width=0.99\linewidth]{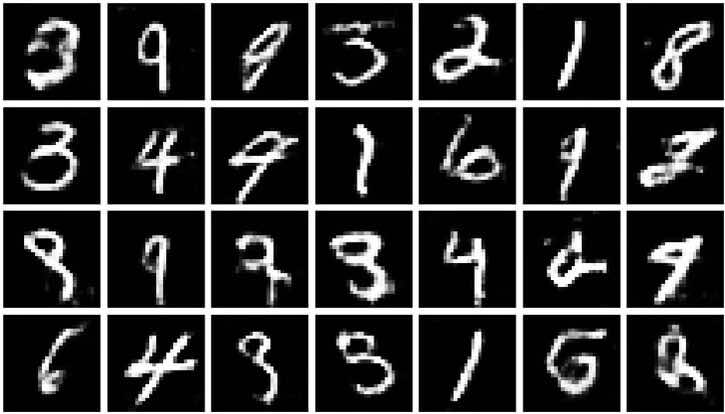}  \\[-1pt]
    (i) Ukiyo-E Faces input & (j) LSUN-Churches input \\[1pt] \hline
  \end{tabular} 
\caption[]{Images generated by the baseline GAN and Spider GAN for various input distributions, with MNIST being the target. Spider GAN trained with Fashion-MNIST input (the friendliest neighbor of MNIST as identified by SID) generates sharper output images.}
\label{Fig_RandMNIST}  
\end{center}
\vskip-1em
\end{figure*}

\begin{figure*}[!thb]
\begin{center}
  \begin{tabular}[b]{P{.38\linewidth}|P{.38\linewidth}}
      \includegraphics[width=0.99\linewidth]{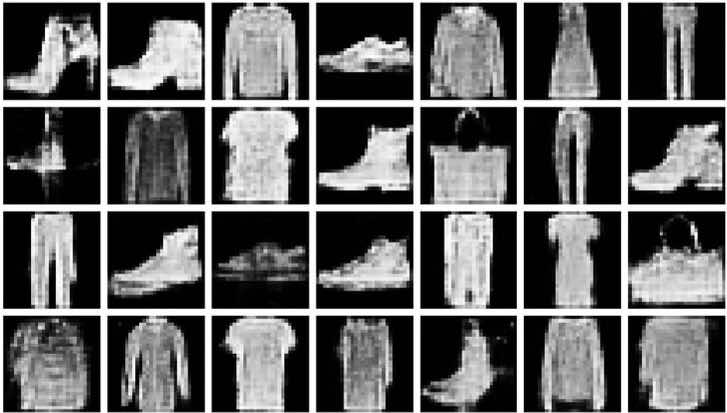} & 
    \includegraphics[width=0.99\linewidth]{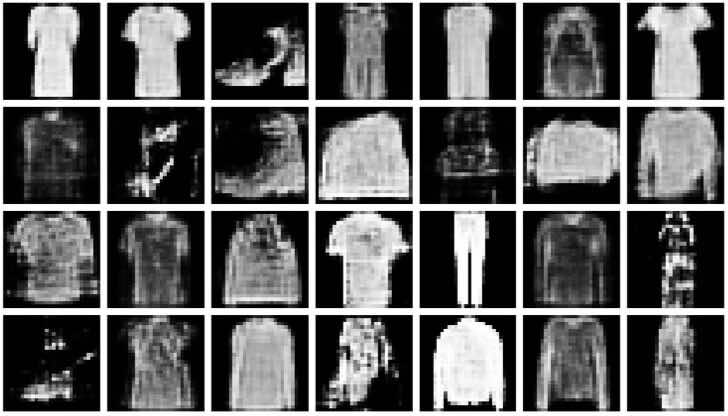}  \\[-1pt]
    (a) Gaussian input & (b) Gamma input \\[1pt] \hline
     \includegraphics[width=0.99\linewidth]{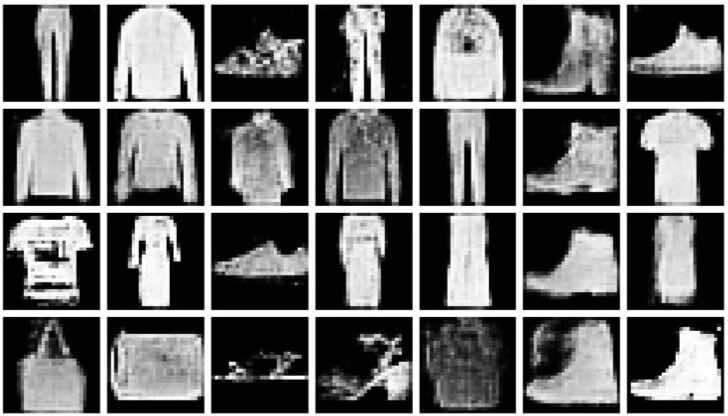} & 
    \includegraphics[width=0.99\linewidth]{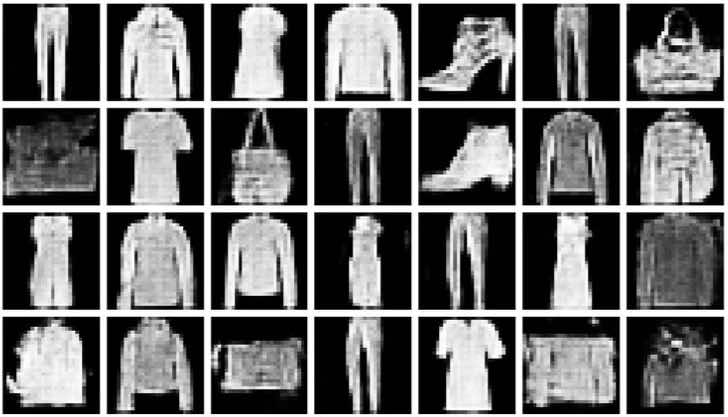}  \\[-1pt]
    (c) Non-Parametric input & (d) MNIST input \\[1pt] \hline
     \includegraphics[width=0.99\linewidth]{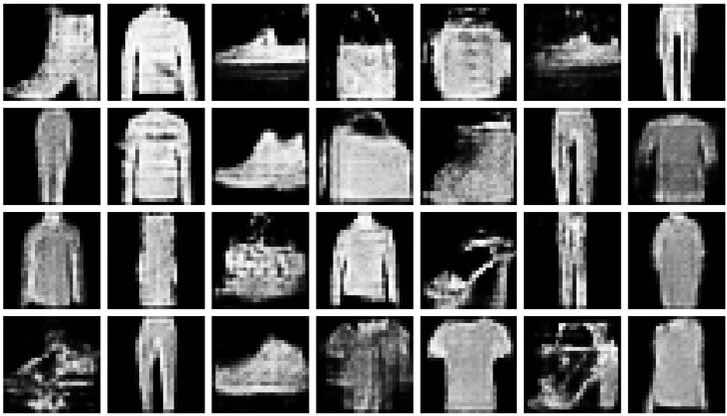} & 
    \includegraphics[width=0.99\linewidth]{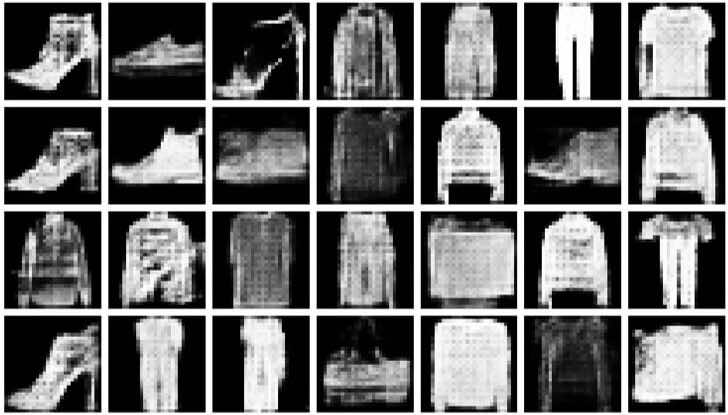}  \\[-1pt]
    (e) SVHN input & (f) CIFAR-10 input \\[1pt] \hline
     \includegraphics[width=0.99\linewidth]{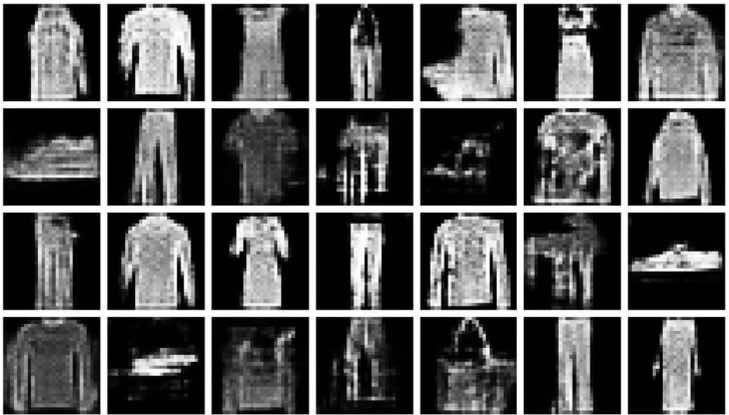} & 
    \includegraphics[width=0.99\linewidth]{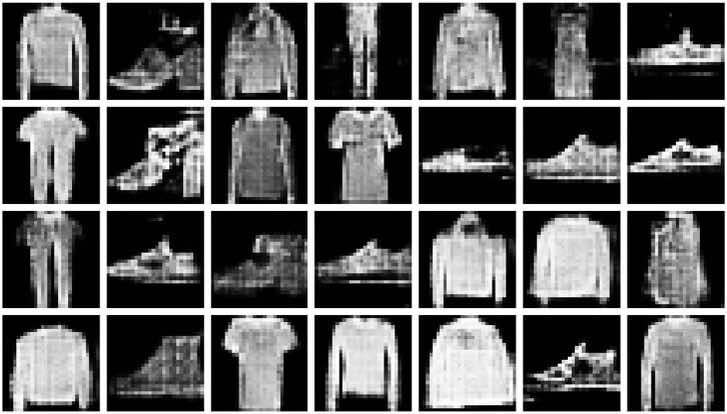}  \\[-1pt]
    (g) Tiny-ImageNet input & (h) CelebA input \\[1pt] \hline
     \includegraphics[width=0.99\linewidth]{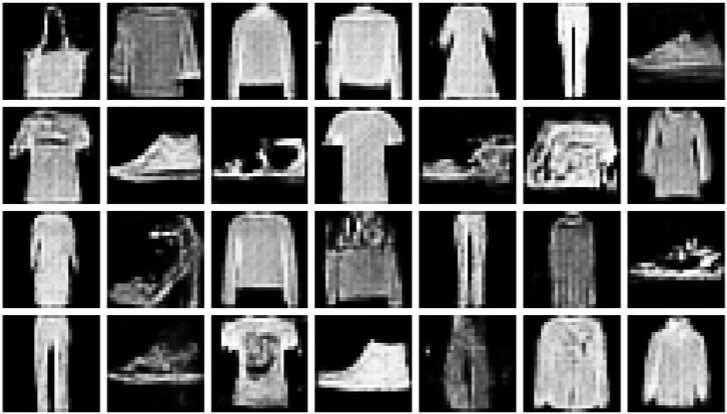} & 
    \includegraphics[width=0.99\linewidth]{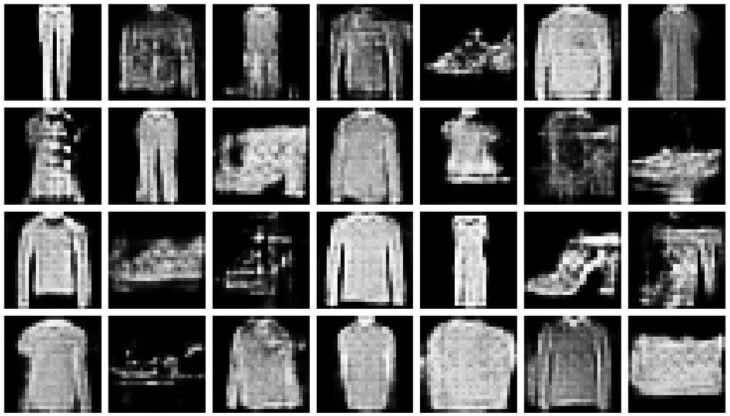}  \\[-1pt]
    (i) Ukiyo-E Faces input & (j) LSUN-Churches input \\[1pt] \hline
  \end{tabular} 
\caption[]{Images generated by the baseline GAN and Spider GAN for various input distributions, with Fashion-MNIST chosen as the target. A poor choice of the input distribution results in a suboptimal generator that outputs low-quality images. For instance, the output generated for inputs coming from CelebA or a non-parametric distribution.}
\label{Fig_RandFMNIST}  
\end{center}
\vskip-1em
\end{figure*}

\begin{figure*}[!thb]
\begin{center}
  \begin{tabular}[b]{P{.38\linewidth}|P{.38\linewidth}}
      \includegraphics[width=0.99\linewidth]{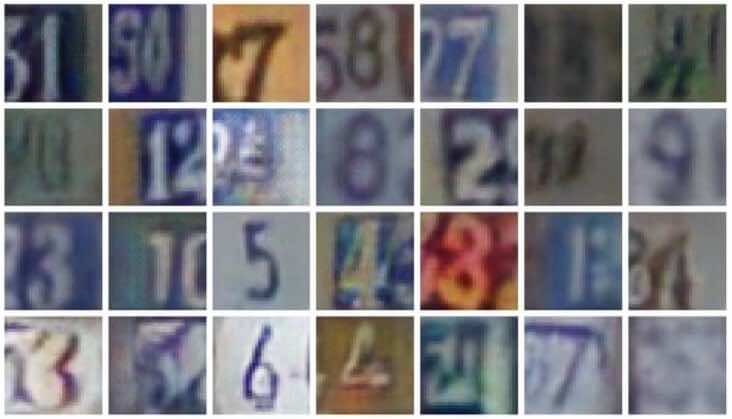} & 
    \includegraphics[width=0.99\linewidth]{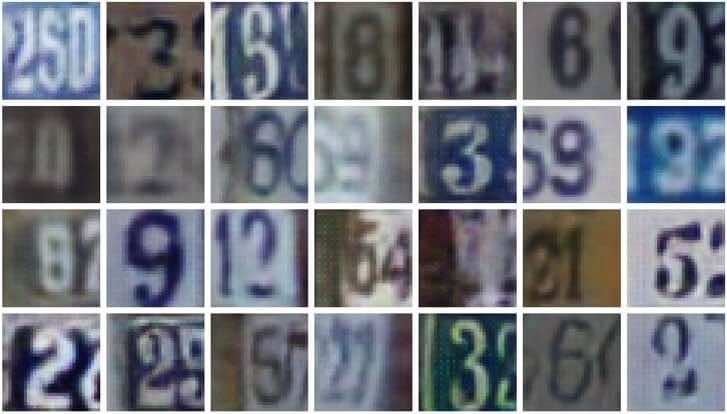}  \\[-1pt]
    (a) Gaussian input & (b) Gamma input \\[1pt] \hline
     \includegraphics[width=0.99\linewidth]{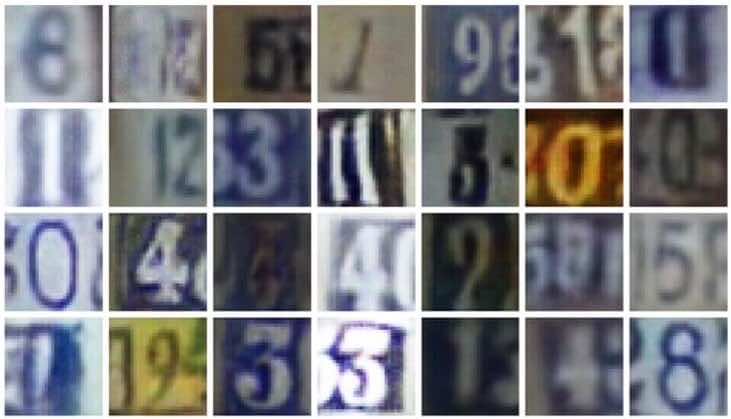} & 
    \includegraphics[width=0.99\linewidth]{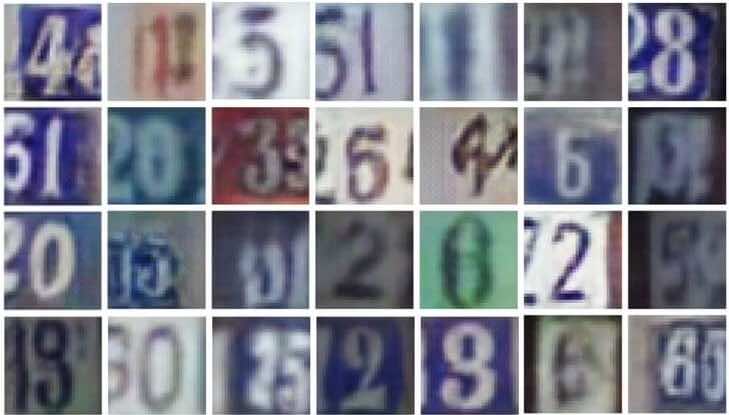}  \\[-1pt]
    (c) Non-Parametric input & (d) MNIST input \\[1pt] \hline
     \includegraphics[width=0.99\linewidth]{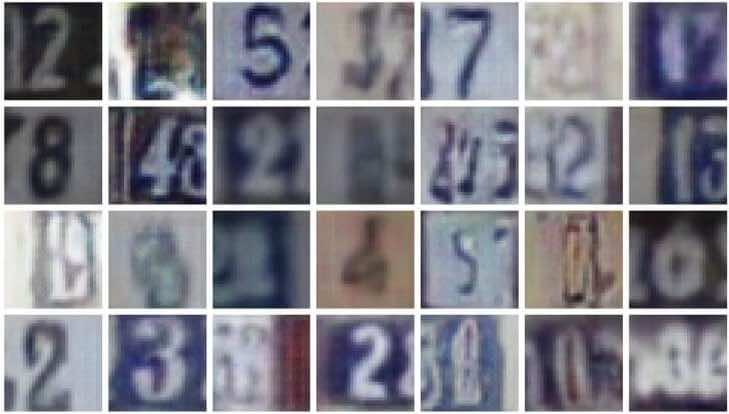} & 
    \includegraphics[width=0.99\linewidth]{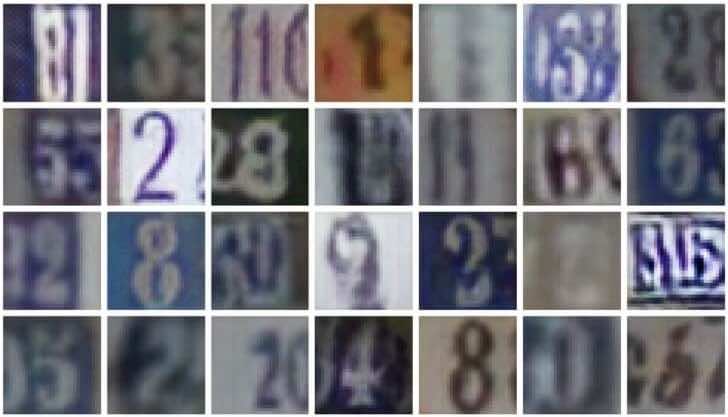}  \\[-1pt]
    (e) Fashion-MNIST input & (f) CIFAR-10 input \\[1pt] \hline
     \includegraphics[width=0.99\linewidth]{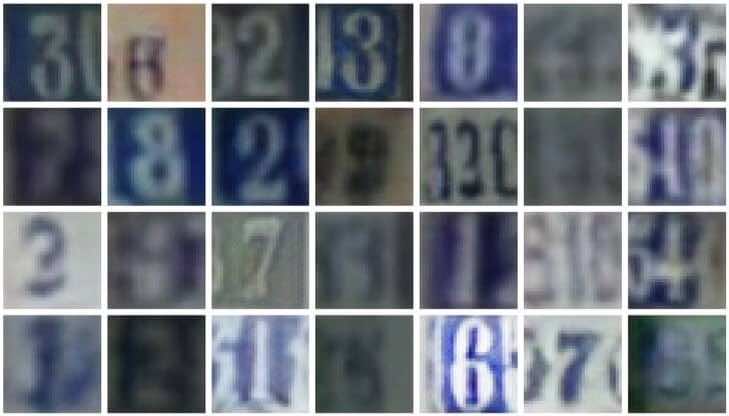} & 
    \includegraphics[width=0.99\linewidth]{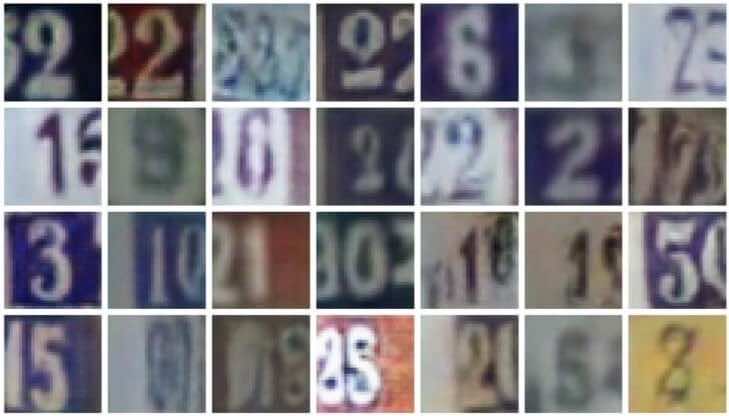}  \\[-1pt]
    (g) Tiny-ImageNet input & (h) CelebA input \\[1pt] \hline
     \includegraphics[width=0.99\linewidth]{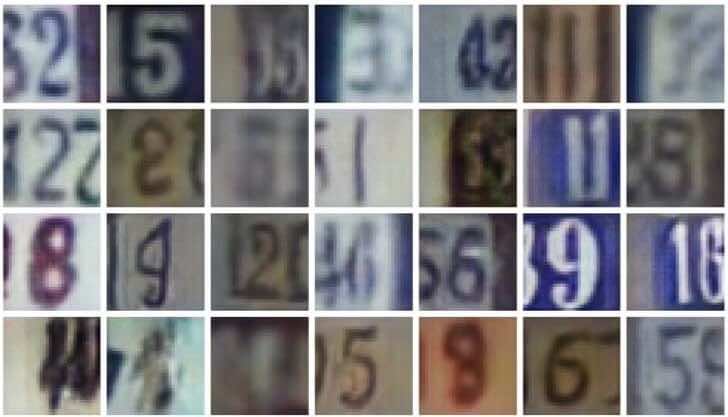} & 
    \includegraphics[width=0.99\linewidth]{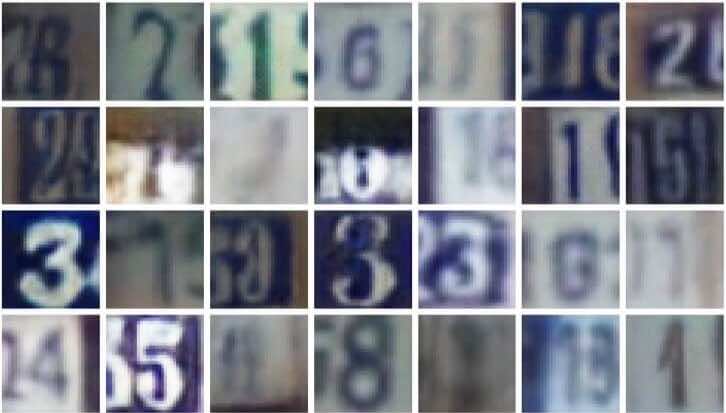}  \\[-1pt]
    (i) Ukiyo-E Faces input & (j) LSUN-Churches input \\[1pt] \hline
  \end{tabular} 
\caption[]{Images generated by the baseline GAN and Spider GAN for various input distributions, when trained with SVHN as the target. A poor choice of the input distribution results in low-quality images output by the generator.}
\label{Fig_RandSVHN}  
\end{center}
\vskip-1em
\end{figure*}

\begin{figure*}[!thb]
\begin{center}
  \begin{tabular}[b]{P{.38\linewidth}|P{.38\linewidth}}
      \includegraphics[width=0.99\linewidth]{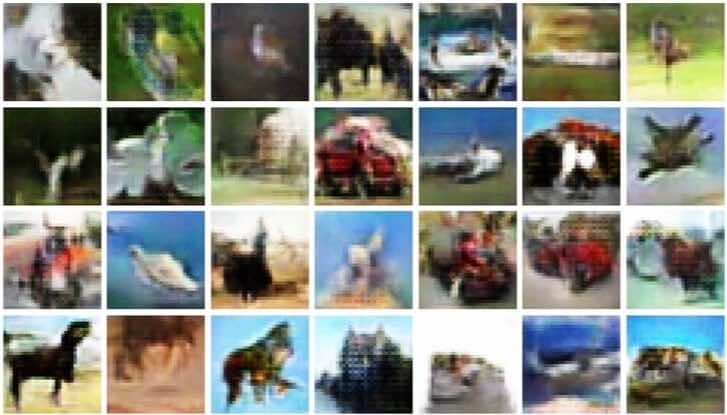} & 
    \includegraphics[width=0.99\linewidth]{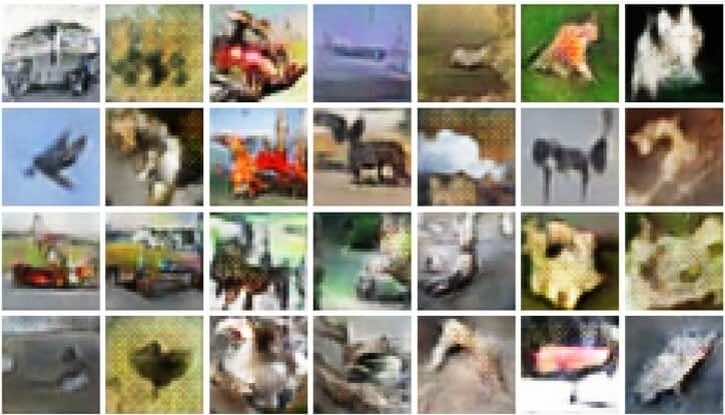}  \\[-1pt]
    (a) Gaussian input & (b) Gamma input \\[1pt] \hline
     \includegraphics[width=0.99\linewidth]{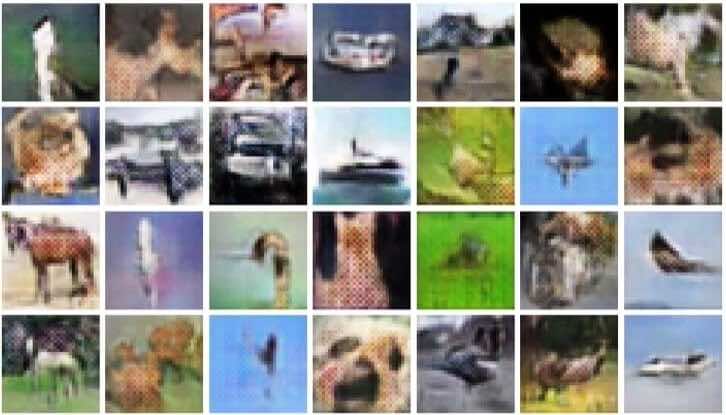} & 
    \includegraphics[width=0.99\linewidth]{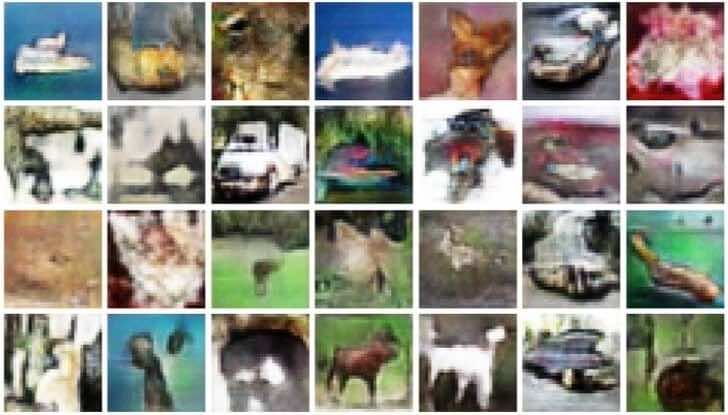}  \\[-1pt]
    (c) Non-Parametric input & (d) MNIST input \\[1pt] \hline
     \includegraphics[width=0.99\linewidth]{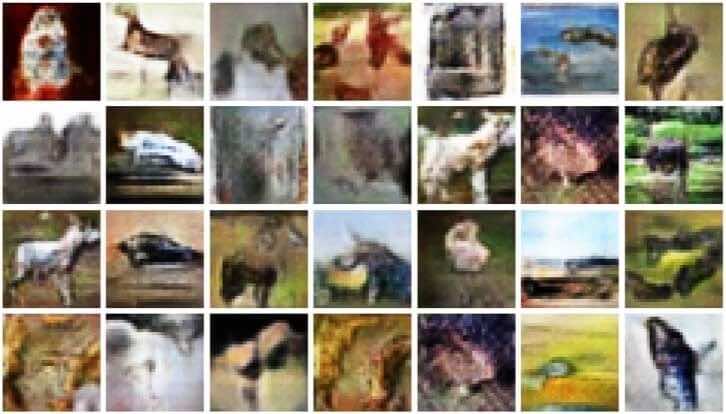} & 
    \includegraphics[width=0.99\linewidth]{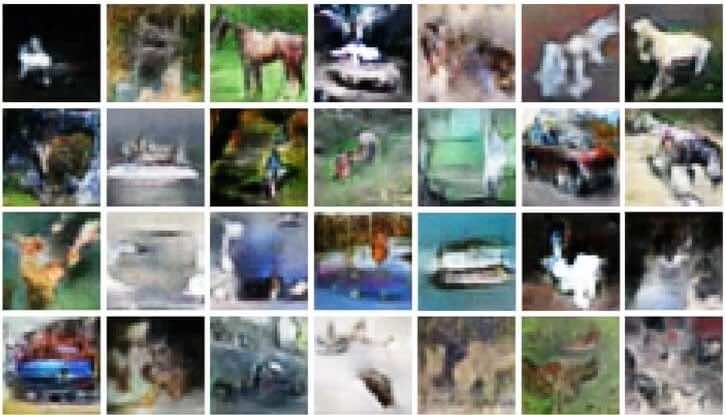}  \\[-1pt]
    (e) Fashion-MNIST input & (f) SVHN input \\[1pt] \hline
     \includegraphics[width=0.99\linewidth]{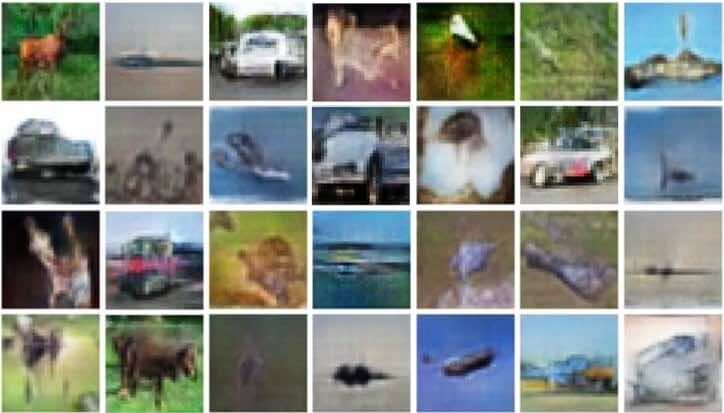} & 
    \includegraphics[width=0.99\linewidth]{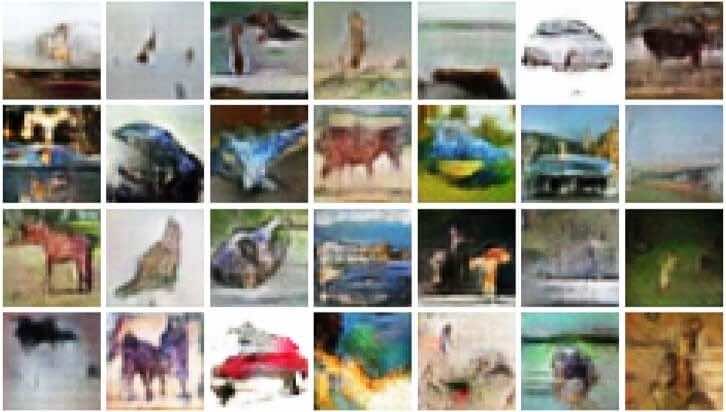}  \\[-1pt]
    (g) Tiny-ImageNet input & (h) CelebA input \\[1pt] \hline
     \includegraphics[width=0.99\linewidth]{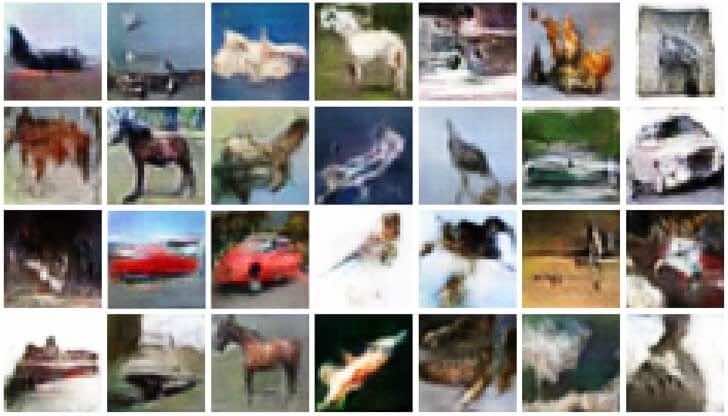} & 
    \includegraphics[width=0.99\linewidth]{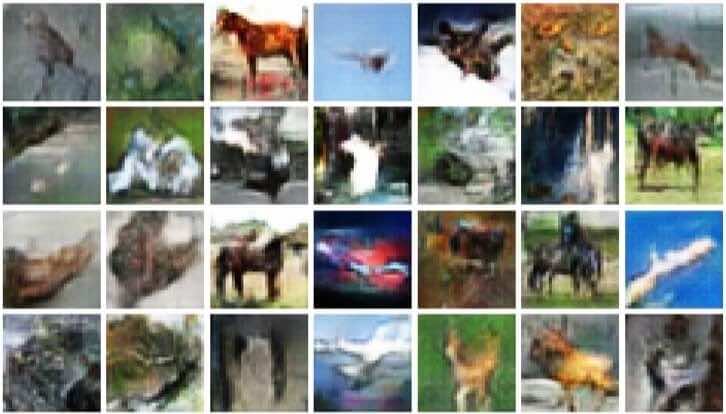}  \\[-1pt]
    (i) Ukiyo-E Faces input & (j) LSUN-Churches input \\[1pt] \hline
  \end{tabular} 
\caption[]{Images generated by the baseline GAN and Spider GAN with CIFAR-10 as the target, for various input distributions. While some classes, such as the {\it horse}, {\it car} or {\it boat} are well generated by all GAN, neither the baseline GANs nor the Spider GANs are able to reliably learn all the classes in CIFAR-10.}
\label{Fig_RandC10}  
\end{center}
\vskip-1em
\end{figure*}

\begin{figure*}[!thb]
\begin{center}
  \begin{tabular}[b]{P{.38\linewidth}|P{.38\linewidth}}
      \includegraphics[width=0.99\linewidth]{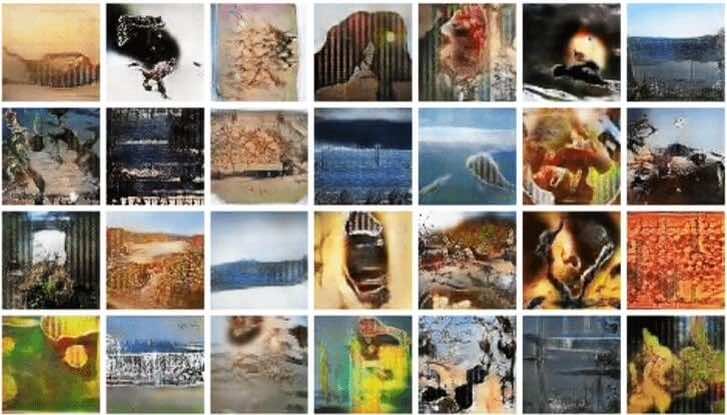} & 
    \includegraphics[width=0.99\linewidth]{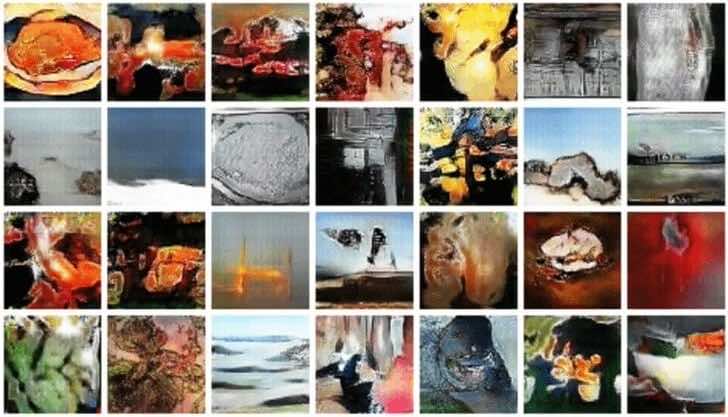}  \\[-1pt]
    (a) Gaussian input & (b) Gamma input \\[1pt] \hline
     \includegraphics[width=0.99\linewidth]{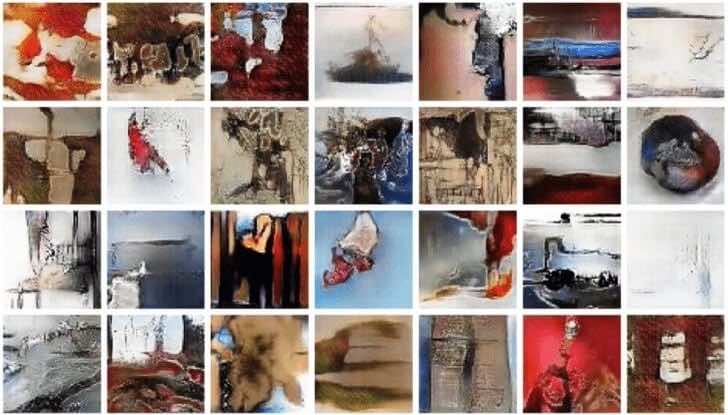} & 
    \includegraphics[width=0.99\linewidth]{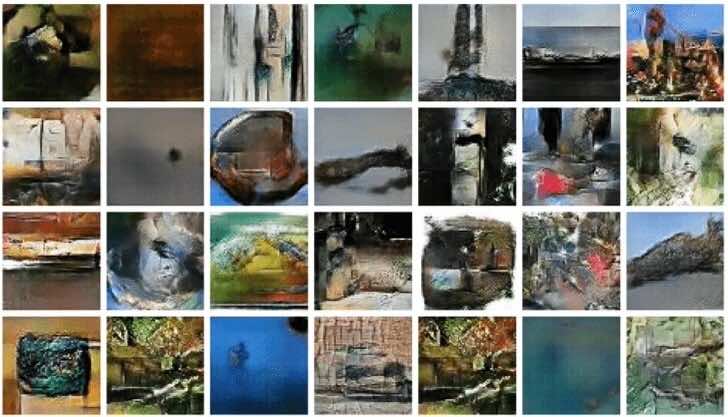}  \\[-1pt]
    (c) Non-Parametric input & (d) MNIST input \\[1pt] \hline
     \includegraphics[width=0.99\linewidth]{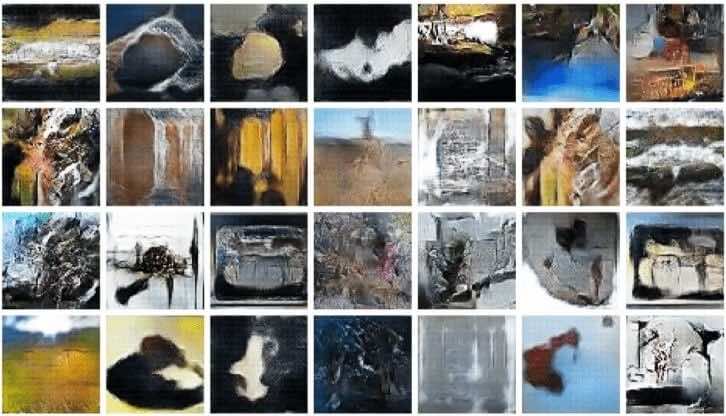} & 
    \includegraphics[width=0.99\linewidth]{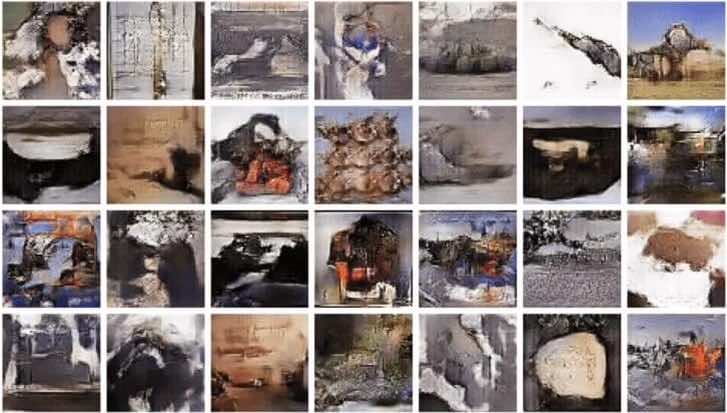}  \\[-1pt]
    (e) Fashion-MNIST input & (f) SVHN input \\[1pt] \hline
     \includegraphics[width=0.99\linewidth]{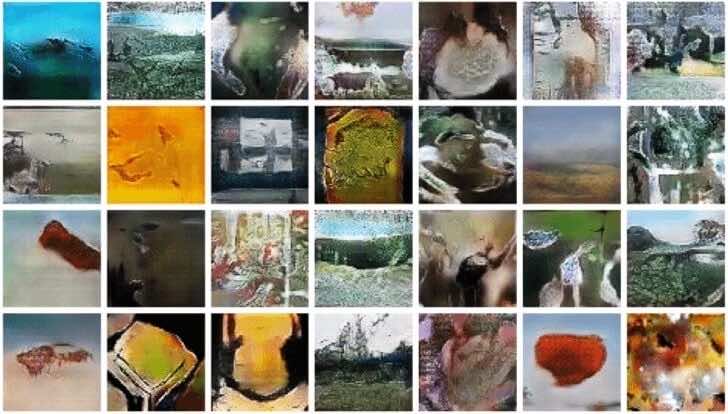} & 
    \includegraphics[width=0.99\linewidth]{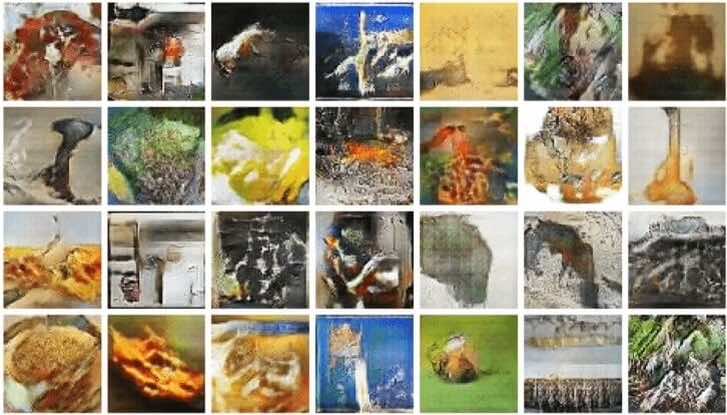}  \\[-1pt]
    (g) CIFAR-10 input & (h) CelebA input \\[1pt] \hline
     \includegraphics[width=0.99\linewidth]{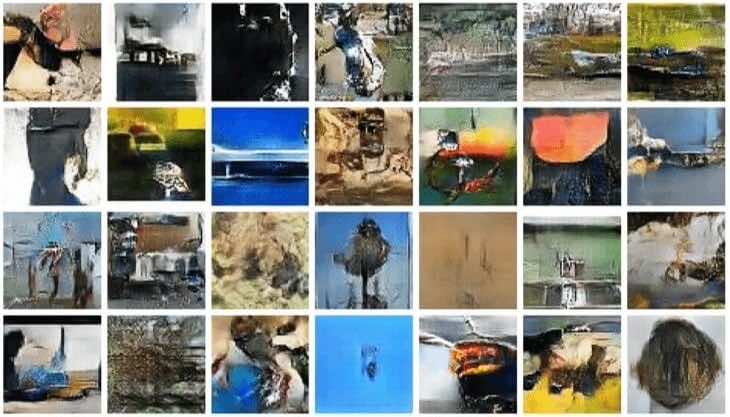} & 
    \includegraphics[width=0.99\linewidth]{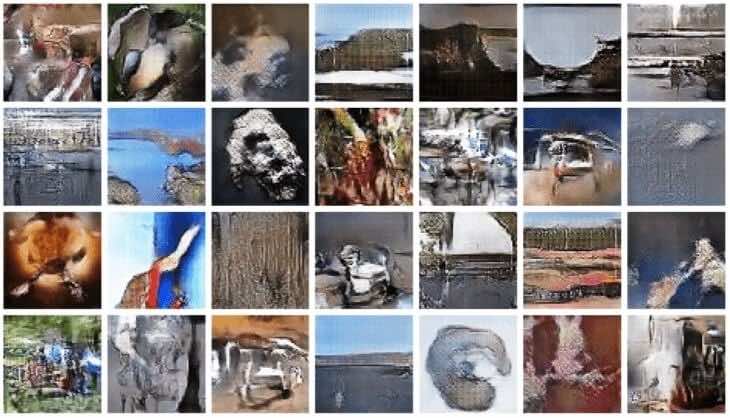}  \\[-1pt]
    (i) Ukiyo-E Faces input & (j) LSUN-Churches input \\[1pt] \hline
  \end{tabular} 
\caption[]{Images generated by the baseline GAN and Spider GAN on Tiny-ImageNet as the target, for various input distributions as indicated. While Spider GAN approaches achieve a lower FID than the baselines on this task, none of the GAN variants generate realistic output images.}
\label{Fig_RandTIN}  
\end{center}
\vskip-1em
\end{figure*}

\begin{figure*}[!thb]
\begin{center}
  \begin{tabular}[b]{P{.38\linewidth}|P{.38\linewidth}}
      \includegraphics[width=0.99\linewidth]{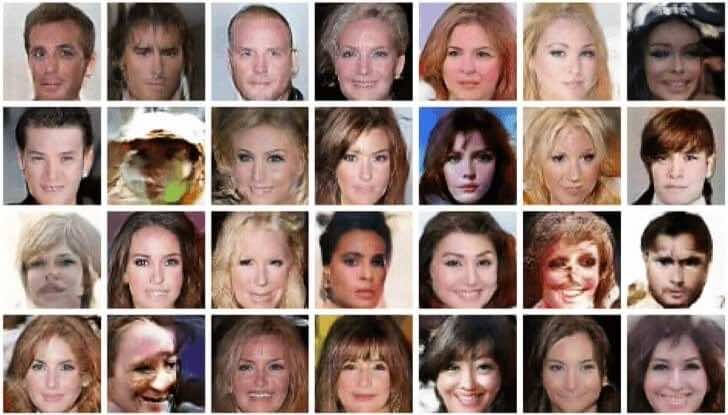} & 
    \includegraphics[width=0.99\linewidth]{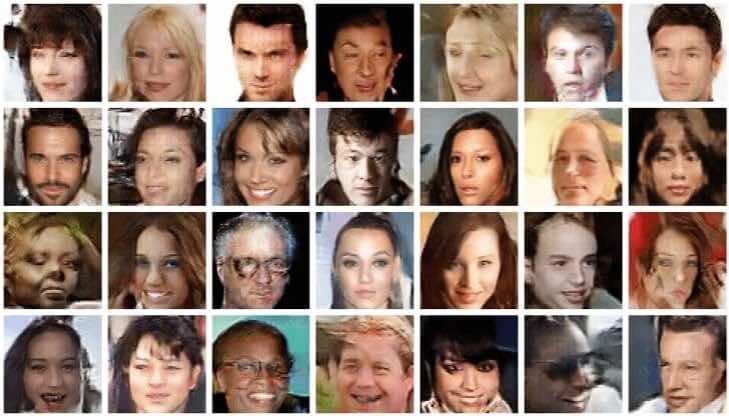}  \\[-1pt]
    (a) Gaussian input & (b) Gamma input \\[1pt] \hline
     \includegraphics[width=0.99\linewidth]{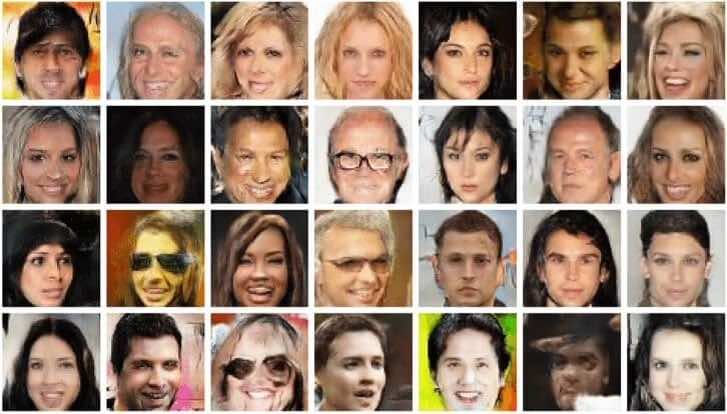} & 
    \includegraphics[width=0.99\linewidth]{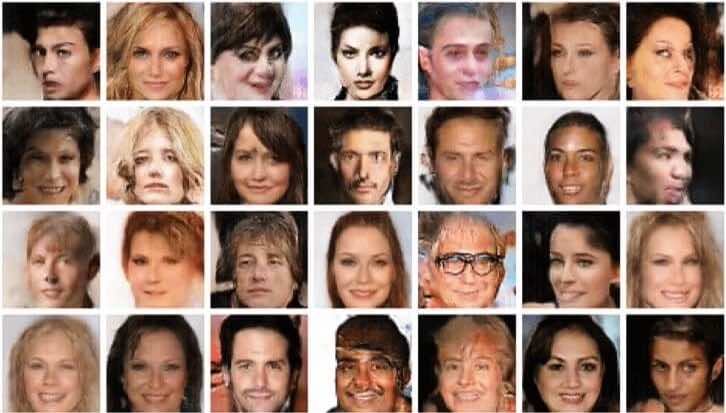}  \\[-1pt]
    (c) Non-Parametric input & (d) MNIST input \\[1pt] \hline
     \includegraphics[width=0.99\linewidth]{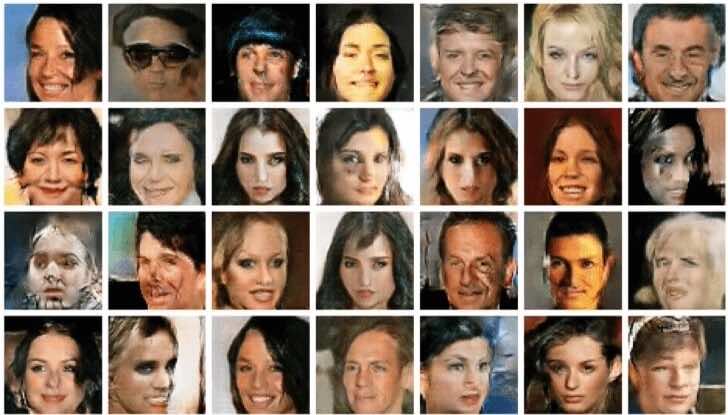} & 
    \includegraphics[width=0.99\linewidth]{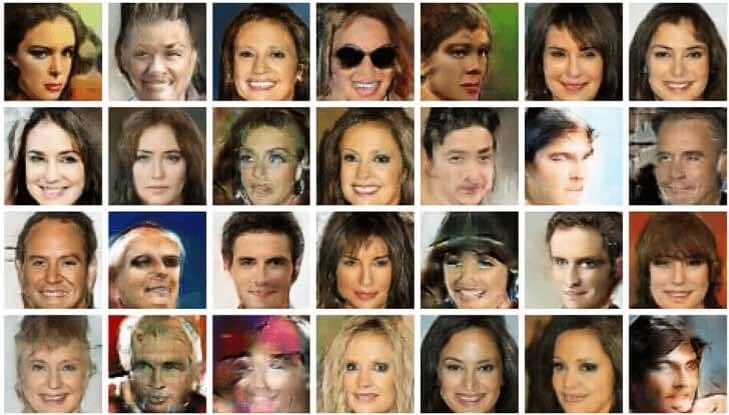}  \\[-1pt]
    (e) Fashion-MNIST input & (f) SVHN input \\[1pt] \hline
     \includegraphics[width=0.99\linewidth]{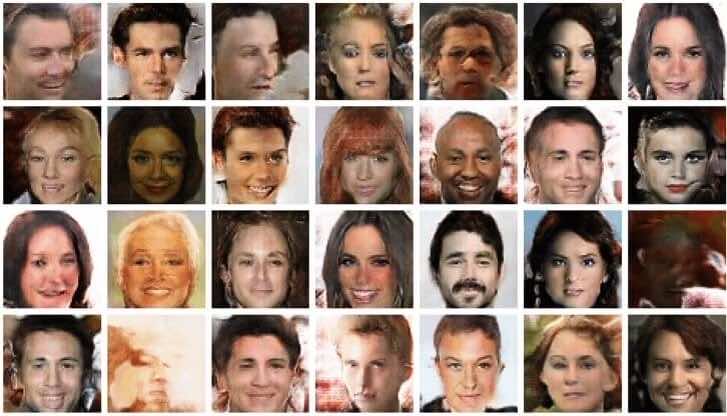} & 
    \includegraphics[width=0.99\linewidth]{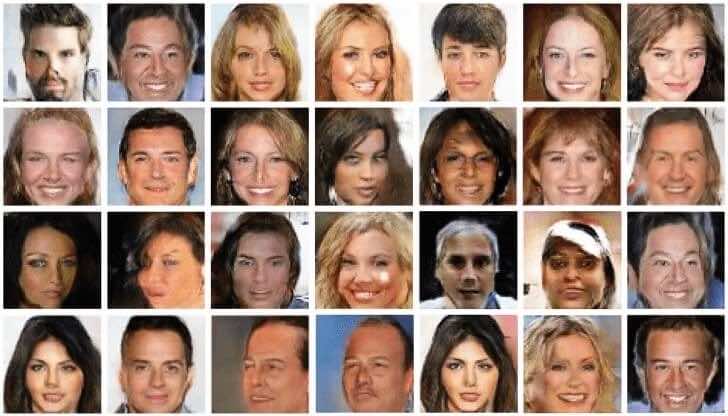}  \\[-1pt]
    (g) CIFAR-10 input & (h) Tiny-ImageNet input  \\[1pt] \hline
     \includegraphics[width=0.99\linewidth]{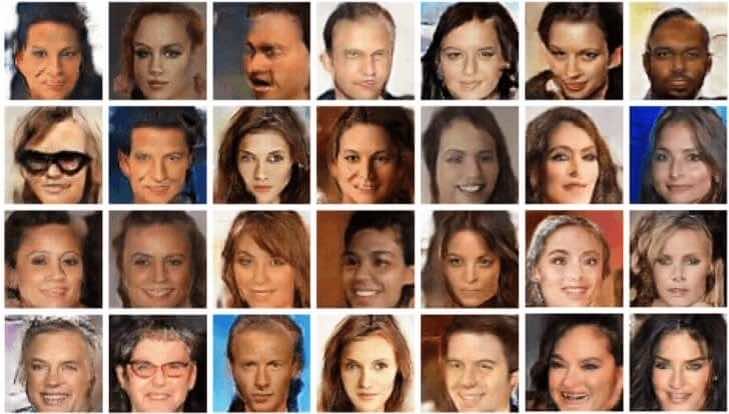} & 
    \includegraphics[width=0.99\linewidth]{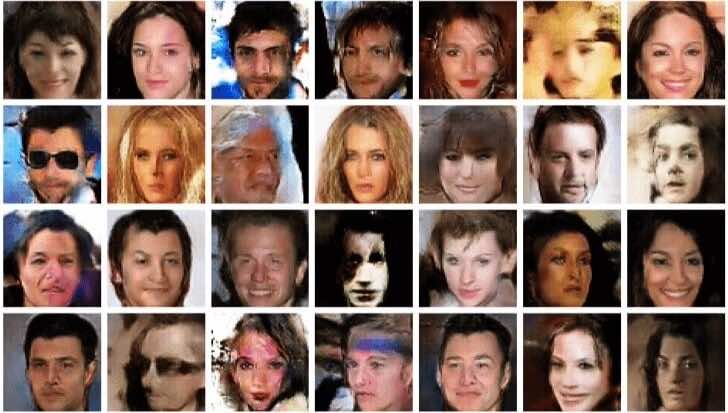}  \\[-1pt]
    (i) Ukiyo-E Faces input & (j) LSUN-Churches input \\[1pt] \hline
  \end{tabular} 
\caption[]{Images generated by the baseline GAN and Spider GAN on the low resolution CelebA (\(64\times64\)), given various input distributions. Images generated by Spider GAN trained with Tiny-ImageNet and Ukiyo-E Faces as the input outperform other GAN flavors.}
\label{Fig_RandCelebA}  
\end{center}
\vskip-1em
\end{figure*}

\begin{figure*}[!thb]
\begin{center}
  \begin{tabular}[b]{P{.38\linewidth}|P{.38\linewidth}}
      \includegraphics[width=0.99\linewidth]{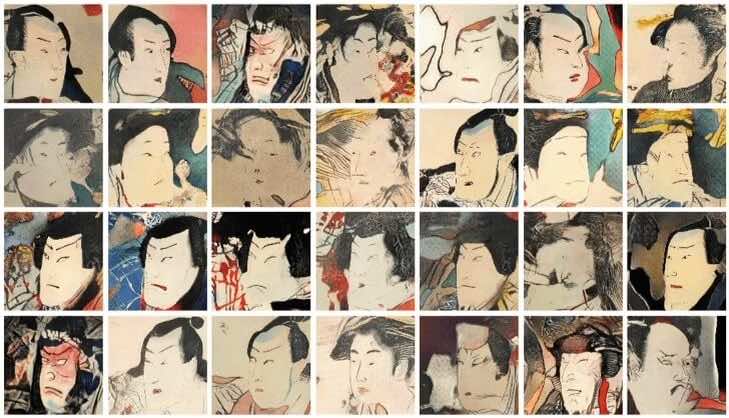} & 
    \includegraphics[width=0.99\linewidth]{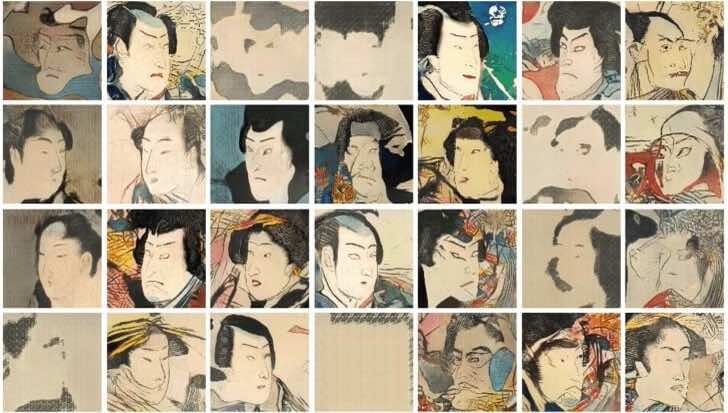}  \\[-1pt]
    (a) Gaussian input & (b) Gamma input\\[1pt] \hline
     \includegraphics[width=0.99\linewidth]{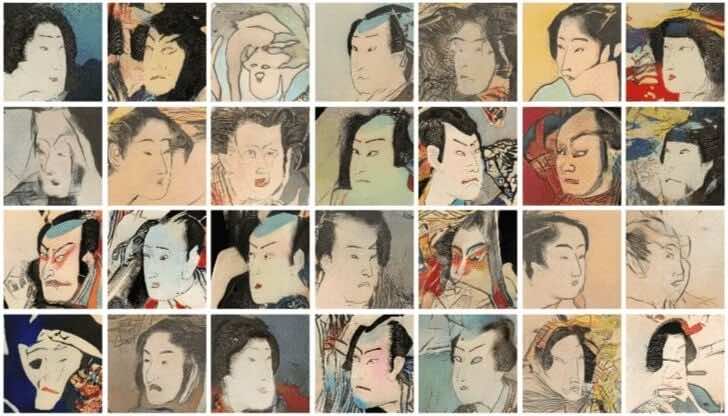} & 
    \includegraphics[width=0.99\linewidth]{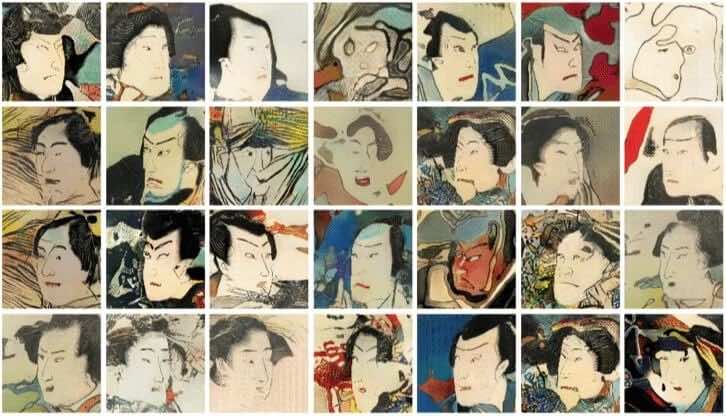}  \\[-1pt]
    (c) Non-Parametric input & (d) MNIST input \\[1pt] \hline
     \includegraphics[width=0.99\linewidth]{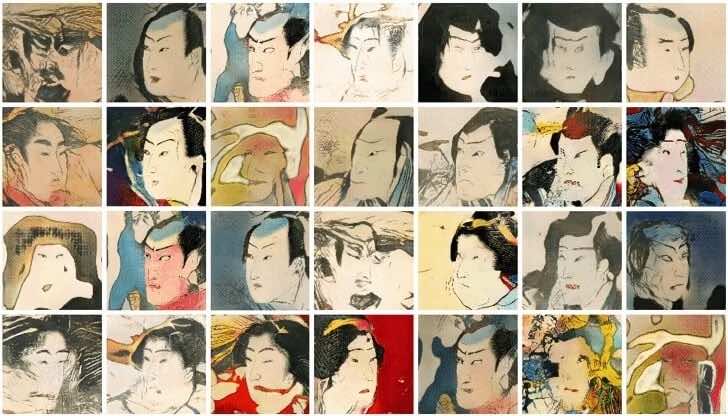} & 
    \includegraphics[width=0.99\linewidth]{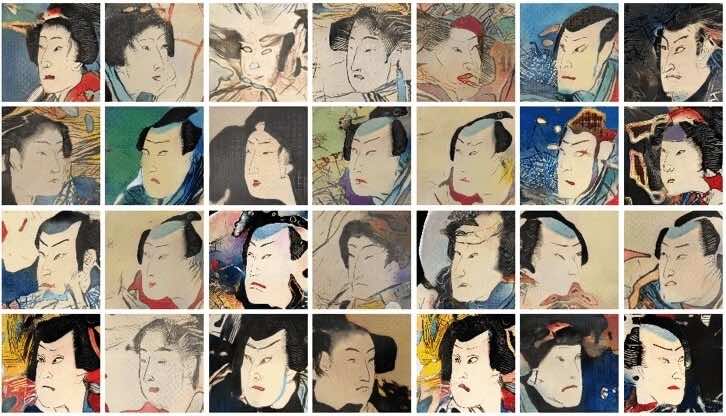}  \\[-1pt]
    (e) Fashion-MNIST input & (f) SVHN input \\[1pt] \hline
     \includegraphics[width=0.99\linewidth]{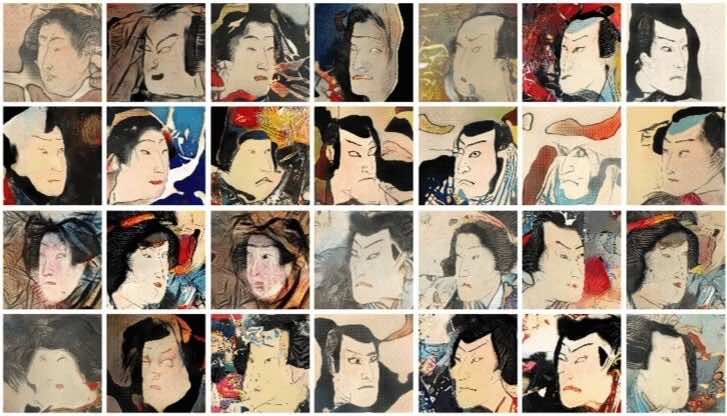} & 
    \includegraphics[width=0.99\linewidth]{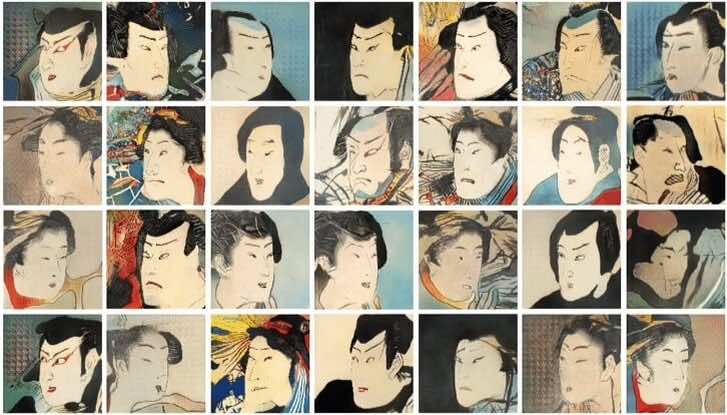}  \\[-1pt]
    (g) CIFAR-10 input & (h) Tiny-ImageNet input  \\[1pt] \hline
     \includegraphics[width=0.99\linewidth]{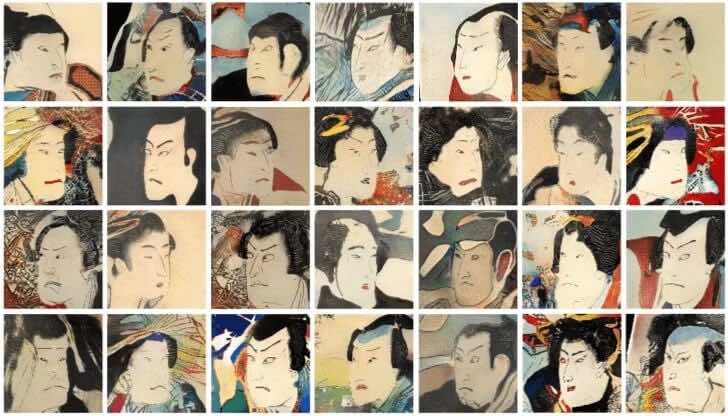} & 
    \includegraphics[width=0.99\linewidth]{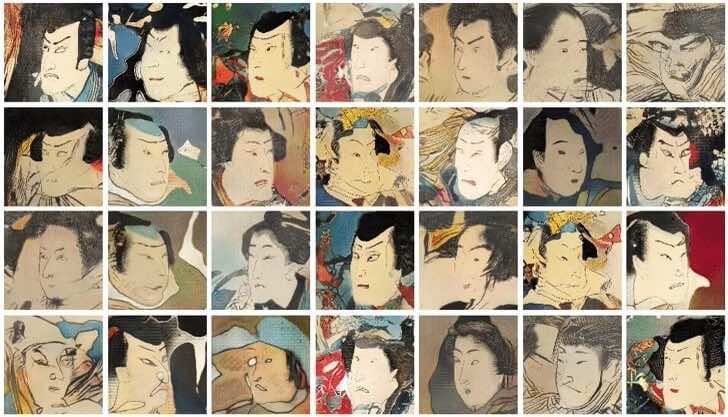}  \\[-1pt]
    (i) CelebA input & (j) LSUN-Churches input \\[1pt] \hline
  \end{tabular} 
\caption[]{Images generated by the baseline GAN and Spider GAN variants on the Ukiyo-E Faces for different inputs to the generator. Images generated by Spider GAN with Tiny-ImageNet or CelebA images as input results in sharper images in comparison to the baselines.}
\label{Fig_RandUki}  
\end{center}
\vskip-1em
\end{figure*}

We now present results on additional experimental validation run on the Spider DCGAN architecture. The experimental setup is the same as the one described in Appendix~\ref{App_ExpSetup}. First, we consider training Spider GAN on Fashion-MNIST, SVHN, Tiny-ImageNet and 64-dimensional CelebA datasets. The FID and KID of the converged models are presented in Table~\ref{Table_FIDKID_Rest}.   On the Fashion-MNIST, SVHN, and CelebA datasets, we observe that the Spider GAN approach with the friendliest neighbor (as identified by FID, KID and CSID\(_m\)), results in improved learning over the baselines. On the Tiny-ImageNet learning task, we observe that a source dataset with less diversity (such as CIFAR-10, as suggested by FID and KID) performs poorly, while a more diverse source dataset, such as CelebA, improves the best-case FID over the baselines. These results validate the {\it friendly neighborhood} of Tiny-ImageNet identified using CSID\(_m\) in Section~\ref{Sec_FriendlyNb}, where CIFAR-10 and LSUN-Churches are less diverse, having a negative CSID\(_m\). Figures~\ref{Fig_RandMNIST}-\ref{Fig_RandUki} present the images generated by Spider GAN and the baseline variants on various datasets considered. Figure~\ref{Fig_FID_vs_Iters_All} presents the convergence of FID as a function of iterations for the remaining source dataset combinations of Spider GAN models considered in Section~\ref{Sec_Exp} and Figure~\ref{Fig_FID_vs_Iters}.\par

\begin{table*}[t!]
\begin{center}
\caption{A comparison of FID and KID in Spider GAN for various noise perturbations considered when the input dataset is Ukiyo-E Faces. Gaussian perturbations such as \( \mcalN(\bm{0},0.25\mbbI)\) and \( \mcalN(\bm{0},0.1\mbbI)\) that are concentrated about their mean result in the best performance improvements over the baseline Spider GAN. } \label{Table_NoisePerturb}  \vskip-0.05in
\begin{tabular}{P{4.8cm}||P{1.5cm}|P{1.5cm}||P{1.5cm}|P{1.5cm}}
\toprule 
\multirow{2 }{*}{Input Distribution}&\multicolumn{2}{c||}{Fashion-MNIST} & \multicolumn{2}{c}{CIFAR-10} \\[1pt] \cline{2-5} \\[-10pt]
 & FID \(\downarrow\) & KID \(\downarrow\) &  FID \(\downarrow\)  &KID \(\downarrow\)\\
\hline\hline &&&& \\[-7pt]
Ukiyo-E Faces 						& 55.1200  & 0.0376 & 74.7085 & 0.0518  \\[4pt]
Ukiyo-E  + \(\mcalN(\bm{0},0.1\mbbI)\) 		& {\bf 47.2873} & {\bf 0.0285} & \uline{70.101} & 0.0488  \\[4pt]
Ukiyo-E  + \(\mcalN(\bm{0},0.25\mbbI)\) 		& \uline{50.2150} & 0.0345 & {\bf 68.7473} & {\bf 0.0473}  \\[4pt]
Ukiyo-E + \(\mcalN(\bm{0},\mbbI)\) 			& 79.8415  & 0.0690 & 71.9181 & 0.0531  \\[4pt]
Ukiyo-E  + Gamma noise			   			& 51.8201  &  0.0343 & 70.362& \uline{0.0476}  \\[4pt]
Ukiyo-E  + Non-parametric  noise				& 50.8536  & \uline{0.0329} & 72.9138  & 0.0495   \\
\bottomrule
\end{tabular}
\end{center}
\end{table*}

\subsubsection{Noise Perturbations on the Input Dataset} \label{App_NoisePerturb}

The SpiderGAN framework relies on the variability present in the chosen input dataset to learn the target better. As discussed in Section~\ref{Sec_Exp}, we considered addition of noise to the dataset input to the Spider GAN generator when the cardinality of the input is small. We observed that CelebA or Tiny-ImageNet are more diverse and perform better than small datasets such as Ukiyo-E Faces. To overcome the lack of diversity in small datasets, we consider additive-noise perturbations to augment the data. While Gaussians are a popular choice, we also consider the Gamma density and non-parametric densities to generate noise, which are known to improve the performance of the GANs on latent-space interpolation. We consider three Gaussian examples: the standard normal \(\mcalN(\bm{0},\mbbI)\), \(\mcalN(\bm{0},0.25\mbbI)\), and \(\mcalN(\bm{0},0.1\mbbI)\). Three variances are considered to highlight the trade-off between generating noisy images (Gaussians with high variance) and low diversity in the input dataset (Gaussians with low variance). We present results on learning MNIST and CIFAR-10 datasets with Ukiyo-E Faces dataset as input. \par
\begin{figure*}[!thb]
\begin{center}
\begin{tabular}[b]{P{.02\linewidth}||P{.42\linewidth}|P{.42\linewidth}}
  
  & Fashion-MNIST  & CIFAR-10 \\[1pt]
  \rotatebox{90}{\footnotesize{ \enskip \(\mfrakD_p = \) Ukiyo-E}} &
    \includegraphics[width=0.99\linewidth]{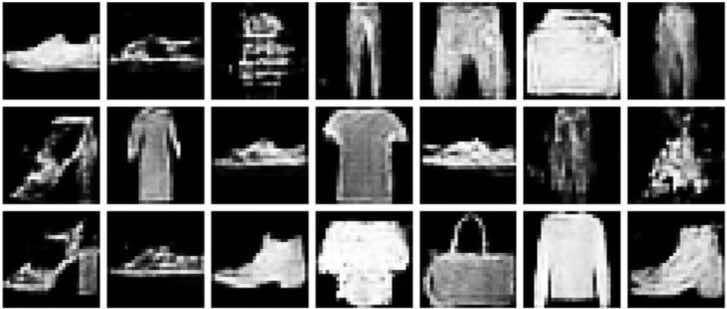} & 
    \includegraphics[width=0.99\linewidth]{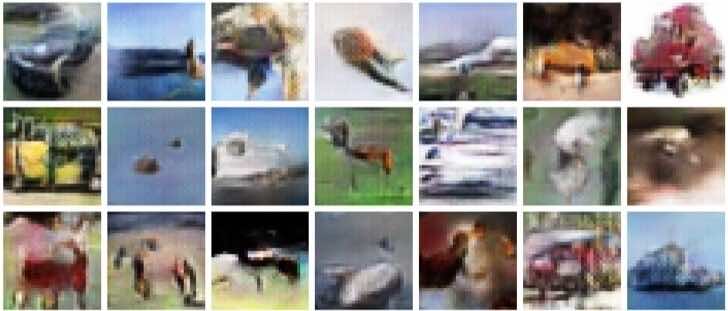}  \\[5pt]
  	\rotatebox{90}{\footnotesize{\quad \(\mfrakD_p + \mathcal{N}(\bm{0}_2, \mathbb{I}_2)\)}} &
      \includegraphics[width=0.99\linewidth]{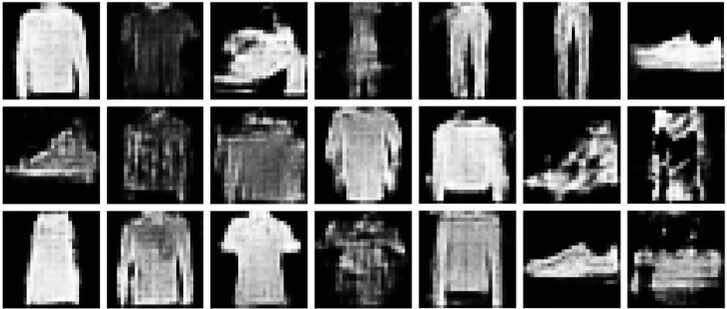} & 
    \includegraphics[width=0.99\linewidth]{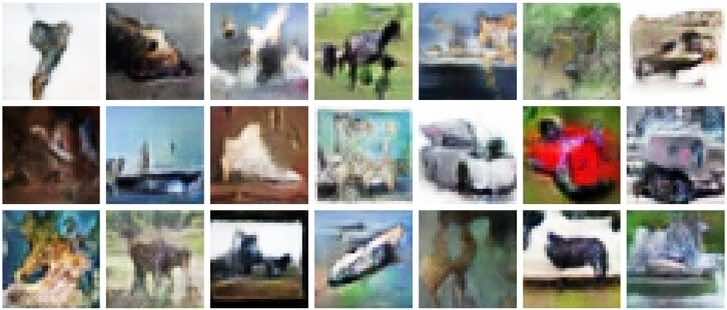}  \\[5pt]
    \rotatebox{90}{\footnotesize{ \(\mfrakD_p + \mathcal{N}(\bm{0}_2, 0.25\mathbb{I}_2)\)}} &
    \includegraphics[width=0.99\linewidth]{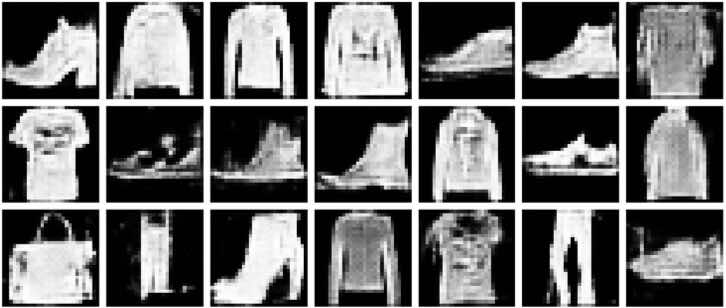} & 
    \includegraphics[width=0.99\linewidth]{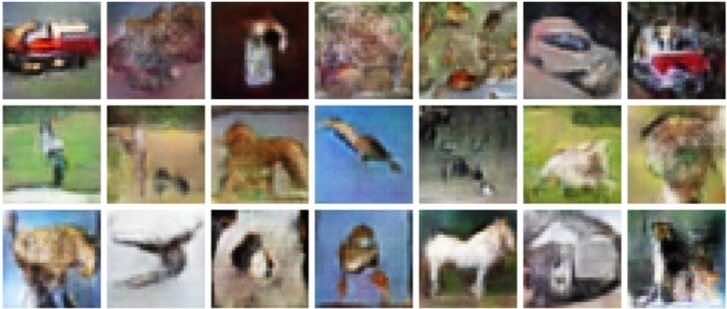}  \\[5pt]
     \rotatebox{90}{\footnotesize{ \(\mfrakD_p + \mathcal{N}(\bm{0}_2, 0.1\mathbb{I}_2)\)}} &
    \includegraphics[width=0.99\linewidth]{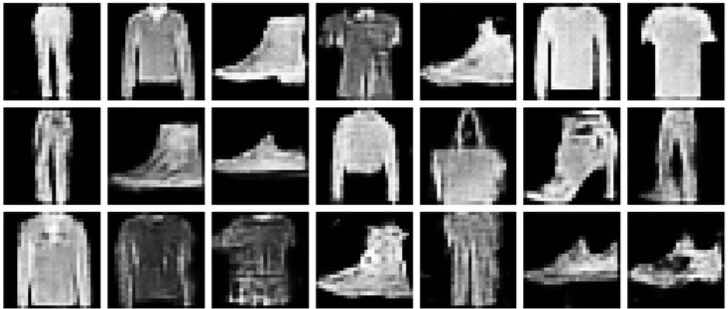} & 
    \includegraphics[width=0.99\linewidth]{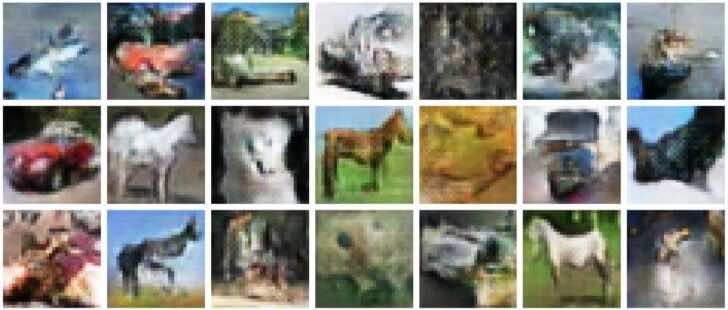}  \\[5pt]
     \rotatebox{90}{\footnotesize{ \quad \(\mfrakD_p +\) Gamma}} &
    \includegraphics[width=0.99\linewidth]{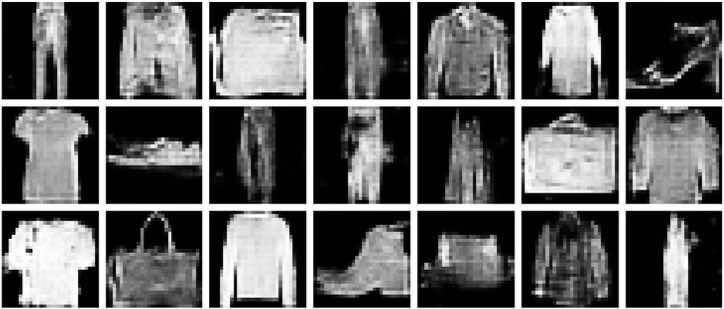} & 
    \includegraphics[width=0.99\linewidth]{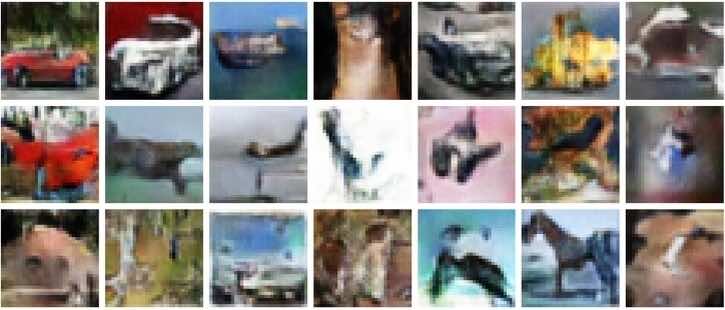}  \\[5pt]
     \rotatebox{90}{\footnotesize{\(\mfrakD_p +\) Non-parametric}} &
    \includegraphics[width=0.99\linewidth]{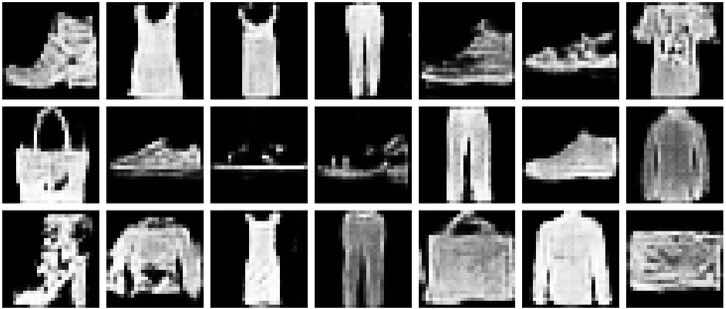} & 
    \includegraphics[width=0.99\linewidth]{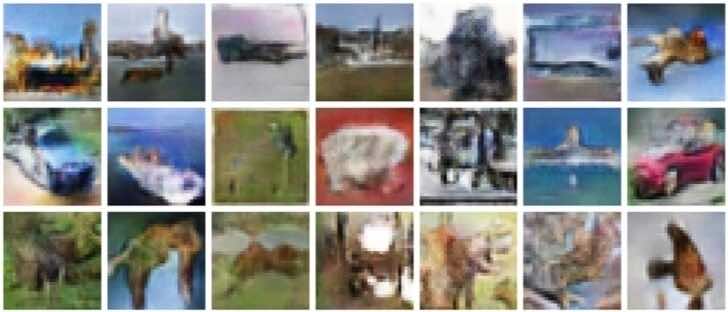}  \\[5pt]
  \end{tabular} 
\caption[]{Images generated by Spider GAN on Fashion-MNIST and CIFAR-10, when trained with the Ukiyo-E Faces as input. Ukiyo-E Faces are perturbed mildly with various parametric noise sources to enhance the input diversity. Gaussian perturbations result in superior image quality compared to the rest.}
\label{Fig_NoisePerturb}  
\end{center}
\vskip-1em
\end{figure*}

\noindent {\bfseries Results:} Figures~\ref{Fig_NoisePerturb}(a)-(f) show the images generated by Spider GAN with various noise perturbations applied to Ukiyo-E Faces. Adding Gaussian noise with a small variance, or Gamma distributed noise results in diverse images and better visual quality of generated images. On the other hand, models trained with the standard normal or non-parametric densities resulted in poor learning, with several out-of-distribution images. The performance of the converged models in presented in Table~\ref{Table_NoisePerturb}. Perturbations that are Gaussian, and concentrated about the mean, such as \(\mcalN(\bm{0},0.1\mbbI)\) or \(\mcalN(\bm{0},0.25\mbbI)\) resulted in the lowest FID and KID. Therefore, Gaussian perturbations with a small variance result in better performance when the input datasets have small cardinality. \par

\begin{table}[!b]
\begin{center}
\caption{Comparison of sharpness metric evaluated on interpolated images in MNIST, CIFAR-10 and Ukiyo-E learning tasks. The benchmark sharpness is computed on target data samples. Spider GAN variants outperform the baselines on CIFAR-10 and MNIST, while being on par with the non-parametric prior on the Ukiyo-E Faces. The values shown in {\bfseries bold} are closest to the benchmark sharpness.} \label{Table_SharpInterpol}  
\vskip1pt
\begin{tabular}{P{0.05cm}P{3cm}||P{2.5cm}|P{2.5cm}|P{2.5cm}}
\toprule 
\multicolumn{2}{c||}{\multirow{2}{*}{Input Distribution}}&\multicolumn{3}{c}{Sharpness of the Interpolated Image} \\\cline{3-5} \\[-12pt] &&&& \\[-8pt]
\multirow{3}{*}{\rotatebox{90}{{\footnotesize\uline{Baselines}\quad \enskip\enskip} }}
&& MNIST & CIFAR-10 & Ukiyo-E  \\[2pt]
\hline\hline
&Gaussian 			& 0.0868 & 0.587 & 1.730    \\[0.5pt]
&Gamma 			    & 0.0536 & 1.217  & 1.981    \\[0.5pt]
&Non-parametric 		& 0.2522 & 0.785  & {\bfseries 2.538}    \\[1pt] \hline
\multirow{6}{*}{\rotatebox{90}{{\footnotesize\uline{\enskip\quad Spider GAN \quad\enskip} } }}
&MNIST 			    & -- & 0.467  & 2.008     \\
&Fashion MNIST  		& {\bfseries 0.1408} & 0.377  & 1.353   \\
&SVHN  			    & 0.0898 & 1.214  & 1.480   \\
&CIFAR-10 			 & 0.0859 & -- &  2.533    \\
&TinyImageNet 		& 0.0623 &  {\bfseries 0.906} & 1.274  \\
&CelebA 			    & 0.1735 & 0.449  & 2.104     \\ \hline
&Benchmark           & 0.1396 & 0.993  & 2.748    \\
\bottomrule
\end{tabular}
\end{center}
\vskip-1pt
\end{table}

\begin{table*}[!b]
\fontsize{9}{12}\selectfont
\begin{center}
\caption{Comparison of {\it Interpolation FID} and {\it Interpolation KID} for the Spider GAN and baseline variants on MNIST, CIFAR-10, and Ukiyo-E Faces datasets. The input provided to the generator \(\z_{in} = \frac{\z_1+\z_2}{2};~\z_1,\z_2 \sim p_Z\) is the mid-point between two samples drawn from the input distribution \(p_Z\), either of parametric form in the case of the baselines, or the {\it friendly neighborhood} datasets, in the case of Spider GAN. The values in the parentheses indicate the relative increase in the FID/KID scores, in comparison to those reported in Table~\ref{Table_FID}. Spider GANs with {\it friendliest neighborhood} input datasets achieve FID and KID scores on par with the best-case baseline.} \label{Table_InterpolFID}  \vskip5pt
\begin{tabular}{P{0.05cm}P{2.9cm}||P{1.1cm}|P{1.1cm}||P{1.1cm}|P{1.1cm}||P{1.1cm}|P{1.1cm}}
\toprule 
\multicolumn{2}{c||}{\multirow{2}{*}{Input Distribution}}&\multicolumn{2}{c||}{MNIST} & \multicolumn{2}{c||}{CIFAR10}& \multicolumn{2}{c}{Ukiyo-E Faces} \\\cline{3-8} 
&& FID & KID & FID & KID & FID & KID  \\
\hline\hline
\multirow{6}{*}{\rotatebox{90}{{\footnotesize\uline{\quad\enskip Baselines\quad \enskip} } }}&\multirow{2}{*}{Gaussian \((\mathbb{R}^{100})\)}				& 25.111 & 0.0181 & 121.198 & 0.0848 & 74.241 & 0.0612 \\
&& (+16.8\%) & (+30.2\%) & (+68.7\%) & (+36.9\%) & (+3.1\%) & (+14.4\%) \\ \cline{2-8}
&\multirow{2}{*}{Gamma 	 \((\mathbb{R}^{100})\)}		    		& 23.564 & 0.0149 & 77.113 & 0.0492 & 70.302 & 0.0558  \\
&& (+11.3\%) & (+12.1\%) & (+6.1\%) & (+1.8\%)  & (+0.4\%) & (+18.7\%)  \\ \cline{2-8}
&\multirow{2}{*}{Non-Parametric  \((\mathbb{R}^{100})\)}		& 22.301 & 0.0142 & 87.478 & 0.0568 & {\bf 66.022} & {\bf 0.0434}  \\[1pt] 
&&  (+6.4\%) & (+3.6\%) & (+16.7\%) & (+7.1\%)  & (+1.0\%) & (+3.0\%)  \\ \hline\hline
\multirow{14}{*}{\rotatebox{90}{{\footnotesize\uline{\quad\quad\quad\quad\quad\quad Spider GAN \quad\quad\quad\quad\quad\quad} } }}
&\multirow{2}{*}{MNIST }			   		 & \multirow{2}{*}{--} & \multirow{2}{*}{--} & 122.084 & 0.0790 & 103.80 & 0.0732 \\
&& && (+71.2\%) & (+47.5\%)  & (+51.4\%) & (+67.1\%)  \\ \cline{2-8}
&\multirow{2}{*}{Fashion MNIST } 			& {\bf 20.644} & {\bf 0.0147} & 113.109 & 0.0731 & 89.901 & 0.0654 \\
&&  (+22.8\%) & (+42.7\%) & (+46.7\%) & (+32.9\%)  & (+23.6\%) & (+43.7\%)  \\ \cline{2-8}
&\multirow{2}{*}{SVHN  	}		    		& 27.630 & 0.0208 & 89.161 & 0.0558 & 77.302 & 0.0542 \\
&&  (+1.8\%) & (+1.5\%) & (+39.1\%) & (+23.7\%)  & (+10.0\%) & (+12.4\%)  \\ \cline{2-8}
&\multirow{2}{*}{CIFAR-10 }				& 30.214 & 0.0305 & \multirow{2}{*}{--} & \multirow{2}{*}{--} & 87.981 & 0.0621 \\
&&  (+3.4\%) & (+38.6\%) & & & (+24.2\%) & (+17.1\%)  \\ \cline{2-8}
&\multirow{2}{*}{TinyImageNet }			& 46.233 & 0.0397 & \uline{86.708} & {\bf 0.0520} & 79.848 & 0.0565 \\
&&  (+41.6\%) & (+50.4\%) & (+47.3\%) & (+70.4\%)  & (+28.9\%) & (+29.5\%)  \\ \cline{2-8}
&\multirow{2}{*}{CelebA }			    		& \uline{21.517} & \uline{0.0152} & {\bf 86.475} & \uline{0.0534} & \uline{68.849} & \uline{0.0449} \\
&&  (+4.6\%) & (+5.5\%) & (+43.9\%) & (+23.0\%)  & (+27.2\%) & (+10.1\%)  \\ \cline{2-8}
&\multirow{2}{*}{Ukiyo-E }			    		& 38.950 & 0.0318 & 98.045 & 0.0671 & \multirow{2}{*}{--} & \multirow{2}{*}{--} \\
&&  (+26.9\%) & (+39.4\%) & (+60.6\%) & (+83.8\%)  & &  \\
\bottomrule
\end{tabular}
\end{center}
\vskip-2.5em
\end{table*}

\subsubsection{Input-space Interpolation with Spider DCGAN} \label{App_Interpol}

Gamma and non-parametric priors were introduced to the GAN landscape to improve the quality of interpolated images in GANs~\cite{Gamma18,NonPara19}. We compare the image interpolation quality of Spider GAN with respect to the gamma and non-parametric baselines. The experimental setup is similar to that in Appendix~\ref{App_ExpSetup}. We compare the visual quality of  images generated by interpolated inputs to the generator. In the baseline GANs, we provide the generator with eight linearly interpolated points between two random samples drawn from the prior densities. In the case of Spider GAN, we draw two random samples from the input dataset, and generate eight linearly interpolated images that are input to the Spider GAN generator. The quality of the interpolation is evaluated in terms of the sharpness metric. We present results on MNIST, CIFAR-10, and Ukiyo-E Faces.  \par
\begin{sidewaysfigure}
\begin{center}
 \begin{tabular}[b]{P{.45\linewidth}|P{.45\linewidth}}
  \toprule
     \includegraphics[width=1\linewidth]{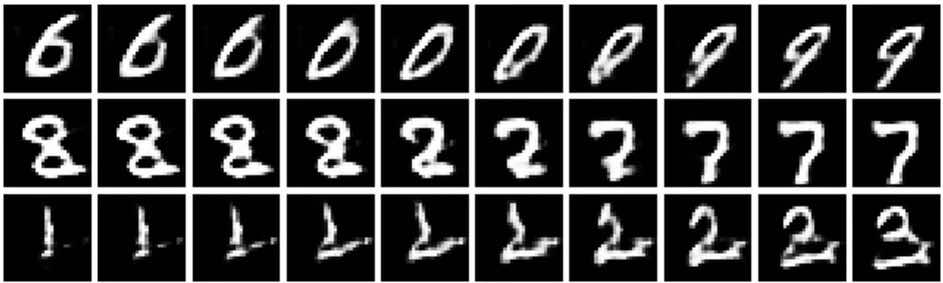} &
    \includegraphics[width=1\linewidth]{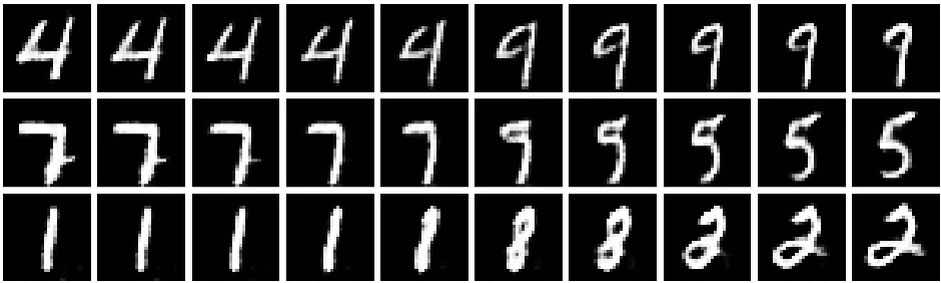}\\ 
    (a)~Gaussian input & (d) Fashion-MNIST input \\
    \midrule\\[-11pt]
    \includegraphics[width=1\linewidth]{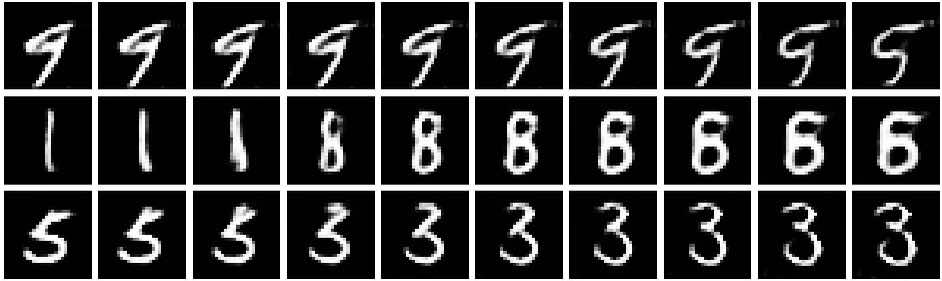} &
    \includegraphics[width=1\linewidth]{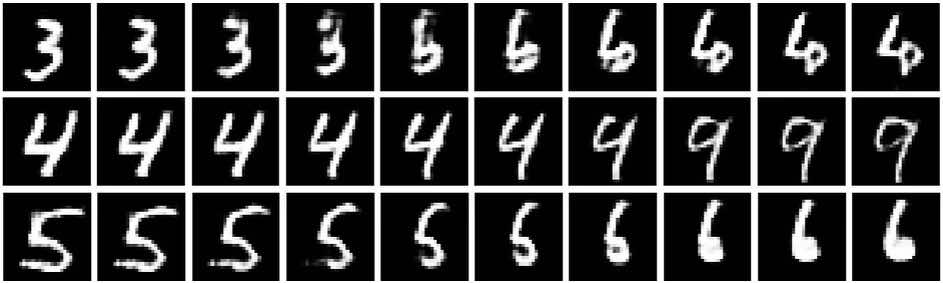} \\
    (b)~Gamma input & (e) CIFAR-10 input \\
    \midrule\\[-11pt]
    \includegraphics[width=1\linewidth]{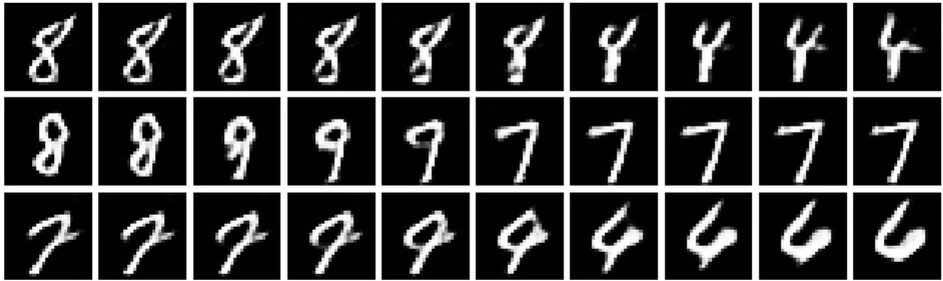} &
    \includegraphics[width=1\linewidth]{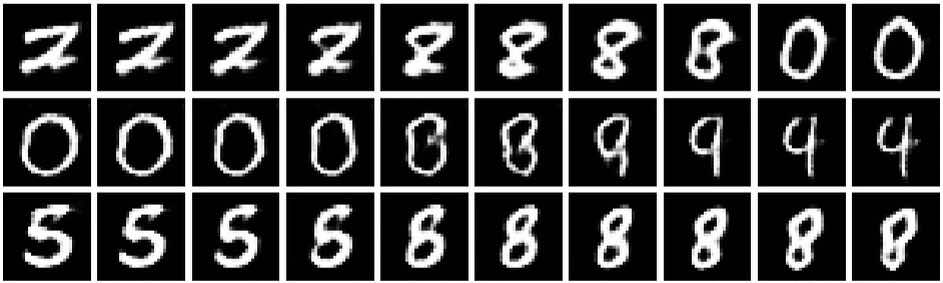} \\
    (c)~Non-parametric input & (f) Tiny-ImageNet input \\
    \bottomrule
     \end{tabular} 
 \caption[]{Input-space interpolation on the baseline and Spider GAN variants trained on the MNIST dataset. Interpolation with Fashion-MNIST input for Spider GAN results in output images that transition smoothly, compared to the baselines.}
 \label{Fig_InterpolMNIST}
 \end{center}
 \vskip-3em
\end{sidewaysfigure}
\begin{sidewaysfigure}
\begin{center}
 \begin{tabular}[b]{P{.45\linewidth}|P{.45\linewidth}}
  \toprule
     \includegraphics[width=1\linewidth]{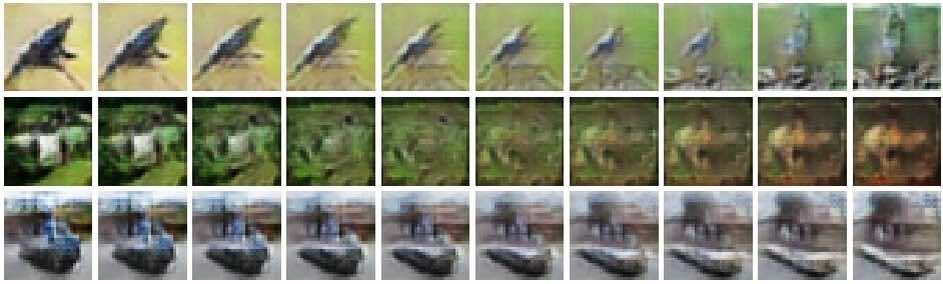} &
    \includegraphics[width=1\linewidth]{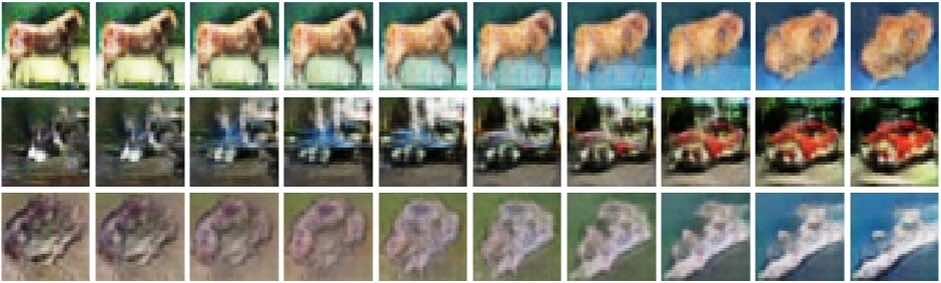}\\ 
    (a)~Gaussian input & (d) Tiny-ImageNet input \\
    \midrule\\[-11pt]
    \includegraphics[width=1\linewidth]{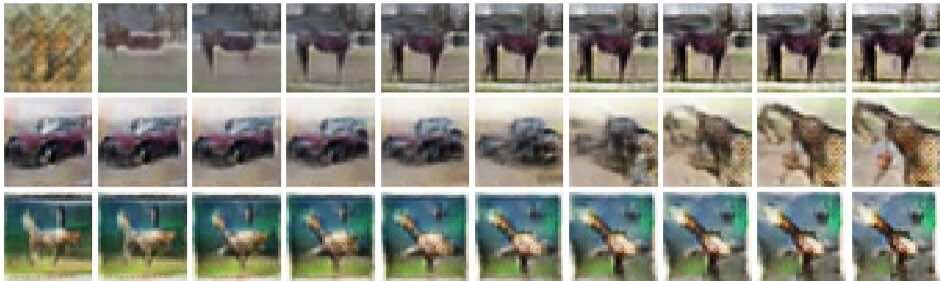} &
    \includegraphics[width=1\linewidth]{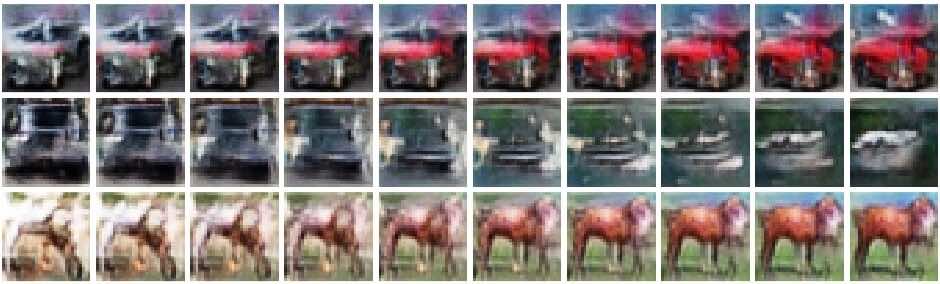} \\
    (b)~Gamma input & (e) CelebA input \\
    \midrule\\[-11pt]
    \includegraphics[width=1\linewidth]{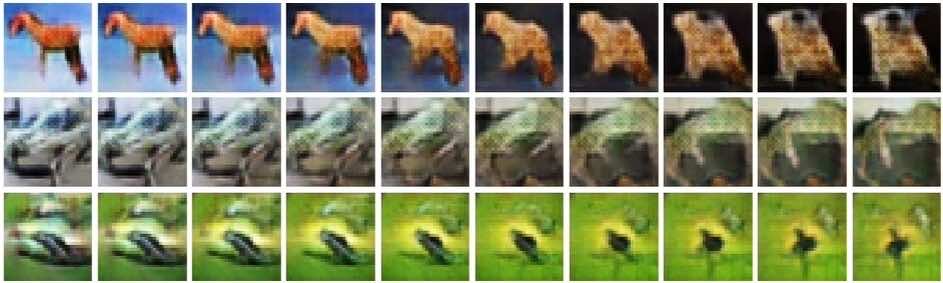} &
    \includegraphics[width=1\linewidth]{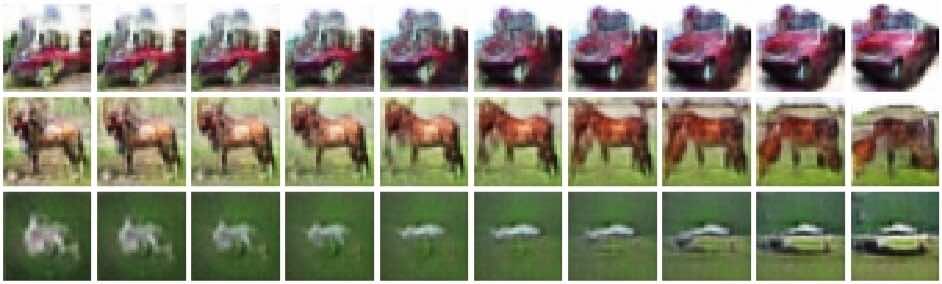} \\
    (c)~Non-parametric input & (f) LSUN-Churches input \\
    \bottomrule
     \end{tabular} 
 \caption[]{(\includegraphics[height=0.012\textheight]{Rgb.png} Color online)~Input-space interpolation on the baseline GANs and Spider GAN trained on the CIFAR-10 dataset. Interpolation with Tiny-ImageNet input for Spider GAN result in output images of superior quality. However, all variants fail to create realistic images when provided with interpolated input samples. }
 \label{Fig_InterpolC10}
 \end{center}
 \vskip-3em
\end{sidewaysfigure}
\begin{sidewaysfigure}
\begin{center}
 \begin{tabular}[b]{P{.45\linewidth}|P{.45\linewidth}}
  \toprule
     \includegraphics[width=1\linewidth]{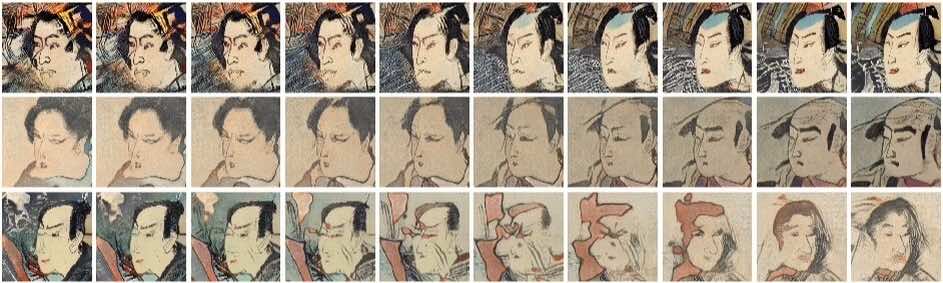} &
    \includegraphics[width=1\linewidth]{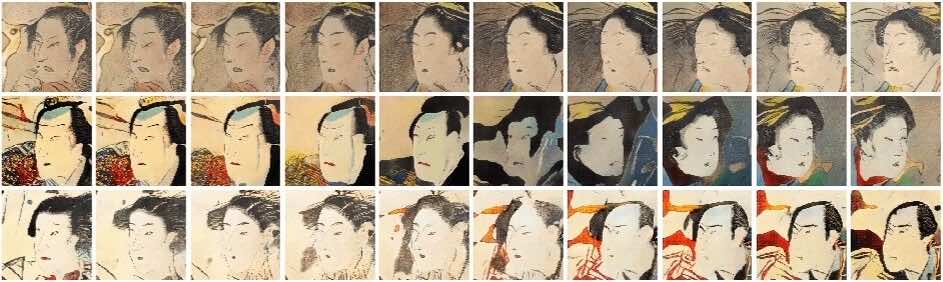}\\ 
    (a)~Gaussian input & (d) Tiny-ImageNet input \\
    \midrule\\[-11pt]
    \includegraphics[width=1\linewidth]{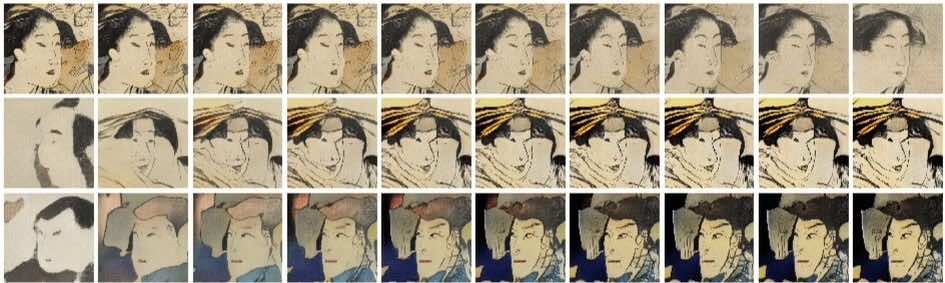} &
    \includegraphics[width=1\linewidth]{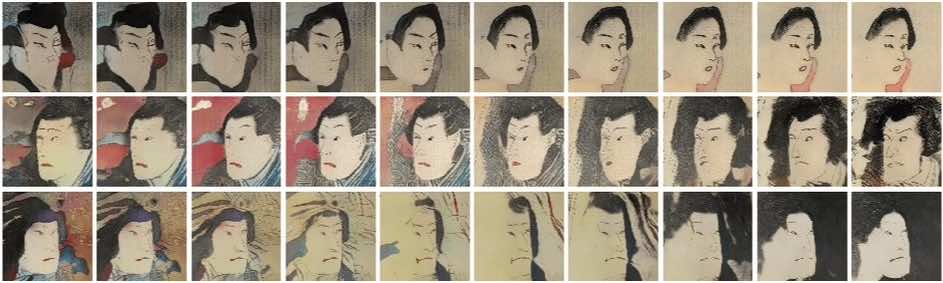} \\
    (b)~Gamma input & (e) CelebA input \\
    \midrule\\[-11pt]
    \includegraphics[width=1\linewidth]{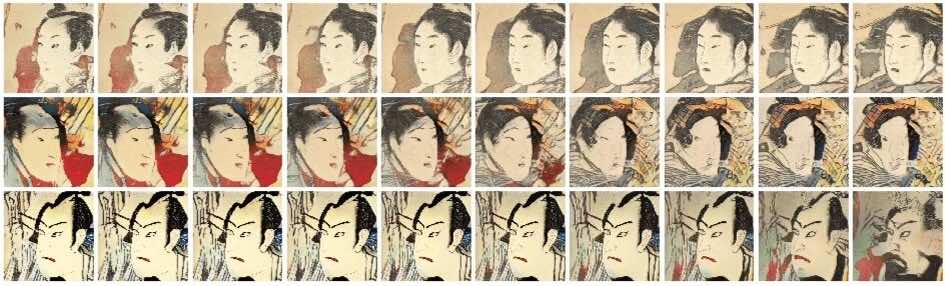} &
    \includegraphics[width=1\linewidth]{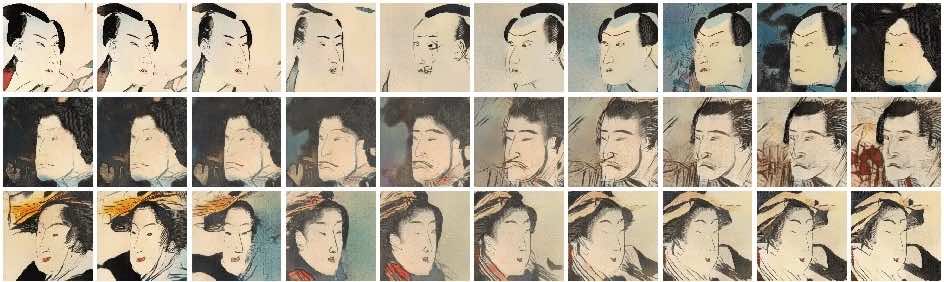} \\
    (c)~Non-parametric input & (f) LSUN-Churches input \\
    \bottomrule
     \end{tabular} 
 \caption[]{(\includegraphics[height=0.012\textheight]{Rgb.png} Color online)~Input-space interpolation on the baseline and Spider GAN variants trained on the Ukiyo-E Faces. Interpolations with CelebA input for Spider GAN are the smoothest. Baseline variants result is sharp fluctuations in the orientation of the faces, which is indicative of a non-smooth generator in the output space.}
 \label{Fig_InterpolUki}
 \end{center}
 \vskip-3em
\end{sidewaysfigure}

Figures~\ref{Fig_InterpolMNIST}-\ref{Fig_InterpolUki} present the images generated by the interpolated input vectors by the three baseline GAN variants and Spider GAN with the three friendliest neighbors as the input datasets. We observe that, Spider GAN, although not trained for the task, is able to generate realistic interpolated images. The visual quality is on par with the non-parametric interpolation scheme in the case of MNIST, and superior to the baselines on the Ukiyo-E Faces learning task. All variants fail to generate realistic images on CIFAR-10. Table~\ref{Table_SharpInterpol} shows the sharpness metric computed on the interpolated images. We observe that Spider GAN variants attain values closer to the benchmark in comparison with the baselines. As discussed in the Main Manuscript, the best performance of Spider GAN is achieved when the input dataset is the {\it friendliest neighbor} of all the target datasets under consideration. Table~\ref{Table_InterpolFID} presents the FID and KID scores of the Spider GAN and baseline variants, when computed on a batch of \(10^4\) samples obtained by proving the mid-point sample \(\z_{in} = \frac{\z_1+\z_2}{2};~\z_1,\z_2 \sim p_Z\) as input to the generator. The inputs \(\z_1\) and \(\z_2\) are samples drawn from parametric distributions as in the case of the baselines, or images from the {\it friendly neighborhood} input dataset as in the case of Spider GAN. Table~\ref{Table_InterpolFID} also shows the relative increase in FID and KID compared to those obtained when unaltered samples drawn from \(p_Z\) are provided as input to the generator (cf. Table~\ref{Table_FID}). Across all the datasets considered, we observe that Spider GAN variants with the {\it friendliest neighbor} input result in a performance comparable with the best-case baselines in terms of FID and KID. However, the baselines GAN with the non-parametric or gamma-distributed priors, which are designed to minimize the interpolation error~\cite{Gamma18,NonPara19}, and consequently, result in lower relative change in the scores. The results suggest that, while Spider GAN is superior to Gaussian latent spaces, a trade-off exists between the interpolated image quality offered by non-parametric or gamma priors, and the overall superior performance offered by Spider GAN. A detailed discussion on the input-space control over the generated images is discussed in the context of Spider StyleGAN2-ADA in Appendix~\ref{App_InterpolStyleGAN} \par

\subsubsection{Impact of Diversity and Dataset Bias on Spider GANs} \label{App_Bias}

The {\it friendly neighbourhood} of a target in Spider GAN is chosen based on the SID metric, which compares the distance between data manifolds. Spider GAN does not enforce image-level structure to learn pairwise transformations.  We therefore expect that the diversity of the source dataset (such as racial or gender bias) should not affect the diversity in the learnt distribution. To demonstrate this, consider the task of learning Ukiyo-E faces dataset with CelebA dataset as input. We consider three variants of CelebA -- (i) The entire dataset of \(2\times 10^5\) images, comprising an even split of the {\it male} and {\it females} classes; (ii) Only the {\it female} class comprising \(10^5\) images; and (iii) A simulated imbalance, created by including the entire {\it male} class and 200 images from the {\it female} class. The input resolution is \(64\times64\), while the output resolution is set to \(128\times128\). The models are trained using the DCGAN architecture with hyperparameters as described in Appendix~\ref{App_ExpSetup}. All the models are trained for \(10^5\) generator iterations. \par

\begin{figure*}[!t]
\begin{center}
  \begin{tabular}[b]{P{.95\linewidth}}
  \toprule
   \includegraphics[width=0.97\linewidth]{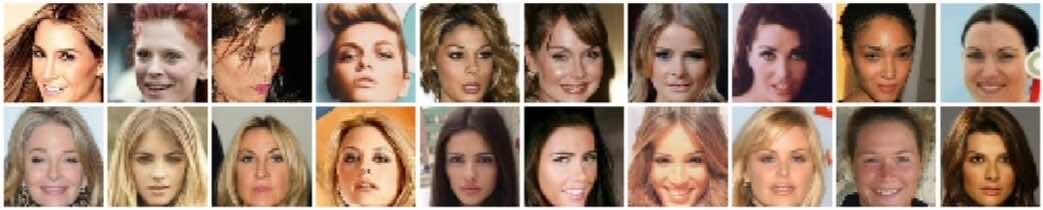}  \\
   (a) Input Samples from the {\it Female Class} of the source CelebA  dataset. \\[3pt]
    \midrule \midrule
    \includegraphics[width=0.97\linewidth]{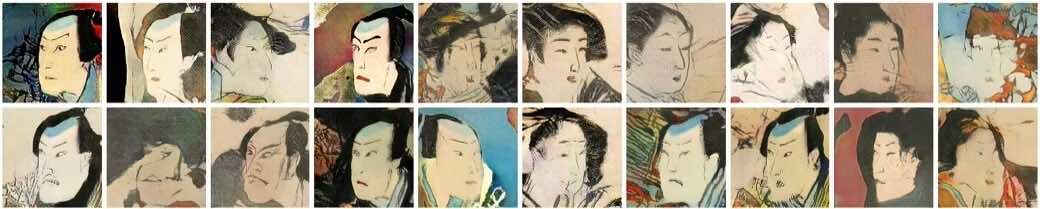}  \\[5pt]
     (a.1) Corresponding outputs for balanced source data. \\[3pt]
      \midrule
    \includegraphics[width=0.97\linewidth]{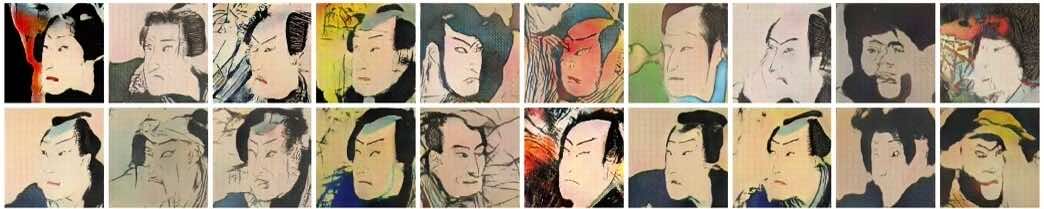}  \\[5pt]
     (a.2) Corresponding outputs for source data bias:  100\% {\it Males} class + 0.2\% {\it Female Class}. \\[3pt]
      \midrule
    \includegraphics[width=0.97\linewidth]{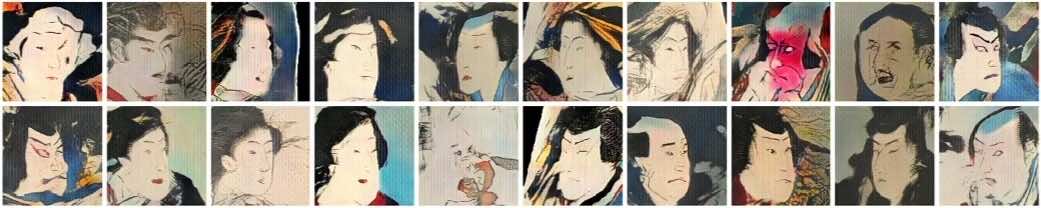}  \\[5pt]
     (a.3) Corresponding outputs for source data bias:   0\% {\it Males} class + 100\% {\it Female Class}.  \\[3pt]
     \bottomrule
  \end{tabular} 
\caption[]{Images generated by Spider GAN when trained on the Ukiyo-E Faces as the target dataset, with varying levels of bias simulated in the source CelebA dataset. The output images (a.1-a.3) correspond to the generator input with the same {\it Females class} CelebA images depicted. The bias in the input dataset does not carry over to the generator outputs in Spider GAN formulation. Irrespective of the class imbalance in the source CelebA images, the generated Ukiyo-E Faces posses sufficient class diversity.  }  
 \label{Fig_DatasetBias}
\end{center}
\vskip-1em
\end{figure*}

The images output by the Spider GAN model in each case are presented in Figure~\ref{Fig_DatasetBias} (a.1-a.3). We did not observe bias in the images generated by the three models. To demonstrate this further, we compared the Spider GAN outputs for the same 20 samples of the {\it female} class images provided as input (cf. Figure~\ref{Fig_DatasetBias} (a)). The results indicate that, while correspondence between images is not learnt, the bias in the source dataset of the generator in Spider GAN does not affect the target diversity. The bias in these datasets is neither leveraged, nor exemplified by Spider GAN.

\subsubsection{Mode Coverage in Spider GANs} \label{App_ModeCoverage}

In order to evaluate the mode coverage in Spider GAN learning, consider the {\it partial MNIST} experiment proposed by Zhong {\it et al.}~\cite{MixGAN19} involving the 11-class augmented Fashion-MNIST dataset consisting of an additional 100 images drawn from from the {\it digit 1} class of MNIST. We train Spider GAN on the Fashion-MNIST dataset with the CAE architecture (cf. Appendix~\ref{App_CAE}). We consider two input datasets:  CIFAR-10 and Tiny-ImageNet. \par

\begin{figure}[!h]
\begin{center}
  \begin{tabular}[b]{P{.45\linewidth}|P{.45\linewidth}}
      \includegraphics[width=0.95\linewidth]{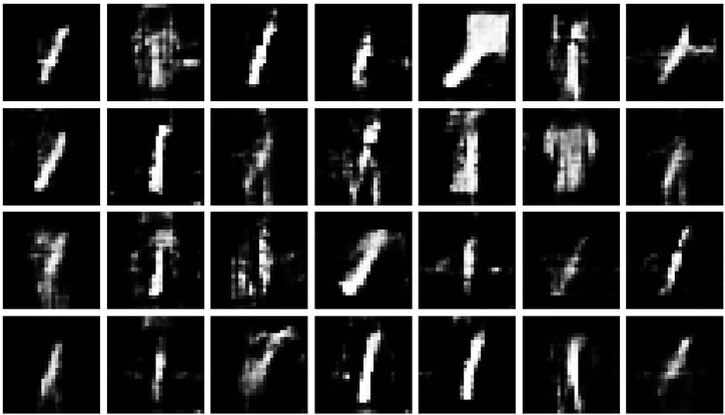} & 
    \includegraphics[width=0.95\linewidth]{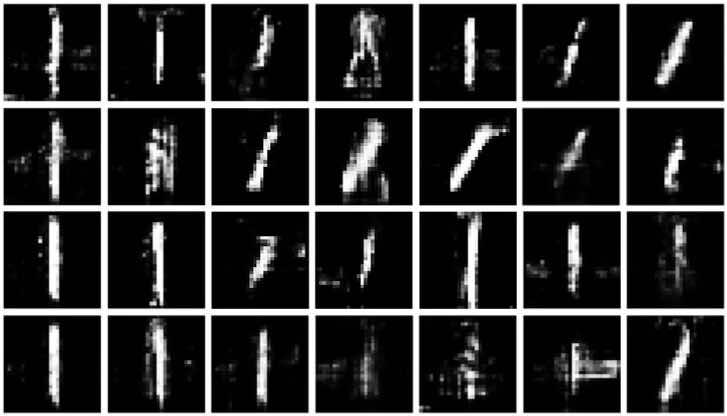}  \\[-1pt]
    (a) & (b) \\[-1pt]
  \end{tabular} 
\caption[]{Images from the {\it digit class} 1 generated by Spider GAN with input images drawn from (a) CIFAR-10, and (b) Tiny-ImageNet datasets. The samples were identified based on the output of a pre-trained 11-class classifier network. Spider GAN with an input class diversity lower  than the target (CIFAR-10 dataset) generated images of inferior quality in comparison to the Spider GAN trained on a more diverse input dataset such as Tiny-ImageNet. }
\label{Fig_Ones}  
\end{center}
\vskip-1em
\end{figure}

In order to evaluate mode coverage, the trained GAN generators are compared on the ability to faithfully generate samples from the underrepresented {\it digit 1} class. For evaluation, an 11-class fully-connected classifier is trained on the augmented dataset consisting of all 10 classes from Fashion-MNIST and the entire {\it digit 1} class from MNIST. Following the approach presented in~\cite{MixGAN19}, the GANs are evaluated by sampling a batch of \(9\times10^5\) images, and computing the number of instances of {\it digit 1} generated, as indicated by the output of the classifier.  We compare against the DCGAN, AdaGAN~\cite{AdaGAN17} and the GAN with mixture of generators (MixGAN)~\cite{MixGAN19}. The images from {\it digit class 1} generated by the Spider GAN variants are presented in Figure~\ref{Fig_Ones}, while Table~\ref{Tabel_Ones} summarizes the performance of the baseline and Spider GAN models. The results highlight the need for class diversity in the input dataset. When Spider GAN is trained with CIFAR-10, consisting of fewer classes than the target, the minority {\it digit 1} class is poorly represented. On the other hand, for Spider GAN with Tiny-ImageNet or CelebA as the input, the minority class is generated faithfully.

\subsubsection{Learning the Identity Mapping} \label{App_Identity}

\begin{figure*}[!b]
\begin{center}
  \begin{tabular}[b]{P{.75\linewidth}}
  \toprule
   \includegraphics[width=0.95\linewidth]{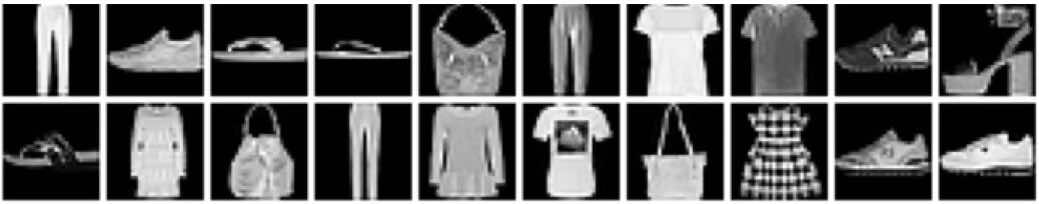}  \\[3pt]
   (a) Input Samples drawn from Fashion-MNIST.   \\[3pt]
    \midrule 
    \includegraphics[width=0.95\linewidth]{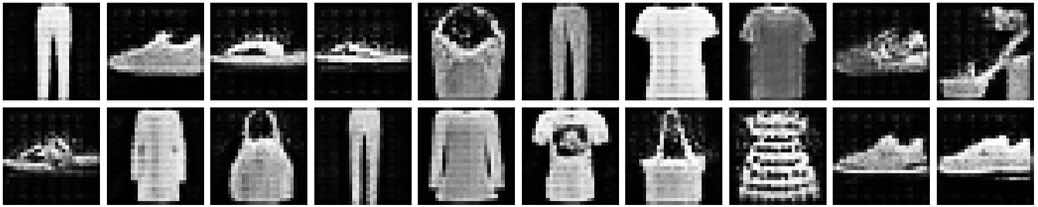}  \\[3pt]
     (a.1) Spider GAN output when trained on noisy Fashion-MNIST as target. \\[3pt]
      \midrule
    \includegraphics[width=0.95\linewidth]{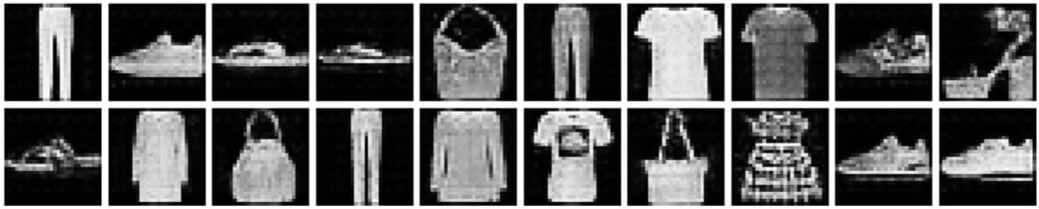}  \\[3pt]
  	(a.2) Spider GAN output when trained on Fashion-MNIST as target.  \\[3pt]
      \midrule \midrule
   \includegraphics[width=0.95\linewidth]{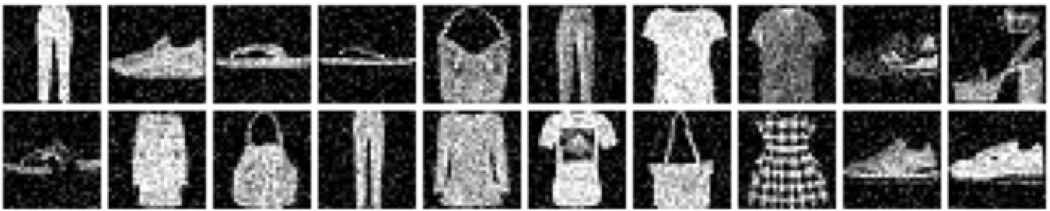}  \\[3pt]
   (b) Input Samples drawn from noisy Fashion-MNIST.   \\[3pt]
    \midrule 
    \includegraphics[width=0.95\linewidth]{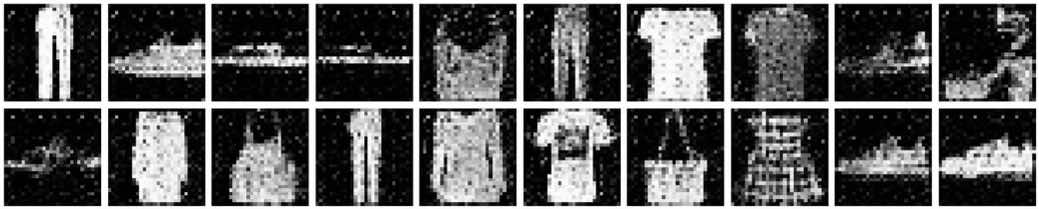} \\[3pt]
     (b.1)  Spider GAN output when trained on noisy Fashion-MNIST as target.  \\[3pt]
      \midrule
    \includegraphics[width=0.95\linewidth]{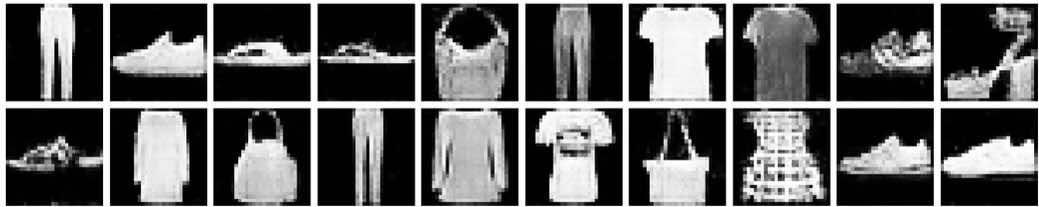}  \\[3pt]
     (b.2) Spider GAN output when trained on Fashion-MNIST as target. \\[3pt]
     \bottomrule
  \end{tabular} 
\caption[]{Images generated by Spider GAN when trained on various combinations of noisy and clean Fashion-MNIST images provided as the input and output to the GAN.  In all scenarios, although pairwise consistency was not explicitly enforced, it was discovered by Spider GAN network.  When the input and output datasets are (a.2) both clean, or (b.1) both noisy, the generator attempts to learn an identity mapping. When the input dataset is clean, but the target dataset incorporates noise (a.1), we observe artifacts in the generated images. Spider GAN with a noisy input dataset and clean target samples learns a denoising network. }  
 \label{Fig_Idenity}
\end{center}
\vskip-1em
\end{figure*}

Based on the intuition that GAN generators perform entropy minimization~\cite{InfoGAN16}, we expect the generator to learn an identity mapping when the same dataset is provided as both input and output. To validate this, we consider the Fashion-MNIST learning task with the DCGAN architecture. We considered all four combinations of adding noise to the input or target datasets. The learnt input-output pairs are presented in Figure~\ref{Fig_Idenity}. In all four scenarios, although pairwise consistency is not explicitly enforced, it was discovered by Spider GAN, resulting in a GAN generator that approximates an identity function. When the input and output datasets are both noisy, the generator attempts to retain the noise in the generated images. However, when the input dataset is clean but the target dataset incorporates noise, artifacts are introduced in the generated images as the models attempts to create noise (which has a higher entropy than the dataset).

\begin{table*}[!h]
\begin{center}
\caption{Mode coverage of Spider GAN in comparison to baseline GANs on the {\it Fashion-MNIST and partial MNIST} experiment. The \(^*\) indicates values reported by Zhong {\it et al.}~\cite{MixGAN19}. The measure {\it\#1s} indicates the number of the samples from the {\it digit class} 1 predicted in a  batch of \(9\times10^5\) samples drawn the generator. {\it Avg. Prob.} denotes the average classification probability of {\it digit class 1} samples output by a pre-trained classifier. Spider GAN trained with an input dataset that posses higher diversity than the target, such as Tiny-ImageNet, outperforms the baselines. } \label{Tabel_Ones} \vskip-0.05in
\begin{tabular}{P{1.7cm}||P{1.5cm}|P{1.5cm}|P{1.5cm}|P{3.5cm}|P{3.5cm}}
\toprule 
Measure \((\uparrow)\) & DCGAN\(^*\) &  AdaGAN\(^*\)  &MixGAN\(^*\) & Spider GAN (CIFAR-10 Source) & Spider GAN (Tiny-ImageNet Source)\\
\hline\hline &&&&& \\[-7pt]
{\it \#1s}	 			& 13  & 60 & 289 & 201 & {\bf 345}  \\[4pt]
{\it Avg. Prob.}			& 0.49 & 0.45 & 0.69 & 0.81 & {\bf 0.89}  \\[4pt]
\bottomrule
\end{tabular}
\end{center}
\vskip-3pt
\end{table*}

\newpage
\FloatBarrier

\subsection{Class-conditional Spider GAN} \label{App_CCSpiderGAN}

\begin{sidewaysfigure}
\begin{center}
  \begin{tabular}[b]{P{.45\linewidth}|P{.45\linewidth}}
  \toprule
   \includegraphics[width=0.99\linewidth]{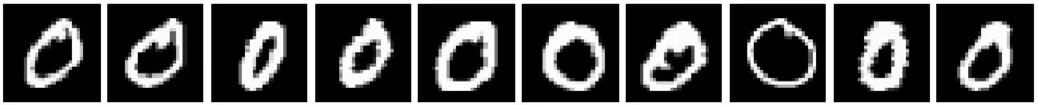} &
   \includegraphics[width=0.99\linewidth]{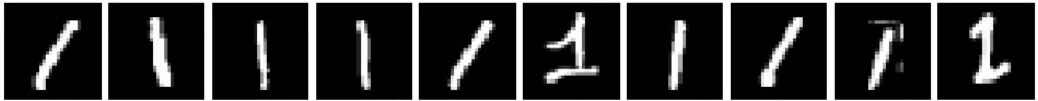}  \\
   \includegraphics[width=0.99\linewidth]{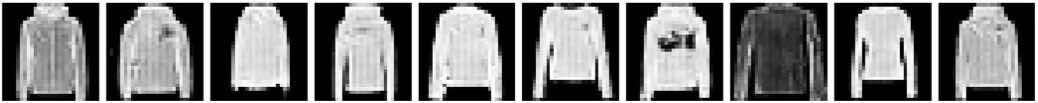} &
    \includegraphics[width=0.99\linewidth]{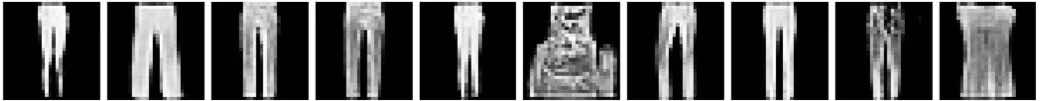}  \\
    \midrule 
     \includegraphics[width=0.99\linewidth]{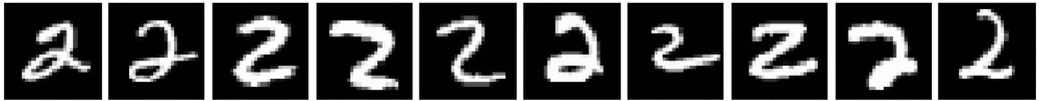} &
   \includegraphics[width=0.99\linewidth]{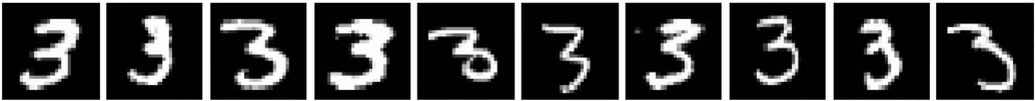}  \\
   \includegraphics[width=0.99\linewidth]{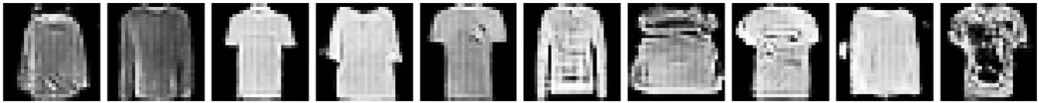} &
    \includegraphics[width=0.99\linewidth]{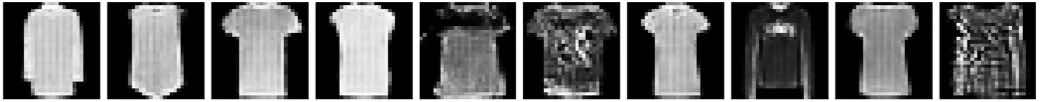}  \\
    \midrule 
     \includegraphics[width=0.99\linewidth]{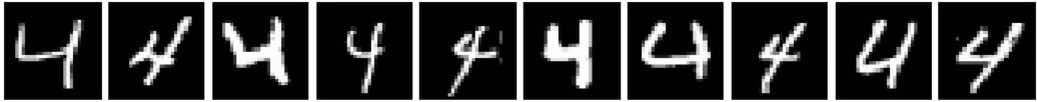} &
   \includegraphics[width=0.99\linewidth]{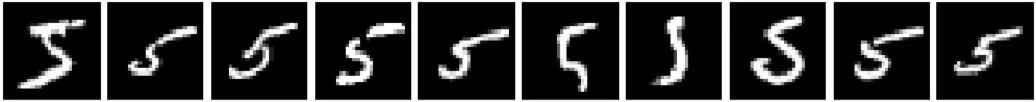}  \\
   \includegraphics[width=0.99\linewidth]{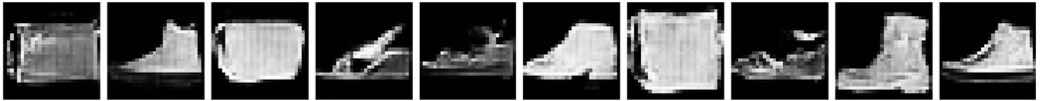} &
    \includegraphics[width=0.99\linewidth]{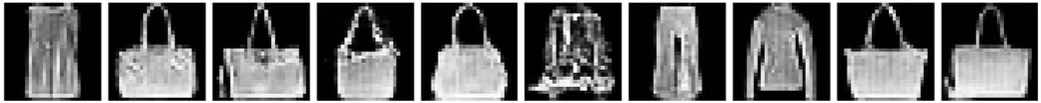}  \\
    \midrule 
     \includegraphics[width=0.99\linewidth]{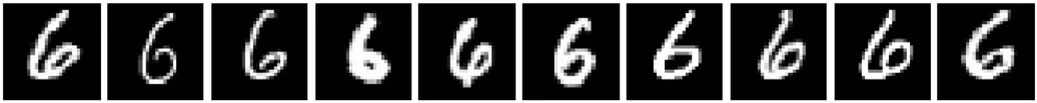} &
   \includegraphics[width=0.99\linewidth]{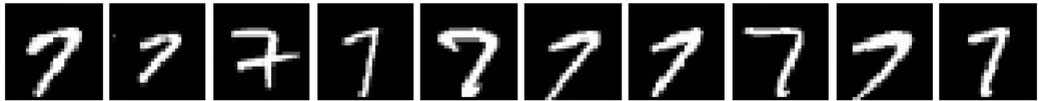}  \\
   \includegraphics[width=0.99\linewidth]{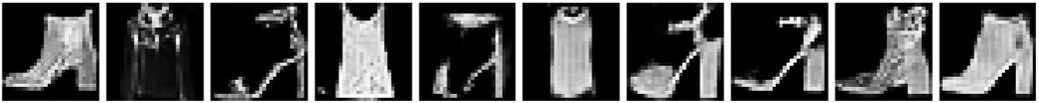} &
    \includegraphics[width=0.99\linewidth]{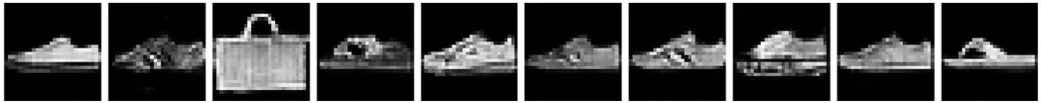}  \\
    \midrule 
     \includegraphics[width=0.99\linewidth]{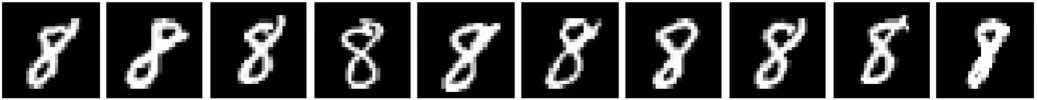} &
   \includegraphics[width=0.99\linewidth]{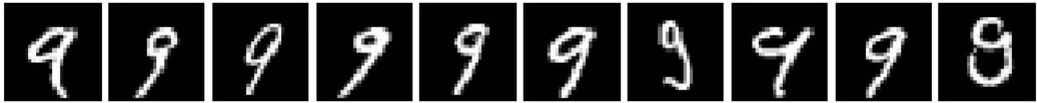}  \\
   \includegraphics[width=0.99\linewidth]{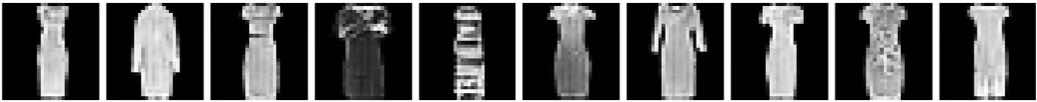} &
    \includegraphics[width=0.99\linewidth]{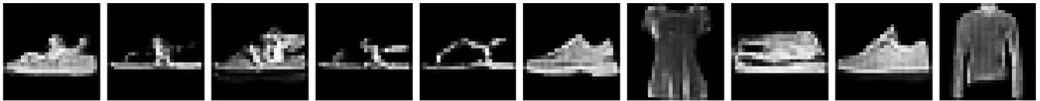}  \\
      \bottomrule
  \end{tabular} 
\caption[]{Images representing the class correspondence in a baseline Spider GAN where class embedding are not provided to the networks. While Style GAN learns to map images based on image-level structure, there is overlap between classes.}  
 \label{Fig_ccBase}
\end{center}
\vskip-1em
\end{sidewaysfigure}

\begin{sidewaysfigure}
\begin{center}
  \begin{tabular}[b]{P{.45\linewidth}|P{.45\linewidth}}
  \toprule
   \includegraphics[width=0.99\linewidth]{Figure282930_Input_00.jpeg} &
   \includegraphics[width=0.99\linewidth]{Figure282930_Input_01.jpeg}  \\
   \includegraphics[width=0.99\linewidth]{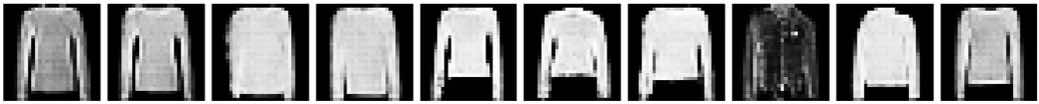} &
    \includegraphics[width=0.99\linewidth]{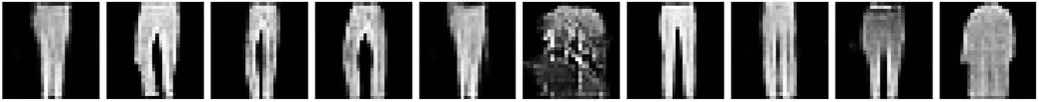}  \\
    \midrule 
     \includegraphics[width=0.99\linewidth]{Figure282930_Input_02.jpeg} &
   \includegraphics[width=0.99\linewidth]{Figure282930_Input_03.jpeg}  \\
   \includegraphics[width=0.99\linewidth]{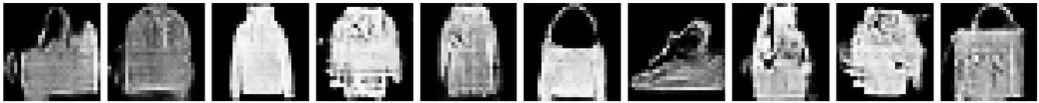} &
    \includegraphics[width=0.99\linewidth]{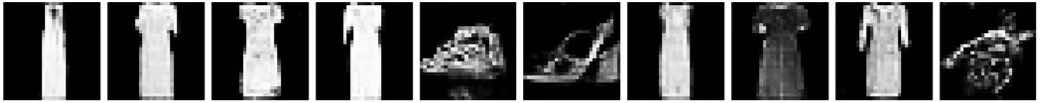}  \\
    \midrule 
     \includegraphics[width=0.99\linewidth]{Figure282930_Input_04.jpeg} &
   \includegraphics[width=0.99\linewidth]{Figure282930_Input_05.jpeg}  \\
   \includegraphics[width=0.99\linewidth]{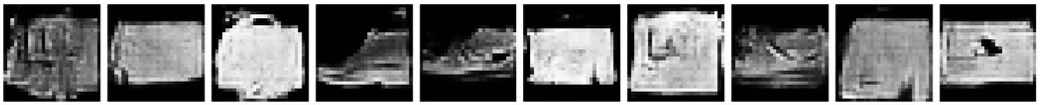} &
    \includegraphics[width=0.99\linewidth]{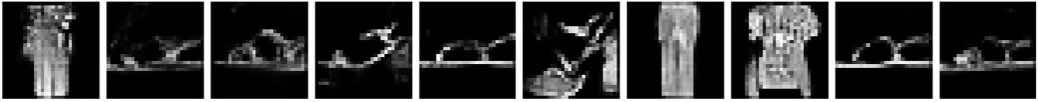}  \\
    \midrule 
     \includegraphics[width=0.99\linewidth]{Figure282930_Input_06.jpeg} &
   \includegraphics[width=0.99\linewidth]{Figure282930_Input_07.jpeg}  \\
   \includegraphics[width=0.99\linewidth]{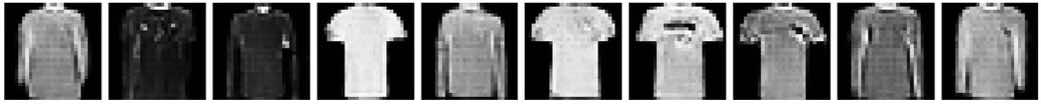} &
    \includegraphics[width=0.99\linewidth]{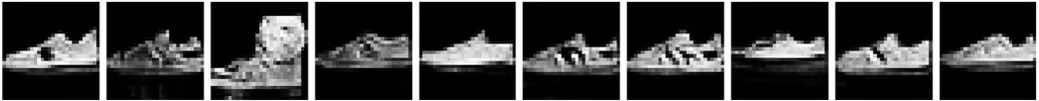}  \\
    \midrule 
     \includegraphics[width=0.99\linewidth]{Figure282930_Input_08.jpeg} &
   \includegraphics[width=0.99\linewidth]{Figure282930_Input_09.jpeg}  \\
   \includegraphics[width=0.99\linewidth]{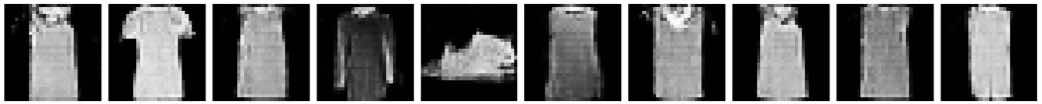} &
    \includegraphics[width=0.99\linewidth]{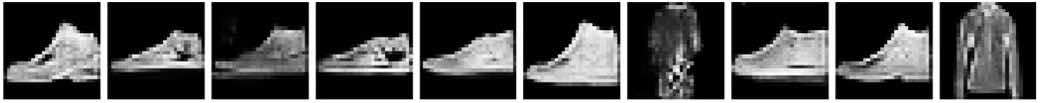}  \\
      \bottomrule
  \end{tabular} 
\caption[]{Images representing the class correspondence between the source and target data in a class-conditional Spider GAN where discriminator is modified to output the class label of the generated images. The Spider GAN generator is not provided class information, but is trained to minimize the classification accuracy. While the model's class-correspondence is superior to that of the baseline Spider GAN, class overlap is not eliminated.}  
 \label{Fig_ccOnlyD}
\end{center}
\vskip-1em
\end{sidewaysfigure}

\begin{sidewaysfigure}
\begin{center}
  \begin{tabular}[b]{P{.45\linewidth}|P{.45\linewidth}}
  \toprule
   \includegraphics[width=0.99\linewidth]{Figure282930_Input_00.jpeg} &
   \includegraphics[width=0.99\linewidth]{Figure282930_Input_01.jpeg}  \\
   \includegraphics[width=0.99\linewidth]{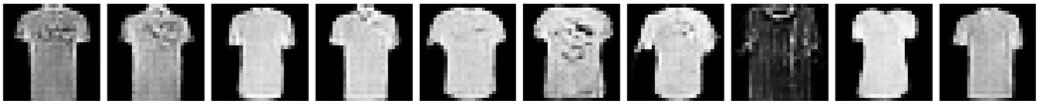} &
    \includegraphics[width=0.99\linewidth]{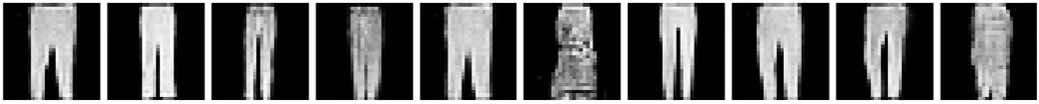}  \\
    \midrule 
     \includegraphics[width=0.99\linewidth]{Figure282930_Input_02.jpeg} &
   \includegraphics[width=0.99\linewidth]{Figure282930_Input_03.jpeg}  \\
   \includegraphics[width=0.99\linewidth]{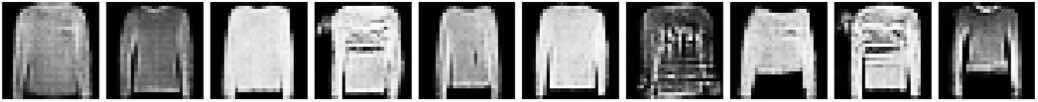} &
    \includegraphics[width=0.99\linewidth]{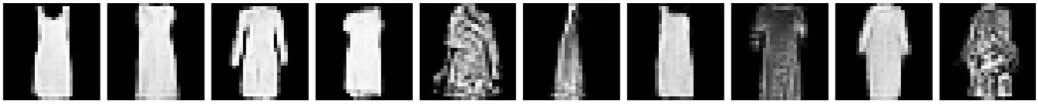}  \\
    \midrule 
     \includegraphics[width=0.99\linewidth]{Figure282930_Input_04.jpeg} &
   \includegraphics[width=0.99\linewidth]{Figure282930_Input_05.jpeg}  \\
   \includegraphics[width=0.99\linewidth]{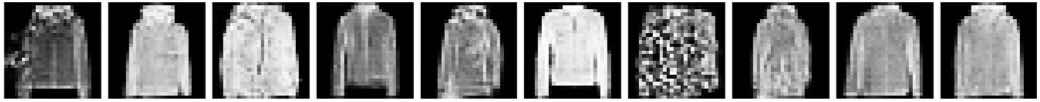} &
    \includegraphics[width=0.99\linewidth]{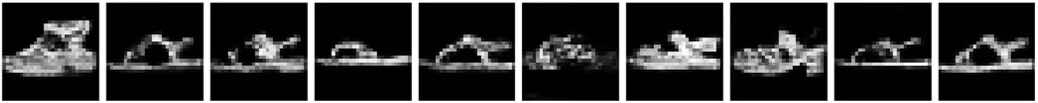}  \\
    \midrule 
     \includegraphics[width=0.99\linewidth]{Figure282930_Input_06.jpeg} &
   \includegraphics[width=0.99\linewidth]{Figure282930_Input_07.jpeg}  \\
   \includegraphics[width=0.99\linewidth]{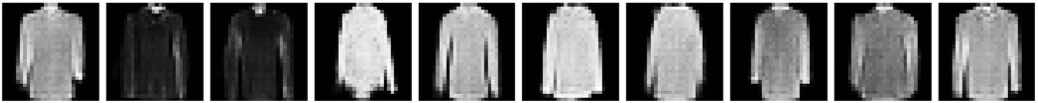} &
    \includegraphics[width=0.99\linewidth]{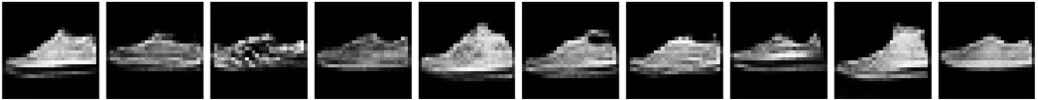}  \\
    \midrule 
     \includegraphics[width=0.99\linewidth]{Figure282930_Input_08.jpeg} &
   \includegraphics[width=0.99\linewidth]{Figure282930_Input_09.jpeg}  \\
   \includegraphics[width=0.99\linewidth]{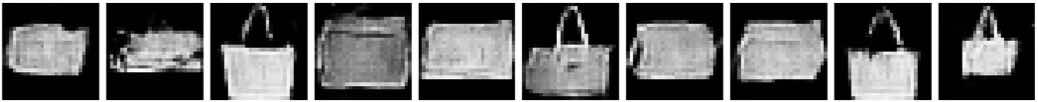} &
    \includegraphics[width=0.99\linewidth]{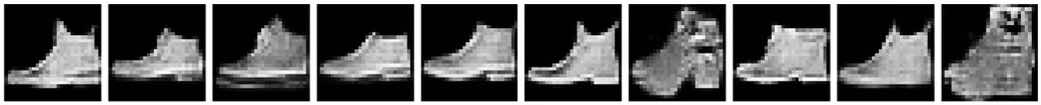}  \\
      \bottomrule
  \end{tabular} 
\caption[]{The class correspondence of the images generated by Spider ACGAN. The discriminator is modified to output the class label of the generated images, while the source class information is provided as an embedding layer to the generator. The Spider ACGAN model achieves superior class disentanglement in comparison to the baselines.}  
 \label{Fig_ccSpiderACGAN}
\end{center}
\vskip-1em
\end{sidewaysfigure}

We present a {\it Spider} counterpart to the auxiliary classifier GAN (ACGAN~\cite{ACGAN17}) formulation, entitled Spider ACGAN. In Spider ACGAN, the discriminator not only provides a {\it real versus fake} classification of its input, but also provides a prediction of the class from which the sample is drawn. The discriminator is trained to minimize both the WGAN loss with the R\(_d\) penalty~\cite{R1R218}, and the classification cross-entropy loss. We consider two variants of the generator, one without class information, and the other with the class label provided as a fully-connected embedding to the input layer. The Spider ACGAN variants are compared with the un-conditional Spider GAN baseline. We present experiments on learning Fashion-MNIST dataset with MNIST as the input. The pairwise correspondences between the input and output images are presented in Figures~\ref{Fig_ccBase}-\ref{Fig_ccSpiderACGAN}. While Spider ACGAN without generator embeddings is superior to the baseline Spider GAN in learning class-level consistency, mixing between the classes is not eliminated. However, with the inclusion of class embeddings in the generator, the disentanglement of classes can be achieved in Spider ACGAN. \par
While this experiment demonstrates the feasibility of employing Spider GAN in class-conditional settings, scenarios involving mismatch between the number of classes in the input and output datasets, is a promising direction for future research.

\begin{figure*}[!t]
\begin{center}
        \includegraphics[width=0.85\linewidth]{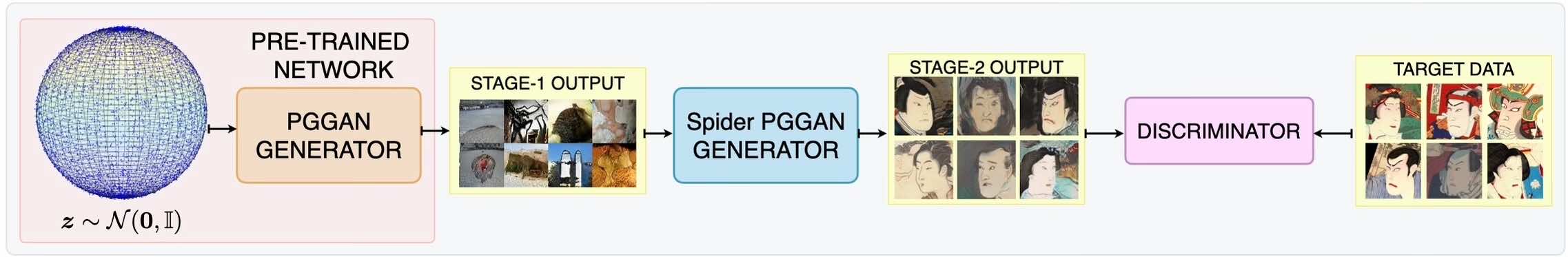} 
  \caption[]{Spider GAN based progressively growing GAN (PGGAN) architecture. The output distribution of PGGAN trained on Tiny-ImageNet data is provided as input to the second Spider PGGAN stage that is trained to learn a high-resolution, small-sized dataset such as Ukiyo-E Faces.} 
   \vspace{-1.em}
  \label{Fig_SpiderPGGAN}  
  \end{center}
  \vskip-1em
\end{figure*}

\subsection{Additional Experiments on Spider PGGAN} \label{App_ExpPGGAN}

\begin{sidewaysfigure}

\begin{center}
    \includegraphics[width=0.99\linewidth]{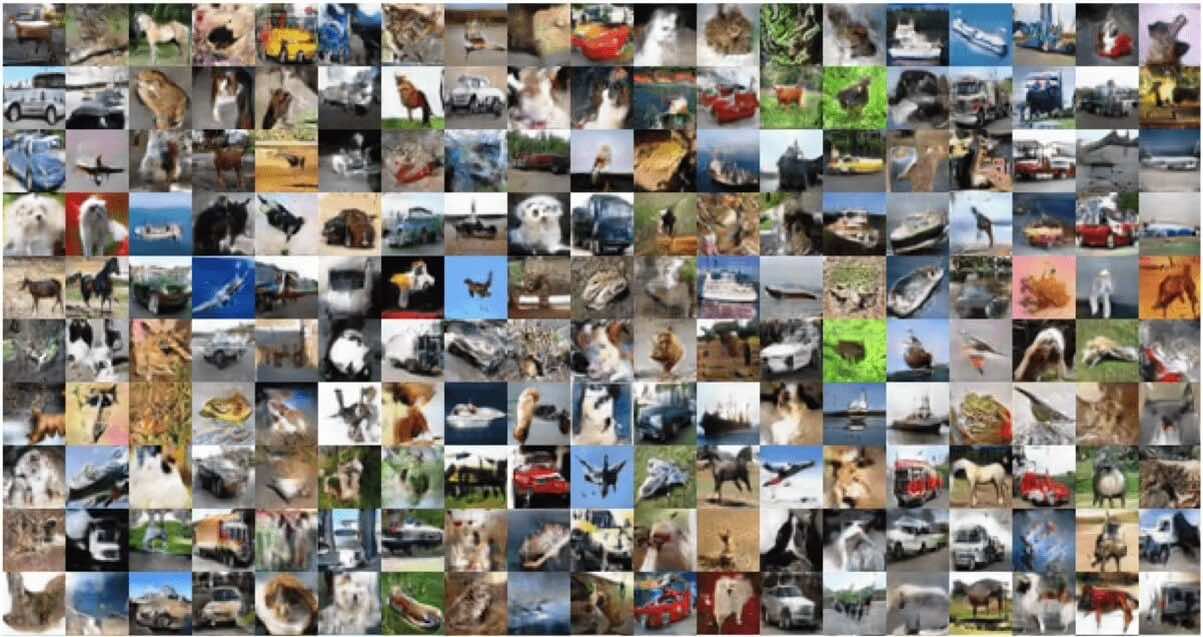}  \\[10pt]
\caption[]{Images generated by Spider PGGAN with CIFAR-10 as the target, when trained with Tiny-ImageNet as the input, which in turn is drawn from a PGGAN trained with a Gaussian input. The weights of the Spider PGGAN are also initialized with the weights of the input PGGAN, resulting in faster training and superior output image quality.}
\label{Fig_SpiderPGGAN_C10Wts}  
\end{center}
\vskip-1em
\end{sidewaysfigure}

We now present additional experiments conducted with the Spider PGGAN architecture, and present the images generated by the Spider PGGAN variants. Figure~\ref{Fig_SpiderPGGAN} depicts the philosophy employed in a two-stage cascaded Spider PGGAN model considered in Section~\ref{Sec_SpiderPGStyleGAN}, where the input-stage PGGAN generated Tiny-ImageNet images, while the second Spider PGGAN stage transforms Tiny-ImageNet into Ukiyo-E Faces. Consider two extensions of the Spider PGGAN training algorithms: (a) The Spider PGGAN is trained on \(32\times32\times3\) CIFAR-10 data with the input images drawn from the output of a PGGAN pre-trained on Tiny-ImageNet. Additionally, weights from PGGAN pre-trained on Tiny-ImageNet are transferred to  Spider PGGAN for all layers but the first because the dimensionality in the first layer does not match. The trained model achieves an FID of 9.56, which is an improvement over the base Spider GAN trained on CIFAR-10 without the weight transfer. Images generated by Spider PGGAN with weight transfer are shown in Figure~\ref{Fig_SpiderPGGAN_C10Wts}. This suggests that other network modifications and augmentations can be used in combination with the Spider GAN framework to improve the performance of PGGAN. (b) We train the Spider PGGAN with multiple cascade layers. The output of a Stage-I PGGAN pre-trained on Tiny-ImageNet is used to train a Spider PGGAN (Stage-II) to generate CIFAR-10 images. The output of the converged second stage model is used to generate high-resolution Ukiyo-E and MetFaces images (Stage-III). The final model achieves an FID of 45.32 on MetFaces (a 12\% improvement over a single-stage Spider PGGAN), and 57.63 on Ukiyo-E Faces (a 10\% improvement over single stage).  The MetFaces images generated by the cascade network, juxtaposed the images generated by the baseline methods are provided in  Figure~\ref{Fig_MetFaces_Compares}.  These results suggest that having multiple stages of pre-trained networks in the Spider PGGAN, and training incrementally results in superior performance than a single-stage Spider PGGAN.

\subsection{Additional Experimental on Spider StyleGAN} \label{App_ExpStyleGAN}

The philosophy behind StyleGAN~\cite{ADAStyleGAN20,StyleGAN321} architectures run parallel to our proposed philosophy, where a {\it mapping} network is used to learn editable intermediate representations of the input noise distribution. A {\it synthesis} network subsequently transforms this representation into an image. The Spider GAN approach can be incorporated readily into any StyleGAN network, by replacing the input noise distribution to the mapping network with samples from the input dataset, drawn from a pre-trained GAN. \par

\begin{figure*}[!thb]
\begin{center}
  \begin{tabular}[b]{P{.01\linewidth}|P{.38\linewidth}|P{.38\linewidth}}
     & (a) Baseline sampling & (b) Sampling with the truncation trick~\cite{BIGGAN18} \\[1pt] \hline
      \rotatebox{90}{\qquad\qquad\qquad\qquad FFHQ} &\includegraphics[width=0.99\linewidth]{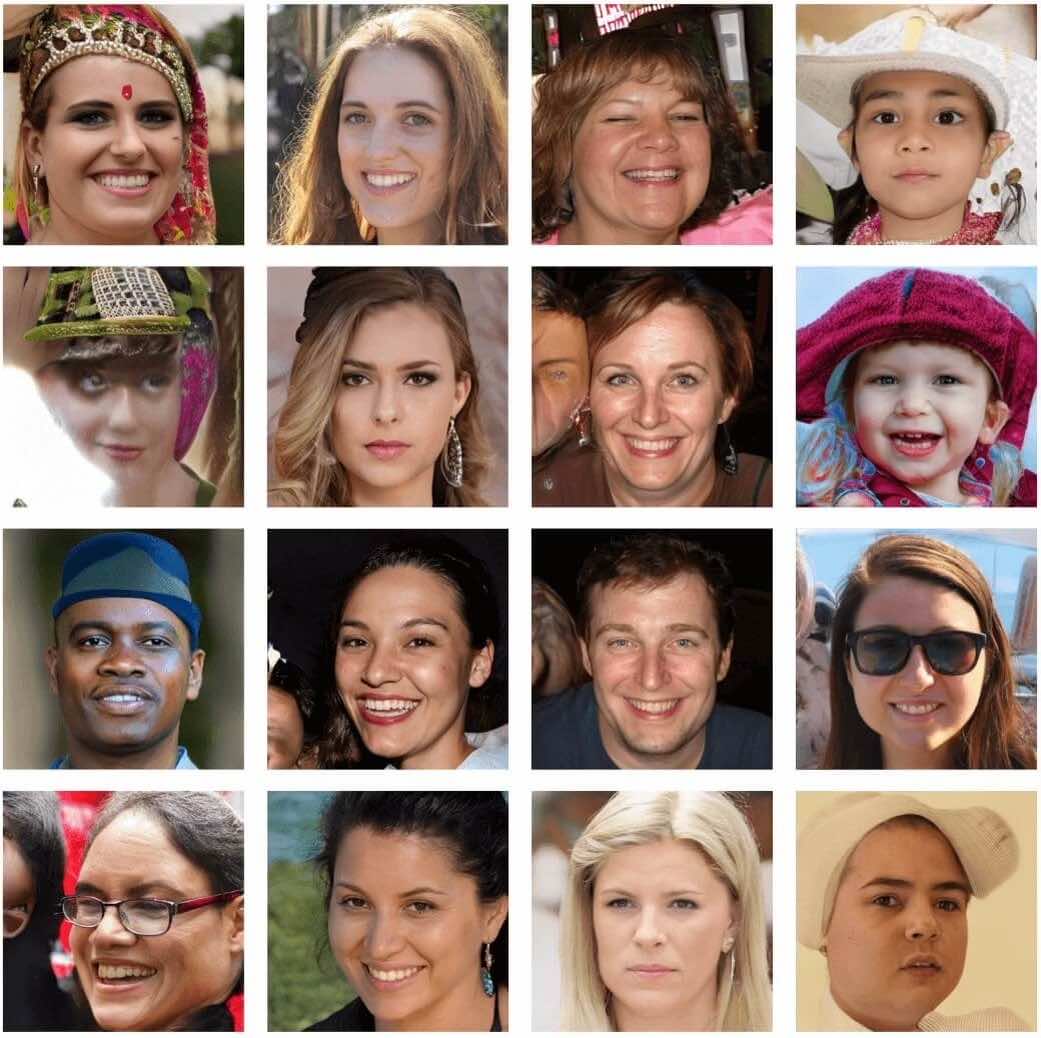} & 
    \includegraphics[width=0.99\linewidth]{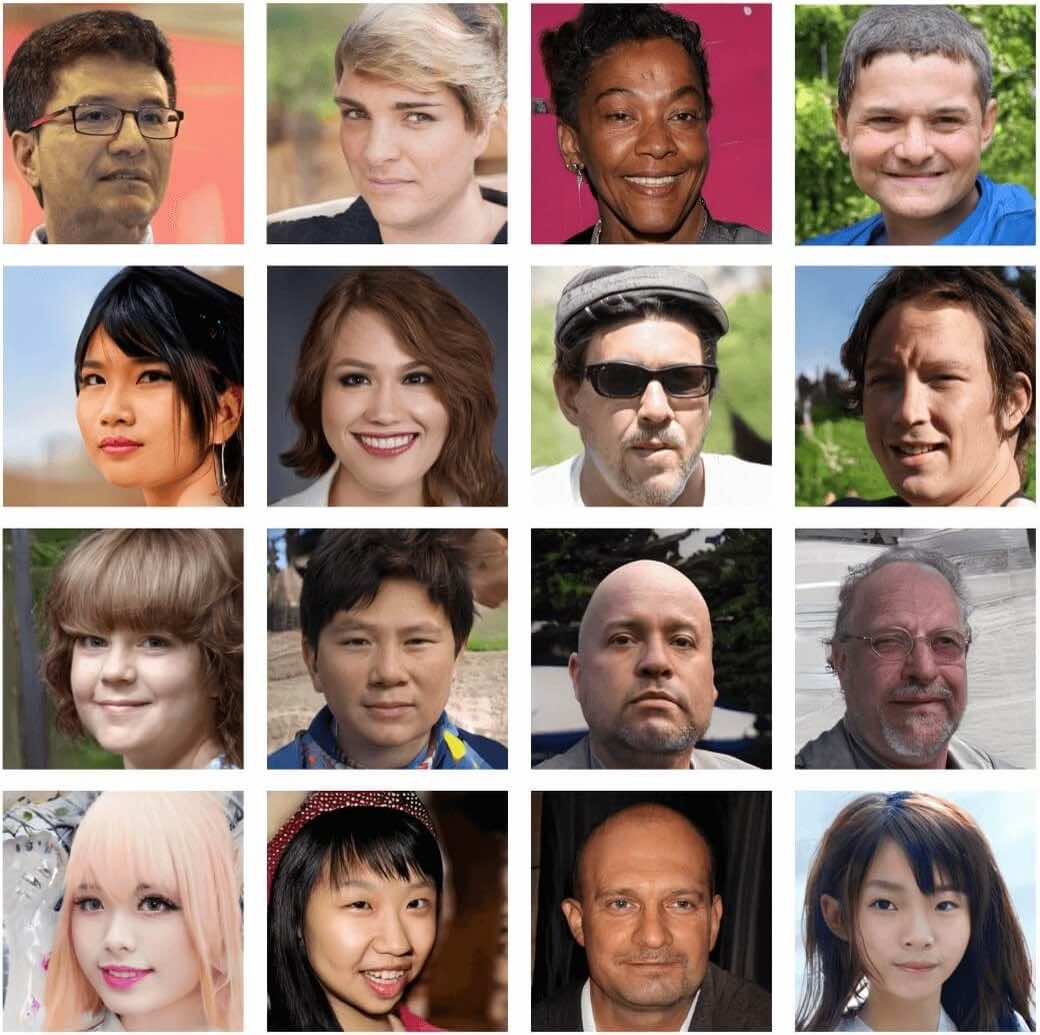}    \\[0.25pt] \midrule \\[-10pt]
     \rotatebox{90}{\qquad\qquad\qquad\quad\enskip Ukiyo-E Faces} &\includegraphics[width=0.99\linewidth]{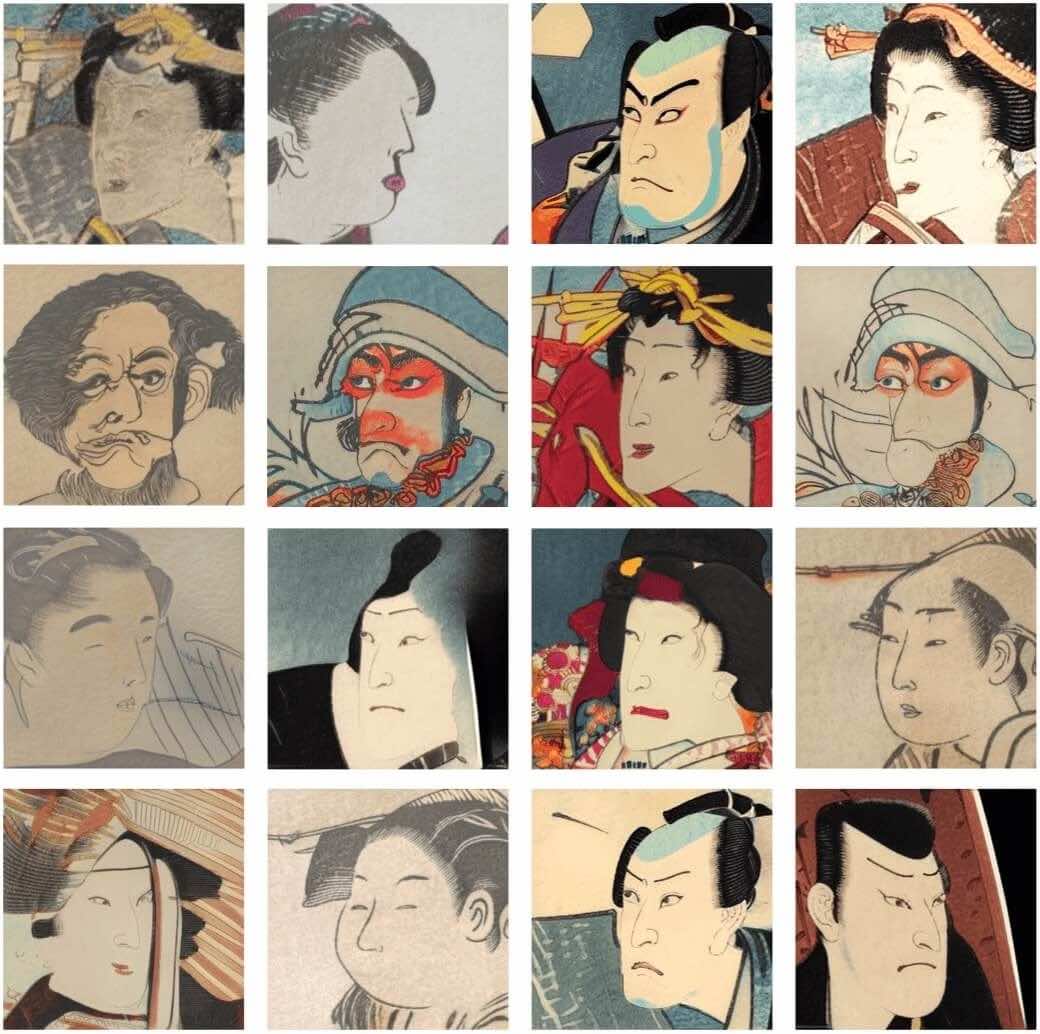} & 
    \includegraphics[width=0.99\linewidth]{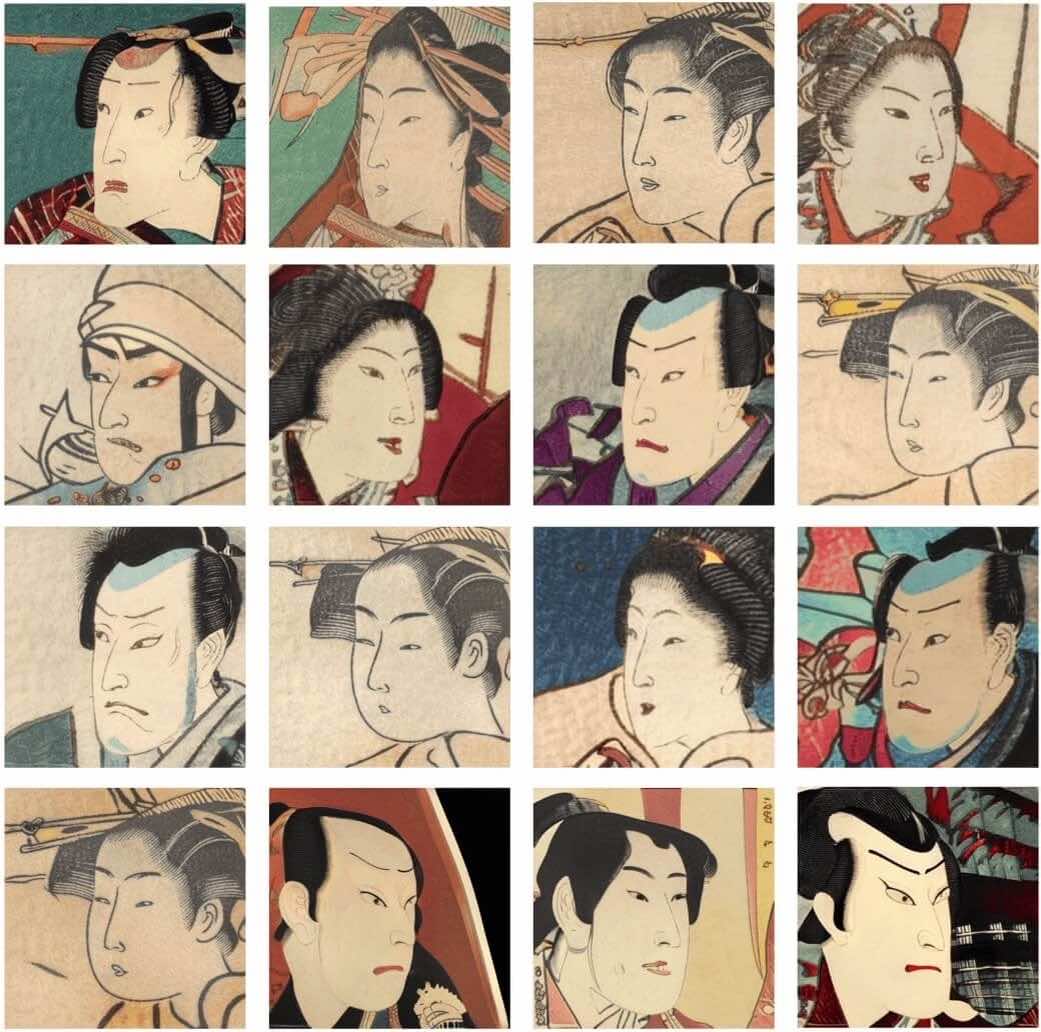}    \\[0.25pt] \midrule \\[-10pt]
     \rotatebox{90}{\qquad\qquad\qquad\quad MetFaces} &\includegraphics[width=0.99\linewidth]{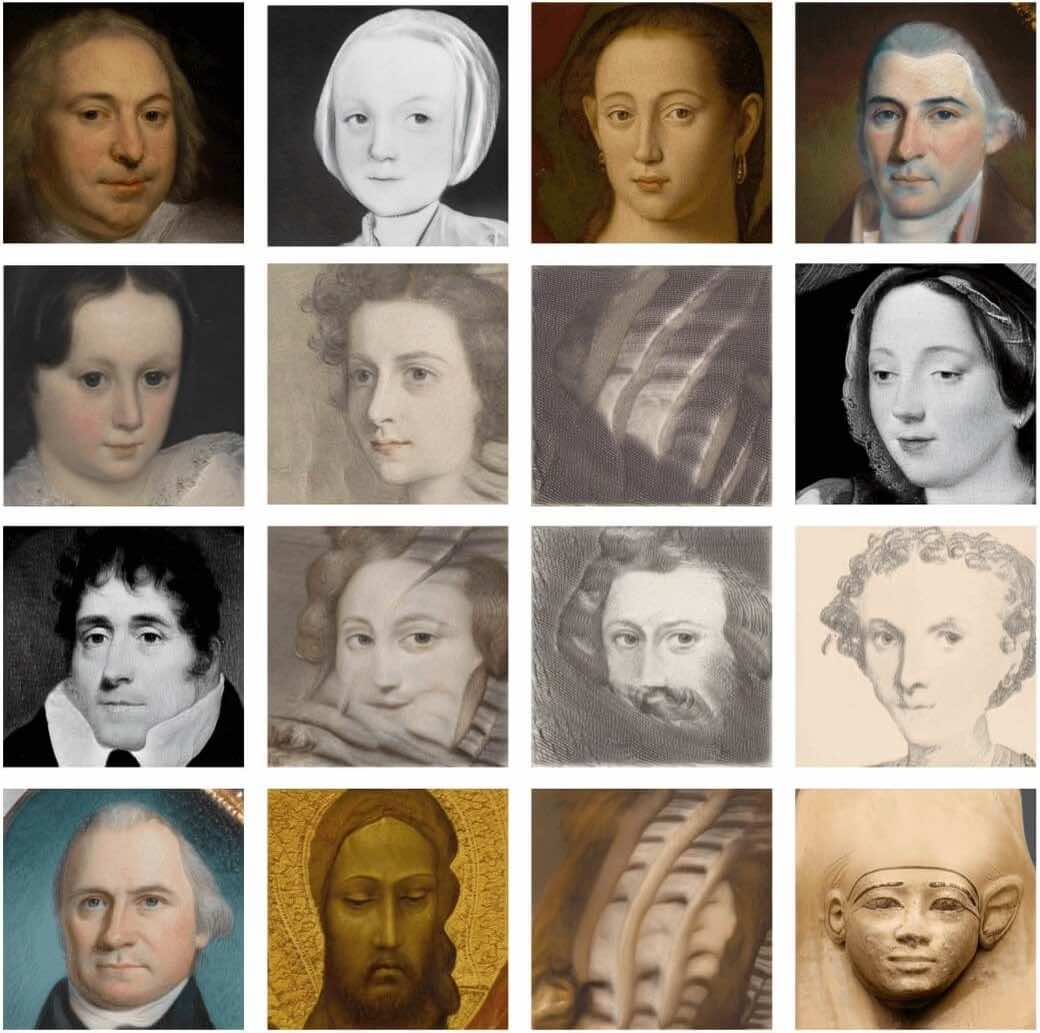} & 
    \includegraphics[width=0.99\linewidth]{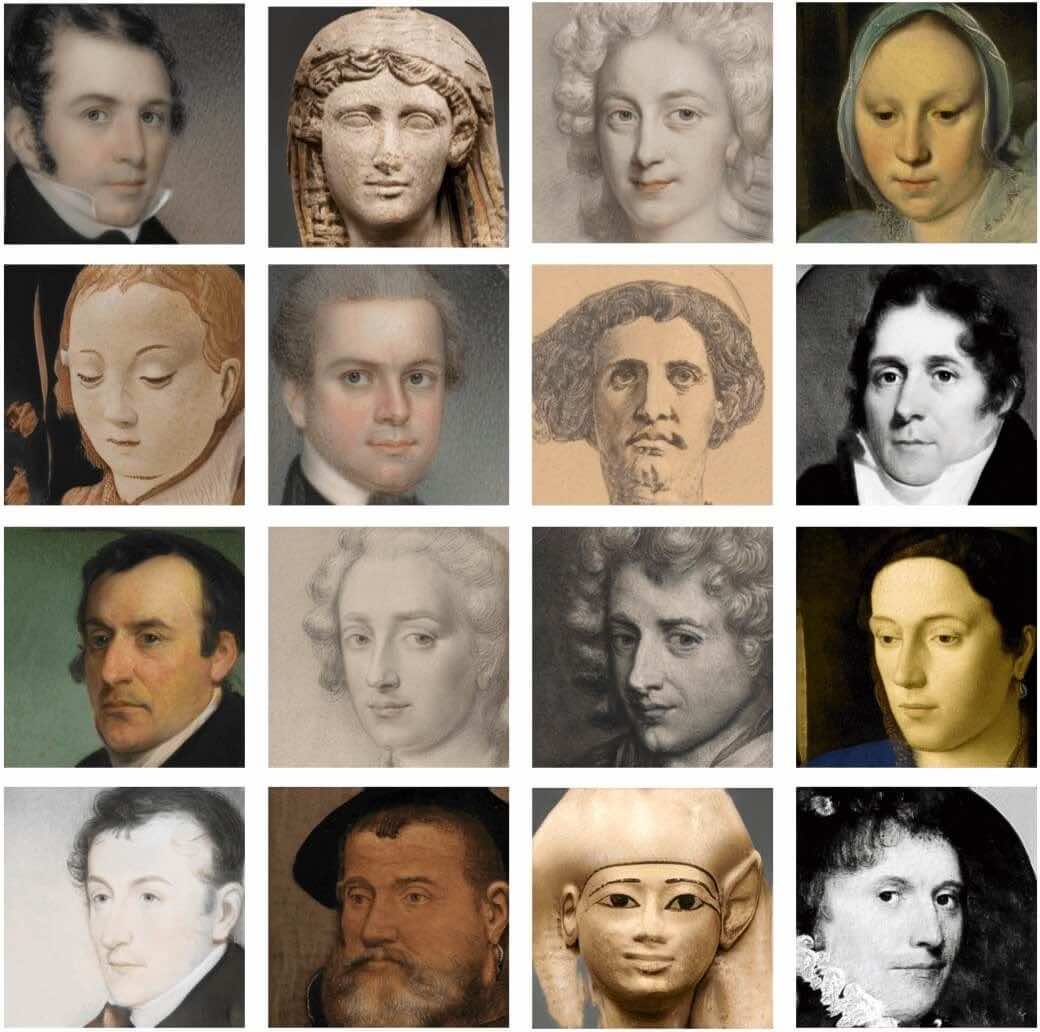}    
  \end{tabular} 
\caption[]{Images generated by cascaded Spider GAN variants when the Gaussian samples provided to the input-stage are (a) retained as-is; and (b) resampled when lying outside of the \(2\sigma\) interval \([-2,2]\) ({\it the truncation trick}~\cite{BIGGAN18}). Images generated using truncated input samples are of a superior visual quality. Baseline sampling results in distorted faces in the case of FFHQ and Ukiyo-E faces datasets, while on MetFaces, poor quality samples resulted in {\it alien} patterns.}
\label{Fig_Trunc}  
\end{center}
\vskip-1em
\end{figure*}

We trained the {\it Spider} variants of StyleGAN2, StyleGAN2-ADA~\cite{ADAStyleGAN20} and StyleGAN3~\cite{StyleGAN321} on the Ukiyo-E Faces, MetFaces, FFHQ, animal faces HQ Cats (AFHQ-Cats) dataset using the various combinations that included adaptive regularization and weight transfer. Across all experiments, two pre-trained networks were employed to generate the input dataset distribution -- (i) A StyleGAN2-ADA network trained on the AFHQ-Dogs dataset of resolution \(32\times32\); and (ii) A StyleGAN2-ADA network trained on the Tiny-ImageNet dataset of resolution \(32\times32\). The outputs are transformed based on the approach described in Appendix~\ref{App_ExpSetup}. To generate higher-quality samples, we adopted the popular {\it truncation trick}~\cite{BIGGAN18} in sampling from the input-stage generator -- The input-stage baseline generator is trained to transform samples drawn from the standard normal distribution to those coming from Tiny-ImageNet, or AFHQ-Dogs datasets. When generating the inputs to the cascaded Spider GAN stage, samples are drawn from a truncated Gaussian, where a sample is re-drawn if it lies outside the \([-2,2]^n\) hypercube (a \(2\sigma\) interval). This was shown to improve the generator output quality at a small cost of marginally reduced sample diversity~\cite{BIGGAN18}. On the experiments on learning Ukiyo-E Faces, MetFaces, and  FFHQ with cascaded Spider StyleGAN2-ADA, the truncation trick resulted in a 10\% improvement in FID on the average. Figure~\ref{Fig_Trunc} presents the images generated by these models considering baseline sampling and the truncation trick.  \par

\begin{figure*}[!thb]
\begin{center}
  \begin{tabular}[b]{P{.9\linewidth}}
    \includegraphics[width=1.\linewidth]{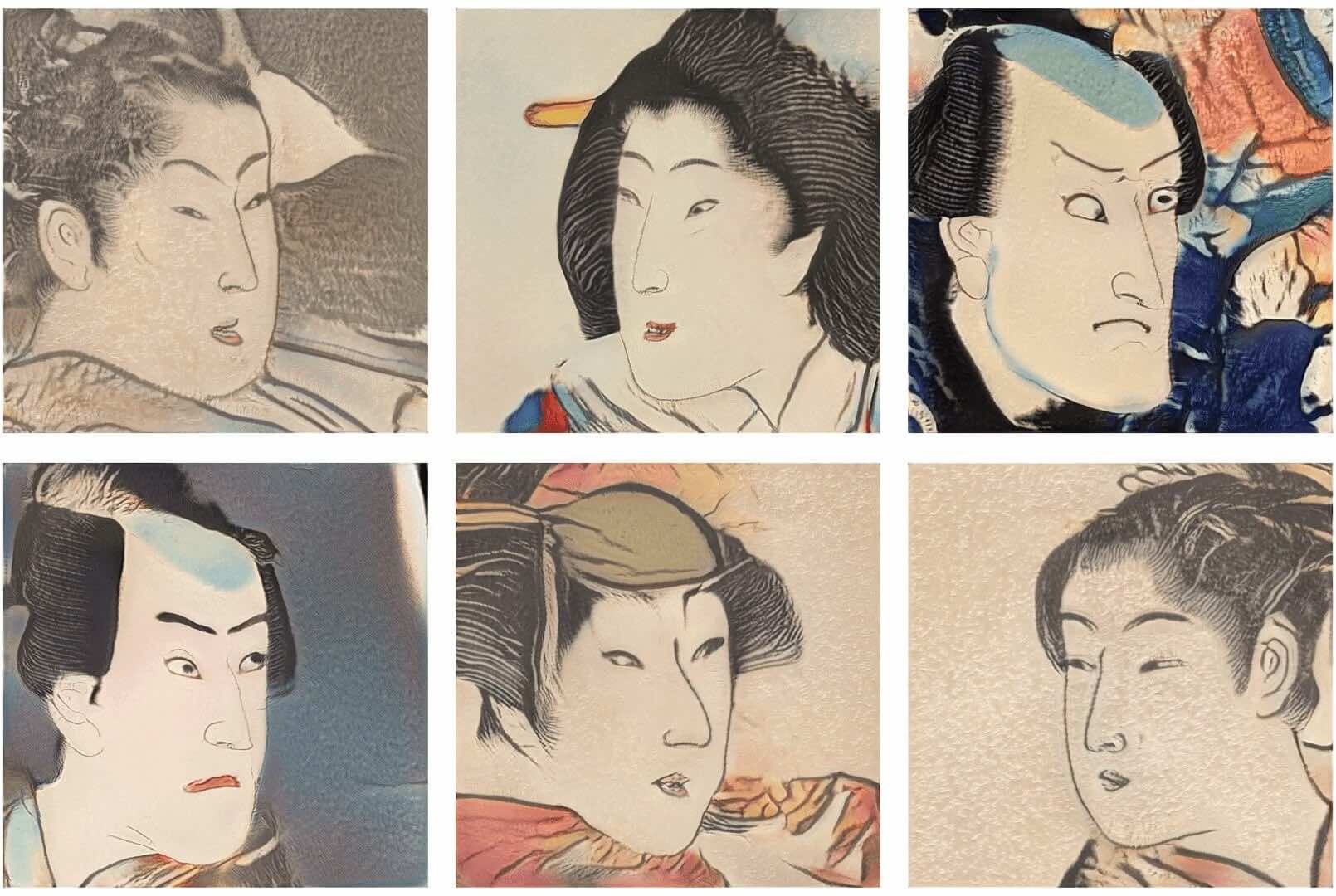}  \\[1.5pt]
    \includegraphics[width=1.\linewidth]{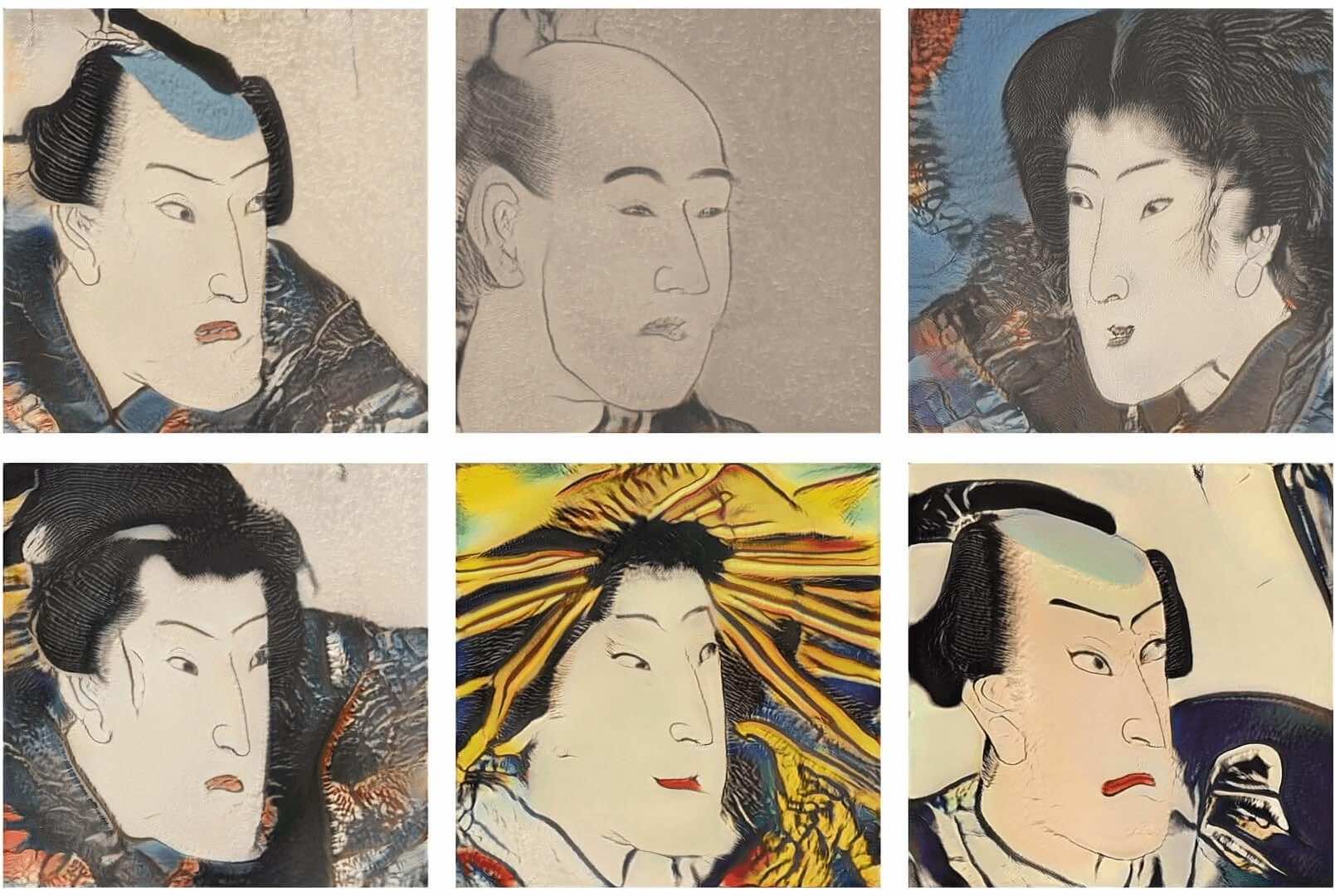}  
  \end{tabular} 
\caption[]{Ukiyo-E images generated by the {\it Spider} variant of StyleGAN2, trained on AFHQ-Dogs input. Since the AGFQ-Dogs dataset has relatively lower diversity than the target, the generated  Ukiyo-E samples are visually sup-par compared to the performance of the baseline StyleGAN2-ADA. }
\label{Fig_UkiyoES2Dogs}  
\end{center}
\vskip-1em
\end{figure*}

\begin{figure*}[!thb]
\begin{center}
  \begin{tabular}[b]{P{.9\linewidth}}
    \includegraphics[width=1.\linewidth]{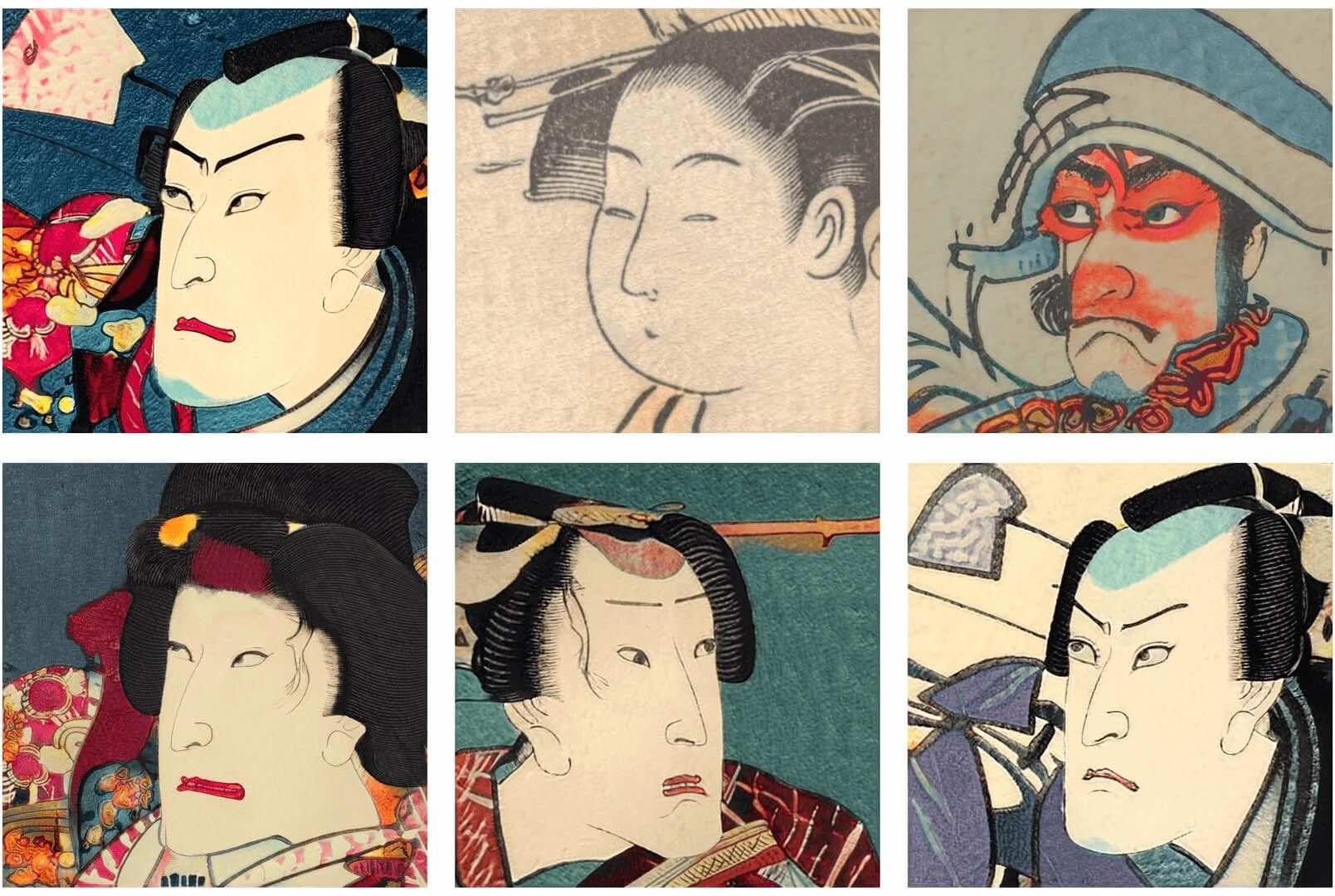}  \\[1.5pt]
    \includegraphics[width=1.\linewidth]{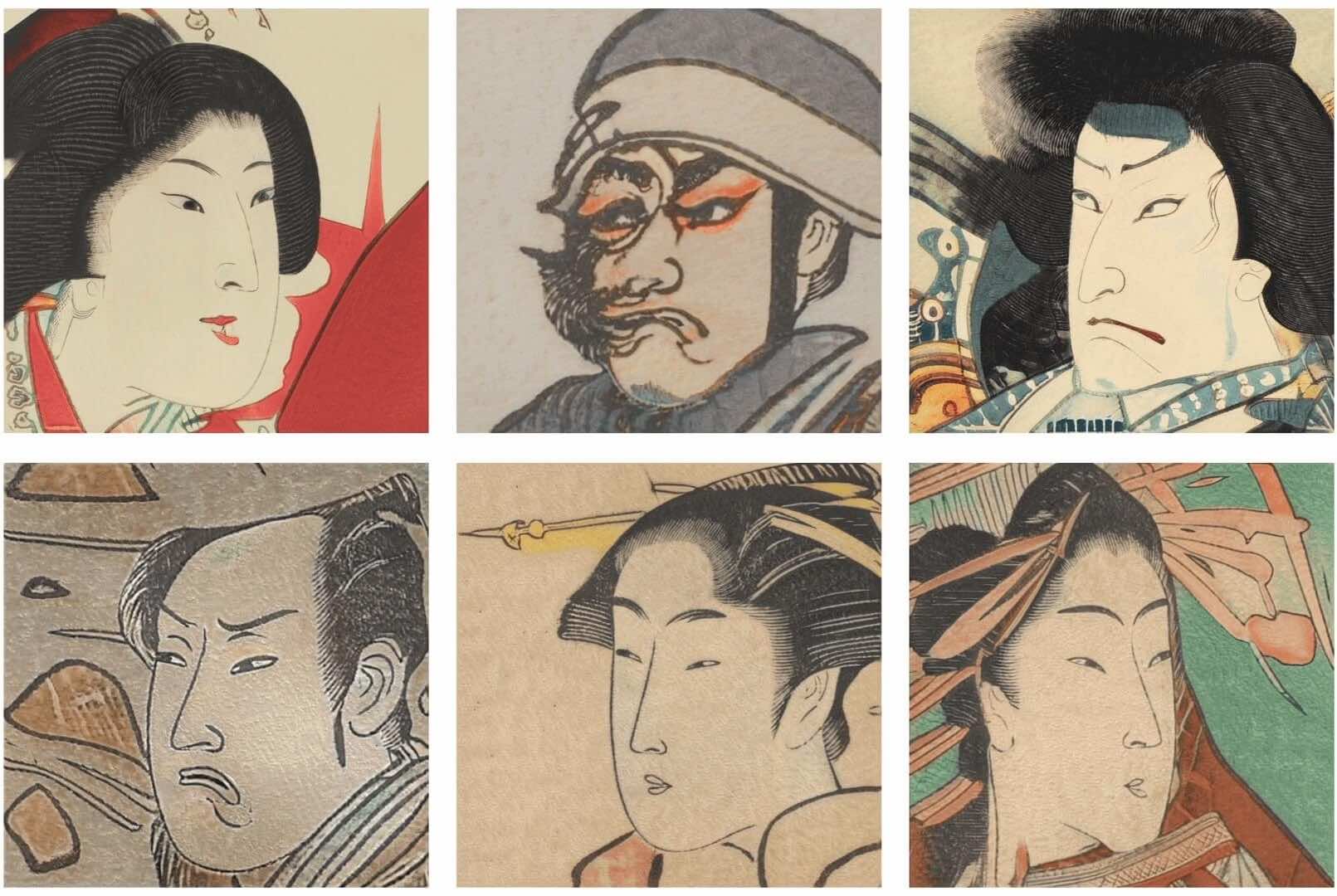}  
  \end{tabular} 
\caption[]{Ukiyo-E face images generated by the {\it Spider} variant of StyleGAN2, trained on Tiny-ImageNet input. The Spider variant achieves state-of-the-art FID of 20.44, compared to 26.74 of the baseline StyleGAN2-ADA (lower FID is better).  }
\label{Fig_UkiyoES2TIN}  
\end{center}
\vskip-1em
\end{figure*}

\begin{figure*}[!thb]
\begin{center}
  \begin{tabular}[b]{P{.02\linewidth}||P{.875\linewidth}}
  \rotatebox{90}{\quad\quad\enskip\footnotesize{StyleGAN2}} &
    \includegraphics[width=1.\linewidth]{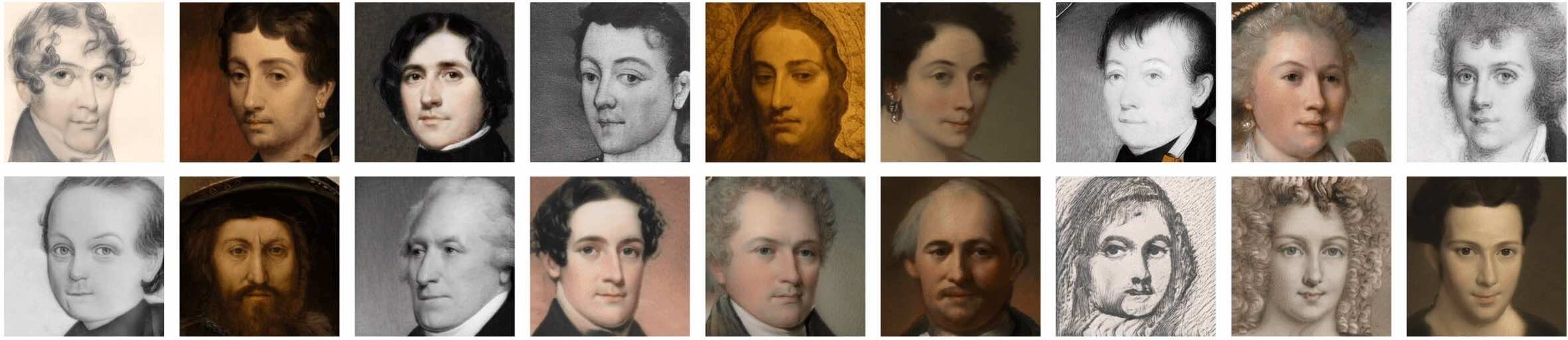}  
    \\[0.25pt] \midrule \\[-10pt]
    \rotatebox{90}{\quad\quad\footnotesize{StyleGAN3-T}} &
    \includegraphics[width=1.\linewidth]{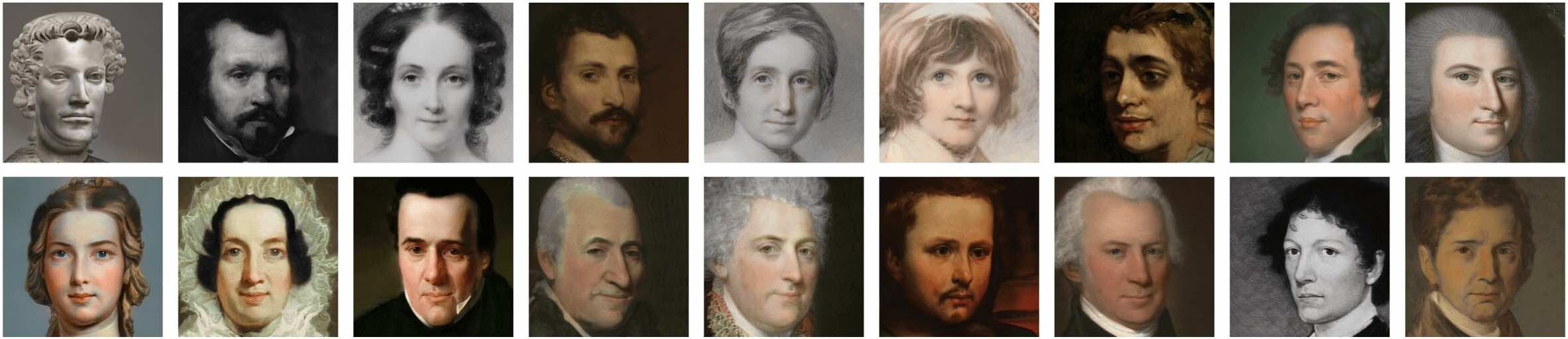}  
    \\[0.25pt] \midrule \\[-10pt]
    \rotatebox{90}{~~\footnotesize{Spider PGGAN (1 Stage)}}&
        \includegraphics[width=1.\linewidth]{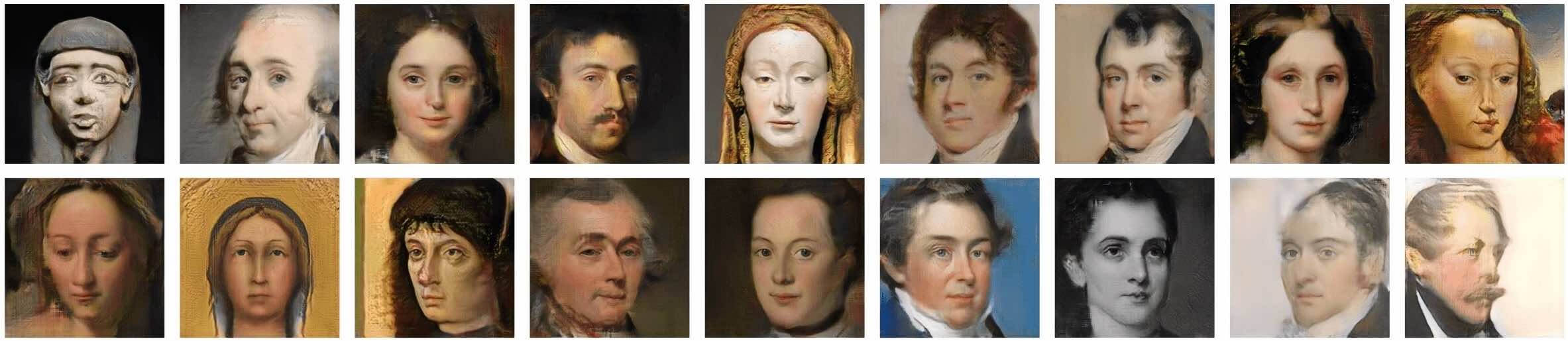}  \\[0.25pt] \midrule \\[-10pt]
        \rotatebox{90}{~~\footnotesize{Spider PGGAN (2 Stage)} }& 
    \includegraphics[width=1.\linewidth]{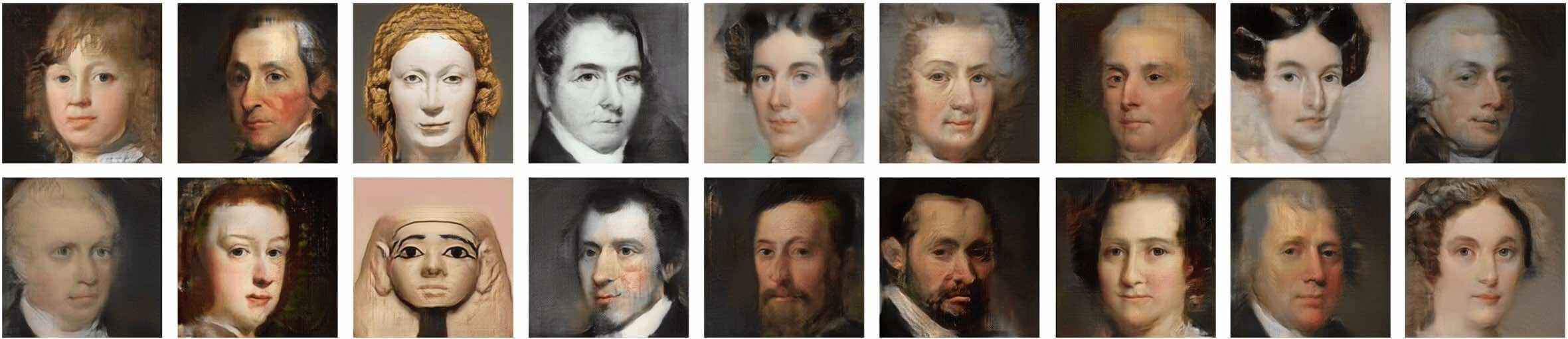}  
    \\[0.25pt] \midrule \\[-10pt]
    \rotatebox{90}{~~\footnotesize{Spider StyleGAN2 (Dogs)}} &
    \includegraphics[width=1.\linewidth]{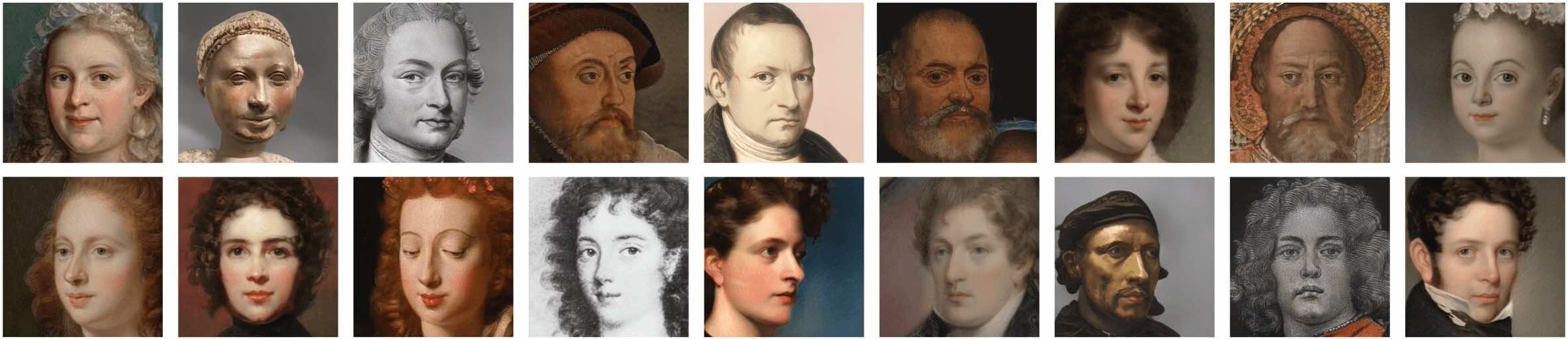}  
    \\[0.25pt] \midrule \\[-10pt]
    \rotatebox{90}{~~\footnotesize{Spider StyleGAN2 (TIN)}} &
    \includegraphics[width=1.\linewidth]{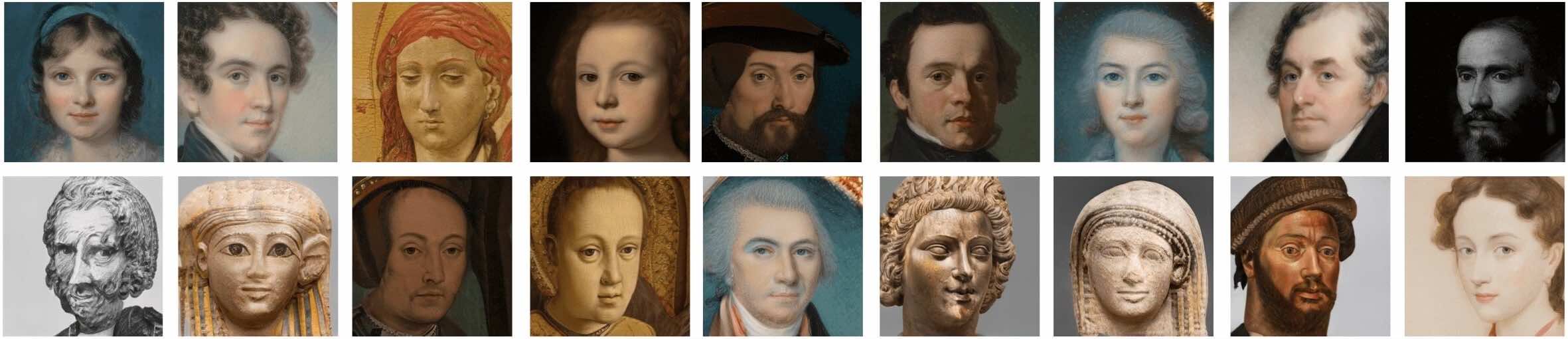}  
    \\[-10pt]
  \end{tabular} 
\caption[]{A comparison of MetFaces images generated by the baseline and Spider GAN variants. Spider StyleGAN2 with the Tiny-ImageNet (TIN) input data outperforms all other variants, generating sharper and more diverse images, achieving a state-of-the-art FID of 15.60 as opposed to an FID of 18.75 achieved by StyleGAN2-ADA.  }
\label{Fig_MetFaces_Compares}  
\end{center}
\vskip-1em
\end{figure*}

\begin{figure*}[!thb]
\begin{center}
  \begin{tabular}[b]{P{.9\linewidth}}
    \includegraphics[width=1.\linewidth]{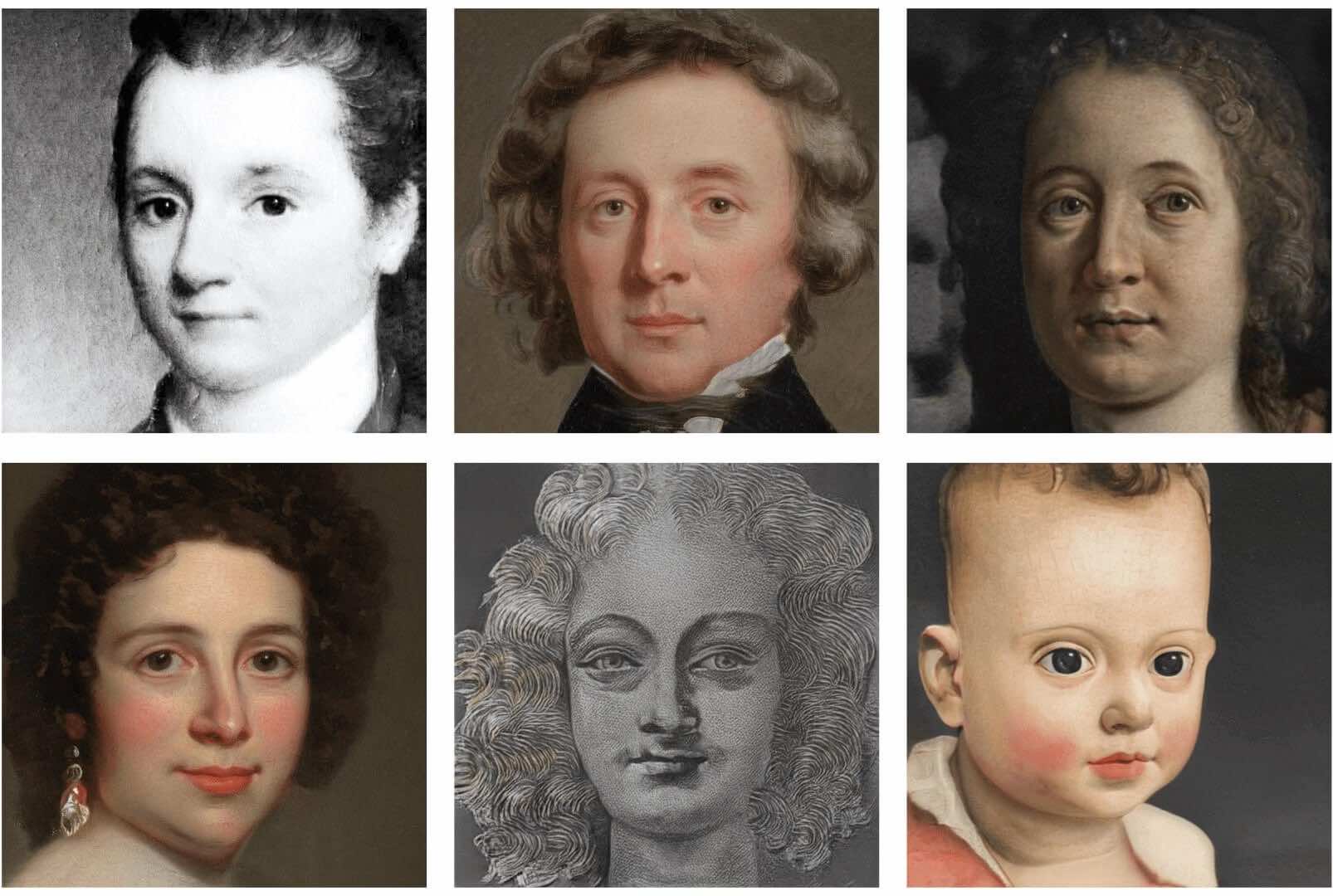}  \\[1.5pt]
    \includegraphics[width=1.\linewidth]{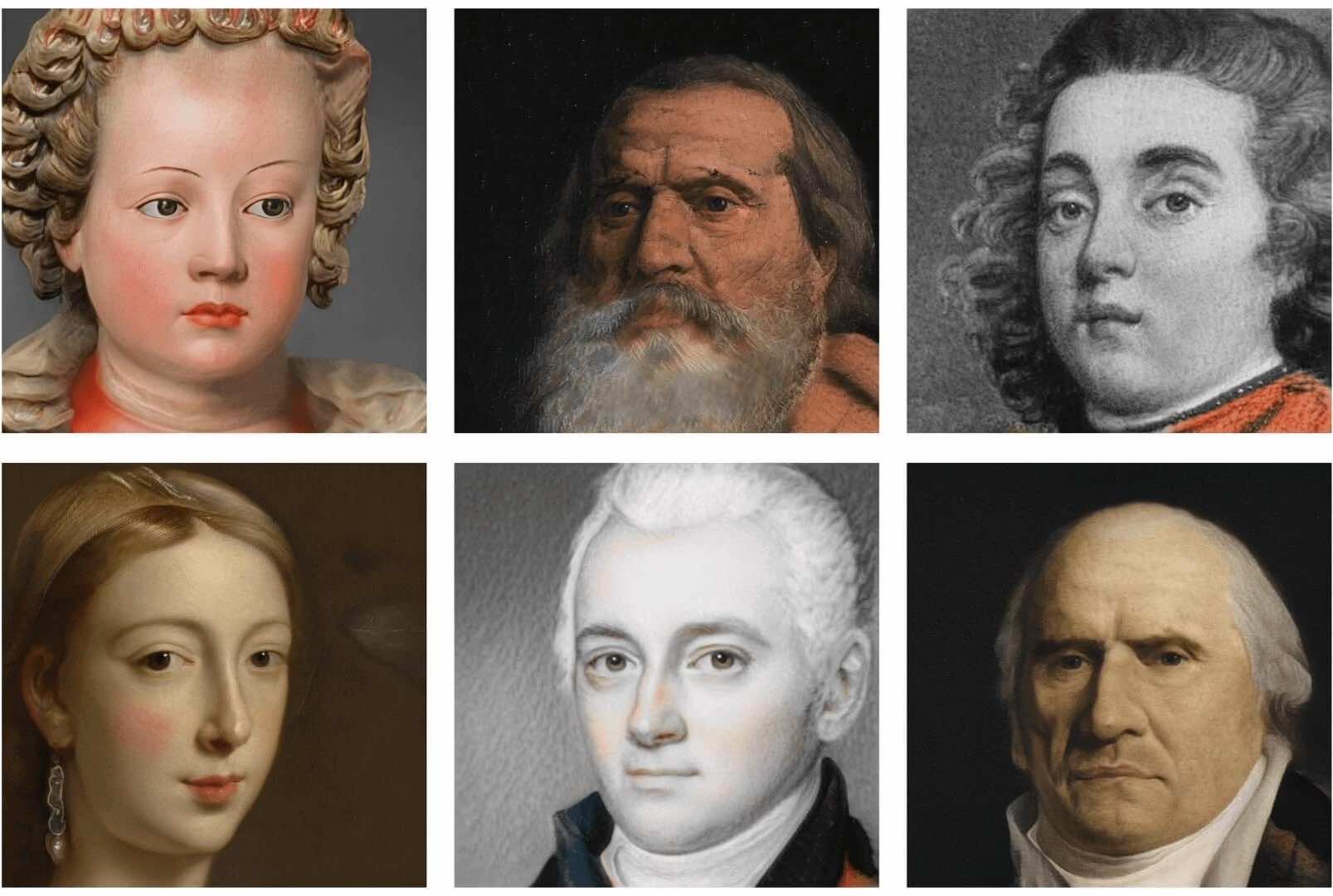}  
  \end{tabular} 
\caption[]{Representative MetFaces images generated by the {\it Spider} variant of StyleGAN2, trained on AFHQ-Dogs input. The model achieved an FID score of 29.82, which is lower than the FID of the StyleGAN2-ADA baseline (18.75). This is expected, as the AFHQ-Dogs is not a {\it friendly neighbor} of the target dataset. }
\label{Fig_MetFacesS2Dogs}  
\end{center}
\vskip-1em
\end{figure*}

\begin{figure*}[!thb]
\begin{center}
  \begin{tabular}[b]{P{.9\linewidth}}
    \includegraphics[width=1.\linewidth]{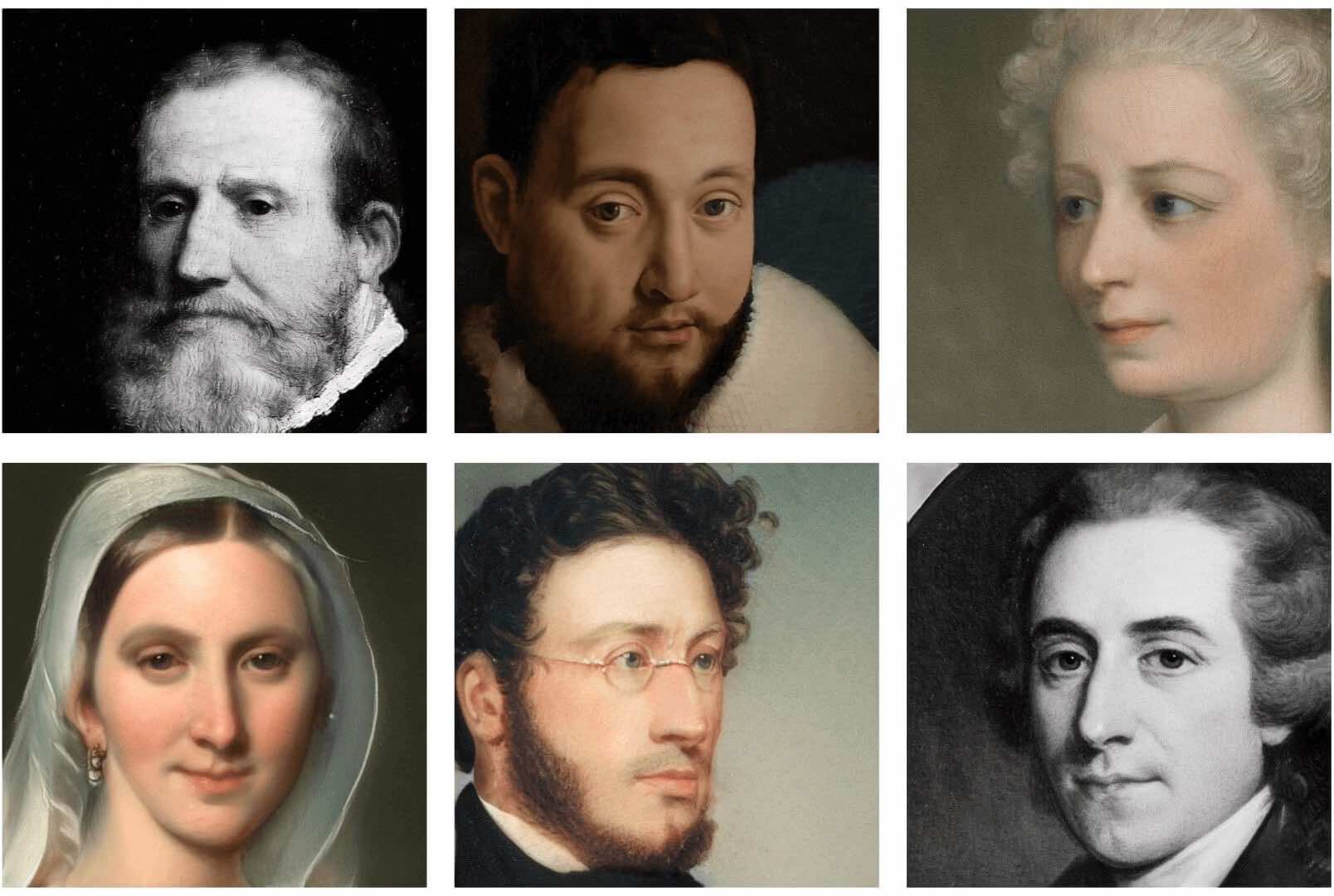}  \\[1.5pt]
    \includegraphics[width=1.\linewidth]{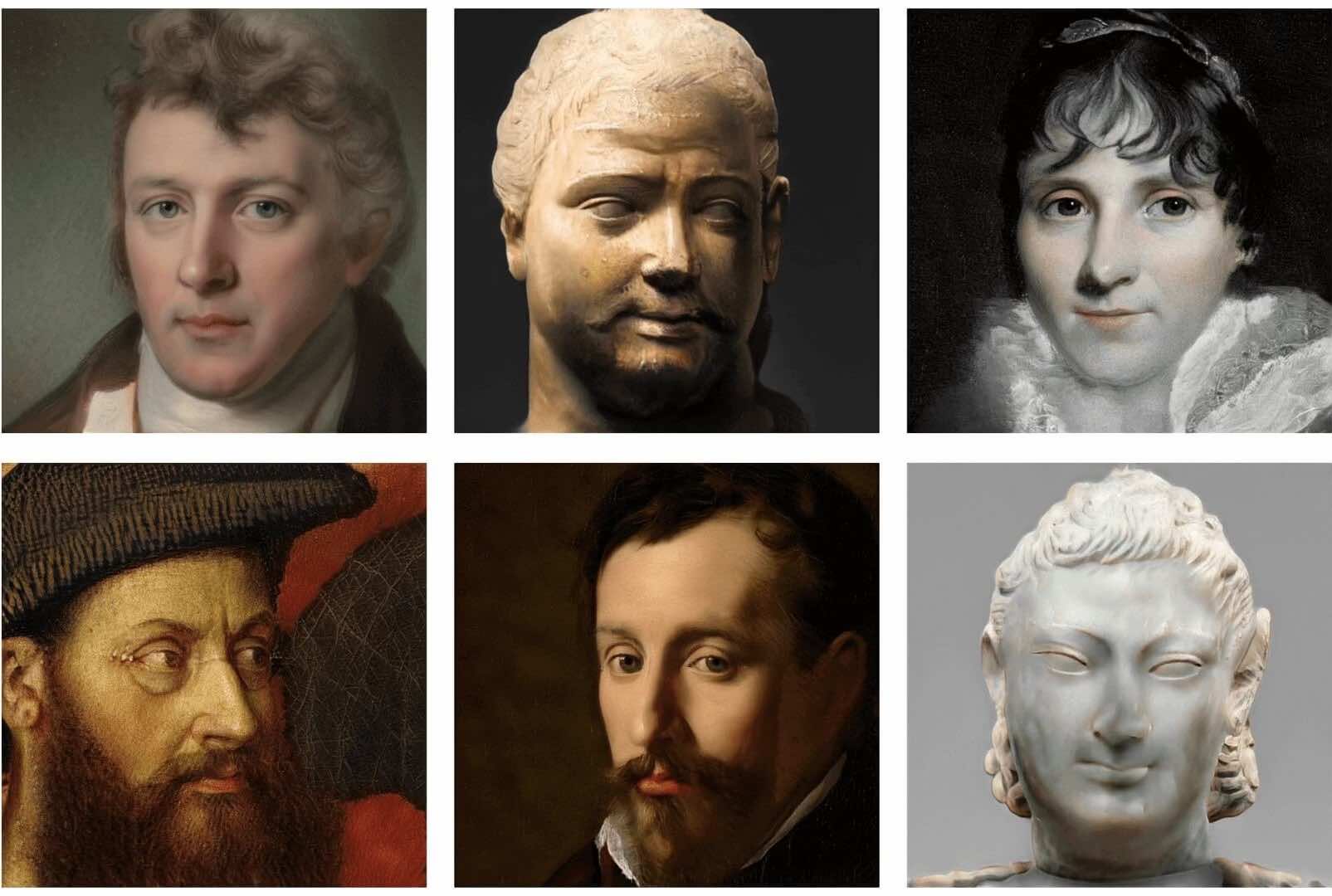}  
  \end{tabular} 
\caption[]{Sample images generated by the {\it Spider} variant of StyleGAN2, trained on Tiny-ImageNet input and MetFaces as output. The Spider StyleGAN variant achieves state-of-the-art FID of 15.60, against an FID of 18.75 achieved by the StyleGAN2-ADA baseline. }
\label{Fig_MetFacesS2TIN}  
\end{center}
\vskip-1em
\end{figure*}

\begin{figure*}[!thb]
\begin{center}
  \begin{tabular}[b]{P{.02\linewidth}||P{.875\linewidth}}
  \rotatebox{90}{\quad\quad\enskip\footnotesize{StyleGAN2}} &
    \includegraphics[width=1.\linewidth]{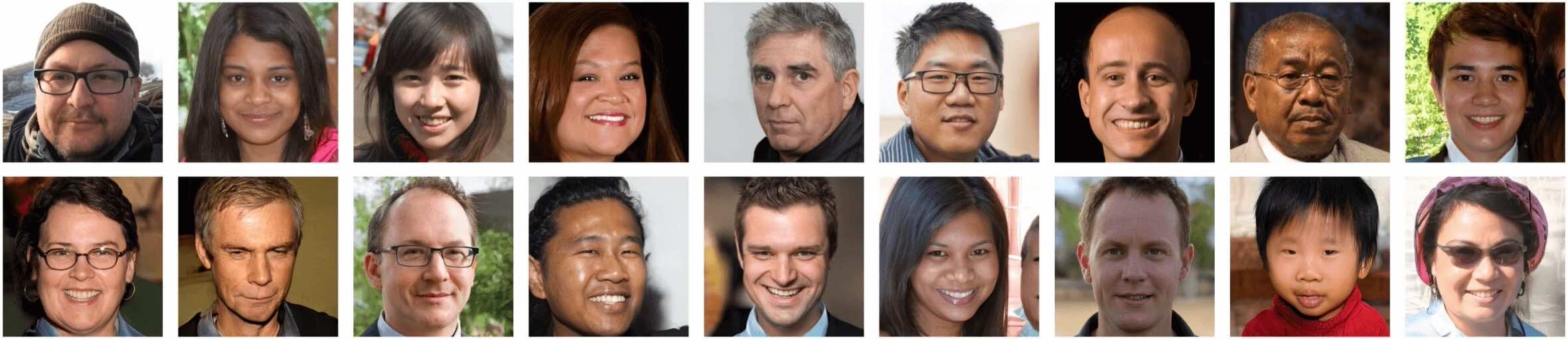}  
    \\[0.25pt] \midrule \\[-10pt]
    \rotatebox{90}{\quad\quad\footnotesize{StyleGAN3-T}} &
    \includegraphics[width=1.\linewidth]{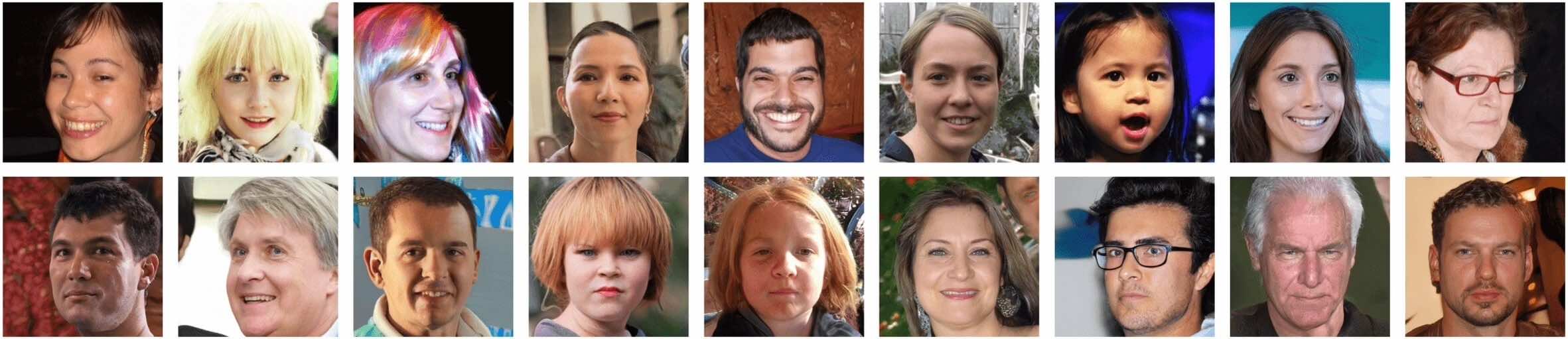}  
    \\[0.25pt] \midrule \\[-10pt]
    \rotatebox{90}{\quad\quad \footnotesize{StyleGAN2-XL}}&
        \includegraphics[width=1.\linewidth]{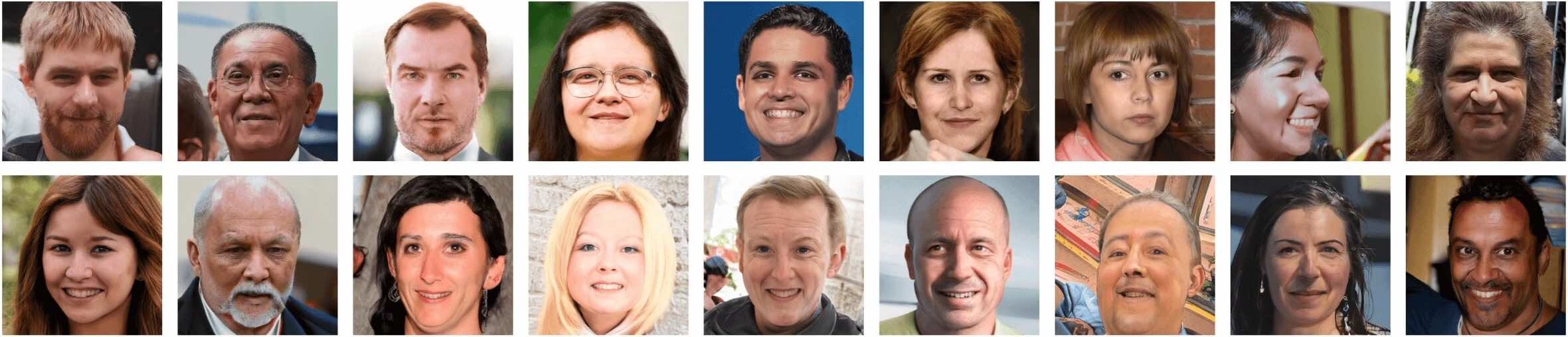}  \\[0.25pt] \midrule \\[-10pt]
        \rotatebox{90}{~~\footnotesize{Spider StyleGAN2 (Dogs)} }& 
    \includegraphics[width=1.\linewidth]{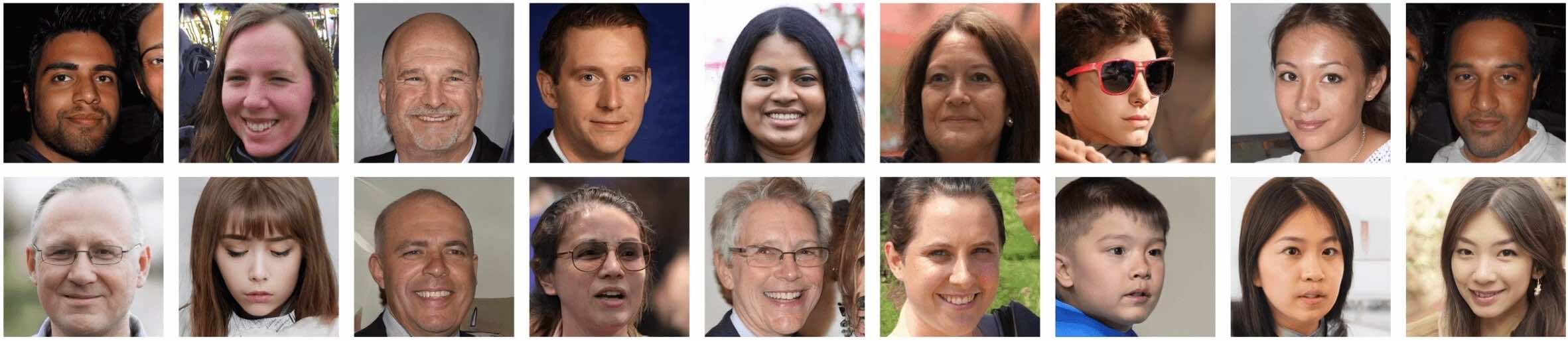}  
    \\[0.25pt] \midrule \\[-10pt]
    \rotatebox{90}{~~\footnotesize{Spider StyleGAN2 (TIN)}} &
    \includegraphics[width=1.\linewidth]{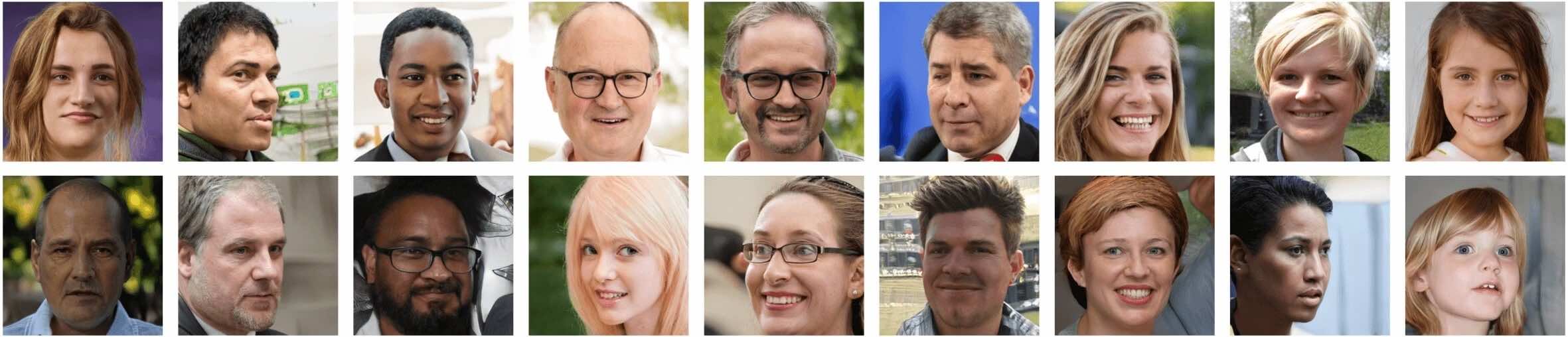}  
    \\[0.25pt] \midrule \\[-10pt]
    \rotatebox{90}{\footnotesize{Spider StyleGAN3-T (TIN)}} &
    \includegraphics[width=1.\linewidth]{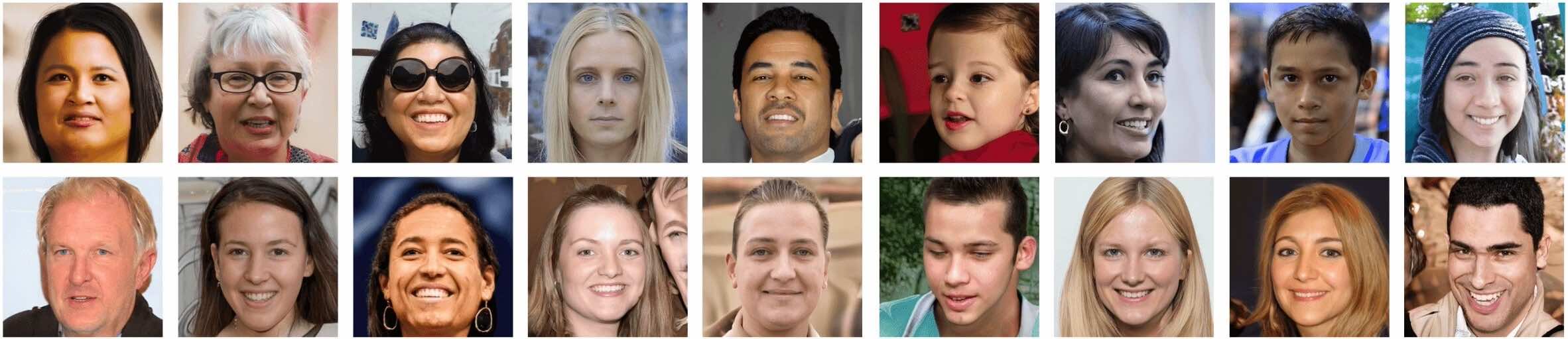}  
    \\[-10pt]
  \end{tabular} 
\caption[]{A comparison of FFHQ ages generated by the baseline and Spider GAN variants trained with AFHQ-Dogs and Tiny-ImageNet (TIN) inputs. Spider StyleGAN2-ADA with the Tiny-ImageNet input performs on par with the StyleGAN-XL baseline (FID of 2.45 for the proposed approach versus FID of 2.07 for the baseline), with a mere one-third of the network complexity. }
\label{Fig_FFHQ_Compares}  
\end{center}
\vskip-1em
\end{figure*}

\begin{figure*}[!thb]
\begin{center}
  \begin{tabular}[b]{P{.9\linewidth}}
    \includegraphics[width=1.\linewidth]{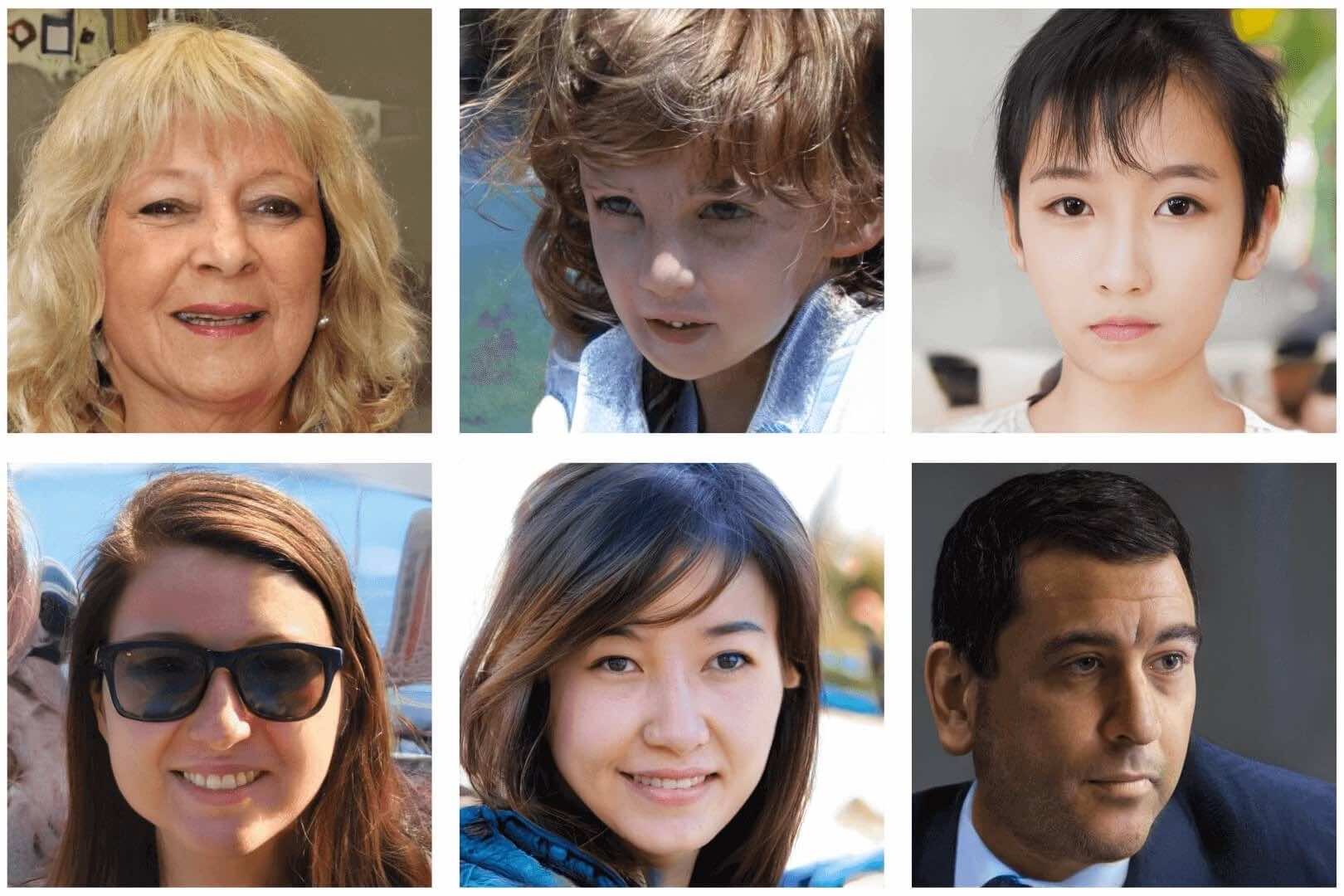}  \\[1.5pt]
    \includegraphics[width=1.\linewidth]{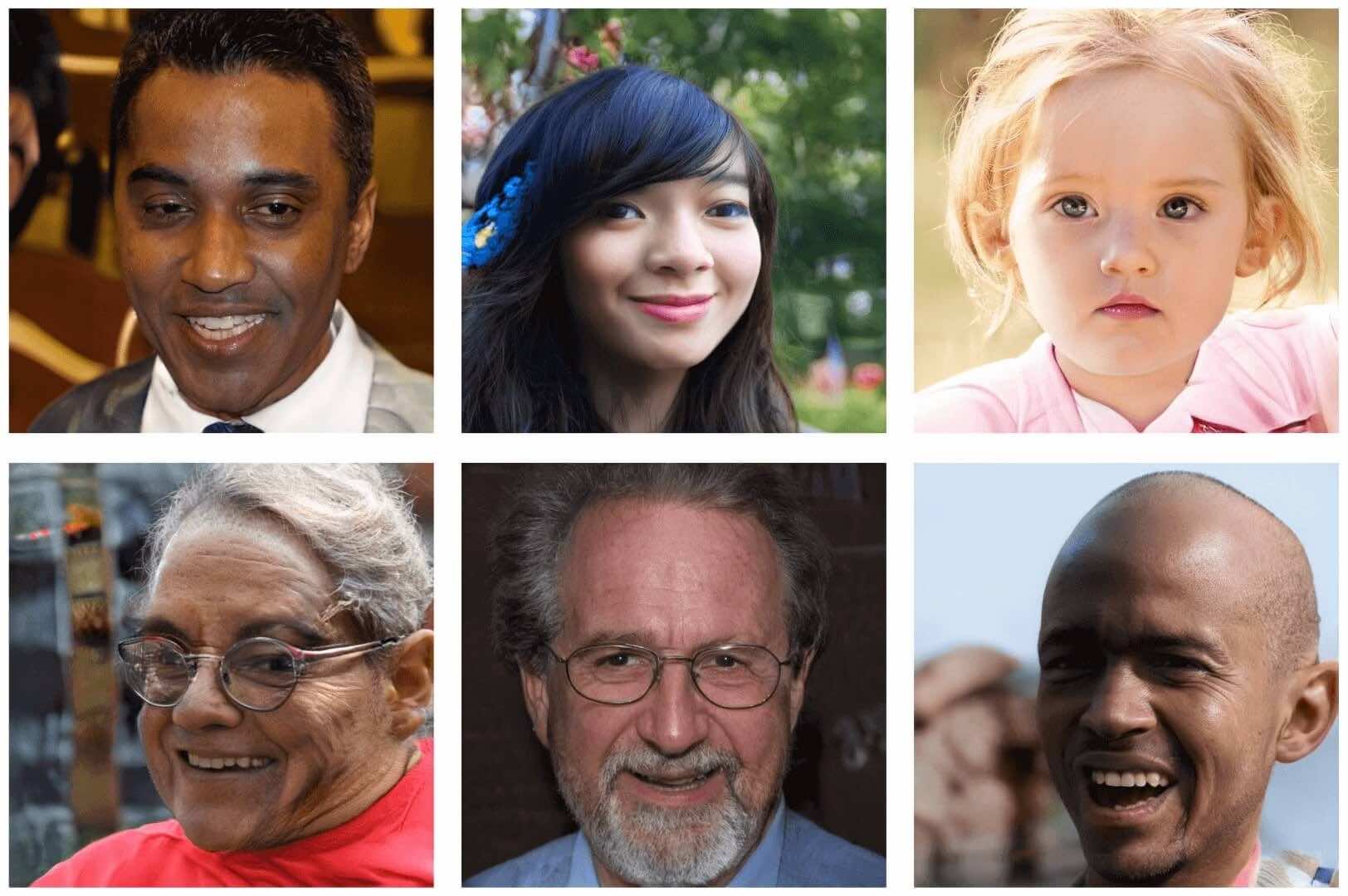}  
  \end{tabular} 
\caption[]{FFHQ images generated by the {\it Spider} variant of StyleGAN2-ADA, trained on a model incorporating weight transfer from the StyleGAN2-ADA model trained on AFHQ-Dogs. The input samples are drawn from a StyleGAN2-ADA model pre-trained on \(32\times32\) AFHQ-Dogs images. The converged model achieved an FID of 3.07 as opposed to 2.70 of the baseline model. The lower FID can be attributed to the choice of a {\it poor neighbor} of the target dataset. }
\label{Fig_FFHQS2Dogs}  
\end{center}
\vskip-1em
\end{figure*}

\begin{figure*}[!thb]
\begin{center}
  \begin{tabular}[b]{P{.9\linewidth}}
    \includegraphics[width=1.\linewidth]{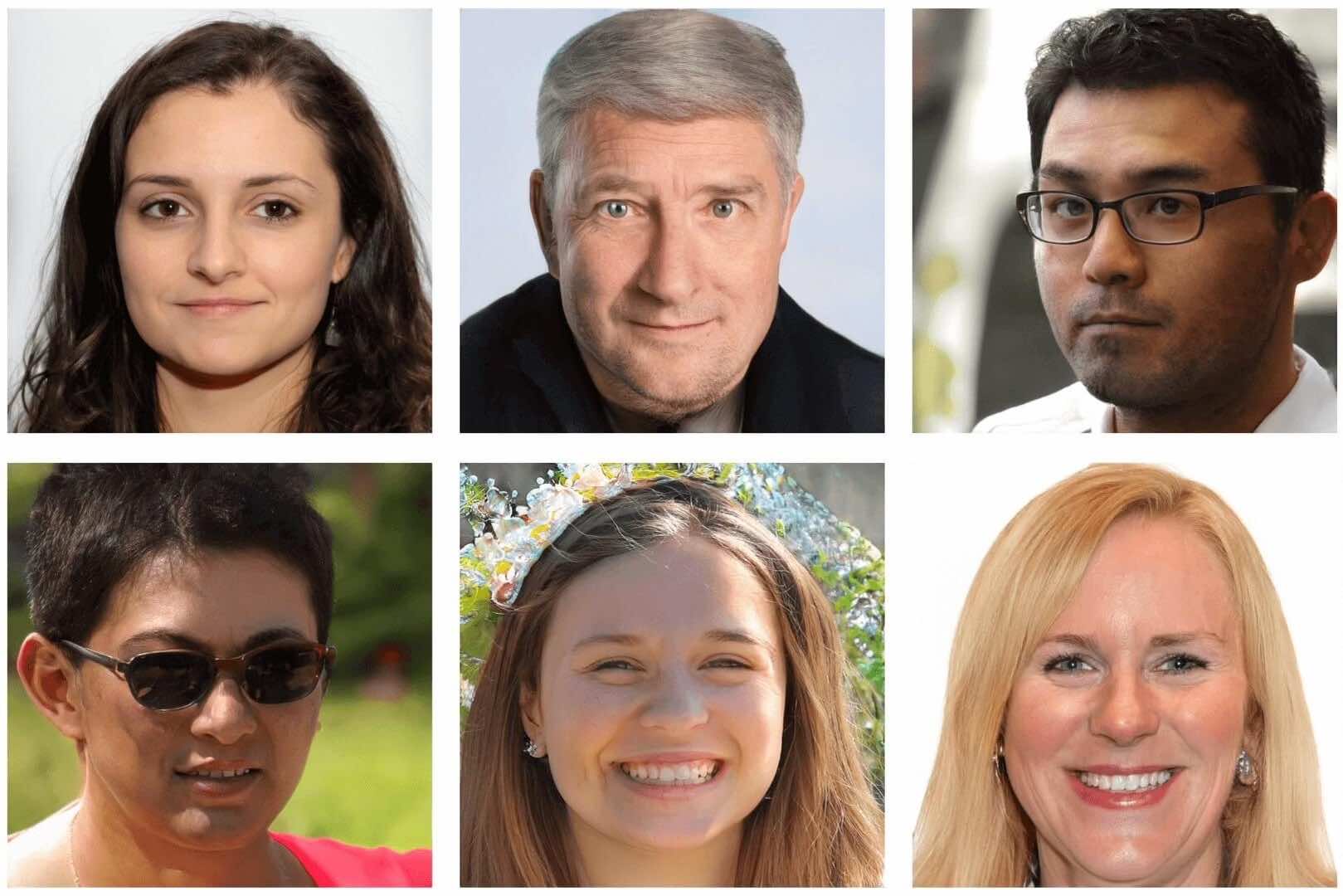}  \\[1.5pt]
    \includegraphics[width=1.\linewidth]{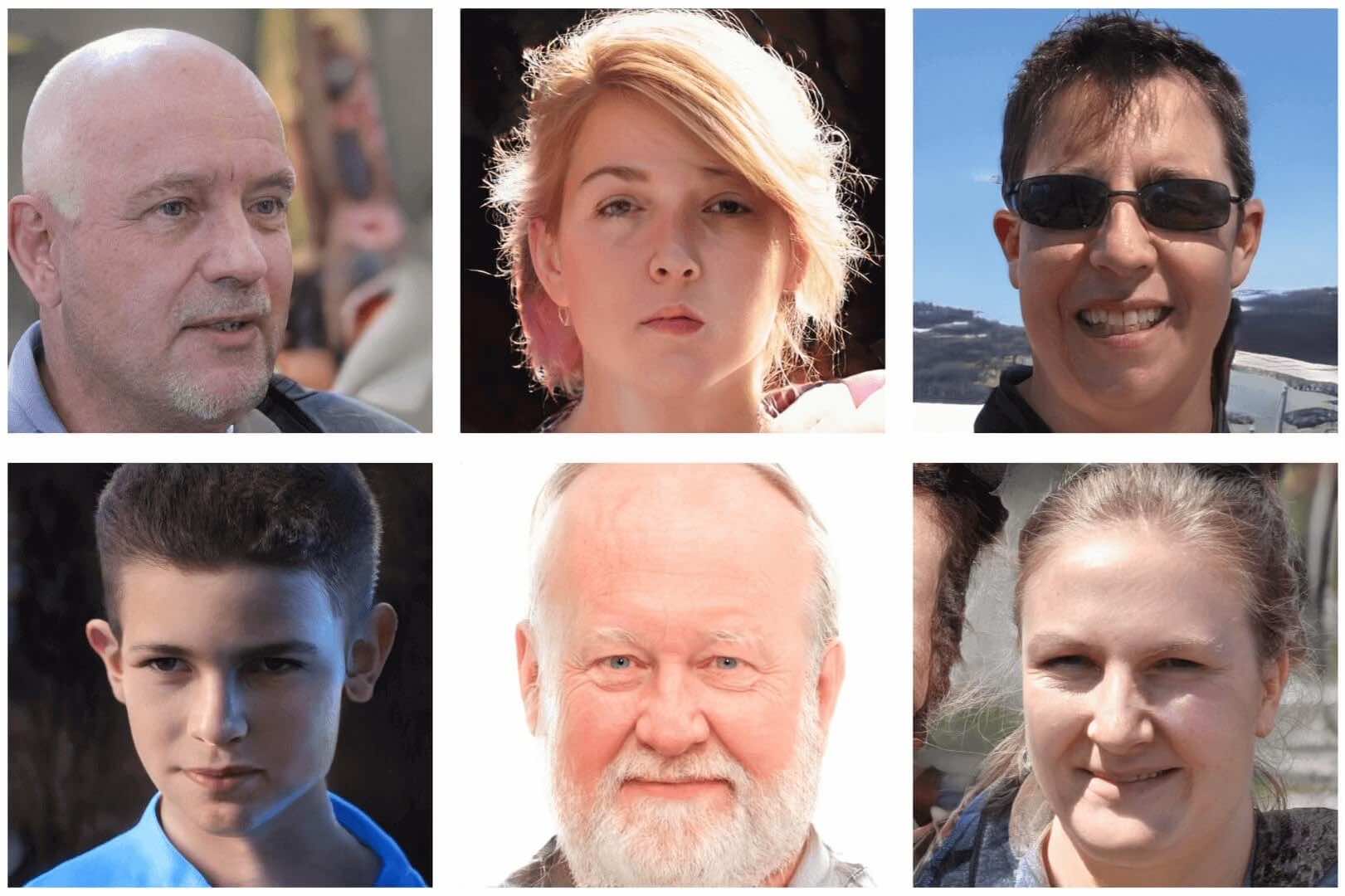}  
  \end{tabular} 
\caption[]{FFHQ images generated by the {\it Spider} variant of StyleGAN2-ADA, trained on a model incorporating weight transfer from the StyleGAN2-ADA model trained on AFHQ-Dogs images. The input samples are drawn from a StyleGAN2-ADA model pre-trained on \(32\times32\) Tiny-ImageNet images. The model achieves an FID of 2.45, superior to the baseline StyleGAN2-ADA, Polarity-StyleGAN2 and MaGNET-StyleGAN2 (with FID scores of 2.70, 2.57 and 2.66, respectively). }
\label{Fig_FFHQS2TIN}  
\end{center}
\vskip-1em
\end{figure*}

\begin{figure*}[!thb]
\begin{center}
  \begin{tabular}[b]{P{.9\linewidth}}
    \includegraphics[width=1.\linewidth]{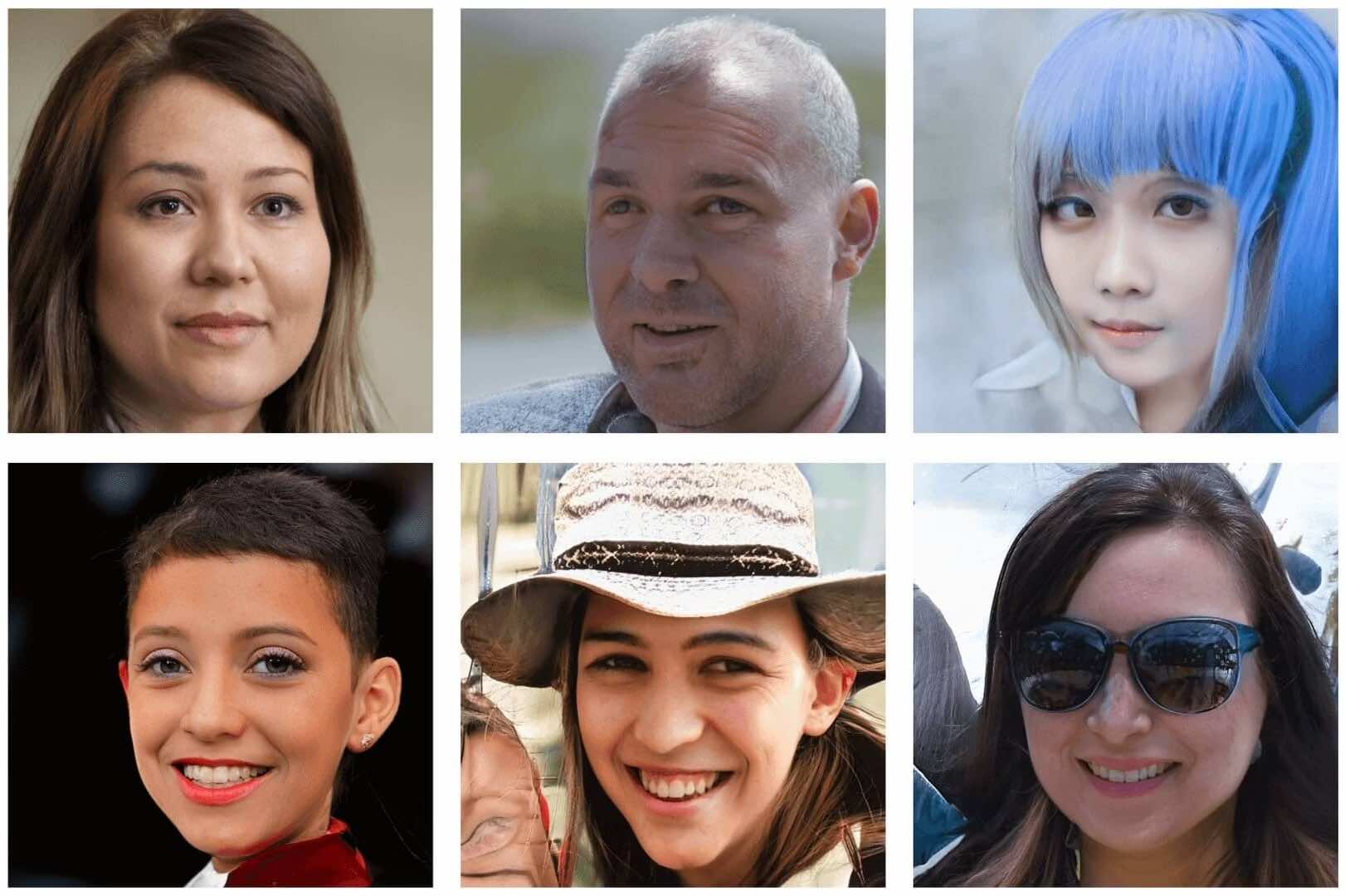}  \\[1.5pt]
    \includegraphics[width=1.\linewidth]{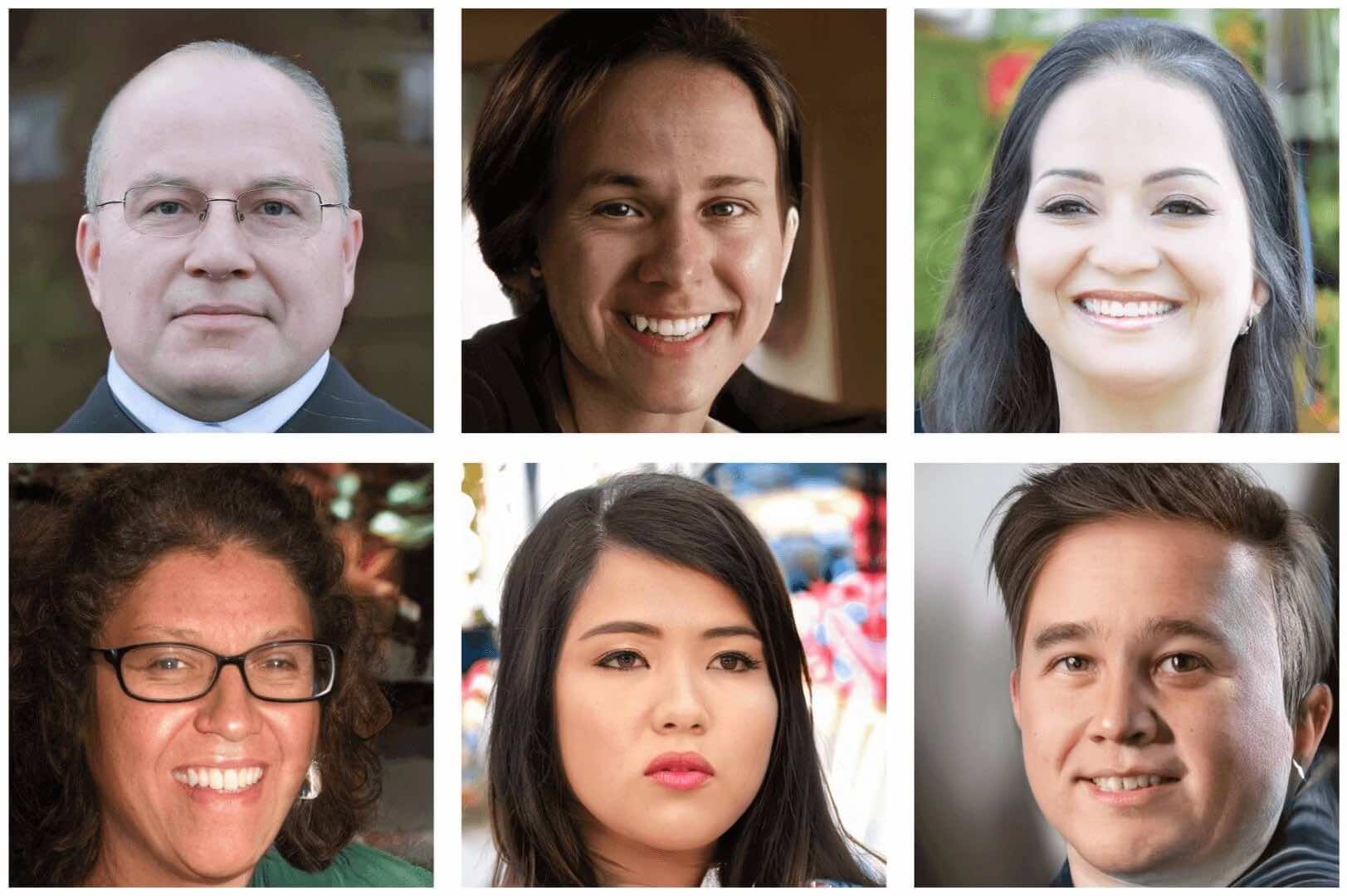}  
  \end{tabular} 
\caption[]{FFHQ images generated by the {\it Spider} variant of StyleGAN3-T, trained on Tiny-ImageNet dataset. The model achieved an FID of 2.86.}
\label{Fig_FFHQS3TIN}  
\end{center}
\vskip-1em
\end{figure*}

\begin{figure*}[!thb]
\begin{center}
  \begin{tabular}[b]{P{.9\linewidth}}
\includegraphics[width=1.\linewidth]{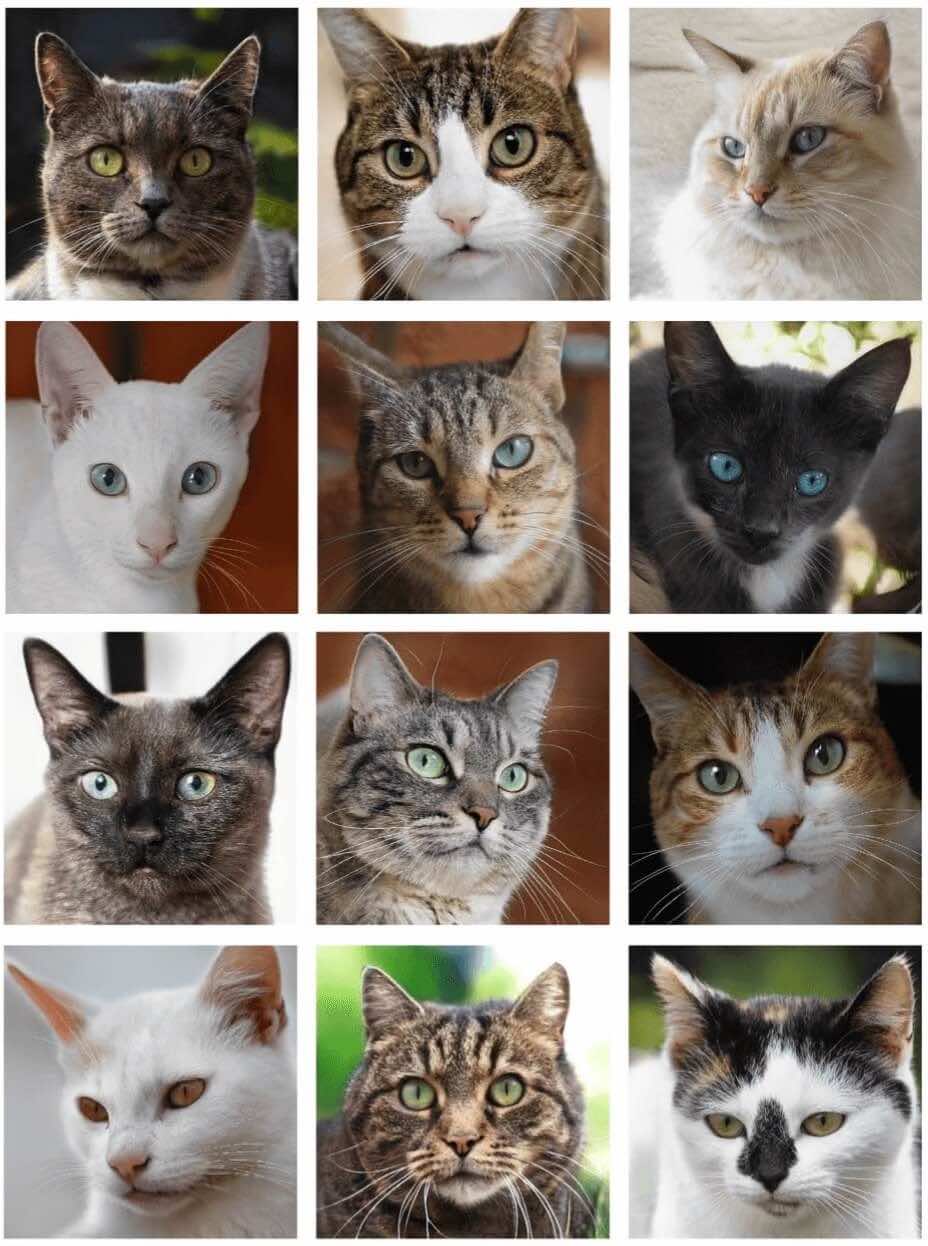}  
  \end{tabular} 
\caption[]{AFHQ-Cat images generated by the {\it Spider} variant of StyleGAN2-ADA, trained on a model incorporating weight transfer from the StyleGAN2-ADA model trained on AFHQ-Dogs. The input samples are drawn from a StyleGAN2-ADA model pre-trained on \(32\times32\) Tiny-ImageNet images. The converged model achieves an FID of 3.86 in one-fifth of the training iterations required by the baseline. }
\label{Fig_CatsS2}  
\end{center}
\vskip-1em
\end{figure*}

\begin{figure*}[!thb]
\begin{center}
  \begin{tabular}[b]{P{.9\linewidth}}
\includegraphics[width=1.\linewidth]{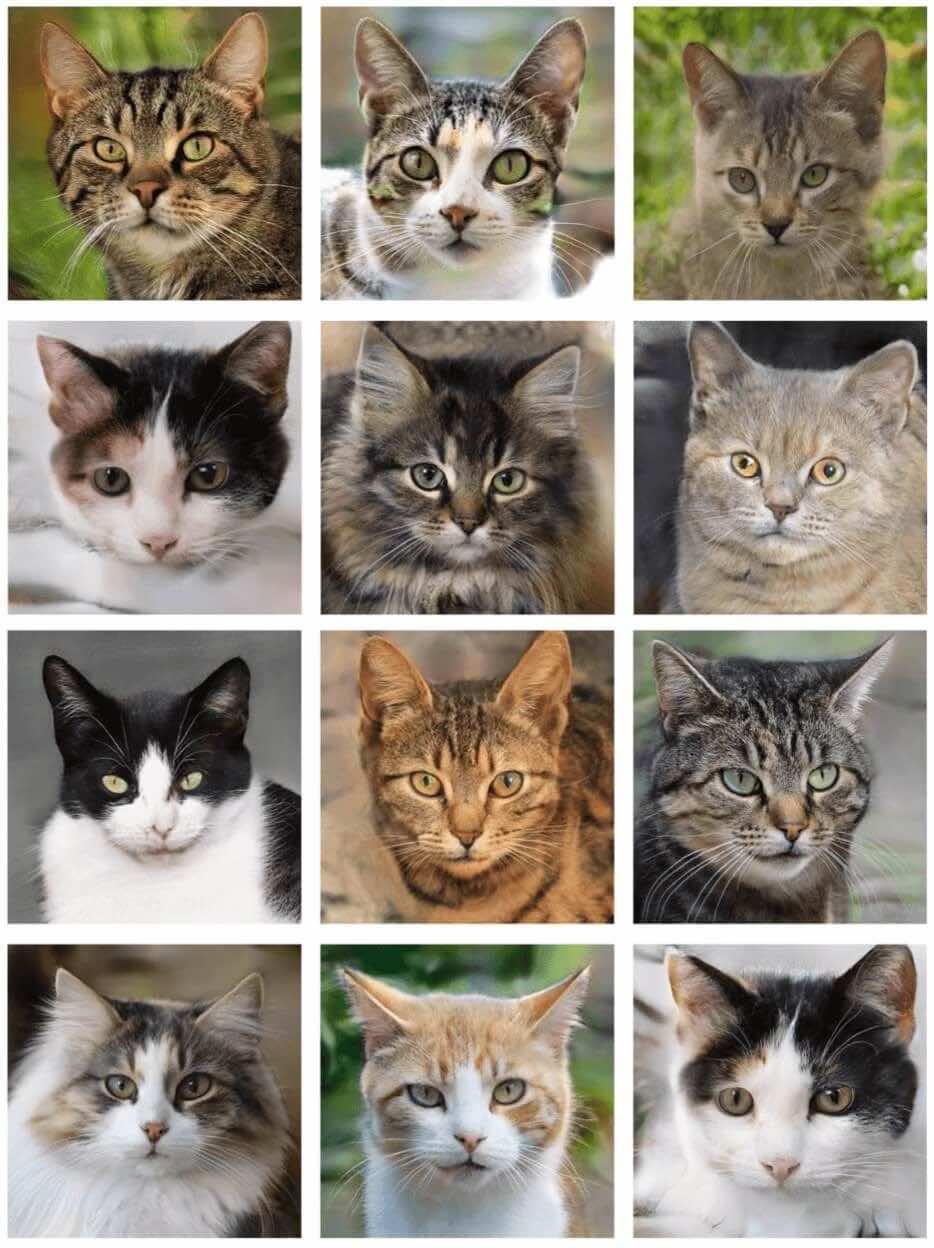}  
  \end{tabular} 
\caption[]{AFHQ-Cat images generated by the {\it Spider} variant of StyleGAN3-T from scratch. The input samples are drawn from a StyleGAN3-T model model pre-trained on \(32\times32\) AFHQ-Dog images. The converged model achieves an FID of 6.29, which is on par with the baselines, in a mere one-fifth of the suggested~\cite{StyleGAN321} training iterations. }
\label{Fig_CatsS3}  
\end{center}
\vskip-1em
\end{figure*}

\begin{figure*}[!thb]
\begin{center}
  \begin{tabular}[b]{P{.9\linewidth}}
\includegraphics[width=1.\linewidth]{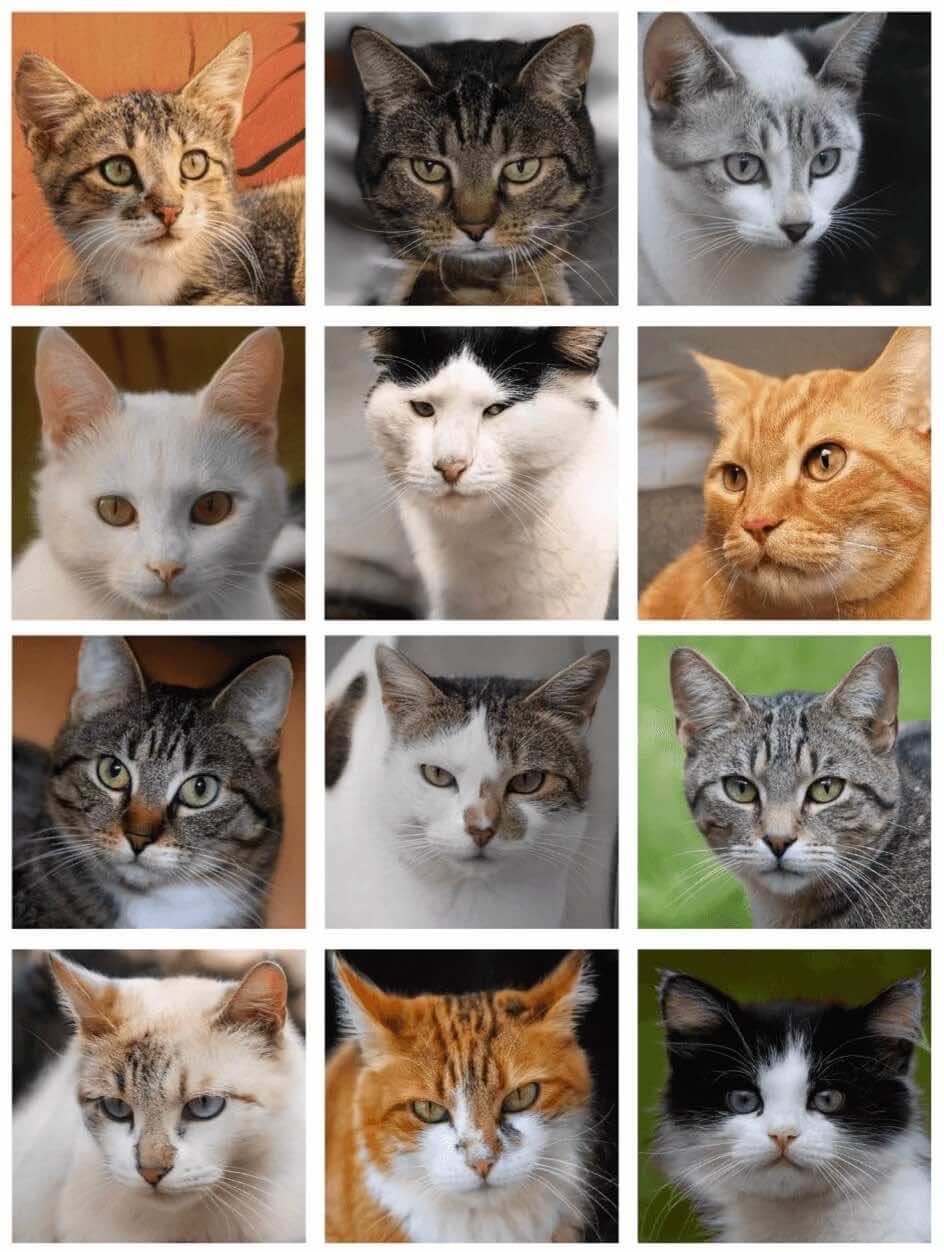}  
  \end{tabular} 
\caption[]{AFHQ-Cat images generated by the {\it Spider} variant of StyleGAN3-T, trained on a model incorporating weight transfer from the StyleGAN3-T model trained on AFHQv2-Dog. The input samples are drawn from a StyleGAN2-ADA model model pre-trained on \(32\times32\) Tiny-ImageNet images. The converged model achieves a state-of-the-art FID of 3.07 and KID of \(0.23\times10^{-3}\) in one-fourth of the training iterations of baseline StyleGAN3~\cite{StyleGAN321}.}
\label{Fig_CatsS3Wts}  
\end{center}
\vskip-1em
\end{figure*}

The comparison of FID and CSID\(_m\) of the StyleGAN variants trained on FFHQ are provided in Table~\ref{Table_SpiderStyleGAN_FFHQ_App}. Spider StyleGAN2-ADA with the Tiny-ImageNet input achieved an FID score on par with StyleGAN-XL, a model with three-fold higher network complexity. However, in terms of CSID\(_m\), Spider StyleGAN2-ADA achieves state-of-the-art performance, which suggests that the diversity of images generated by Spider StyleGAN2-ADA is superior to that of StyleGAN-XL. The Spider StyleGAN3 model with weight transfer achieves a state-of-the-art FID of 3.07 on AFHQ-Cats, with one-fourth of the training iterations as the baselines. The accelerated convergence can be attributed to the superior {\it initialization} in the Spider GAN framework, as opposed to initializing with high-dimensional Gaussian inputs. Figures~\ref{Fig_UkiyoES2Dogs}-~\ref{Fig_CatsS3Wts} show the images generated by the various models and side-by-side comparison of the images generated by Spider StyleGAN and baseline variants.

\subsubsection{Interpolating with Spider StyleGAN3}\label{App_InterpolStyleGAN}

\begin{figure*}[!thb]
\begin{center}
  \begin{tabular}[b]{P{.9\linewidth}}
    \includegraphics[width=0.35\linewidth]{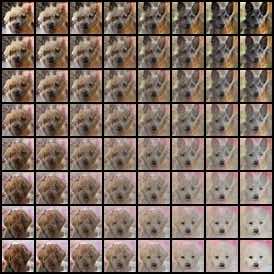}  \\[1.5pt]
    \includegraphics[width=1.\linewidth]{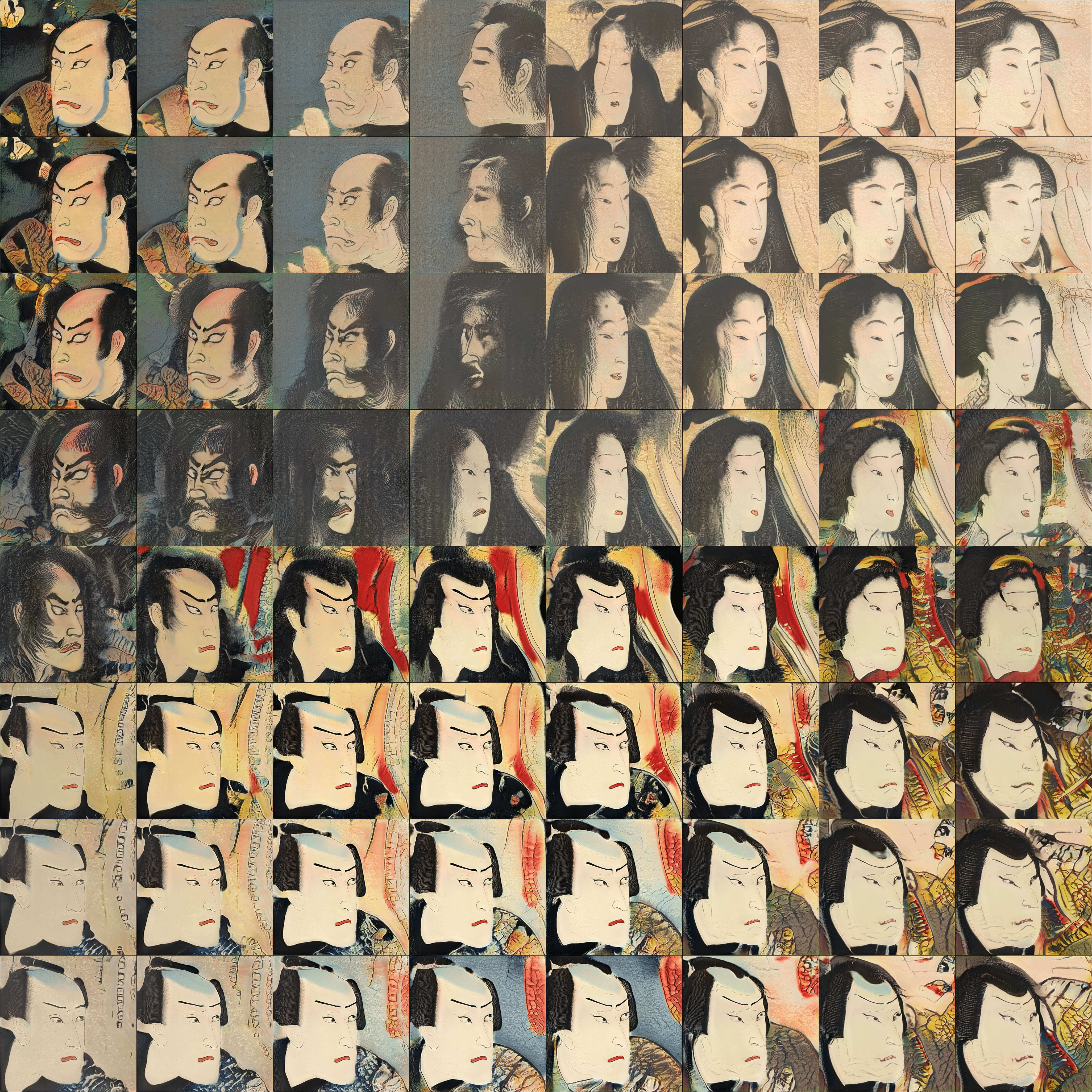}  \\[-7pt]
  \end{tabular} 
\caption[]{A grid of interpolated Ukiyo-E images generated by Spider StyleGAN2-ADA, trained on AFHQ-Dogs. Images are generated by transforming Gaussian noise to AFHQ-Dogs images via an input-stage model, whose subsequent outputs serve as the input to Spider StyleGAN2-ADA. 
The interpolation is performed in the AFHQ-Dogs space, and provided as input to Spider StyleGAN2-ADA. We observe smooth transitions between the interpolated  images, which allows for fine-grained control of the features. }
\label{Fig_UkiyoES2DogsMidInterpol}  
\end{center}
\vskip-1em
\end{figure*}

\begin{figure*}[!thb]
\begin{center}
  \begin{tabular}[b]{P{.9\linewidth}}
    \includegraphics[width=0.33\linewidth]{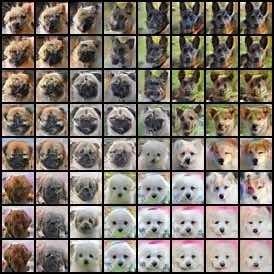}  \\[1.5pt]
    \includegraphics[width=0.97\linewidth]{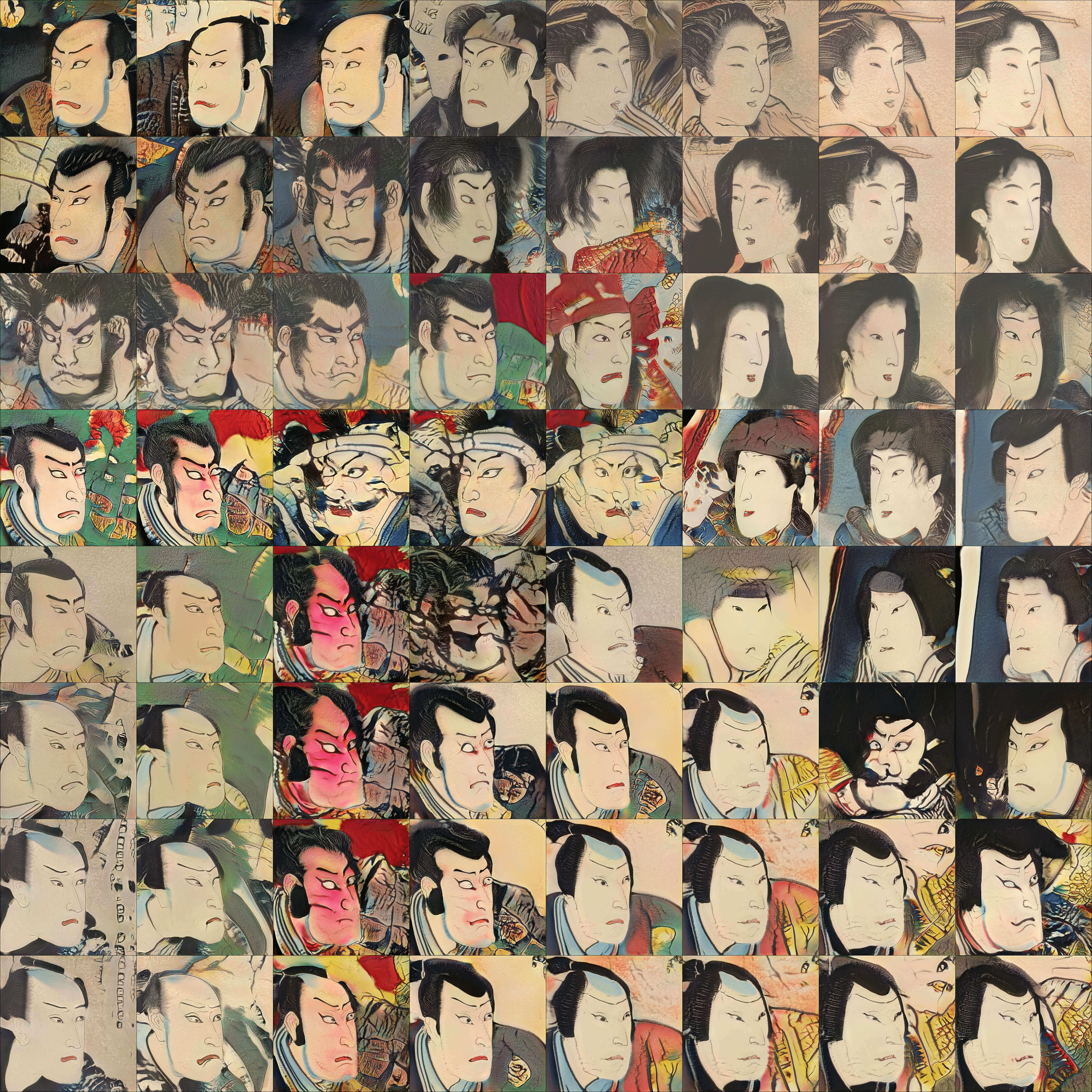}   \\[-7pt]
  \end{tabular} 
\caption[]{A grid of interpolated Ukiyo-E images generated by the Spider StyleGAN2-ADA, trained on AFHQ-Dogs. Images are generated by transforming Gaussian noise to AFHQ-Dogs images via an input-stage model, whose subsequent outputs serve as the input Spider StyleGAN2-ADA. In this case, the interpolation is performed in the Gaussian space and fed to the input-stage pre-trained StyleGAN. The corresponding AFHQ-Dogs images generated are provided as input to the Spider StyleGAN2-ADA. We observe abrupt and unnatural transitions between images. Some images also appear to be unrealistic, which is not surprising, as the interpolation of points drawn from a Gaussian manifold have an extremely low probability of lying on the manifold.  }
\label{Fig_UkiyoES2DogsInputInterpol}  
\end{center}
\vskip-1em
\end{figure*}

\begin{figure*}[!thb]
\begin{center}
  \begin{tabular}[b]{P{.9\linewidth}}
    \includegraphics[width=0.35\linewidth]{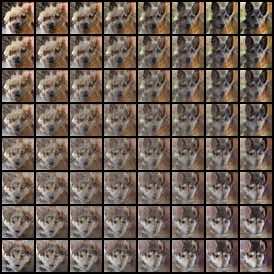}  \\[1.5pt]
    \includegraphics[width=1.\linewidth]{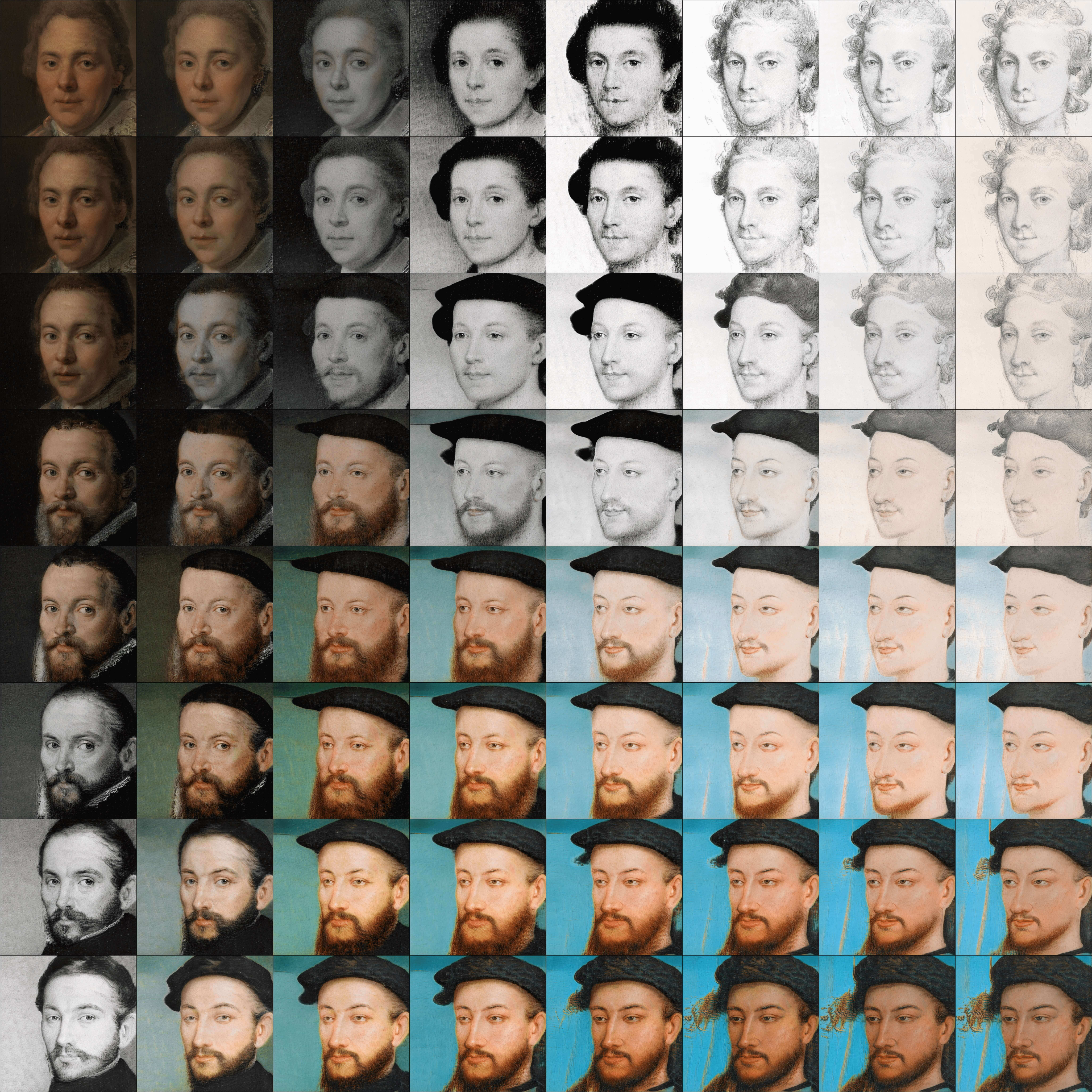}  
  \end{tabular} 
\caption[]{Interpolations on the MetFaces  images generated by the Spider StyleGAN2-ADA, trained on AFHQ-Dogs. The inputs to the Spider StyleGAN are  linearly interpolated AFHQ-Dogs images. We observe smooth and gradual transitions between the color- and sketch-based images generated by Spider StyleGAN, which is highly desirable for feature manipulation. }
\label{Fig_MetFacesS2DogsMidInterpol}  
\end{center}
\vskip-1em
\end{figure*}

\begin{figure*}[!thb]
\begin{center}
  \begin{tabular}[b]{P{.9\linewidth}}
    \includegraphics[width=0.35\linewidth]{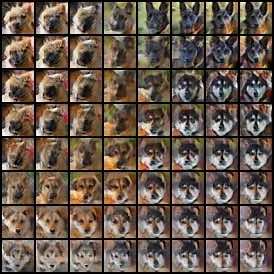}  \\[1.5pt]
    \includegraphics[width=1.\linewidth]{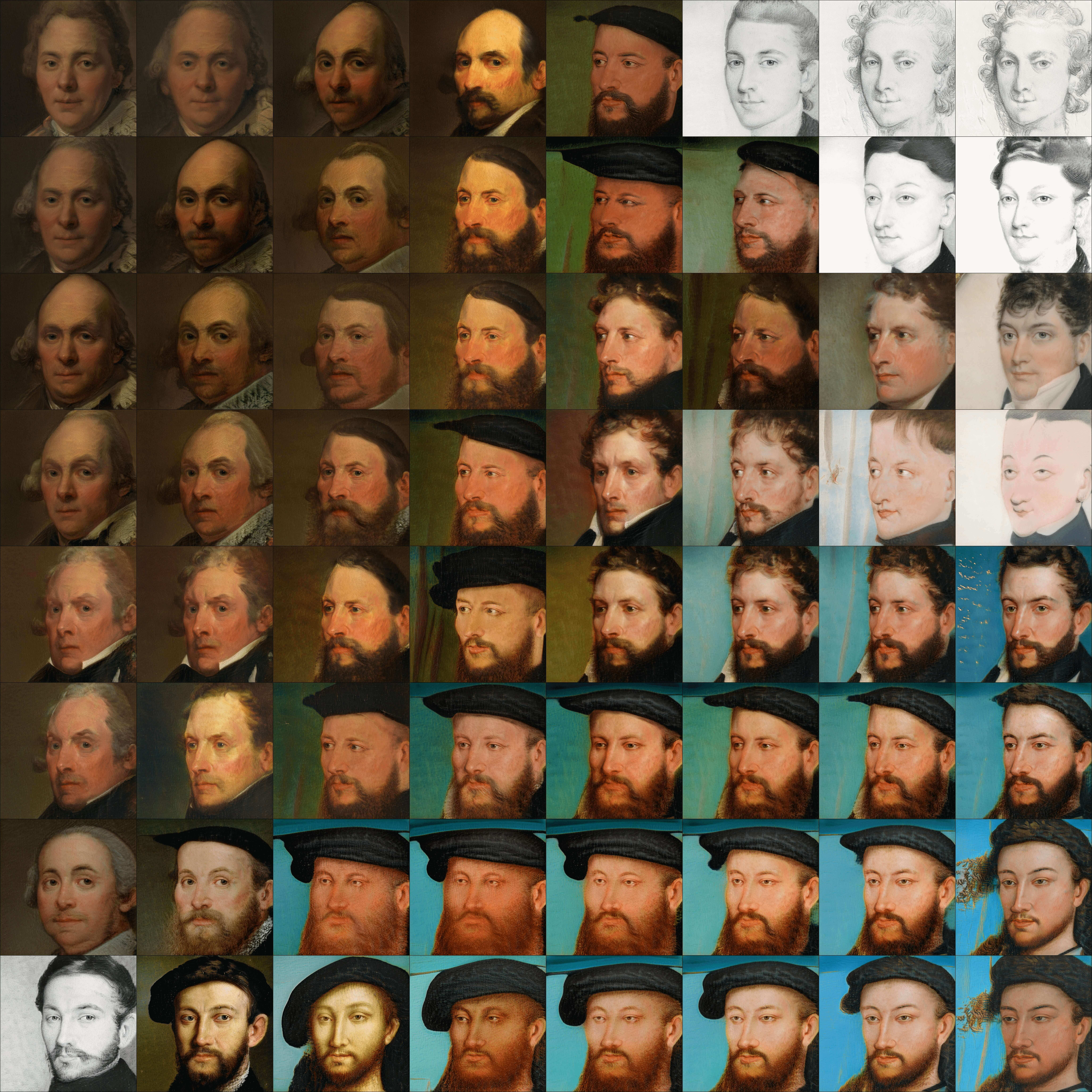}  \\[-3pt]
  \end{tabular} 
\caption[]{Interpolated MetFaces images generated by the Spider StyleGAN2-ADA, trained on AFHQ-Dogs. The interpolation is carried out in the Gaussian fed to the pre-trained input-stage StyleGAN. The corresponding AFHQ-Dogs images generated are given as input to Spider StyleGAN2-ADA. We observe unnatural and discontinuous transitions between the color and sketch images which can be attributed to the disconnected manifold structure of the dataset. }
\label{Fig_MetFacesS2DogsInputInterpol}  
\end{center}
\vskip-1em
\end{figure*}

\begin{figure*}[!thb]
\begin{center}
  \begin{tabular}[b]{P{.9\linewidth}}
    \includegraphics[width=0.35\linewidth]{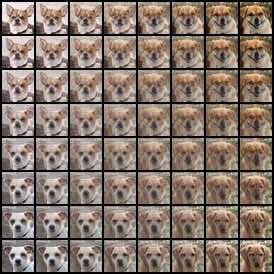}  \\[1.5pt]
    \includegraphics[width=1.\linewidth]{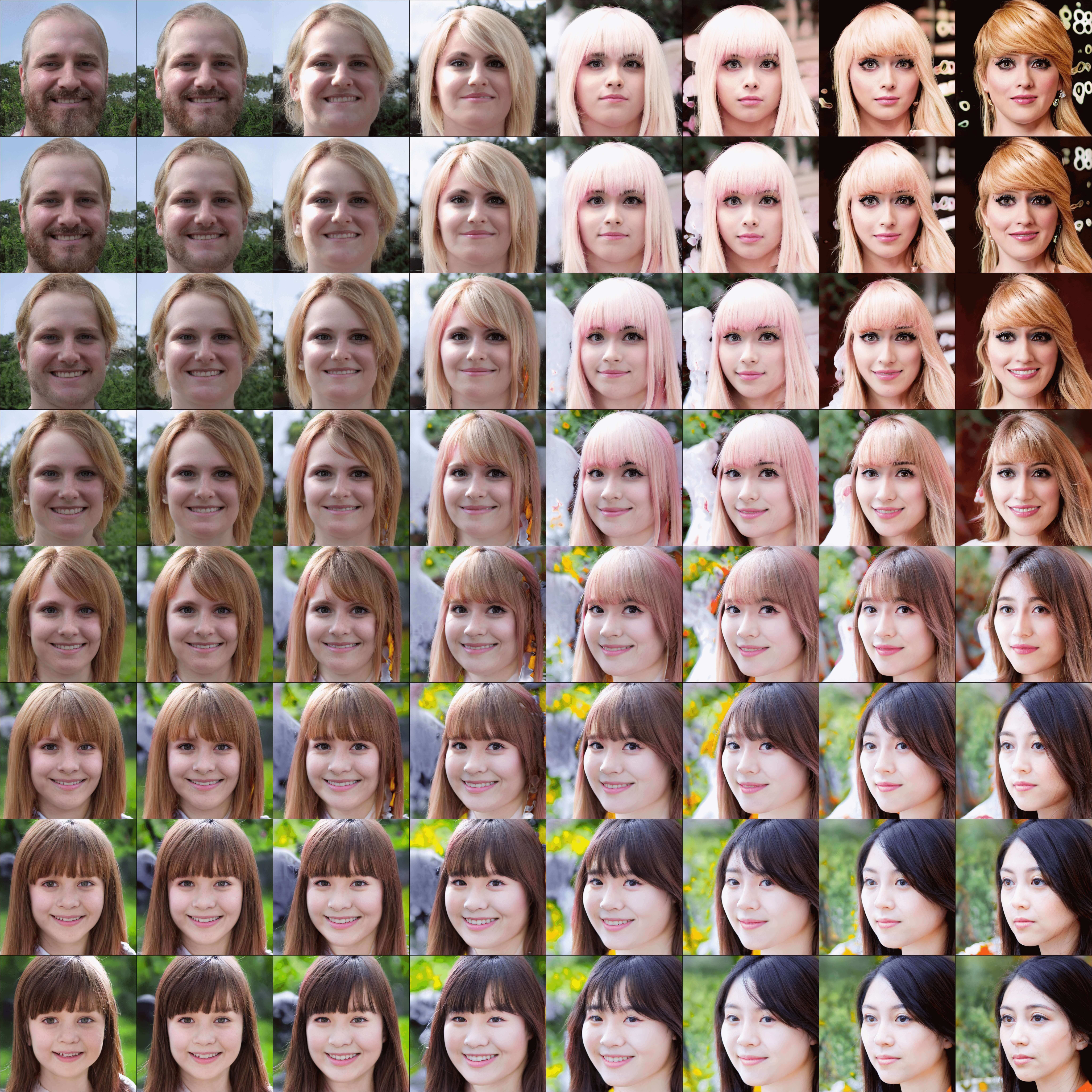}  
  \end{tabular} 
\caption[]{Interpolations on the FFHQ  images generated by the Spider StyleGAN2-ADA, trained on AFHQ-Dogs. The inputs to the Spider StyleGAN are linear interpolates computed on the AFHQ-Dogs images. We observe that the proposed Spider variant generates smooth and gradual transitions with fine-grained facial features allowing for superior control of the image generation. }
\label{Fig_FFHQS2DogsMidInterpol}  
\end{center}
\vskip-1em
\end{figure*}

\begin{figure*}[!thb]
\begin{center}
  \begin{tabular}[b]{P{.9\linewidth}}
    \includegraphics[width=0.35\linewidth]{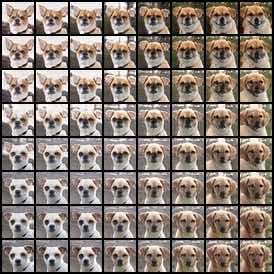}  \\[1.5pt]
    \includegraphics[width=1.\linewidth]{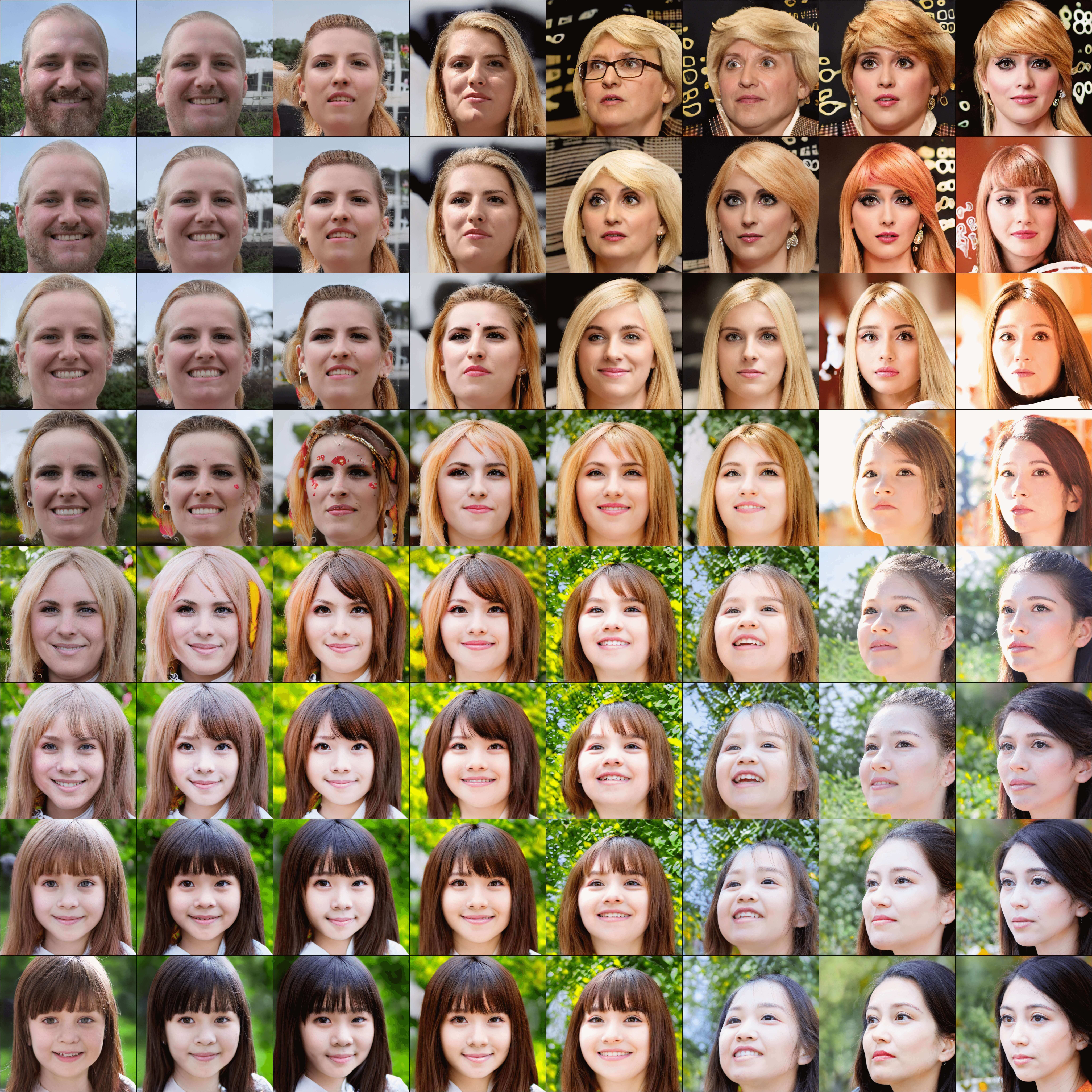} \\[-7pt]
  \end{tabular} 
\caption[]{Interpolated FFHQ images generated by the Spider StyleGAN2-ADA, trained on AFHQ-Dogs. Interpolation is performed in the Gaussian space of the input-stage StyleGAN, which generate a set of AFHQ-Dogs images, which in turn serve as the input to the Spider StyleGAN2-ADA. We observe discontinuous transitions in the hair, color, and other features of the generated images. Some images are also noisy, as they correspond to inputs drawn from outside of the training manifold. }
\label{Fig_FFHQS2DogsInputInterpol}  
\end{center}
\vskip-1em
\end{figure*}

In order to better understand the control over representations that the {\it Spider} framework provides, we consider interpolation experiments on cascaded Spider StyleGAN2-ADA. A pre-trained SpiderStyleGAN2 with Gaussian distributed input and AFHQ-Dogs as output forms the input-stage network. The outputs of this network serve as the input to Spider StyleGAN2-ADA. As discussed in Section~\ref{Sec_UnderstandSpiderGAN}, we consider the following two interpolation schemes:
\begin{itemize}
\item  Scheme-1, where interpolation is carried out between the AFHQ-Dogs images generated by the input-stage GAN, and subsequently fed to cascaded Spider GAN stage. Figures~\ref{Fig_UkiyoES2DogsMidInterpol},~\ref{Fig_MetFacesS2DogsMidInterpol} and~\ref{Fig_FFHQS2DogsMidInterpol} present the outputs of the first- and second-stage GANs, when trained on Ukiyo-E Faces, MetFaces and FFHQ images, respectively. 
\item  Scheme-2, where linear interpolation is performed in the Gaussian space. The corresponding samples are used to generate AFHQ-Dogs images, which are fed as input to the Spider GAN stage. Figures~\ref{Fig_UkiyoES2DogsInputInterpol},~\ref{Fig_MetFacesS2DogsInputInterpol} and~\ref{Fig_FFHQS2DogsInputInterpol} show the intermediate AFHQ-Dogs and Spider GAN outputs for this configuration, when trained on Ukiyo-E Faces, MetFaces and FFHQ images, respectively. 
\end{itemize}
Across all datasets, we observe that Scheme-1 results in superior control over the features, with gradual, fine-grained transitions between the images. On the other hand, images generated by Scheme-2 are affected by the known caveats of Gaussian-space interpolation~\cite{Gamma18,NonPara19}. Interpolations of Gaussian-distributed points have a very low probability of lying on the Gaussian manifold. Consequently, the generated AFHQ-Dogs images, and the subsequent target-dataset images possess unnatural discontinuities, appearing unrealistic. In the case of generating FFHQ and Ukiyo-E Faces, this results in the generation of noisy images at intermediate locations.

\begin{table*}[!bht]
\fontsize{10}{12}\selectfont
\begin{center}
\caption{A comparison of StyleGAN2-ADA and StyleGAN3 variants, in terms of FID, KID and CSID\(_m\), on learning FFHQ. A \({\dagger}\) indicates a reported score. Spider StyleGAN2-ADA performs on par with the state-of-the-art StyleGAN-XL (three fold higher network complexity)~\cite{StyleGANXL22} in terms of FID and KID. However, Spider StyleGAN2-ADA variants achieved the best (lowest) CSID\(_m\) scores, which suggests that the {\it Spider} variants learnt more diverse representations of the target dataset when compared against the baselines. }\label{Table_SpiderStyleGAN_FFHQ_App}  \vskip-0.05in
\begin{tabular}{P{5.15cm}|P{3.05cm}||P{2.6cm}|P{2.6cm}|P{1.6cm}}
\toprule 
Architecture &  Input  & Clean-FID~\cite{CleanFID21} & Clean-KID~\cite{CleanFID21} & CSID\(_m\) \\
\hline\hline \\[-10pt]
StyleGAN2-ADA \cite{ADAStyleGAN20} 	&  Gaussian & 2.70\(^\dagger\)  & \(0.906 \times 10^{-3}\) &  2.65    \\[3pt]
StyleGAN3-T \cite{StyleGAN321} 	&  Gaussian & 2.79\(^{\dagger}\)  & \(1.031 \times 10^{-3}\) & 2.95  \\[3pt]
StyleGAN-XL~\cite{StyleGANXL22} & Gaussian & \(\bm{2.02}\)\(^{\dagger}\) &\(\bm{0.287} \bm{\times} \bm{10^{-3}}\) & 3.94  \\\midrule
Spider StyleGAN2-ADA {\bf (Ours)}		&  TinyImageNet  & \uline{2.45}  & \(0.915 \times 10^{-3}\) & \(\bm{1.99}\) \\[3pt]
 Spider StyleGAN2-ADA {\bf (Ours)} 		&  AFHQ-Dogs  & 3.07  & \uline{\(0.795 \times 10^{-3}\)} & \uline{2.55} \\[3pt]
Spider StyleGAN3-T {\bf (Ours)} 	&  TinyImageNet & 2.86  &  \(1.162 \times 10^{-3}\) & 3.25 \\[3pt]
\bottomrule
\end{tabular}
\end{center}
\vskip-12pt
\end{table*}

\begin{sidewaysfigure}
\begin{center}
      \includegraphics[width=0.99\linewidth]{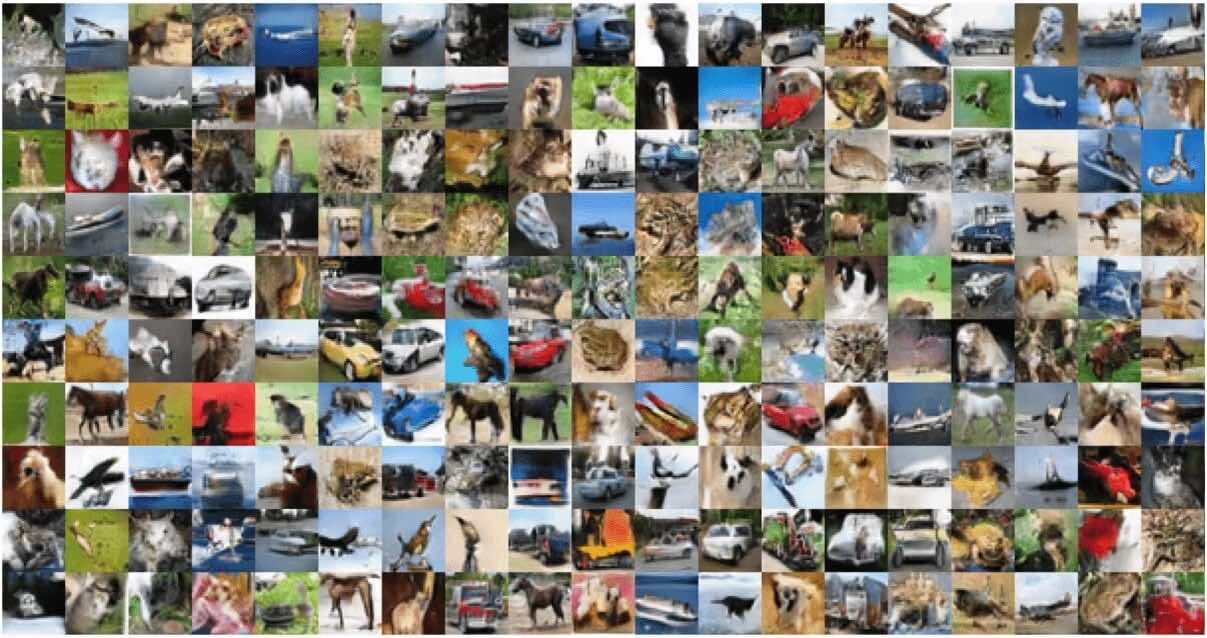} \\[10pt]
\caption[]{Images generated by Spider PGGAN with CIFAR-10 as the target, when trained with Tiny-ImageNet as input, which in turn, is the target for a PGGAN trained with Gaussian noise as the input.}
\label{Fig_SpiderPGGAN_C10}  
\end{center}
\vskip-1em
\end{sidewaysfigure}

 \FloatBarrier
\newpage

 \section{GitHub Repository and Code Release} \label{App_GitHub}
 
The codebase for implementing Spider GAN, Spider PGGAN, Spider StyleGAN and SID has been included as part of the Supplementary. The baseline non-parametric prior~\cite{NonPara19} was implemented using the publicly released {\it .mat} file. PGGAN~\cite{PGGAN18} and StyleGAN2 and StyleGAN3 were implemented using publicly available GitHub repositories, with modification included to implement their respective {\it Spider} variants. \par
 The implementation for CSID\(_m\) and SID are based on the Inception features provided by the Clean-FID~\cite{CleanFID21} library. In order to maintain uniformity, SID can also be computed by providing the path to existing source and target image folders, akin to FID and KID. \par
 
 An implementation of SID atop the Clean-FID~\cite{CleanFID21} backbone, with associated animations of the experiments presented in this manuscript are available at \url{https://github.com/DarthSid95/clean-sid}. The TensorFlow-based source code for Spider GANs built atop the DCGAN architecture, and associated pre-trained models are available at \url{https://github.com/DarthSid95/SpiderDCGAN}. The PyTorch-based source code for implementing the {\it Spider} variants of PGGAN, StyleGAN2, StyleGAN2-ADA and StyleGAN3, with the corresponding pre-trained models are available at \url{https://github.com/DarthSid95/SpiderStyleGAN.} The GitHub repositories also include the full-resolution versions of the images provided in the {\it Main Manuscript} and {\it Appendices}, and animations associated with the {\it Supplementary Material}.

\end{document}